# Ultra-Reduced-Impact-Encased-Logging (URIEL): propose a new method for selective sustainable logging and post-harvest silvicultural treatment in tropical forest using airborne robotics systems

Daniel Albiero[1]*, Gelton Fernando de Morais[1], Daniela Han[1], Flávio Roberto de Freitas Gonçalves[1], Artur Vitório Andrade Santos[1], Wesllen Lins de Araújo[1,2], Alessandra Maia Freire[1], Cláudio Kiyoshi Umezu[1], Mateus Peressin[1], Francesco Toscano[3], Admilson Írio Ribeiro[4] Alfeu J. Sguarezi Filho[5], Américo Ferraz Dias Neto[1], Angel Pontin Garcia[1].

[1] School of Agricultural Engineering, University of Campinas (UNICAMP), Campinas, 13083-875, Brazil;

[2] School of Mechanical Engineering, University of Campinas (UNICAMP), Campinas, 13083-875, Brazil;

[3] Depart. of Agricultural, Forestry, Food and Environmental Sciences, University of Basilicata, Viale Dell'Ateneo Lucano 10, 85100, Italy

[4] Sorocaba Environmental Engineering, São Paulo State University (UNESP), Sorocaba, 18087-180, Brazil;

[5]Center for Engineering, Modeling and Applied Social Sciences, Federal University of ABC (UFABC), Santo André, 09280-560, Brazil;

* Corresponding author: dalbiero@unicamp.br

**Abstracts:** Tropical forests worldwide are under intense deforestation pressure driven by economic and political interests, and scientific evidence suggests this deforestation contributes to climate change. This paper proposes a novel logging method for tropical forests, Ultra-Reduced-Impact-Encased-Logging (URIEL). This new method is based on heli-logging techniques combined with intensive use of robotics and AI integrated with post-harvest silvicultural treatments performed by drones. The concept of appropriate equipment for this method was developed, dimensions were determined, details were completed in a digital proof of concept, and an effective digital simulation and economic feasibility analysis were carried out for various helicopter-timber-distance combinations. The results demonstrated that a URIEL method has high economic viability and makes it possible to virtually eliminate collateral damage to forests while maintaining ecosystem services. The main conclusion of this paper is that, despite the satisfactory scientific and technological results, the feasibility of a Uriel method depends on the integration of stakeholders intrinsic to the context: high-tech industry; political governments; certified logging companies; and native populations.

## Introduction

Tropical forests play an important role in moderating local, regional, and global climate through their impacts on energy, water, and the carbon cycle, which control local and regional rainfall patterns. The evapotranspiration of their plants can contribute 41% of the rainfall in the Amazon and 50% in the Congo, so the rapid deforestation of tropical forests has warmed the climate on local and global scales [1]. In addition, tropical forests have played a critical role in changing atmospheric carbon concentration, acting as a source of carbon emissions since the beginning of the industrial era, and this role has accelerated in the 21st century [2].

In a regional context, the Amazon Rainforest is a tropical forest made up of a complex interconnected system of species, ecosystems, and human culture that contributes to the well-being of people globally. The Amazon Rainforest represents more than 10% of the planet Earth's biodiversity, stores the carbon equivalent of 20 years of global $CO_2$ emissions, and has a cooling effect due to the evapotranspiration of its plants, which helps stabilize the Earth's climate [3].

If all the tropical forests in the world are considered, it becomes clear that the values of the parameters of ecosystem services (biodiversity, $CO_2$ stock, energy balance, atmospheric and fluvial flows, among others) used to measure the consequences of their deforestation on climate change are immense, perhaps one of the fulcrums of global warming. Thus, the indiscriminate and uncontrolled deforestation of tropical forests is a very serious global problem that must be considered scientifically within the common sense of the state of the art.

Understanding the risks of catastrophic behaviors of tropical deforestation in forest terms requires addressing complex factors that shape ecosystem resilience, the biggest question is whether a large-scale collapse of a system like the Amazon Rainforest could occur and whether it could be associated with a specific tipping point [3].

In recent decades, globalization and new technologies have brought about a rapid change in the use of forests, threatening ecosystem services, and to avoid destructive events it is essential to understand the conflicts between the uses of these services [4]. An example of a harmful change is presented by [5] who studied the effect of logging on bird nest density in relation to the microsite (tree density and vegetation cover) and found that due to the changes that logging produced in the forest structure, a negative influence was generated on nest density.

Thus, the adoption of sustainable forest management practices (FMS) is the best way to balance the various ecosystem services of forests, as it applies management practices that aim to obtain products and services from the forest without affecting its capacity and functions, guaranteeing future generations the same opportunities to use forests [6]. The concept of sustainable forest management (SFM) appeared in the early 1990s and its principle is based on the management of the social, ecological and economic resources of forests and currently also focuses on policies related to rural development, biodiversity conservation, water, territories and bioenergy, in short, everything that has influenced expectations regarding forests [7].

Considering the complex of factors involved in an SFM, the concept of resilience has gained prominence in the forestry sector, defined as the capacity of a social or ecological system to absorb disturbances while retaining its basic structure and functionalities, enabling its self-organization and capacity to adapt to change and stress [8]. [9], studying salvage logging of areas that have suffered severe disturbances, such as fires or wind damage, found that areas that underwent salvage logging had more microsite variety with higher temperatures and larger canopy openings, indicating that it is necessary to determine the consequences of salvage before carrying it out, depending on the severity of the natural disturbance.

Logging systems that enable increased resilience of forest systems and reduced disturbances can help to decrease the deleterious effects of harvesting natural tree species of commercial interest, so the development of a logging system with these requirements may be a good solution within the context of SFM to maintain ecosystem services unchanged as advocated by [10].

One of the techniques that can be used is Reduced Impact Logging (RIL), defined by ITTO - International Tropical Timber Organization (2025) [11] as an intensively planned and carefully controlled implementation of timber harvesting operations to minimize environmental impacts on forest stands and soils. It involves forest inventory and individual tree mapping; planning roads and trails to minimize soil impact; cutting vines in areas with high density of vines connecting trees; using cutting, felling and handling techniques to avoid impacting other plants; and finally, removing the harvested material while minimizing damage to the remaining forest.

According to [12] and [13], RIL is a prerequisite for Sustaining Timber Yield (STY), particularly in SFM, but it cannot be confused with these systems. [14] clarifies that STY requires longer cutting cycles (60 years or more) and lower logging intensity than prescribed by law, as well as effective low-impact logging practices and finally silvicultural treatment to promote timber stock recovery. STY fits perfectly into two UN SDGs: goal 13 Climate Action and goal 15 Life on Land [14], because by keeping forests in good condition, ecosystem services automatically help combat climate change, goal 13; and with sustainable timber management, there is sustainable forest management, goal 15.

According to [12], central to all RIL guidelines is the use of felling techniques that increase worker safety, reduce wood waste, and direct the fall of trees to facilitate extraction and protection of forestry. Within RIL there is the Selective Logging (SL) technique which, according to [15], involves the selective extraction of trees from natural forests where efficient management of selective logging is crucial to overcome the challenges of timber demand, biodiversity conservation and carbon sequestration.

Although there is strong evidence that SL integrated into RIL is much more interesting in terms of forest protection and recovery than the Conventional logging (CL) system [16,17]. Extensive research shows that they are not perfect techniques, [18] demonstrated that after 30 years of RIL in the Brazilian Amazon only 50% of the commercial stand was recovered, causing a drastic reduction in harvesting intensity in the second cutting cycle, this according to the authors shows that RIL alone is not sufficient to achieve sustainable forest management.

In addition, there is the economic issue that[12] state that the financial burden is the biggest constraint inhibiting the large-scale adoption of RIL, as loggers who implement RIL have to adopt new work techniques, invest in new equipment, safety systems and personnel training, such requirements impose incremental costs that CL operators avoid. [19] in a literature review found data that present comparative costs between CL and RIL, with only 3 cases where RIL had lower costs versus 6 cases where it had much higher costs.

Another problem according to [20] is that SL integrated into an RIL can alter the biota (presence of seed sources) as well as the distribution of resources (light and soil), modifying the natural seeding of trees of economic interest. These authors concluded that over time the habitat becomes less viable for the regeneration of the logged areas. These results indicate that RIL may have more effects on tree regeneration than previously assumed. In this issue [21] they go even further stating that RIL has little effect on natural regeneration rates of key commercial species in logged areas of neotropical forests.

[22] find the same results stating that there is a lack of natural regeneration of commercial species in the Eastern Amazon.

A very serious issue with RIL is the residual stand damage that generates extensive damage to the remaining forest, represented by three categories depending on the tree fall: snapped at base, severely bent, and damaged cambial tissue, which occur mainly on slopes greater than 17° [12]. [18] reported a damage rate on stand trees of 20.5%, which can reach 39.5% depending on the intensity of management, and the proportion of destroyed trees with a mean of 16% reaching 30%. Another important damage generated by RIL is soil degradation by skid trails, as [18] report that it can occupy 7% of forest soil area. This proportion is very significant in the sense that the soil in the skid trail area is recognized as highly degraded with excessive soil compaction, nutrient removal, and high erosion rates [23].

In this way, to minimize these problems, new methods should be considered for management plans [20]. The use of RIL alone does not guarantee SFM [12,13]. Thus, combining RIL techniques by limiting logging intensities and extending cutting cycles can guarantee the production and maintenance of forests [12]. But for this, the use of silvicultural interventions as advocated by [18,24] is essential.

Peña-Claros et al. (2008) [25] applying silvicultural interventions related to liana cutting and girdling techniques of competing trees ensured a 60% increase in the growth of Future Crop Trees (FCTs). The authors also applied the following techniques: pre-harvest inventory of merchantable commercial trees with DBH of 50-70 cm; skid trial planning; retention of 20% merchantable commercial trees as seed trees; directional felling; pre-harvest marking of FCTs; post-harvest liberation of FCTs from overtopping non-commercial trees by girdling; liana cut on FCTs; Soil scarification in felling gaps; post-harvest girdling of non-commercial trees >40 cm DBH.

Schwartz et al. (2017) [22] states that post-harvesting silvicultural treatments (HST) are a path to improve regeneration of commercial species in managed forests, and among these treatments, logging gap management promotes the regeneration of commercial species, provided there are a high number of seedlings belonging to commercial species. Another technique described by the authors is the opening of canopy gaps by felled trees, which offer microclimatic conditions for seedling establishment. Enrichment by planting, along with reducing competition and maintaining favorable seedling growth conditions, is also important [26]. These authors teach that silvicultural treatments should not begin long after logging actions, as the gaps are quickly closed by surrounding trees, pioneer species, and lianas.

In this context, considering the broad spectrum of a STY-based FSM by SL integrated with RIL with post-harvest silvicultural treatments (HST), this paper aims to propose a new logging method: Ultra-Reduced-Impact-Encased-Logging (URIEL).

This new method is an RIL based on airborne robotic systems. It is an Ultra-Reduced Impact system because, being airborne, it completely avoids impacts on the ground from skid trails, and being robotic practically eliminates residual stand damage because it maximizes the actions of directional feller and liana cutting. It is an Encased system because it optimizes post-harvest silvicultural treatments by encapsulating all actions in the open harvest gap.

The key point of the URIEL method is the helicopter-based aerial transport of timber (Heli-logging – HL) operations. Helicopter logging is a method that uses helicopters to remove felled trees from forests by lifting them with cables attached to the helicopter. This logging method reduces the level of infrastructure required to perform logging and greatly reduces the impacts on the environment [27,28] [27,28]. The first flight with the explicit objective of performing logging took place in 1971 in California, USA [29].

HL is based on the individual selection of trees, which is carried out by personnel specialized in silvicultural knowledge. The trees are selected according to species and specific size, this size being defined according to the volume of wood in economic terms and the lifting capacity of the helicopter [30].

Then the trees are climbed by arborists who cut the branches and top the tree, at the base of the tree a deep perimeter cut is made partially, the trunk is prevented from falling by placing stabilizing wood wedges in the cut, the helicopter approaches and with an appropriate apparatus hanging from a cable grabs the trunk by the top and then it is pulled until the supporting wood in the partial cut breaks [30]. A simpler and safer variation of this method is operationalized by felling the trees before the helicopter approaches, then on the ground these are delimbed and topped, with the arrival of the aircraft the trees are picked up and lifted [31].

HL can be a dangerous and very costly operation [27,29,30,32–34], so an excellent opportunity for development is the intensive and extensive use of agricultural robotics which has as its main characteristic the exponential increase in efficiency and the significant decrease in costs in agricultural/forestry processes [35–42].

According to [43], robots are not new in agriculture and forestry; there is much research being developed, some of it very advanced, and already with actual applications in the field. Agricultural robotics is an overwhelming trend, as the papers by [38,41,44] show. [45] in an extensive study state that robotics techniques not only increase efficiency, but also solve problems of decreasing adequate labor supply. Agricultural robotics is synonymous with the frontier of knowledge in the agricultural area [41], which fits into the concept of agriculture 4.0 [39] through the interface between IoT [46], Connectivity [47], and AI [48].

All this, combined with advances in computer vision technologies [49], sensors [50], and electric mobility [51], presents the trend in technological development in the coming years. Within the topic of logging, some researchers have developed AI processes and techniques that can be implemented, such as those presented in an extensive literature review by [43]. [52] present a linguistic pipeline for semantic processing of robotic commands that combine discriminative structured learning, [53] an interesting paradigm shift as opposed to deep learning techniques. They suggest, instead of using “big data for small tasks,” changing to “small data for bit tasks,” they propose this change through the use of AI systems that use “common sense.”

In more practical terms, [54] present the development of an automatic system for forestry logging processes with cranes. They developed two platforms for testing control systems and motion planning algorithms in real time, presenting the results of this development by providing an overview of our trajectory-planning algorithm and motion-control method. Following this development, La Hera et al. (2024) [55] in an excellent paper present the development of the world's first unmanned machine designed for autonomous forestry operations. A platform equipped with essential hardware components necessary for performing autonomous forwarding tasks was developed. Through the use of computer vision, autonomous navigation, and manipulator control algorithms, the machine is able to pick up logs from the ground and maneuver through a range of forest terrains without the need for human intervention. The results demonstrate the potential for safe and efficient autonomous extraction of logs in the cut-to-length harvesting process.

A very strong perception of the authors of this paper is that all the technologies necessary to operationalize the URIEL method in a concrete way are already in advanced development and can be classified as TRL (Technology Readiness Level) 7, and most are already prior art in commercial production, TRL 9 [56], therefore they are not state of the art.

It is important to differentiate these two terms: prior art is what has become publicly accessible before the filing date of a patent application (publications, previous patents, presentations at conferences or even the sale of a product) [57], it is an absolute and worldwide term. The state of the art, on the other hand, represents the highest level of development of a device, technique or scientific field achieved to date [58]. While the prior art focuses on "what already existed" (to prove novelty), the state of the art focuses on "what is best" (to prove excellence or evolution). In general, the state of the art is not economically viable.

Therefore, the objective of this paper is to present a proposal for the engineering development of the robotic systems necessary for a URIEL System (helicopter + URIEL Pod assembly): feller, delimbing, topping, lift and stem handling, as well as the integrated robotic systems for post-harvest silvicultural treatments (HST). The URIEL System is implemented through two distinct modules housed in a Pod airborne by a large helicopter integrating: the harvesting module (MH) and the HST module (MHST), both are carried in a pod attached to a helicopter. The hypothesis of this work is that the concept of a URIEL System is economically viable after an economic analysis of the logging operation based on this possible System.

As a final word in this introduction, it's worth noting a fortuitous coincidence in the acronym that gives this new method its name. Uriel is one of the archangels in Hebrew and Christian tradition [59,60]. Uriel is the guardian of the gates of paradise; those of good character see him as a beautiful bird, the wicked as a terrible warrior with a flaming sword [61]. According to esotericism, Uriel is the angel of Earth's vegetation [62]. The authors of this paper consider the name of this new logging method to be very appropriate. The winged guardian of the tropical paradise of terrestrial vegetation.

# Methods

### Study Area

For the purposes of developing the design premises and preparing the calculations for sizing the systems and preparing the technical feasibility study, the Tapajós-Arapiuns Extractive Reserve (RESEX-TA) was chosen. RESEX-TA is located in the municipalities of Santarém and Aveiro, in the western region of the State of Pará-Brazil, Figure 13, with an area, according to the creation decree, of 647,610 ha, between the geographical coordinates 02º 20’ to 03º 40’ South, and 55º 00’ to 56º 00’ West, altitude varying from 2 to 216 m above sea level [117]. It was created by Presidential Decree on November 6, 1998, consisting of 74 communities with an estimated population of 13 thousand people [118].

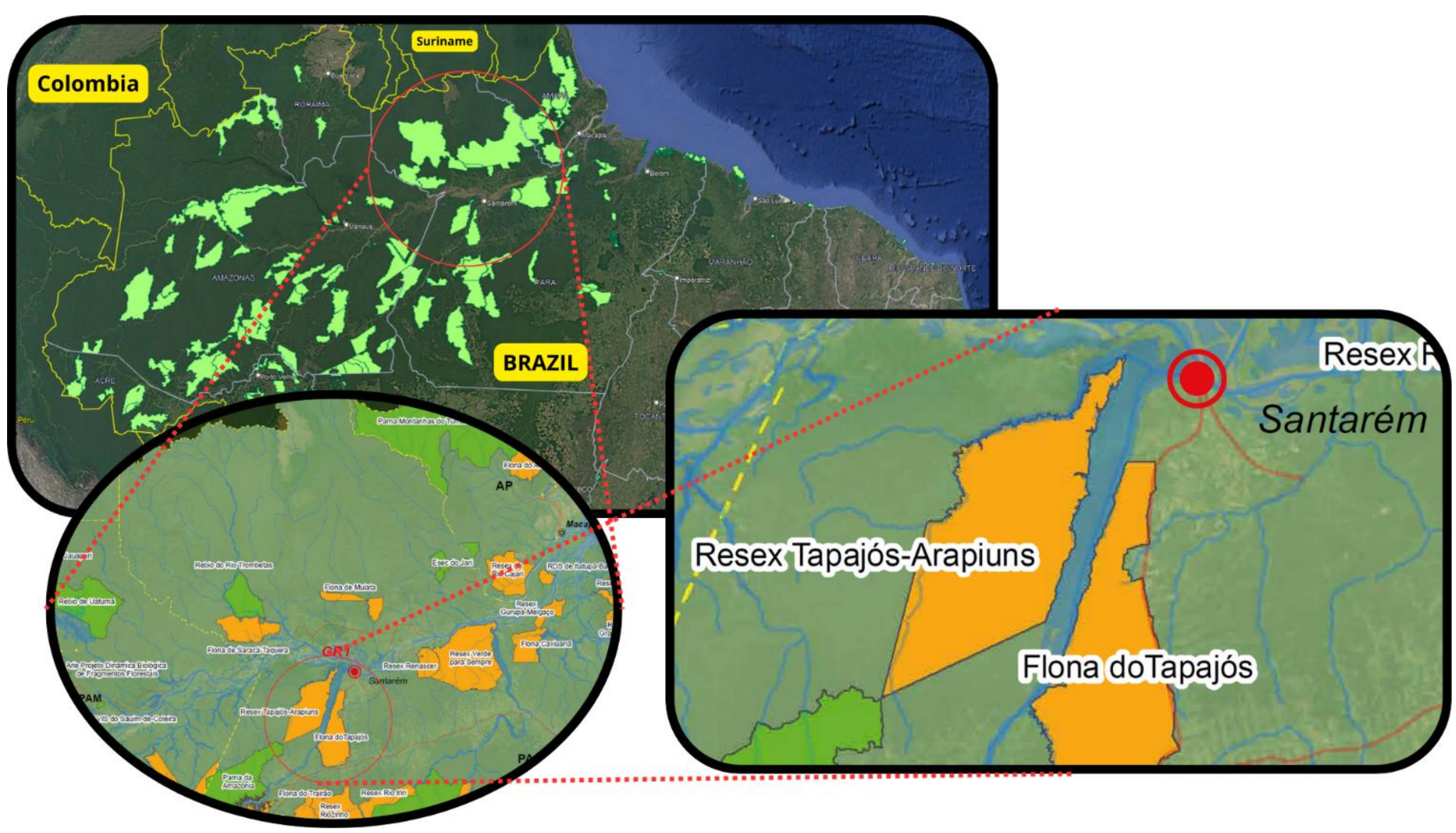

Figure 13. Location of the study area: Tapajós-Arapiuns Extractive Reserve. Sources: Adapted [117,118].

Approximately 34% of the Tapajós-Arapiuns RESEX area is located in the Municipality of Aveiro/PA, which corresponds to approximately 194,283 ha, and the remaining 66% is located in the Municipality of Santarém/PA, equivalent to 453,327 ha. The Tapajós-Arapiuns RESEX is bordered by the Arapiuns, Maró, and Mentae rivers. To the west, the RESEX-TA is bordered by the Mamuru and Nova Olinda land parcels. To the east, it is delimited by the Tapajós River. The fact that it is located in the interfluve and confluence of the Tapajós and Arapiuns rivers gave rise to the name of the RESEX-TA [118,119].

The climate of the Tapajós-Arapiuns RESEX is the equatorial continental megathermal humid climate of Central Amazonia, which is characterized by the combination of high heat (average annual temperatures between 26 and 28ºC) with high humidity (average annual values between 80 and 85%), with average annual rainfall totals varying between 2,000 and 2,800 mm. The rainy season lasts on average around 7 to 9 months, extending from November to July [119].

Geologically, the RESEX is located within the large geotectonic unit constituted by the Amazon Sedimentary Basin, within the Middle Amazon sub-basin, which is limited by structural highs of the basement with the adjacent sub-basins. Predominantly, the Tapajós-Arapiuns RESEX is composed of Holocene alluvium and detrital-lateritic cover. The basins and sedimentary covers of the RESEX TA are predominantly Phanerozoic. In some regions of the RESEX Tapajós-Arapiuns, mainly in the region where the Tapajós and Arapiuns rivers meet, we find Quaternary sedimentary deposits [120].

The main soils of the RESEX-TA come from the Barreiras Formation: Yellow Latosols, which cover most of the RESEX, are developed from clayey or sandy-clayey sedimentary materials. There is also a large area of Yellow Argisols, which are located in the eastern portion of the RESEX. Red-Yellow Argisols are located in the southernmost part of the RESEX, occupying a small portion, and Dystrophic Fluvisols occupy the entire Inhambú River drainage [120].

The predominant vegetation type in RESEX-TA is Dense Ombrophilous Forest, which occurs in 88% of the total area of the Unit, covering approximately 591,420.00 hectares and is characterized by large trees, the presence of woody lianas and epiphytes in abundance, its main ecological characteristics are high temperatures (averages of 25° C) and high precipitation, well distributed throughout the year (from 0 to 60 dry days), which determines a bioecological situation with practically no dry period [119–121].

According to [120] in RESEX-TA in the sampled forest areas, timber species considered to have high commercial value had low abundance. However, the authors of this paper attribute this fact to the sampled area being extremely small and statistically insufficient for abundance estimates. According to [119–121]residents pointed out the need to develop community timber management plans and cited the following species as abundant and with potential for commercial exploitation: angelim, araraúba, cedro, cedrorana, cumaru cupiuba, guariuba, ipê, itaúba, jacarandá. Jarana, jatobá, louro, maçaranduba, maracatiara, piquiá, tatajuba and ucuuba.

In this paper, due to the commercial value linked to Brazilian and international demand, the following woods were chosen for the sizing of the systems and evaluation of economic viability: Cedro (Cedrela fissilis), Ipê (Handroanthus impetiginosus), Jatobá (Hymenaea courbaril).

## Species Considered

The sizing of robotic airborne harvesting systems is directly influenced by the characteristics of the trees considered. In consideration by the authors of this proposal, commercial trees that are in demand and at the same time possess a plant architecture that facilitates robotic harvesting by air were chosen. In this context, Cedar (Cedrela fissilis), Pink Ipê (Handroanthus impetiginosus), and Jatobá (Hymenaea courbaril) were selected, as in a global analysis considering the commercial woods available in RESEX-TA, these choices offer the best economic viability.

***Cedar (Cedrela fissilis), Figure 14.***

Deciduous tree, 10 to 25 m tall and 40 to 80 cm DBH [122] in adulthood around 50 years [123,124]. Cylindrical habit, straight or slightly tortuous, with an absence of buttresses or, when present, poorly developed, dichotomous branching, tall crown, densely foliated, multiple, corymbiform, typical. The natural occurrence is delimited between latitudes 12º N in Costa Rica to 32º S in Brazil, in Rio Grande do Sul. The northern limit of the species in Brazil occurs at approximately 1º S in Pará, Figure 15, the altitudinal variation is 5 m, on the coast of the South and Southeast Regions to 1,800 m altitude, in Campos do Jordão-SP-Brazil [122].

Ecologically, it is an early secondary, late secondary species, reaching climax, commonly found in the Submontane Dense Ombrophilous Forest (Atlantic Forest), in the Montane and Submontane formations, and Dense Ombrophilous Forest (Amazon Forest), its occurrence being restricted in Pará, in the terra firme forests; in the Mixed Ombrophilous Forest (Araucaria Forest), where it is common, in the Alluvial, Submontane, Montane and High-Montane formations; in the Semi-deciduous Seasonal Forest, in the Riparian and Submontane formations, where it is also common; and in the Deciduous Seasonal Forest, in the Montane and Lower-Montane formations [122–125].

Its frequency in the forests of Southern Brazil varies from 1 to 7 trees per hectare and its regeneration method, due to its ecophysiological characteristics, as it presents greater productivity under less intense light conditions, cedar is suitable for mixed plantings in association with Syzygium cumini (synonym: Syzygium jambolanum) in the Northeast Region and with chinaberry (Melia azedarach), as a way to reduce the incidence of cedar borer. In arboreal matrix vegetation, planting should be done in open strips in secondary forests and in exploited forests it should be planted in rows, at a density never exceeding one hundred trees per hectare, cedar sprouts after cutting, especially when young [122].

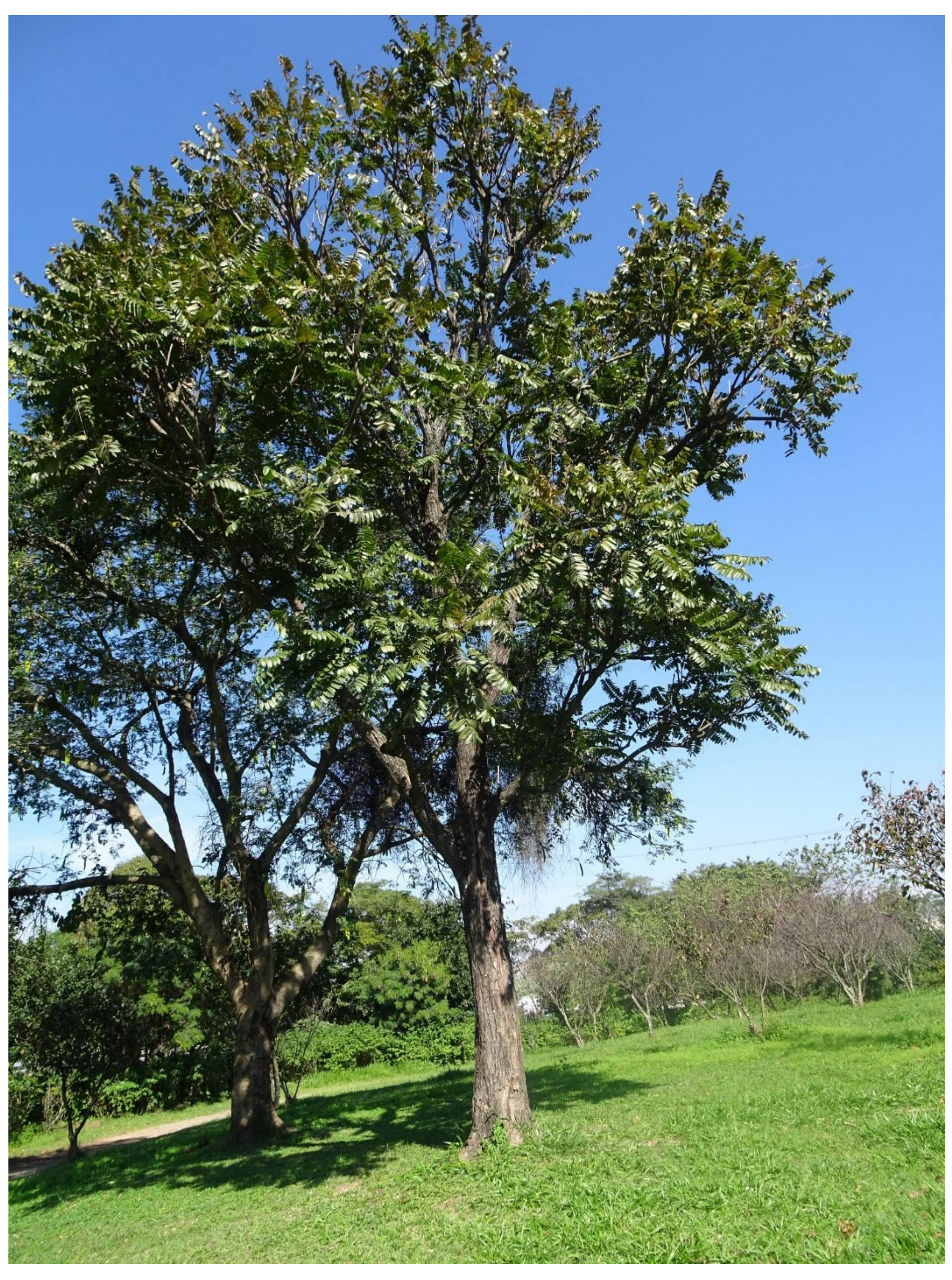

Figura 14. Cedrela fissilis [126].

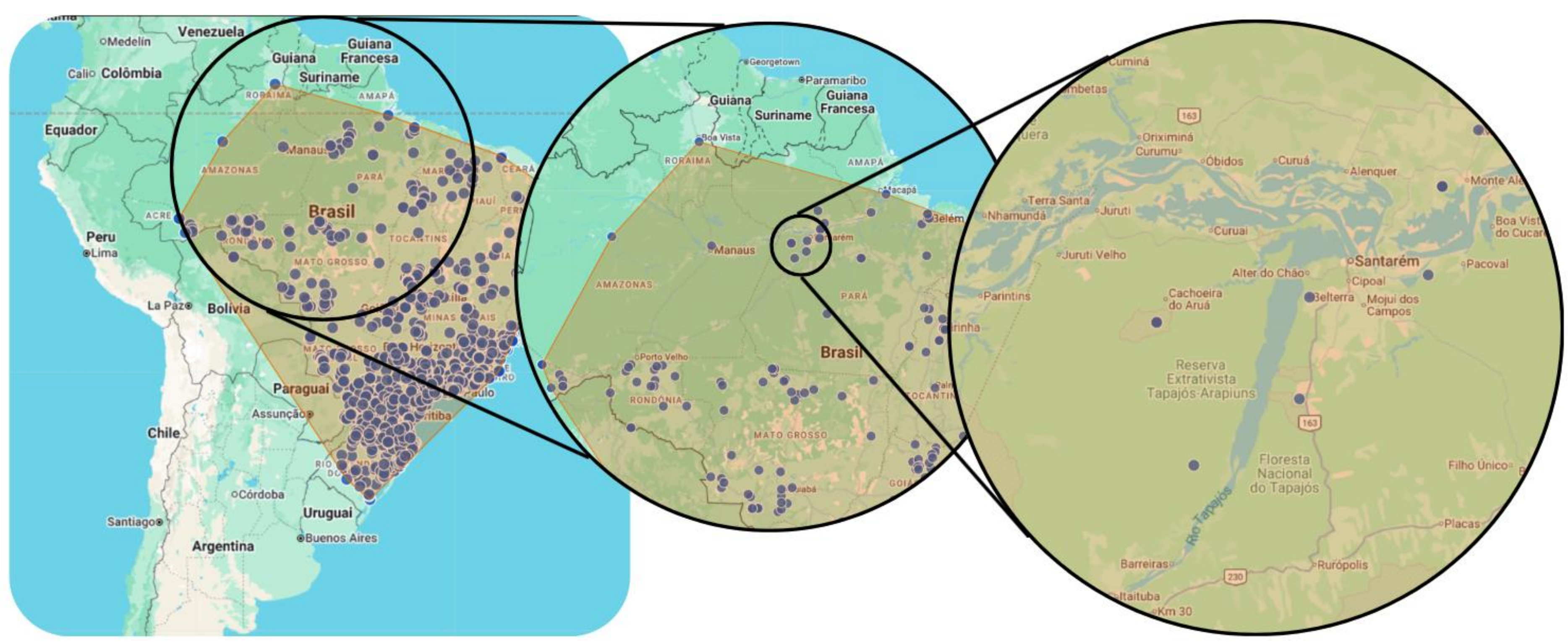

Figure 15. Geographic area of occurrence of Cedrela fissilis in Brazil. Source: Adapted [125].

Apparent specific gravity: Cedar wood is dense (0.53 g.cm-3) at 15% moisture content [127]. Cedar wood is particularly suitable for civil construction, in the making of shutters, skirting boards, trims, linings, frames, windows and paneling; in shipbuilding, it is used in decorative interior finishes and in the hull of light vessels; interior parts of fine furniture, decorative veneer sheets, plywood, decorative packaging, picture frames, foundry models, carving work, office supplies, musical instruments and broom handles [122].

***Pink Ipê (Handroanthus impetiginosus), Figure 16.***

There was a confusion of nomenclature regarding the Pink Ipê, initially it was scientifically named Tabebuia, but based on the latest taxonomic revisions, the Tabebuias temporarily includes 99 species and three genera, including 30 species of Handroanthus [128].

Deciduous tree, reaching up to 50 m in height and 100 cm in DBH in the Amazon [122], at adult age 80 years [107], frequently tortuous, with individuals of straight and cylindrical habit being found.

The natural occurrence is delimited by latitude 20º N in northeastern Mexico to 28ºS in northeastern Argentina. In Brazil, the southern limit of this species is 22º45' S, in Paraná. Altitudinal variation ranges from 10 m on the coast of the Brazilian Northeast Region to 1,400 m altitude in the State of São Paulo [122,129]. Ecologically, it is found naturally in various vegetation formations, mainly in the Submontane Semideciduous Seasonal Forest; in the Dense Ombrophilous Forest (Amazon Forest and Atlantic Forest), Figure 17; in the Deciduous Seasonal Forest, in Mato Grosso do Sul; in the Cerradão, in the Open Forest without babassu [122].

Its frequency in southeastern Brazil in Perdizes-MG is three individuals per hectare, the pink trumpet tree is a heliophilous species, which tolerates medium intensity shading in the young phase, the sympodial growth habit, the regeneration methods can be planted in full sun, in mixed planting, associated with pioneer species, and in matrix vegetation, in open strips in secondary forests or thickets and planted in rows [122].

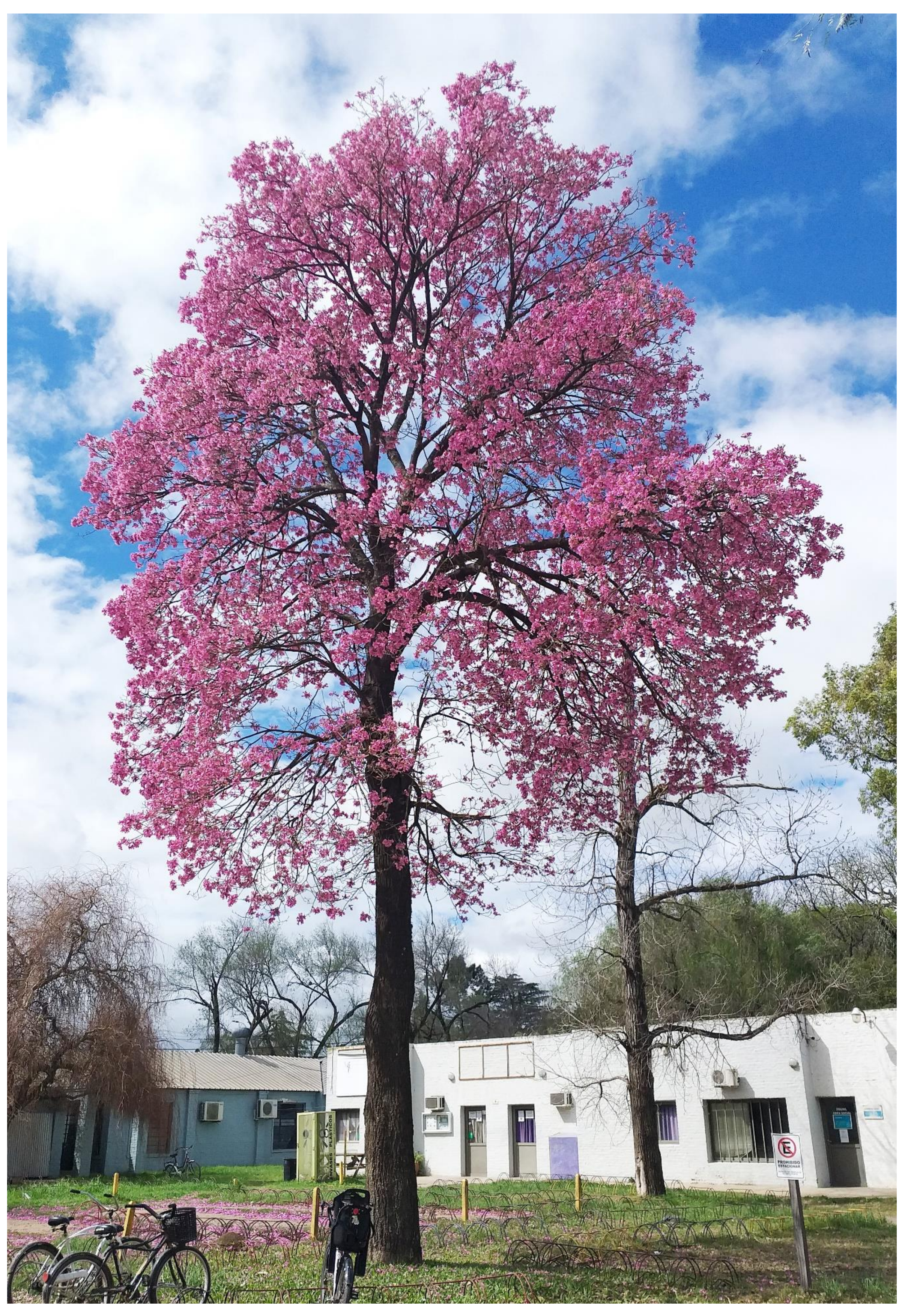

Figure 16. Handroanthus impetiginosus [130].

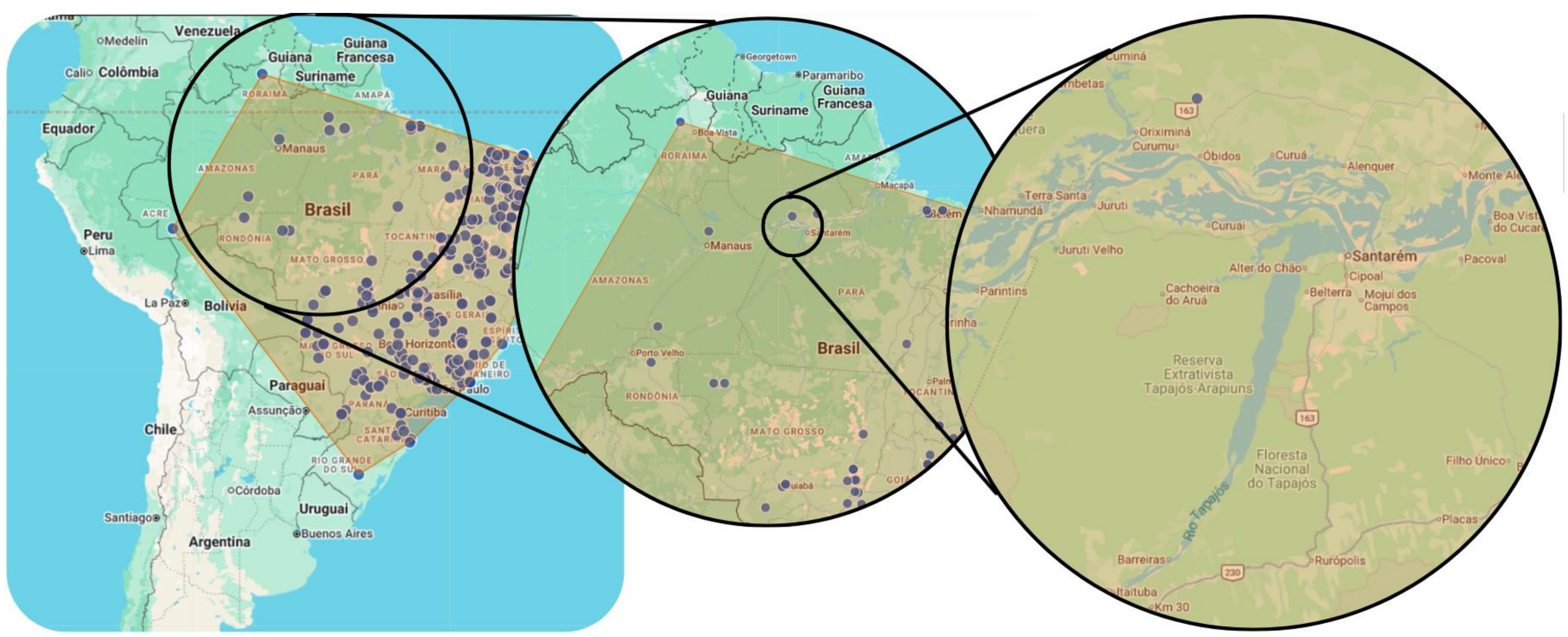

Figure 17. Geographic area of occurrence of Handroanthus impetiginosus. Source: Adapted [129].

Apparent specific gravity: the wood of the Ipê-Rosa is dense (0.96 g.cm-3), at 15% moisture content [131]. Ipê-Rosa wood can be used in interior finishing, making sporting goods, tool handles and agricultural implements; in exterior construction it is used as structures, sleepers and crossbeams; joinery and paneling; turned pieces, blocks and boards for flooring, wagons, carriages and musical instruments; in civil construction it is used as rafters, lining, laths, beams, and stair treads and posts [122].

***Jatobá (Hymenaea courbaril), Figure 18.***

Semi-deciduous tree, 8 to 15 m tall and 40 to 80 cm in DBH, reaching up to 25 m in height, in forests of Central Brazil [122], with ages around 80 years [108]. The natural occurrence is delimited at latitude 2º 30’ S in Maranhão to 25º 19’ S in Paraná-Brazil, Figure 19, the altitudinal variation is from 30 m, in Espírito Santo-Brazil to 1,300 m altitude, in Minas Gerais-Brazil [122].

Ecologically, it is a late secondary or light-demanding climax species, the jatobá is characteristic of the interior of the primary forest, in the forest, the individuals are found distanced from each other. It is a long-lived tree, it is a characteristic species of the Semideciduous Seasonal Forest, in the Submontane formation, where it occupies the dominant stratum, also found in the Dense Ombrophilous Forest (Atlantic Forest), in the Deciduous Seasonal Forest in the Paranã River Valley [122].

The frequency of trees in the Southeast Region is between 1 and 6 trees per hectare. Jatobá is a semi-heliophilous species, suitable for planting from edges and clearings to canopy closure. Its habit is inherently sympodial branching, irregular and variable. Jatobá can regenerate when planted in pure stands, in full sun, under dense spacing; however, the silvicultural behavior of this species is better in mixed stands than in pure stands in full sun, associated with pioneer species. However, in arboreal matrix vegetation, in open strips in secondary forests, and planted in enrichment in rows or groups, jatobá grew less when planted in shade or partial shade than when planted in full sun. This species sprouts from the stump after cutting [122].

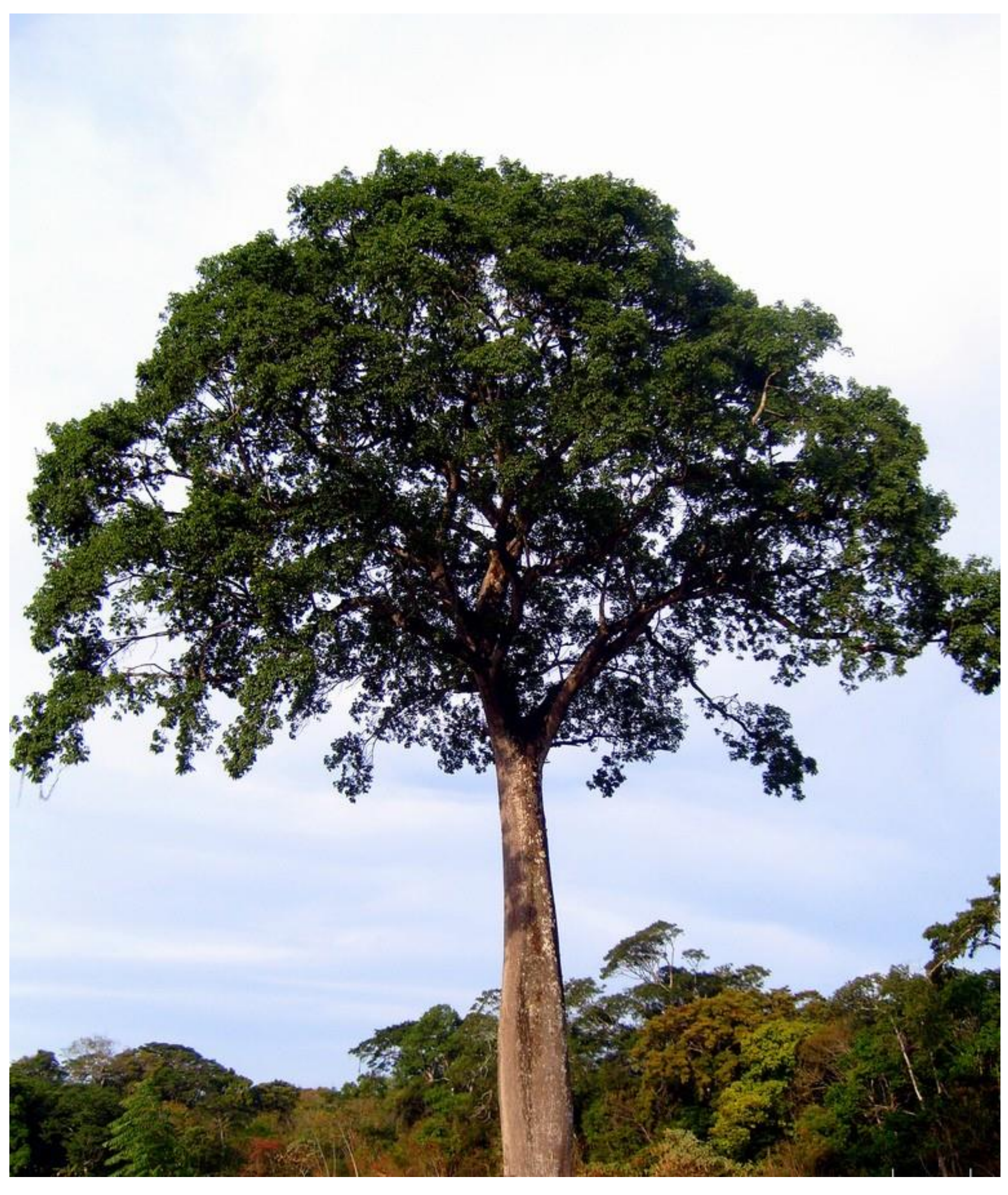

Figure 18. Hymenaea courbaril [132].

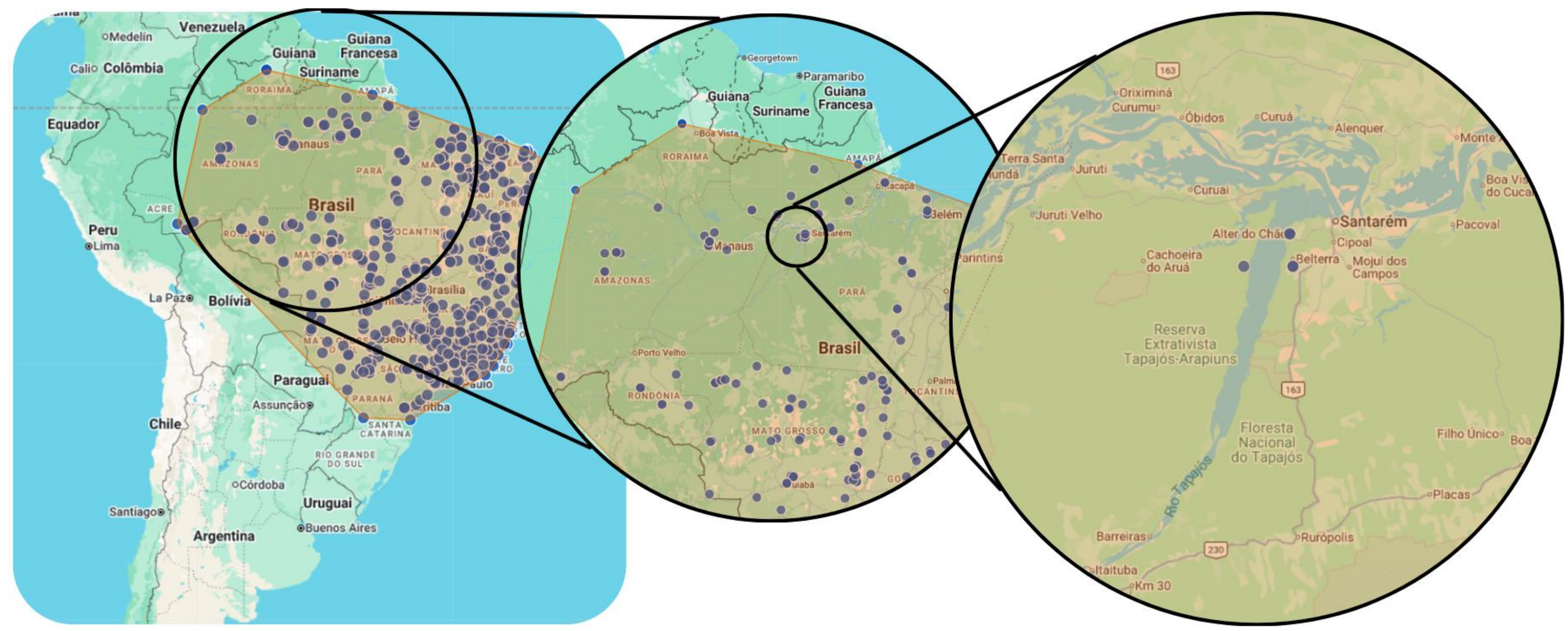

Figure 19. Geographic area of occurrence of Hymenaea courbaril. Source: Adapted [133].

Apparent specific gravity: Jatobá wood is dense (0.96 g.cm-3) at 15% moisture content [127]. Jatobá wood can be used in civil construction and carpentry in general; in interior finishes such as beams, rafters, battens, door jambs, flooring blocks, sporting goods, tool handles and agricultural implements; exterior constructions such as hydraulic works, posts, sleepers, crossbeams and joinery; decorative veneers, furniture, turned pieces, truck bodies, wagons, mills and barrels [122].

*Specific stem volume*

The specific volume of wood will be calculated based on the Schumacher-Hall mathematical model defined as follows:

$$\ln(V_s) = \beta_0 + \beta_1 . \ln(d) + \beta_2 . \ln(h) + \varepsilon \quad (1)$$

Where:

Vs is the predicted volume of wood (m3);
$\beta_{0, 1, 2}$ are specific constants for each type of wood (dimensionless);
d is the diameter at breast height of the tree (cm);
h is the height of the trunk (m);
ε is an estimation error.

According to [134–137], the dimensionless coefficients for the woods considered are:

Table 7. Dimensionless coefficients for the Schumacher-Hall model.

| Wood | $\beta_0$ | $\beta_1$ | $\beta_2$ | Source |
|---|---|---|---|---|
| **Cedar** | -9.99 | 2.19 | 0.57 | [134] |
| **Ipê** | -9.40 | 1.94 | 1.01 | [138] |
| **Jatobá** | -7.21 | 1.77 | 0.44 | [136] |

Based on the definition of mathematical models for cubing the wood of the chosen species, taking as a basis hypothetical trunks of 80 cm DBH and 25 m height, the volume of wood of a trunk was determined as follows: Cedar = 4.22 m3; Ipê = 10.50 m3; Jatobá = 7.11 m3. If we consider trunks of 100 cm DBH and 30 m height: Cedar = 7.64 m3; Ipê = 19.47 m3; Jatobá = 11.44 m3.

**Green wood**

Considering that the URIEL System harvests wood extracted directly from the standing forest, the load required for lifting by helicopter must be dimensioned considering green wood. Thus, a method to estimate the weight of green wood is to find the density of green wood from the specific gravity of the wood through the equation of relationship between densities defined by [139–141]:

$$\rho = 1000. G_m. (1 + \frac{M}{100}) \quad (2)$$

Where:

ρ is the desired moisture density (kg m-3);

Gm is the specific gravity of the wood at 15% moisture when the desired moisture content is greater than 30% (g cm-3);

M is the desired moisture content (%).

Direct measurements of green wood moisture content for Hymenaea courbaril, Cedrela fissilis, and Handroanthus impetiginosus are limited, but available studies and reviews on tropical hardwoods provide relevant data. For Hymenaea courbaril (jatobá), green wood moisture content often exceeds 60–100% (based on dry mass) [140,141]. Cedrela fissilis and Handroanthus impetiginosus are less frequently studied, but similar tropical hardwoods typically show green wood moisture contents in the range of 60–120%, depending on species and site conditions [139–141]. In this paper, for safety reasons, a moisture content of 100% will be considered for the green wood of all selected species. The density of the green wood was found using equation 2.

Table 8. Densities considered for the selected woods.

| Wood | Specific gravity at 15% moisture content | Green density at 100% moisture content (kg $m^{-3}$) |
|---|---|---|
| **Cedro** | 0.53 | 1060 |
| **Ipê-Rosa** | 0.96 | 1920 |
| **Jatobá** | 0.96 | 1920 |

Thus, considering the volume and density of green wood for each species in a configuration of 80 cm DBH and 25 m height, the required lifting load for the helicopter is: Cedar = 4,473 kg; Ipê = 20,160 kg; Jatobá = 13,651 kg.

If a configuration of 100 cm DBH and 30 m height is considered (normal configuration for Amazonian trees), the load for the helicopter to carry is: Cedar = 8,098 kg; Ipê = 37,382 kg; Jatobá = 21,964 kg.

**Helicopters**

The main focus of the URIEL methodology is the use of helicopters, which enable the safe transport of the harvesting module and the HST module through the forest. Helicopters also vertically transport large logs, preventing falls that would severely damage the surrounding vegetation.

However, given the magnitude of the loads in terms of the weight of the chosen timber logs, it is clear that only large helicopters have the capacity to transport the timber to locations outside the forest. In this context, three commercially available helicopter models currently in production were selected, possessing operational capabilities consistent with the challenge of transporting the modules along with the timber.

***Boeing CH-47 Chinook, Figure 20.***

The Boeing CH-47 Chinook has been produced since 1961, its origins dating back to the CH-46 Sea King, which in 1956, at the request of the US Army, was asked for a larger helicopter with the same configuration of tandem twin rotors. The rotors rotate in opposite directions to cancel torque, and power is provided by two externally mounted turboshafts meshed in such a way that only one of the engines can drive both rotors [142]. The Chinook is the standard medium helicopter of the US Army and has been deployed worldwide and is capable of operating in any environment, such as Antarctica, being the logistics and rescue operator of the Argentine Air Force on the icy continent [142].

The twin rotor design provides greater stability and control, maximum agility, ease of loading and unloading, and superior performance in windy conditions. This design also allows the Chinook to fly up to 6,000 meters altitude. The current production version (2025) is the H-47 Chinook Block II, which is equipped with state-of-the-art technologies, redesigned fuel tanks, a reinforced fuselage, and an improved transmission system [143]. The H-47 Chinook's engine consists of two Honeywell T55-GA-714A turboshaft engines with 4,777 shp each. The Chinook's technical characteristics are [144]:

Table 9. Technical data of the H-47 Chinook [144].

| Technical Parameter | Value |
|---|---|
| Rotor Diameter | 18.29 m |
| Length (Rotors Turning) | 30.14 m |
| Height | 5.68 m |
| Fuselage Width | 3.78 m |
| Fuel Capacity | 4,088 L |
| **Performance** | **(Standard Day, Sea Level)** |
| Maximum Speed | 302 km/h |
| Cruise Speed | 291 km/h |
| Mission Radius | 306 km |
| Service Ceiling | 6,096 m |
| Max Gross Weight | 24,494 kg |
| Useful Load / Payload | 12,565 kg |

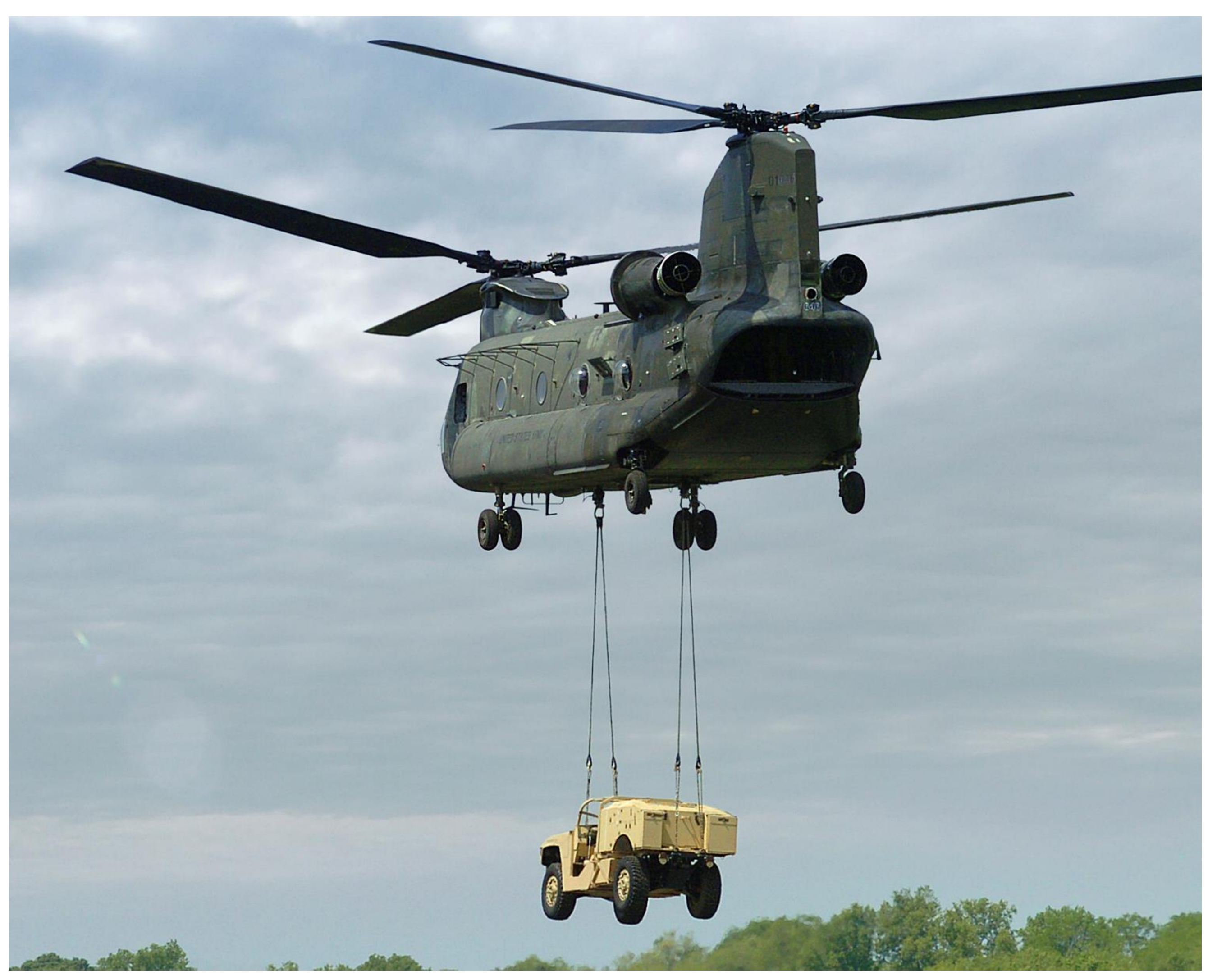

Figure 20. Boeing CH-47 Chinook on-the-go [145].

***Sikorsky CH-53 Stallion, Figure 21.***

The Sikorsky CH-53 Stallion was designed to meet a need of the United States Marine Corps for a helicopter that could carry very heavy loads, the first CH-53 version A had two engines and were called Sea Stallion and began operations in 1967 [146]. The dynamic parts: rotors, gearboxes and control systems were inherited from the S-64 Skycrane, but the main rotor had folding titanium blades [146]. In 1974 the Stallion became the Super Stallion with three engines and was designated CH-53E [146].

In 2015, Lockheed Martin acquired Sikorsky and in 2015, starting from the CH-53E, developed the current production version, the Sikorsky CH-53K King Stallion, which has a new digital system with a node in distributed sensor networks, in addition to being able to work with unmanned systems. The CH-53K helicopters have high-authority fly-by-wire flight controls, which allow them to maintain precise hovering flight in wind gusts of 74 km/h [147].

The CH-53K's engine consists of three GE T408 GE38-1B turboshaft engines with 7,500 shp each. The technical characteristics of the King Stallion are:

Table 10. Technical data of the CH-53K King Stallion [147].

| Technical Parameter | Value |
|---|---|
| Rotor Diameter | 24.1 meters |
| Length (Rotors Turning) | 30.2 meters |
| Height | 8.6 meters |
| Fuselage Width | 9.1 meters |
| Fuel Capacity | 8,423 liters |
| **Performance** | **(Standard Day, Sea Level)** |
| Maximum Speed | 315 km/h |
| Cruise Speed | 270 km/h |
| Mission Radius | 200 km |
| Service Ceiling | 4,900 meters |
| Max Gross Weight | 39,916 kg |
| Useful Load / Payload | 16,329 kg |

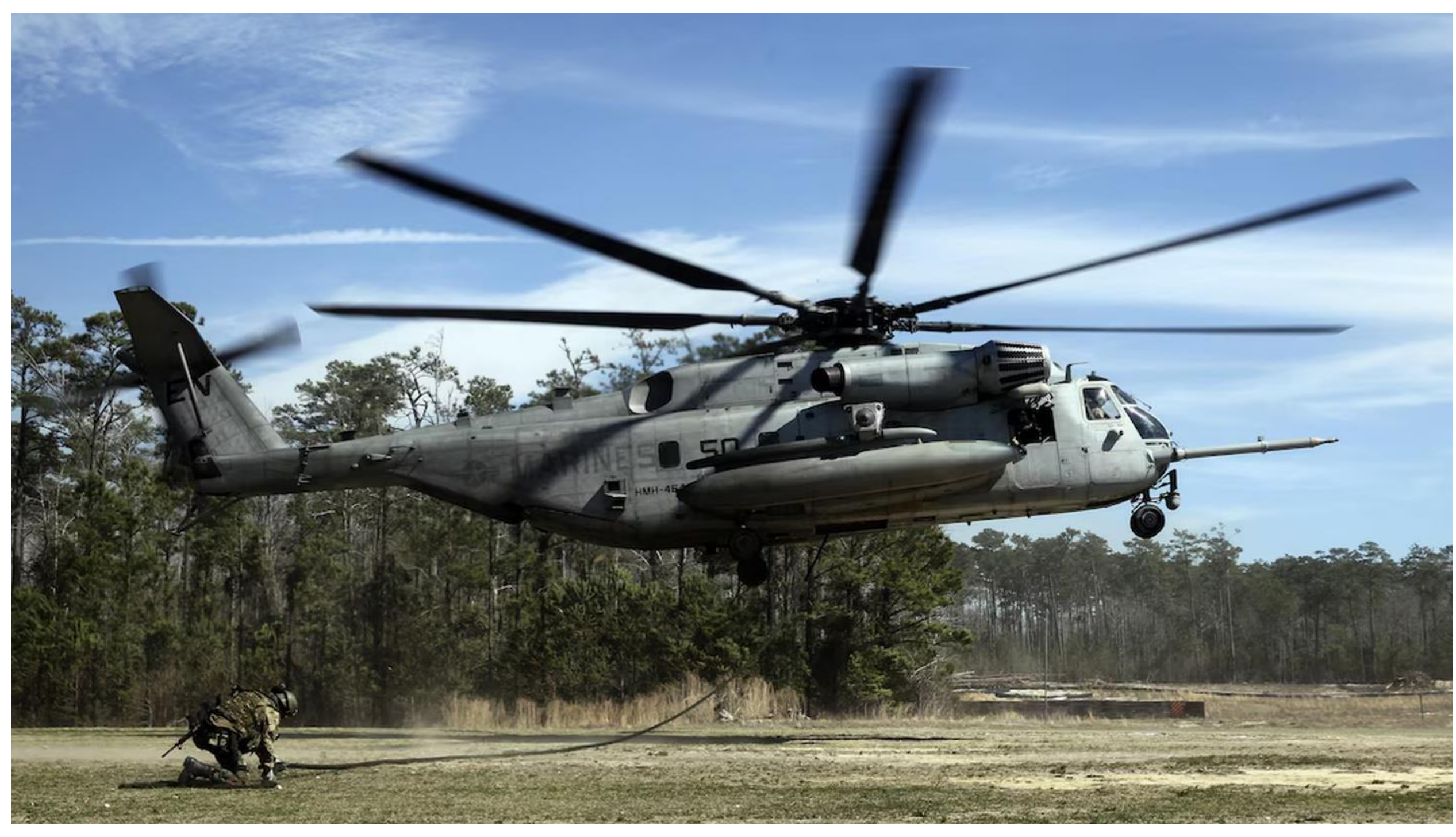

Figure 21. Sikorsky CH-53 on-the-go [148].

***Mil Mi-26 Letayushchaya Korova, Figure 22.***

The Mil Mi-26 Letayushchaya Korova is the world's heaviest production helicopter and the first to operate with an eight-bladed main rotor, designed for heavy transport by the Soviet Air Force and Aeroflot. The Mi-26 Letayushchaya Korova has a conventional configuration with a 5-bladed tail rotor mounted on the right side of the fin. The fuselage can accommodate up to 90 people. The Letayushchaya Korova has an automatic hovering flight system. The first Mi-26s entered service in 1982 [149].

The Mi-26T2 is the current production version and is a modernized version of the Mi-26T. It features a new BREO-26 "glass cockpit" onboard avionics complex with five multi-function LCD indicators, as well as a new digital autopilot and navigation system with NAVSTAR/GLONASS support [150]. The MI-26T2's engine consists of two Lotarev D-136-2 turboshaft engines with 11,400 shp each. The technical characteristics of the Letayushchaya Korova are:

Table 11. Technical data of the Mi-26T2 Letayushchaya Korova [150].

| Technical Parameter | Value |
|---|---|

| Rotor Diameter | 32 meters |
|---|---|
| Length (Rotors Turning) | 40 meters |
| Height | 8.15 |
| Fuselage Width | 6.15 meters |
| Fuel Capacity | 12,000 liters |
| **Performance** | **(Standard Day, Sea Level)** |
| Maximum Speed | 295 km/h |
| Cruise Speed | 255 km/h |
| Mission Radius | 400 km |
| Service Ceiling | 4,600 meters |
| Max Gross Weight | 56,000 kg |
| Useful Load / Payload | 20,000 kg |

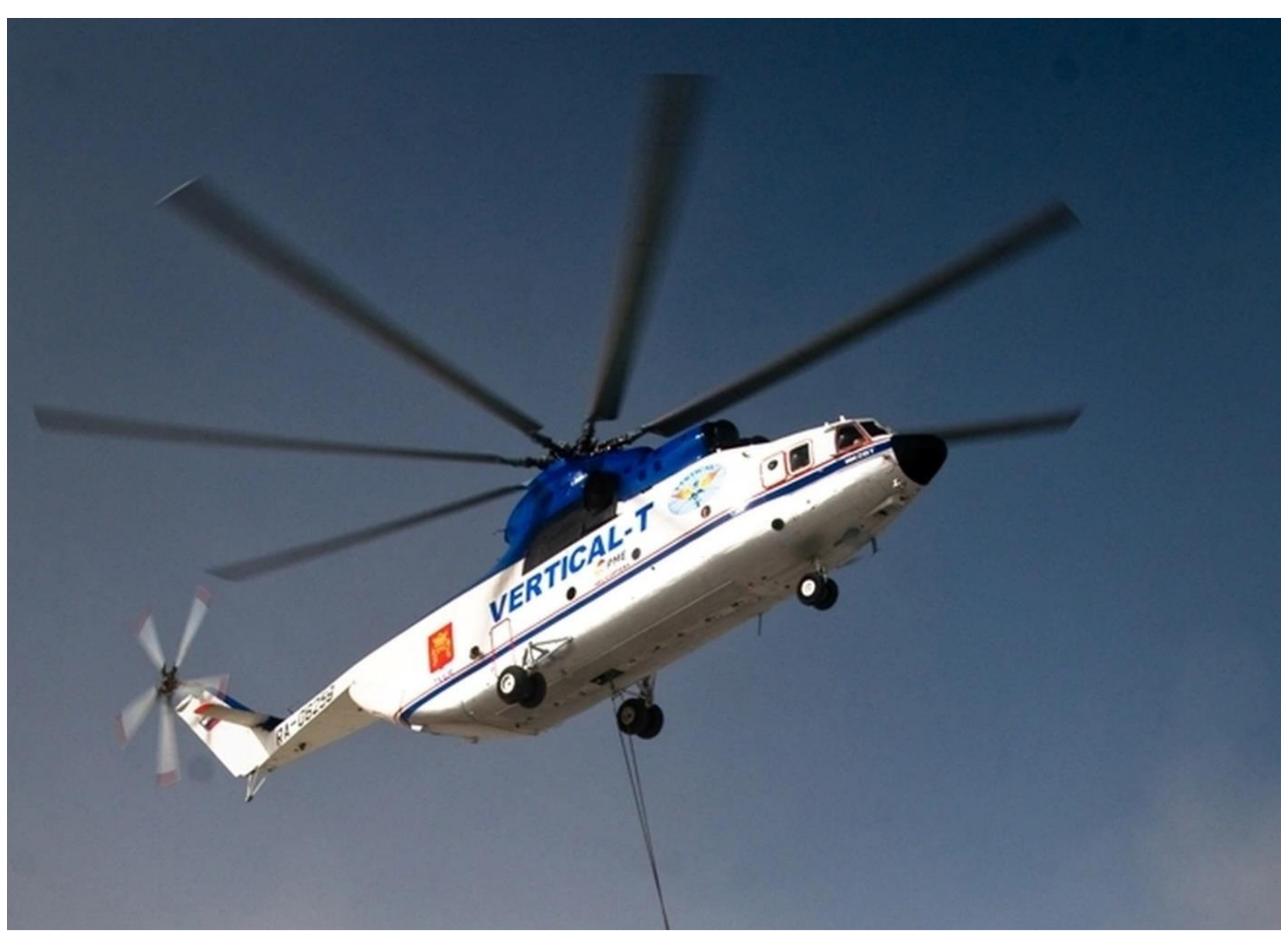

Figure 22. Mil Mi-26 on-the-go [151].

**Engineering project**

The study for development begins with an exploratory analysis of the requirements and constraints for reduced-impact logging, considering the edaphoclimatic and biome specificities of the study area. This analysis will be based on the dictates of [152]. Based on the needs analysis of these systems, a morphological matrix of solutions was developed [153], and from this, a synthesis of exploratory propositions will be carried out, indicating possible directions that lead to solutions for the problems [64].

In terms of project engineering and agricultural/forest machinery designs, the recommendations of [64,152,154–158] will be followed. To operationalize the robotic system of a URIEL, recommendations from [159] will be followed in the areas of:

a) Computer vision: in the present development, it can be subdivided into acquisition, preprocessing, feature extraction, segmentation, pattern recognition, and image processing [160];

b) Plant identification intelligence: using computer vision technology, artificial intelligence algorithms will be used to analyze the plant morphology of trees, aiming at identification and characterization by leaves and stems [161,162];

c) Robot programming: use of robot programming language for the movement and interaction of the robot with trees and forests, using terminal tools [163,164];

d) Plant manipulation tool: customized manipulator to hold the stem in the most suitable location, with electric or pneumatic actuation [159,165];

e) Instrumentation, automation and control: technology needed for interfacing with the image capture device, robot movement control, interface with safety systems, actuation in the positioning and movement system of the devices [159,165].

For the design of robotic systems, a very challenging aspect of this project's robotics is the artificial intelligence required by the mechatronic systems involved in every function needed for harvester, logging, and post-harvest silvicultural treatments. This robot is designed to make decisions in non-repetitive or unforeseen situations. The methodology recommended by [166], [161], [167], and [163] was followed, as AI reasoning technology involves inference, planning, and learning, which are based on symbolic reasoning using first-order predicative logic (FOPL) and Bayesian Probability Theory. [166] define systems for this project aims to infer a large number of facts that would be very difficult to achieve otherwise without artificial intelligence. Therefore, a specific logic will be selected to operate in the processor's power supply system.

The main challenge in developing the experimental apparatus is the robotic system for harverster, but second challenge is HST autonomous system, whose core will be the artificial intelligence (AI) logic implemented in the digital controllers. This logic will enable the system to determine the best rates based on variable and sometimes discrepant information from the crushing, sectioning, and control systems. Methodologically, the best KR representation will be defined within some inferential logic based on reasoning, as, according to [168], automatic deduction is a subfield of logic, and AI offers a large number of very powerful deduction systems that can be used.

There are several logics that can be used, including pure FOPL, Prolog, OWL (Descriptive Logic), LTL, RCC, LISP, and Bayesian Networks. The authors prefer Bayesian networks, so we will initially attempt to implement them in the project, but not limit ourselves to them. A reasoning diagnosis [168] will be performed based on the observation of all parameters and hidden logic elements of the feeding system. We will attempt to list all variables in the processes both harverst and HST, including operational variables.

From this reasoning diagnosis, a Bayesian network will be implemented based on the potential causes and possible effects, a probability density grid will be implemented. Based on all the interrelations between the physical and operational variables, the network will be structured [169].

**Economic feasibility analysis**

The study of the technical-economic feasibility of URIEL method follows the methodology defined by [170,171] and will be based on the study of the Net Present Value (NPV) of the economic yield gains and the Payback, equations 3 and 4 respectively.

$$NPV = -I + \sum_{t=1}^{n} \frac{FC_t}{(1+k)^t} \quad (3)$$

Where:

I is the investment value;

FCt is cash flow;

k is the interest rate;

t is the investment period.

$$\sum_{t=1}^{Payback} FC_t \quad (4)$$

In this project, a maximum payback period of 5 years was considered. Then, from these, the Internal Rate of Return (IRR), equation 5, will be calculated and compared with the Minimum Attractive Rate of Return (MARR), considered as the SELIC rate (Special Settlement and Custody System) of the Brazilian Central Bank + 8% [101], since in Brazil in 2025 the SELIC is 10%, so the MARR for this study is 18%.

$$0 = -I + \sum_{t=1}^{n} \frac{FC_t}{(1+IRR)^t} \quad (5)$$

Where:

IRR é a Internal Rate of Return.

For the economic feasibility calculations, the following values per cubic meter of the chosen species will be considered: Cedar = US$1,059.00; Ipê = US$1,446.00 and Jatobá = US$868.00 [172,173].

The capital values of the selected helicopters are: H-47 new = US$39,000,000.00 [174]; CH47 used = US$10,000,000.00 at US$20,000,000.00 [175]; CH-53K = US$87,100,000.00 [176] and Mi-26T2 = US$25,000,000.00 [177]. The cost per flight hour for the selected helicopters is: H-47= US$ 6,705.00 [178]; CH-53K= US$ 34,497.00 [178] and Mi-26T2= US$ 15,000.00 [179].

## Results

Considering the geographical specificities and the biome of the Tapajós-Arapiuns Extractive Reserve (RESEX-TA), the dendrometric and ponderal characteristics of the trees Cedrela fissilis, Handroanthus impetiginosus and Hymenaea courbaril and the performance of the H-47, CH-53K and Mi-26 helicopters, based on an exploratory analysis defined by [60] the design premises of the robotic harvesting system and the robotic HST were defined:

a) Identification and location of trees for harvesting carried out by native residents of RESEX-TA;

b) Transfer of bird nests, animal burrows and orchids that are in the area to be managed by native residents;

c) Absence of machine traffic in the area to be managed;

d) Absence of people on the ground in the area to be managed at the time of the harvesting and HST action;

e) Stable atmospheric conditions and no rain (air temperature < 30°C; Relative air humidity > 60%; Wind speed between 3 km/h < V < 10 km/h; Cloud ceiling of at least 3,000 m and minimum visibility of 10 km);

f) Trees with cylindrical and straight trunk geometry with a maximum value of 100 cm DBH (Diameter at Breast Height)) and 30 meters in length;

g) Maximum distance to a medium-sized airport of 400 km;

h) Maximum distance between the management area and the support base in the RESEX-TA of 200 km;

Based on the premises, flowcharts of the processes were developed as prescribed by [63] for the Harvest Module (MH) and for the post-Harvest Silvicultural Treatment Module (MHST), Figure 1 and Figure 2 respectively.

**Harvest Module (MH)**

Considering the process flowchart, Figure 1, and the morphological matrix and solution synthesis methodologies described by [64,65] used by [65–67], it was possible to find optimal solution paths and synthesize the concept of the Harvesting Module (HM).

Figure 1. Flowchart of the harvesting process with a URIEL.

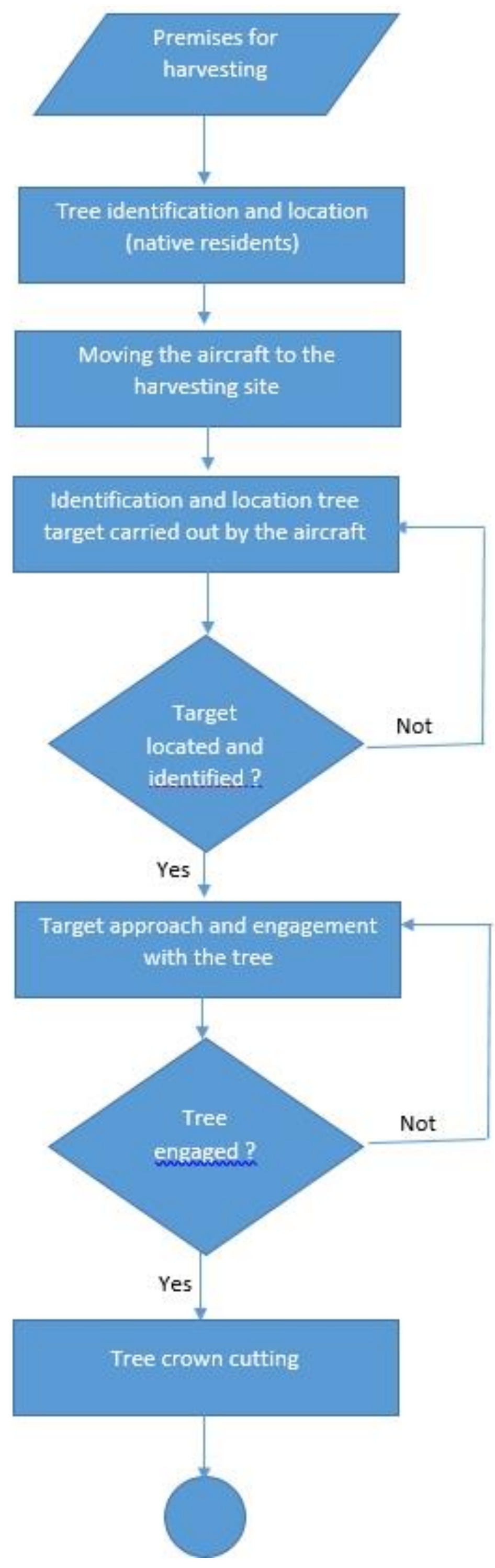

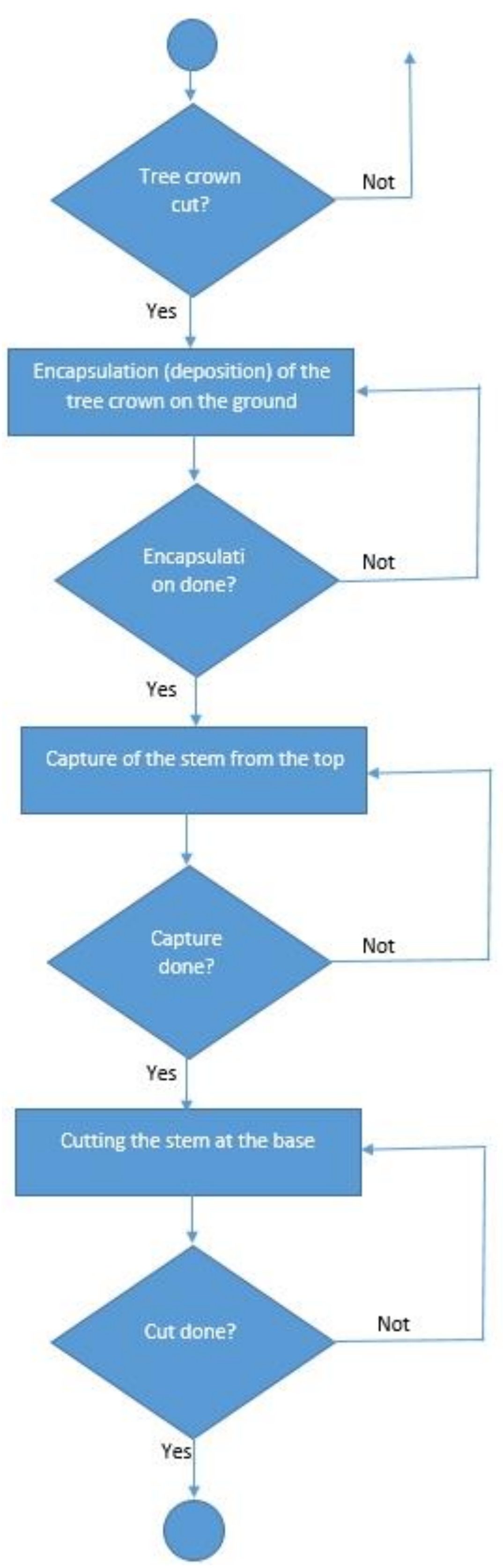
Tree crown
cut?
Not
Yes
Encapsulation (deposition) of the
tree crown on the ground
Encapsulati
on done?
Not
Yes
Capture of the stem from the top
Capture
done?
Not
Yes
Cutting the stem at the base
Cut done?
Not
Yes

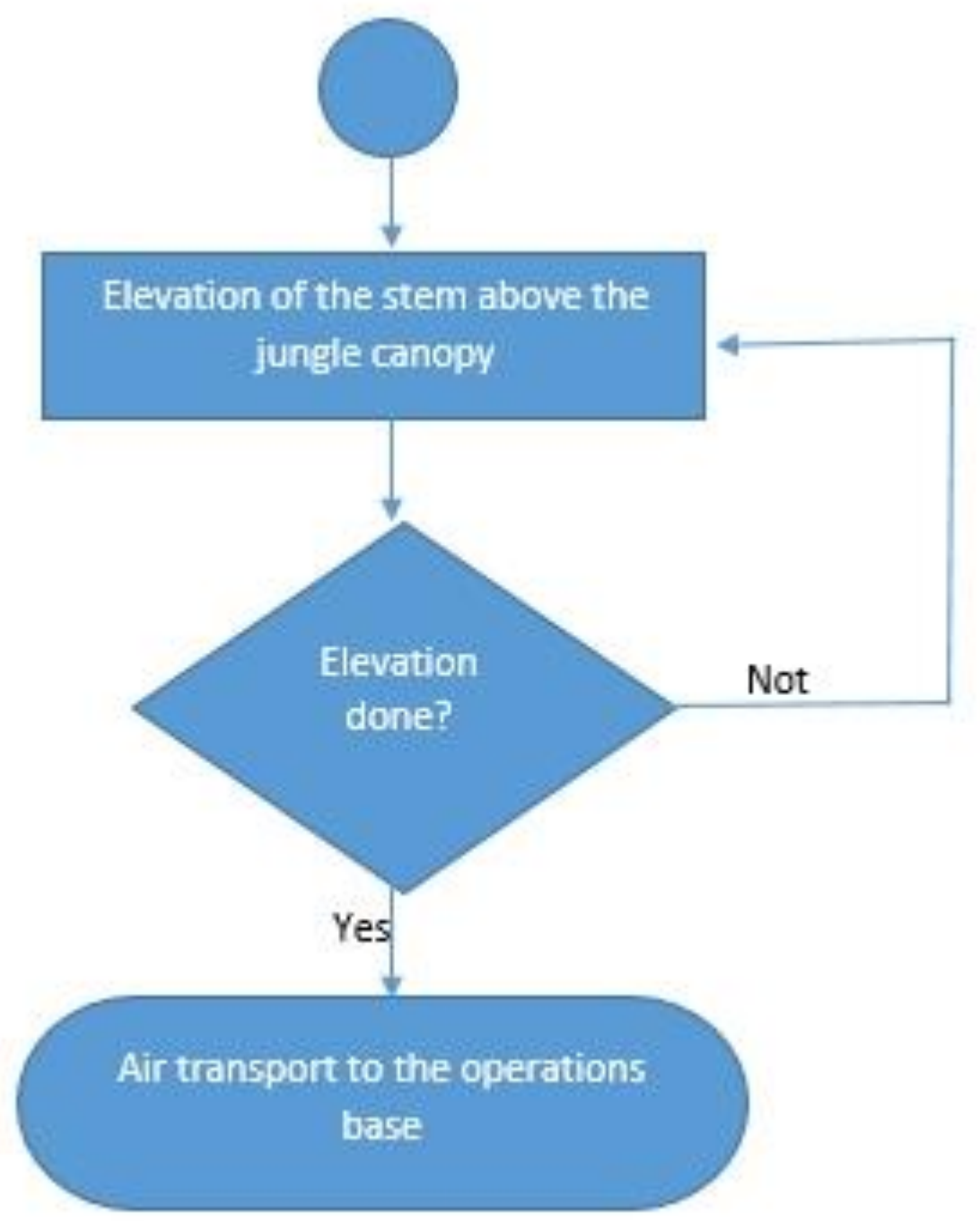

The Harvesting Module (HM) consists of three subsystems, Figure 2: stabilization subsystem (1), decoupling subsystem (2), and stem cutting subsystem (3). All these subsystems are integrated and operated with a lithium battery power source. Operation is autonomously controlled by advanced AI in a 3T architecture divided into three algorithmic cores in the planning layer, one for each specific subsystem, each acting separately but configured in a heterogeneous collaborative system dominated by the stabilization subsystem. These three subsystems follow the prior art in: stabilization subsystem: crane leveling and control systems [68–70]; decoupling subsystem: Feller-Buncher type forest harvesters [71,72]; and stem cutting system: Harvesters type forest harvesters [73,74]. Detailed drawings of each subsystem are presented in the supplementary material Supplementary Figures.

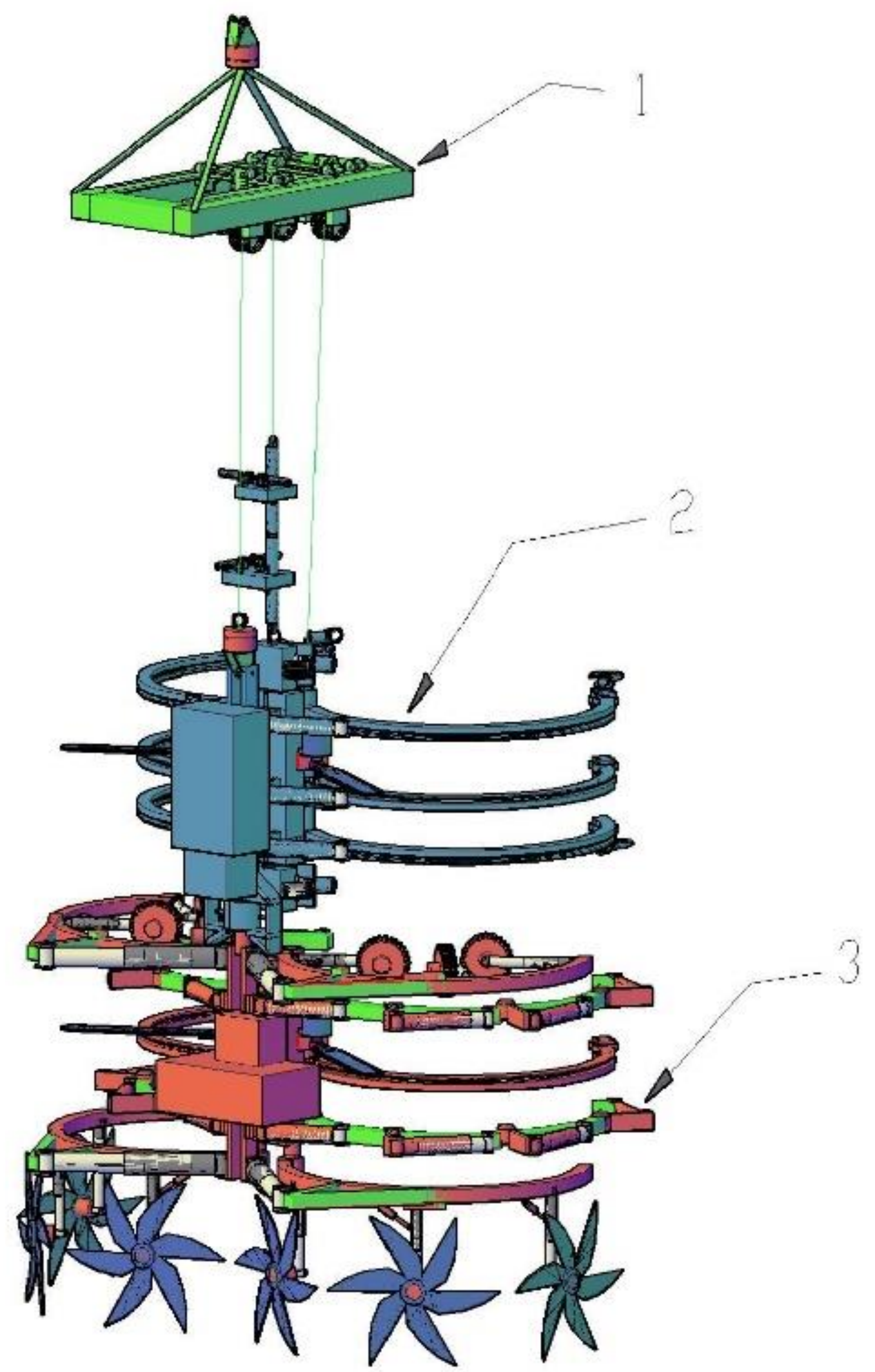

Figure 2. Harvesting Module: 1-stabilization subsystem; 2-decoupling subsystem; 3-stem cutting subsystem.

The stabilization subsystem, Figures 3 a) and b), consists of a rectangular frame with longitudinal and transverse rails supporting three locomotion systems with metal wheels driven by electric motors. This frame is supported by four steel bars that converge at a vertex, the top of which has a rotator with one degree of freedom (Z). Each locomotion system has two degrees of freedom (X, Y) and carries a set of pulleys for steel cables. These pulleys are driven by high-power electric motors. The steel cables are coupled to the decoupling system in such a way that the vector movement of these cables allows control of the orientation of the MH (horse-mounted vehicle), control of the crown's deposition on the ground, and also control of the positioning and stabilization of the timber load. An AI-controlled stabilization system is responsible for activating the locomotion systems and pulleys in such a way that the steel cables are maneuvered to operate the MH and stabilize the load.

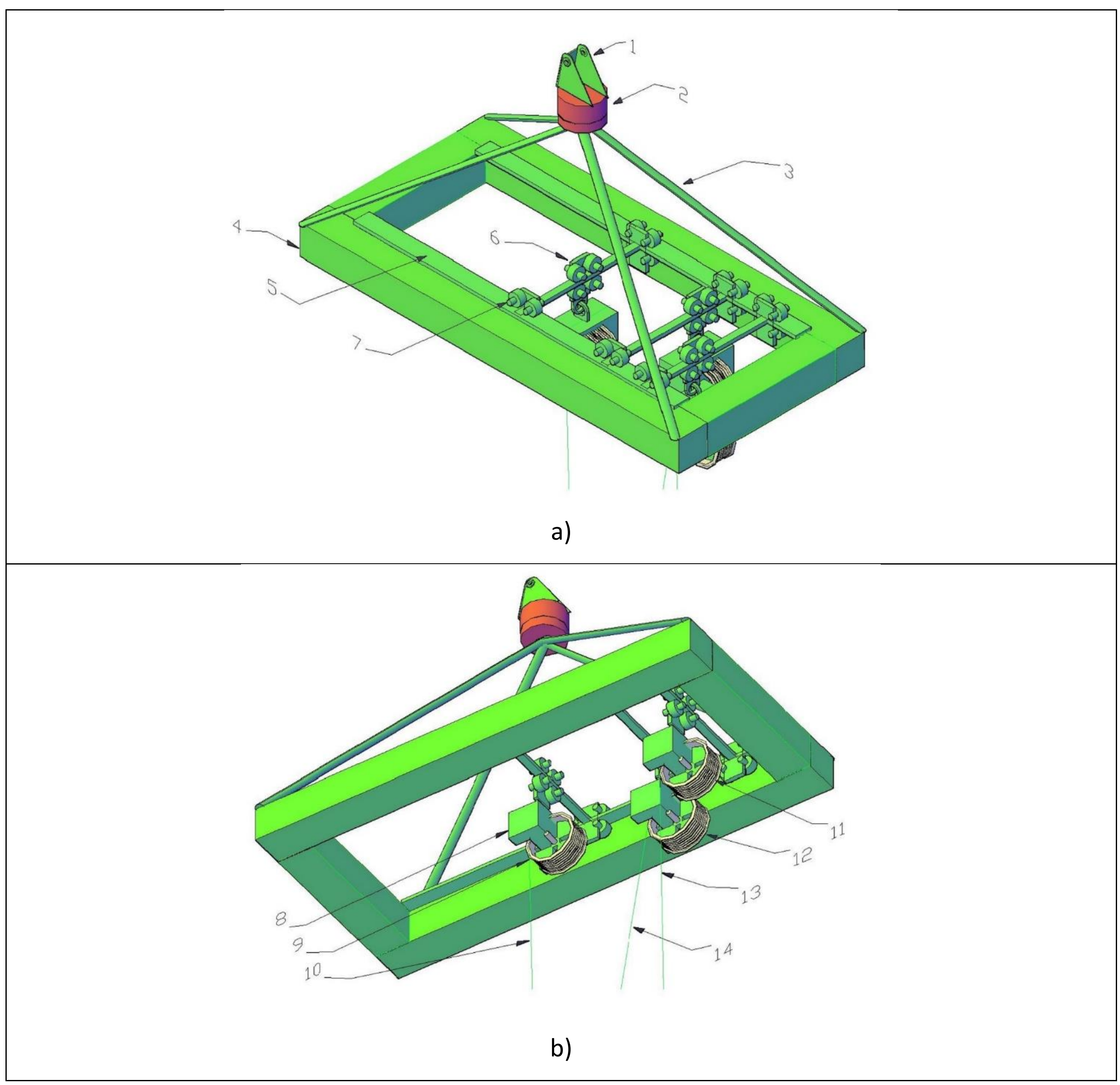

Figure 3. Stabilization subsystem. a) Top view: 1-Rotation pivot for transport position; 2-Electric rotator; 3-Support rods; 4-Frame; 5-Pulley bridge movement rail; 6-Overhead crane; 7-Longitudinal displacement wheel assembly. b) Bottom view: 8-Battery chassis, reducer and electric motor; 9-Stabilizing pulley 1; 10-Stabilizing cable 1; 11-Stabilizing pulley 3; 12-Stabilizing pulley 2; 13-Stabilizing cable 2; 14-Stabilizing cable 3.

The decoupling subsystem consists of three subassemblies, Figure 4: (A)-Grapple subassembly; (B)-Alignment and load subassembly and (C)-Rotation and support subassembly. The rotation and support subassembly consists of a rigid main chassis, where, on the left side of this chassis, a high-capacity lithium battery is housed, and the hydraulic system for a rotator, the power inverter systems, the digital controllers, the industrial computer of the planning layer, and the computer vision system are also coupled. The entire subsystem is supported by a hydraulic rotator from a harvester head located on the upper part of the chassis. This rotator is operated by an oil flow from a hydraulic pump driven by an electric motor. The rotator is coupled to the main support steel cable of the MH. At the bottom of the chassis is the coupling mast of the stem cutting subsystem.

On the right side of the decoupling subsystem chassis, there are bearings and shafts and a main beam that support three claws, each composed of two pincers. Each pincer is actuated by a high-force electric linear actuator. The upper and lower claws have an automatic hook and loop system at their tips. These hooks and loops are coupled to steel cables, which are attached to reel systems located next to the main shaft supporting the claws and are driven by high-power electric motors. The intermediate claw has two chainsaw cutting systems coupled to it, driven by electric motors, and in the center of the support mast there is a magnetic cannon loaded with a metal arrow.

At the top of the main beam of the claws, there is a system consisting of a high-load-capacity chain, where two magnetic cannon support carriages are coupled. Each carriage has two magnetic cannons oriented diametrically to each other. These vehicles move using chain-driven gears coupled to a chain and powered by electric motors. The magnetic cannons are loaded with metal arrows that, at the rear after the steering fins, are attached to small-diameter steel cables. These magnetic cannons have an automatic laser aiming system to guide them to target the center of the tree trunk at the height of the canopy.

The stabilization and decoupling subsystems communicate with each other and with the control center on board the helicopter via a CAN wired network.

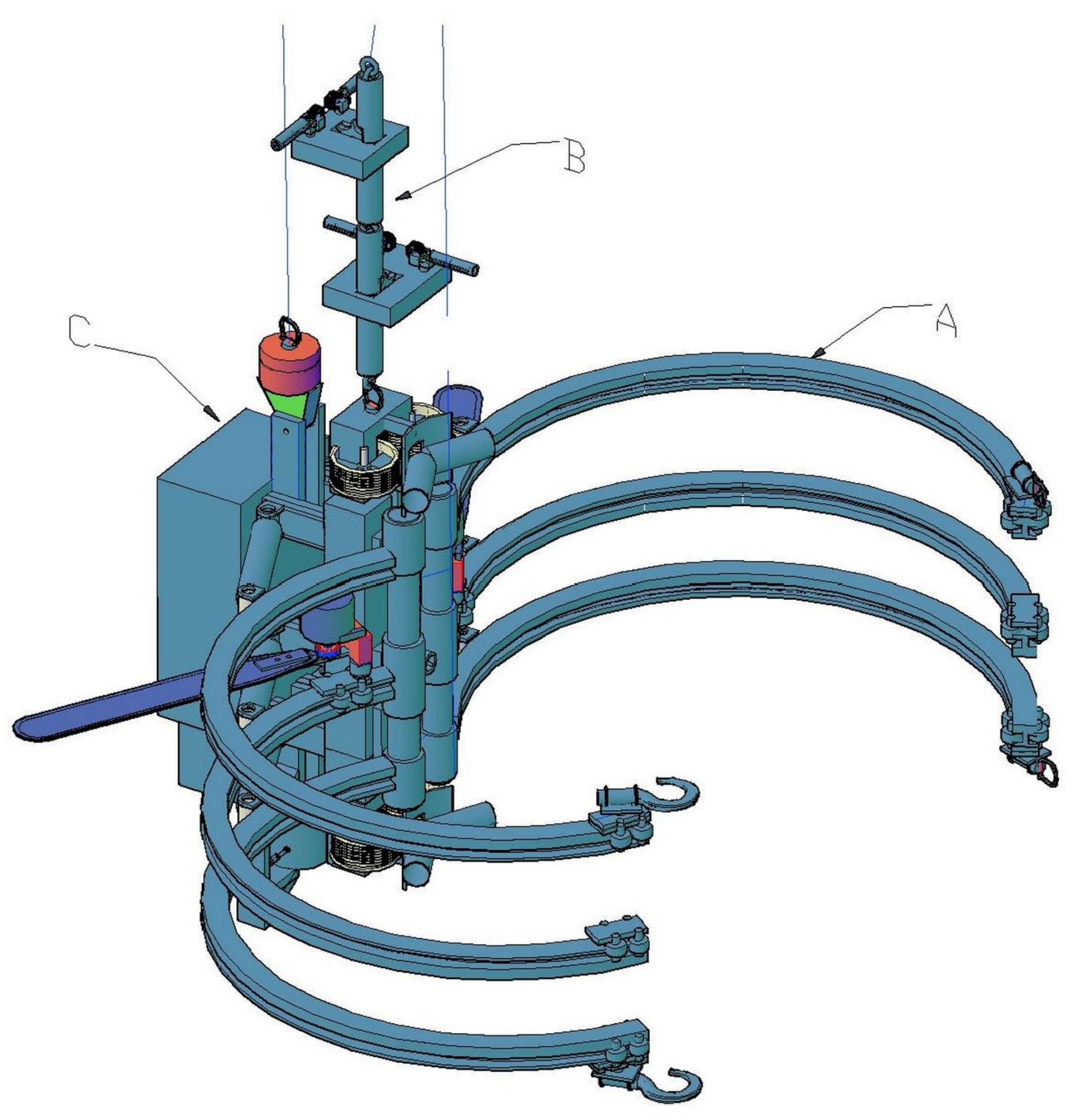

Figure 4. Decoupling subsystem: A; Gripper subassembly; B-Load/alignment subassembly; C-Rotation and support subassembly.

The stem cutting subsystem, Figure 5, consists of a stem traversing subset (A), a cutting subset (B), and a delimbing subset (C). A main chassis, on the left side, houses a high-capacity lithium battery, and also connects the canopy clearing system, the stem traversing system, the power inverter systems, the digital controllers, the industrial planning layer computer, and the computer vision and LoRa communication systems.

The shaft walking system consists of a gripper located at the top of the system, composed of two tongs. Each tong is actuated by a high-force electric linear actuator. On each tong, there are three metal wheels with sharp, prismatic, tungsten carbide tips, each driven by high-power electric motors. These wheels and motors are located at the ends of a telescopic system actuated by a high-force electric linear actuator. Each wheel/motor/actuator assembly is deflected at an angle radially oriented to the arc of the gripper by electric linear actuators.

The delimbing system is located at the bottom of the chassis and consists of a claw composed of two pincers, each pincer is driven by an electric linear actuator, in each pincer there are three high-speed circular cutting systems with sharp carbide blades oriented longitudinally to the shaft, these blades are driven by high-speed electric motors, and are supported by a telescopic system driven by an electric linear actuator and also has a deflection angle controlled by another electric linear actuator. At the top of the chassis there is the coupling claw to the coupling mast of the delimbing subsystem.

On the right side of the shaft walking subsystem chassis there are bearings and shafts and a main beam that support three claws composed of two pincers each. The intermediate pincer is driven by a high-force electric linear actuator. This claw has two chainsaw cutting systems coupled to it, driven by electric motors. The upper and lower claws are segmented into four sections, each section is actuated by high-force electric linear actuators.

The shaft cutting subsystem communicates with the stabilization and decoupling subsystems and with the control center on board the helicopter via a LoRa wireless network.

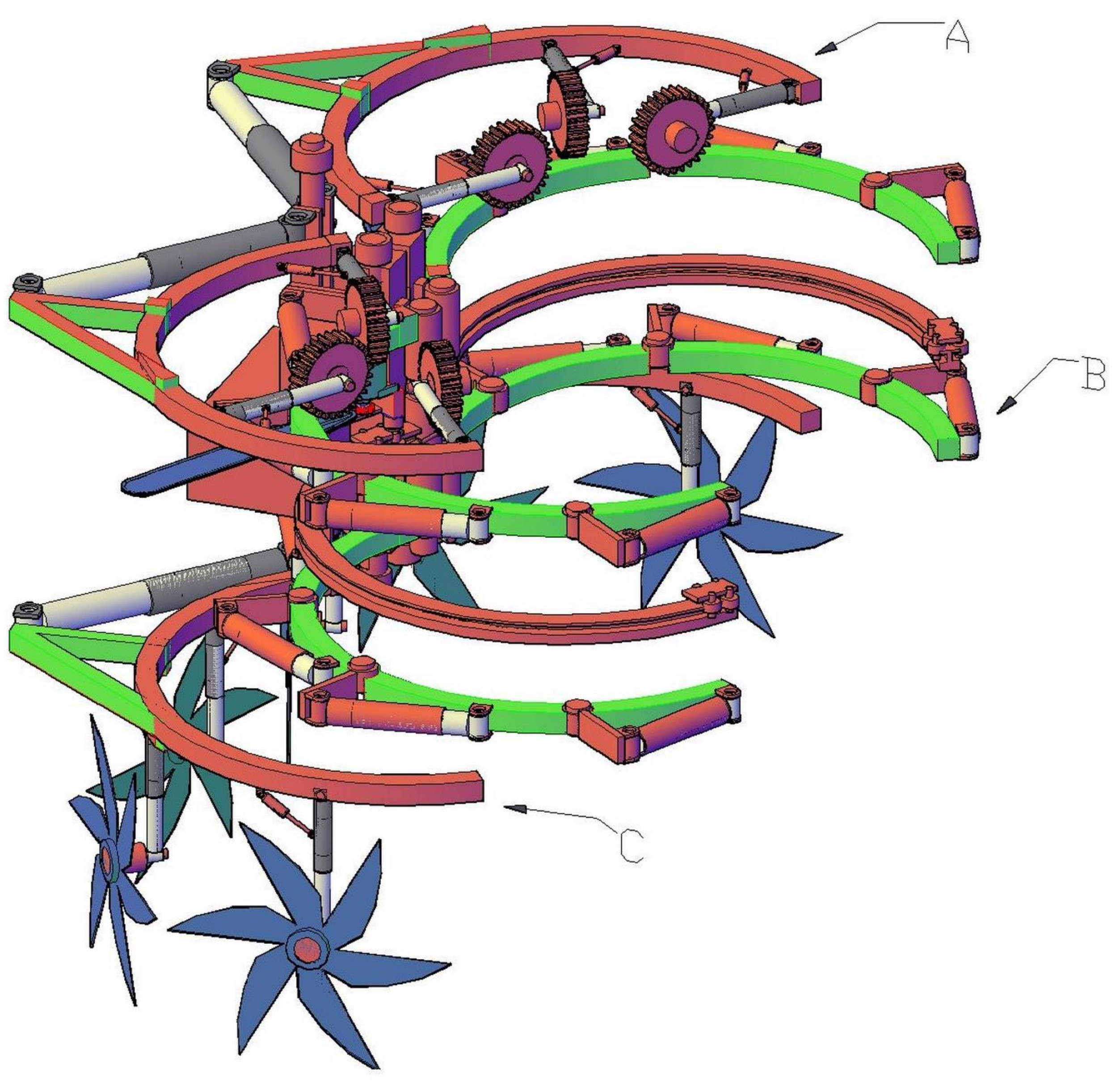

Figure 5. Stem cutting subsystem: A) Stem walking subassembly; B: cutting subassembly and C) delimbing subassembly.

**Module Harvesting Silvicultural Treatments**

Considering the process flowchart, Figure 6, and the morphological matrix and solution synthesis methodologies described by [64,65] used by [65–67] and [75], it was possible to find optimal solution paths and synthesize the concept of the HST Module (MHST).

Figure 6. Flowchart of the HST process with a URIEL.

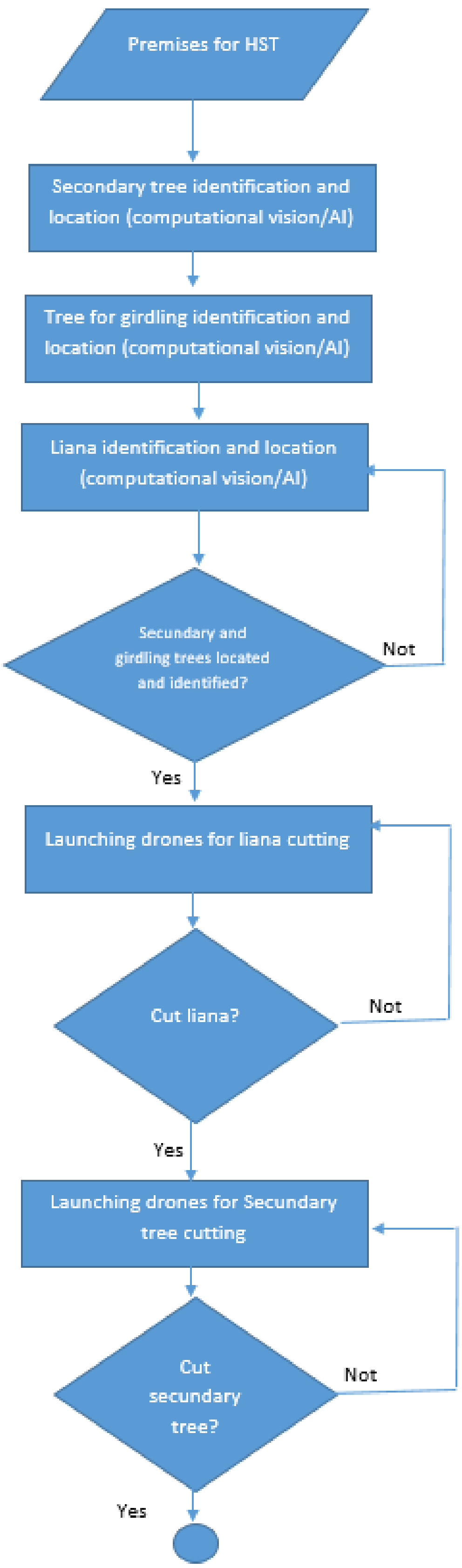

Premises for HST
Secondary tree identification and location (computational vision/AI)
Tree for girdling identification and location (computational vision/AI)
Liana identification and location (computational vision/AI)
Secundary and girdling trees located and identified?
Not
Yes
Launching drones for liana cutting
Cut liana?
Not
Yes
Launching drones for Secundary tree cutting
Cut secundary tree?
Not
Yes

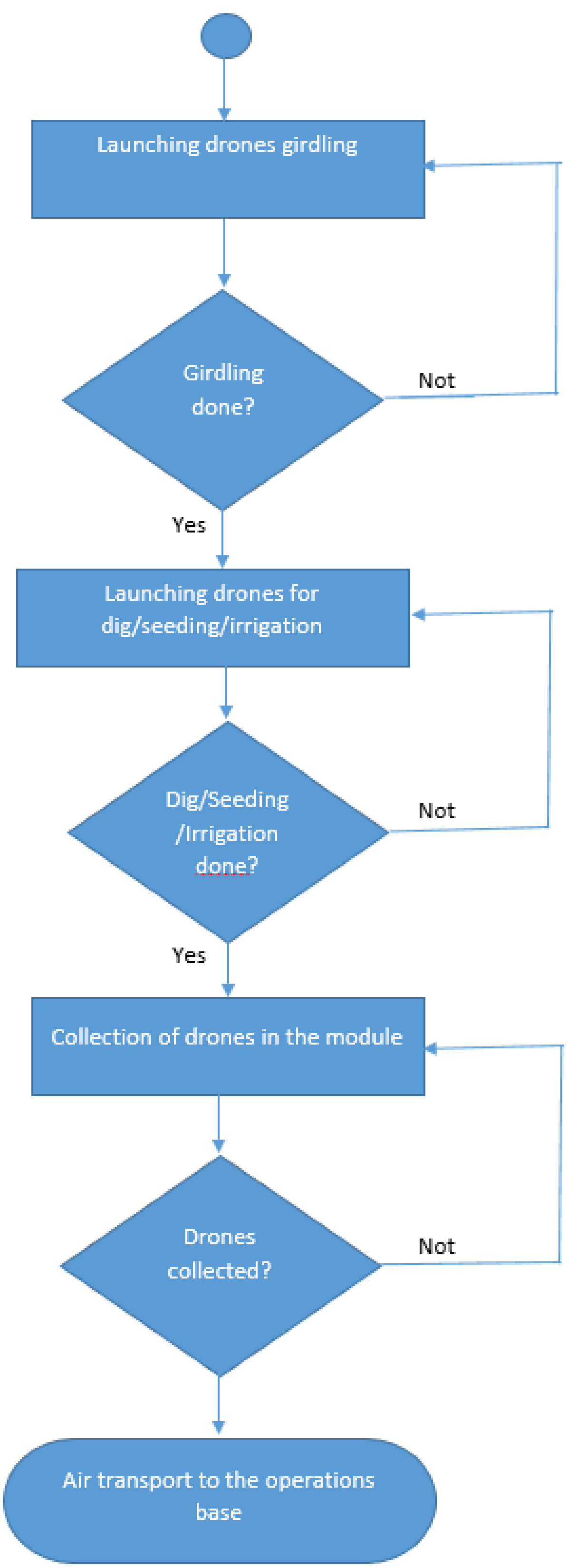
Launching drones girdling
Girdling done?
Not
Yes
Launching drones for dig/seeding/irrigation
Dig/Seeding /Irrigation done?
Not
Yes
Collection of drones in the module
Drones collected?
Not
Air transport to the operations base

The concept of the HST module is based on a drone-carrying system, Figure 7, which transports specific drones responsible for each of the main missions of a Harvest Silvicultural Treatment system: Liana cutting; Cutting of small secondary trees; girdling of large secondary trees; preparation of planting pits; irrigation and planting of seedlings.

It is important to emphasize that there are already drones that airlift commercial systems (TRL9) or in an advanced stage of development (TRL7) for several of these proposed HST actions: Drone for liana cutting [76]; cutting of small secondary trees [77]; pit preparation [78]; irrigation [79]; planting/sowing [80]. And the arms that carry the specific terminal tools with various degrees of freedom are also already a mastered technology [81].

The drone carrier system consists of a hexagonal star-shaped chassis. Each arm of this star has a rail where a locomotion system, consisting of a hitch and wheels, attaches to the star arm. The power source is electric. This hitch supports a coupling mast consisting of a female gripper actuated by linear actuators. Each drone has a male pivot that is suitable for gripping by the female gripper; this pivot is passive and connects to the central structure of the drone. Detailed drawings of the MHST and each drone are presented in the supplementary material Supplementary Figures.

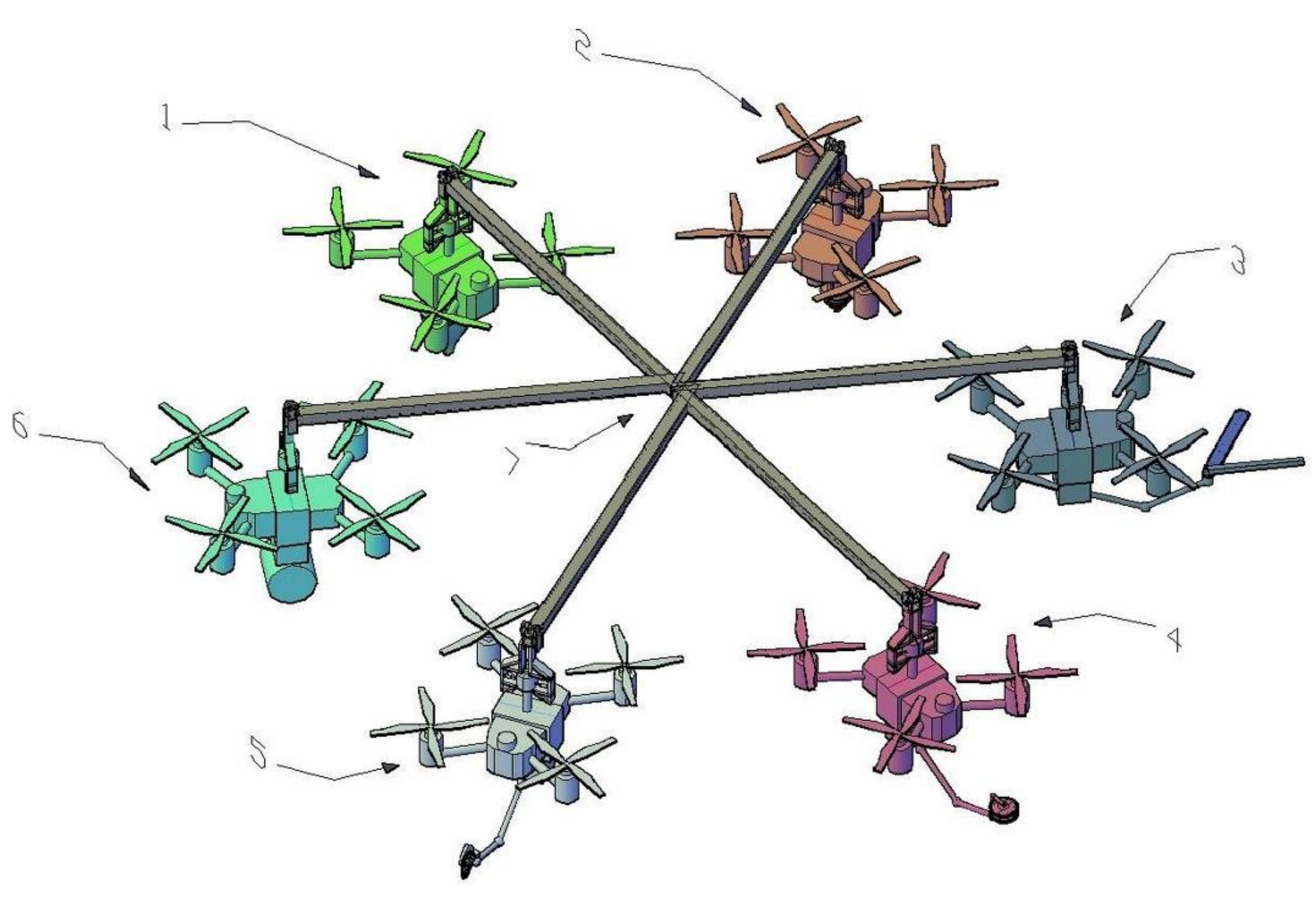

Figure 7. Drone carrier system. Open star (a): 1-Planting drone; 2-Hole digging drone; 3-Liana cutting drone; 4-Girdling drone; 5-Secondary tree cutting drone; 6-Irrigation drone; 7-Hexagonal drone carrier chassis with closed star. Detailed projects in Supplementary material (SupplementaryFigures).

The drones are all of the same model (for example, the DJI Agras 70P), specified as a quadcopter with conventional computer vision systems for drones consisting of RGB cameras, radars, and LIDAR, and a GPS-RTK navigation system (the RTK antenna is mounted on the helicopter). The payload is 70 kg and the flight autonomy is 15 minutes. Each drone can have a robotic arm with 9 degrees of freedom diagrammed as two links (arm and forearm) and three joints (rotations in x, y, and z). For each HST mission there is a different system attached to the drone (robotic arm or structure on the fuselage) namely:

a) Liana cutting drone (DCL), Figure 7(3): there is a terminal tool attached to the robotic arm consisting of a chainsaw bar and its respective cutting chain, this bar mounted on a cylindrical support that houses a high-speed electric motor that drives a gear that moves the cutting chain. On the same support there is a counter-blade consisting of a rigid rectangular support located geometrically just above the bar. There is a servomotor coupled between the saber/gear/motor system and the cylindrical support in such a way that, under command, controlled rotation occurs in terms of angle and angular velocity of the saber on the cylinder axis, generating a mixed cutting action (cutting chain plus counter-knife). This action is important considering the specificities of lianas entangled between trees and between other lianas. The combined cutting action causes a bending moment between the biological elements in the form of a "rope" and the cutting system, thus enabling the cutting of loose elements without a support base, such as lianas suspended in trees.

b) Drone for cutting small secondary trees (DCA), Figure 7(5): there is a terminal tool coupled to the robotic arm consisting of a circular saw with Wydia teeth and a support cylindrical saw. This circular saw/support cylindrical saw assembly is mounted on a cylindrical support. The rotation of the circular saw occurs due to a high-speed motor coupled to the hub of the circular saw and locked onto the fork of the circular saw shaft. In the cylindrical support there is a servomotor coupled in such a way that, under command, controlled rotation occurs in terms of angle and angular speed of the circular saw/support hammer assembly.

c) Drone for girdling (ringing) large secondary trees (DAN), Figure 7(4): there is a terminal tool coupled to the robotic arm consisting of two circular saws with Wydia teeth arranged in parallel and between them a movable debarking blade, this whole assembly is mounted on a cylindrical support. The rotation of the circular saws occurs due to a high-speed motor coupled to the hub of the circular saws and locked onto the fork of the circular saw shaft. In the cylindrical support there is a servomotor coupled in such a way that, under command, controlled rotation occurs in terms of angle and angular speed of the assembly. The debarking blade is positioned in such a way that its cutting edge aligns 5 millimeters with the base of the circular saw teeth, so that when the double cut is made by the saws, this blade immediately penetrates the generated bark strip. Then, a high-torque servomotor mounted on the saw shaft block generates a penetration action of this blade into the doubly sectioned bark. In this action, the robotic arm maneuvers, together with the positioning and orientation of the drone, in such a way that the tree trunk is circled, thus performing the ring strip cut by the saws and immediately on-the-go debarking of this ring by the debarking blade. Of all the operations planned for the MHST, this is by far the most complex and the only one for which there is no system in the available literature.

d) Pit digging drone (DCO), Figure 7(2): the pit digging drone does not have a robotic arm. The drilling system consists of a treated alloy steel helical conical drill bit coupled to a shaft connected to a geared motor system driven by an electric motor. The geared motor assembly generates high torque at the medium/low speeds necessary for drilling holes in hard ground. The entire assembly is mounted in a highly shear- and bend-resistant, yet lightweight, aluminum alloy cylinder tube. This structure is attached to the drone via a lower ring support and four support rods that are fixed to a chassis attached to the drone's fuselage.

e) Planting Drone (DPL), Figure 7(1): the planting drone does not have a robotic arm. The planting system is a commercial system developed by the company AirSeeds Technologies and consists of an encapsulated seed ejector module. These seeds are stored in a reservoir attached to the ejector module. This assembly is attached to the drone via rods that connect the module to the drone's support chassis.

f) Irrigation Drone (DIR), Figure 7(6): the irrigation drone does not have a robotic arm. The irrigation system consists of a water storage tank (40 liters) that has a conductor tube where at the end there is an interchangeable liquid fragmentation/jet launch assembly, thus there is the choice of spraying water in droplets through a conventional atomizer system or launching a continuous jet of water. The hydraulic pump is housed in a box above the tank, which also contains lithium batteries.

This supplementary power module is necessary to power the drone due to the energy consumption of the pump and irrigation system, since using only the drone's batteries would significantly reduce its range. The entire assembly is attached to the drone's fuselage by rectangular supports bolted to the drone's chassis beams, given the considerable weight of the assembly and its dynamic actions during flight maneuvers.

**Pod URIEL**

Harvesting modules (MH) and silvicultural treatment modules (MHST) are large pieces of equipment with elements that can be damaged if they impact tree branches, birds, or even due to the aerodynamic drag force from air displacement in helicopters flying at more than 200 km/h. Thus, it is necessary to design a transport pod that must be fitted ventrally to the helicopter fuselage.

As inspiration for aircraft coupling, aerodynamic geometry and structural characteristics, the AN/ALQ-131 ECM (Electronic Counter-Measures) pod used by the McDonnell Douglas RF-4C Phantom II reconnaissance fighter was used [82]. The transport pod, Figure 8, was structured with a chassis of stringers in aeronautical aluminum beams and its fuselage follows an aerodynamic geometry to avoid excessive drag. It consists of two continuous sections with different dimensions, one for the MH system and the other for the MHST. The section that houses the MH system contains the Uriel System command post, which is operated by a person who manages all the activities of the two modules, the ground movement operations of the Pod, and also the final approach of the helicopter to the target. This section is closed by ventral and movable sealing hatches. The MHST section follows right behind the MH section and is ventrally open.

Due to the size of the MH and MHST modules and their longitudinal arrangement within the transport Pod, it was necessary to design a landing gear system, as the transport Pod must be coupled and uncoupled to the helicopter while it is hovering over the Pod. This is because its dimensions prevent this coupling with aircraft on the ground. This system consists of two sets, one front (next to the MH module) and one rear (next to the MHST module). Each set is formed by a tandem system that supports two axles, each axle with 2 wheels. Each set is directional and has electric traction motors coupled to the wheel hubs. The power source for moving the Pod using the wheels and also for steering the tandems is external, consisting of a lithium battery pack that is towed at the rear of the Pod.

The URIEL System operator: When the Pod is positioned at the coupling point with the helicopter and the helicopter approaches in hovering flight over the Pod, the ground crew couples the Pod to the helicopter via the front, intermediate, and rear pylons. At this moment, the Pod's electrical system is coupled to the helicopter's electrical power source, and the ground crew disconnects the external source and removes it from the flight path. Details of the URIEL Pod are presented in the supplementary material Supplementary Figures.

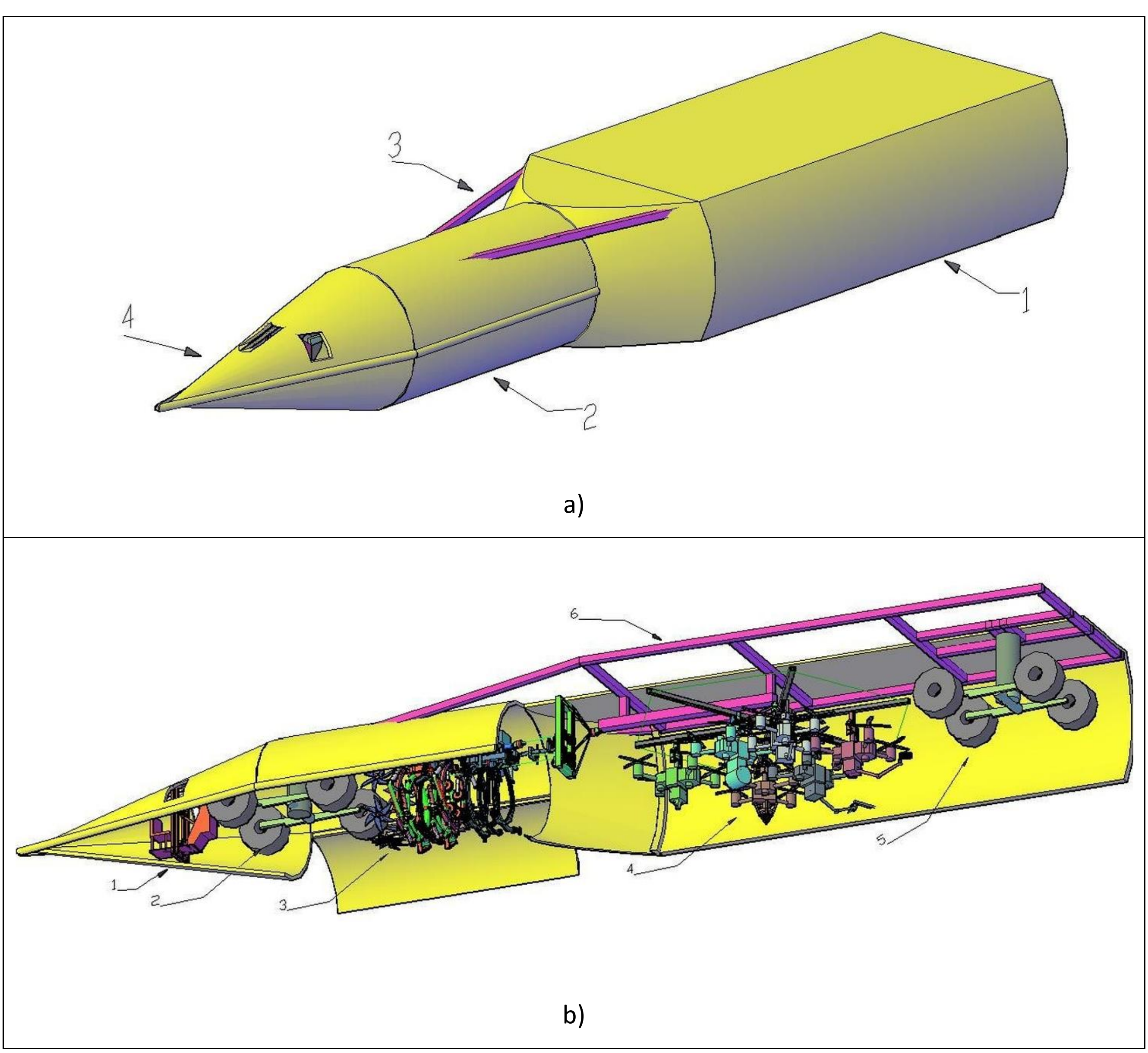

Figure 8. URIEL transport pod. Front view a): 1-MHST transport section; 2-MH transport section; 3-Structural chassis spar; 4-Aerodynamic cone. Ventral view of complete assembly b): 1-URIEL control station; 2-Front landing gear retracted; 3-Harvesting module (MH); 4-Silvicultural treatment module (MHST); 5-Rear landing gear retracted; 6-Main chassis spar.

**Simulations**

Based on the detailed technical drawings, a digital proof of concept (POC) of the URIEL System was developed using a digital model of the system. Additionally, a three-dimensional forest environment was modeled. With the digital model and the 3D forest environment, it was possible to perform simulations of the operation of the URIEL System (Figures 9, 10, 11, and 12). A complete simulation with all phases of harvesting and post-harvest silvicultural treatment is presented in the supplementary material (Supplementary Figures).

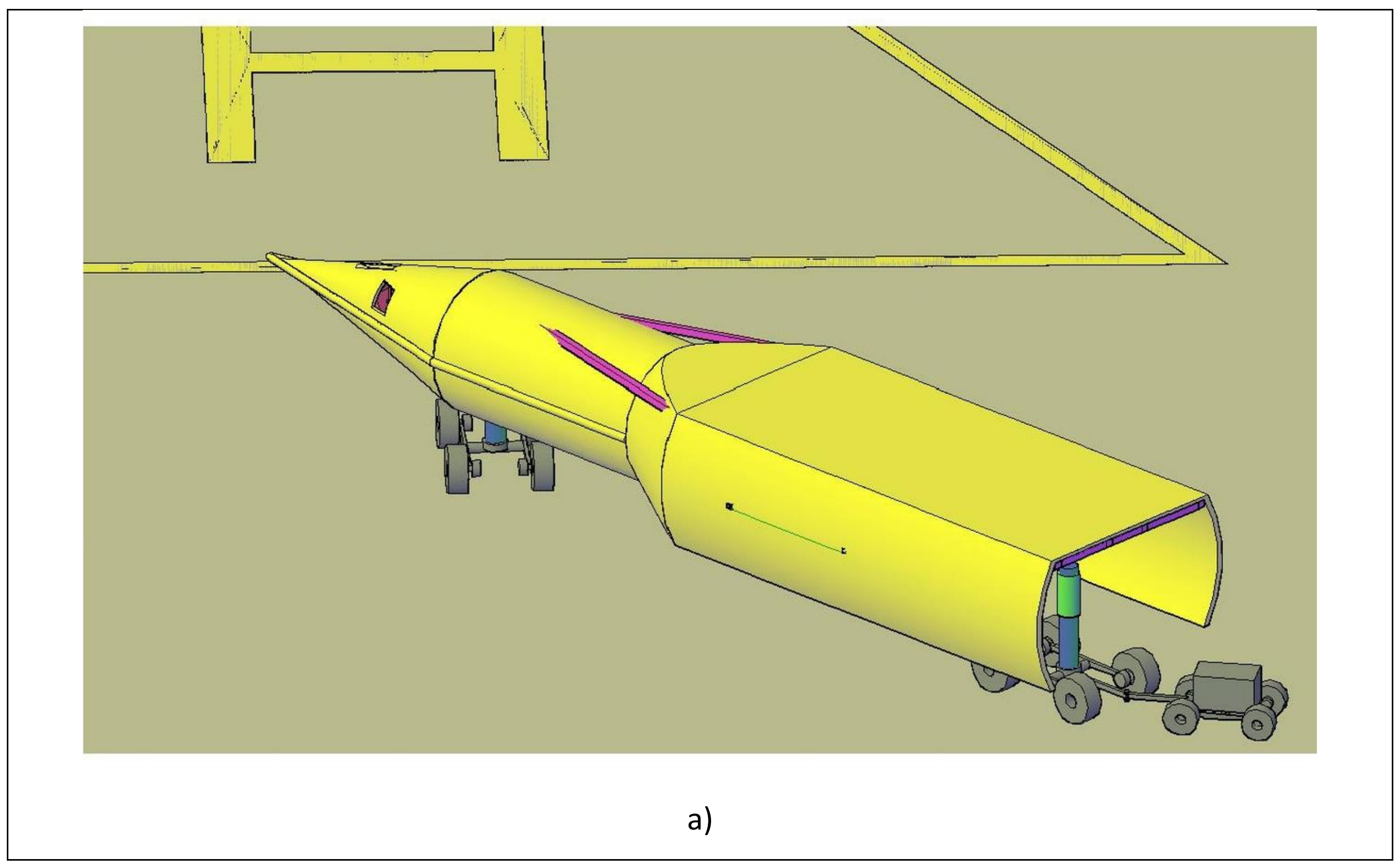

a)

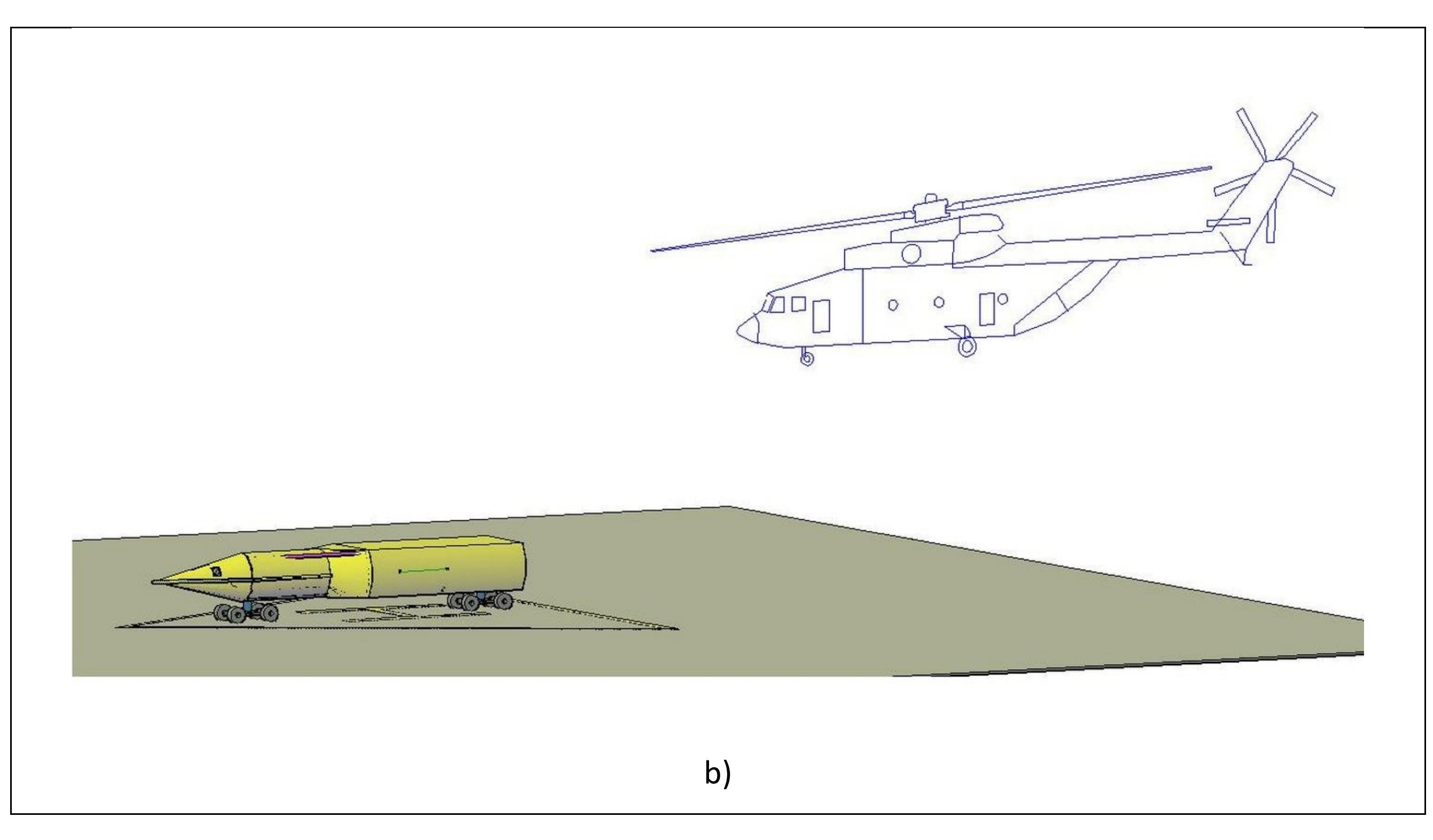

b)

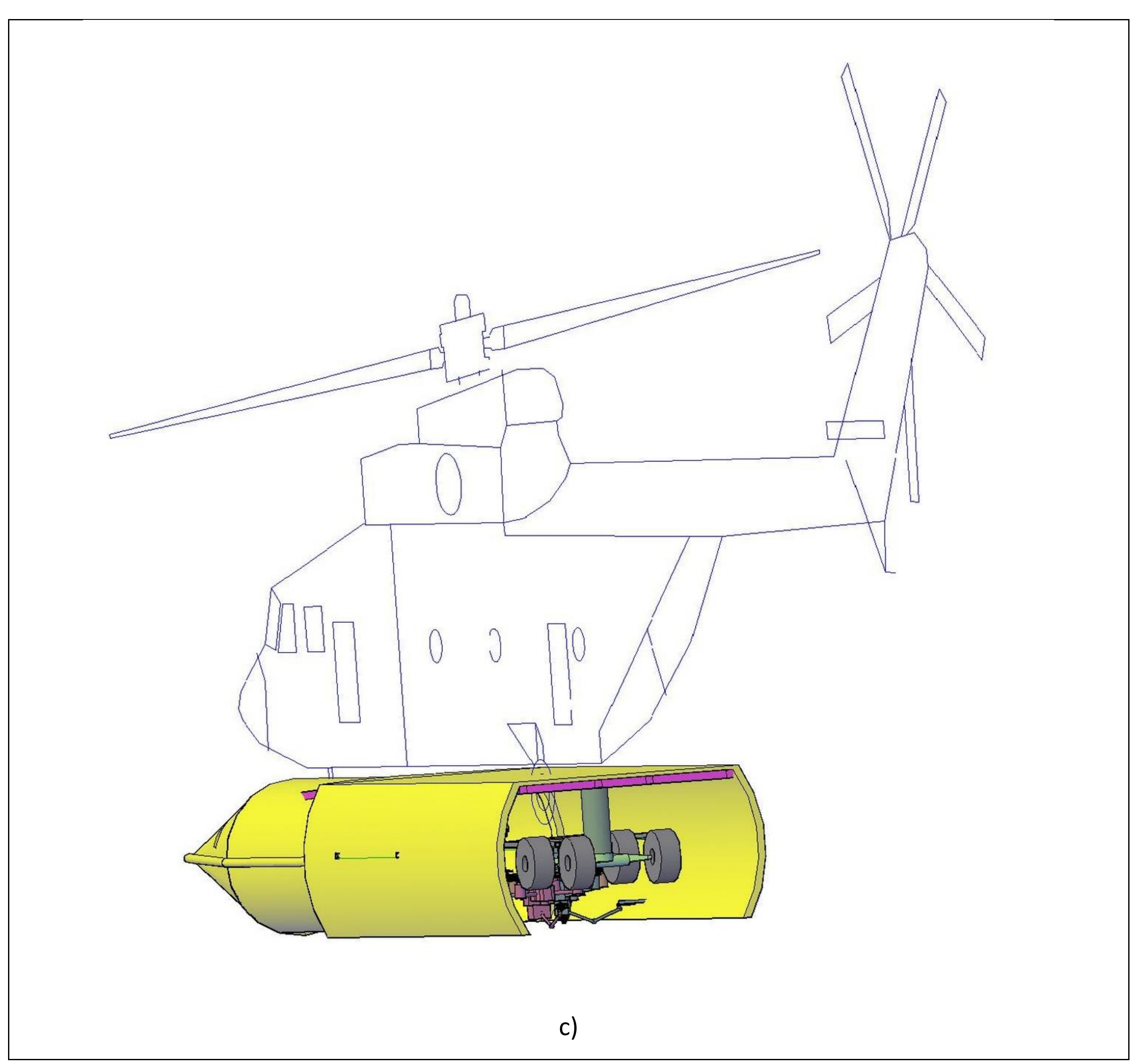

Figure 9. Ground movement simulation of the URIEL Pod with coupling to a Mi-26: a) URIEL Pod moving from the hangar to the helipad; b) Slow-flying approach of a Mil Mi-26; c) URIEL system in flight to the target area, rear view of the URIEL system getting underway.

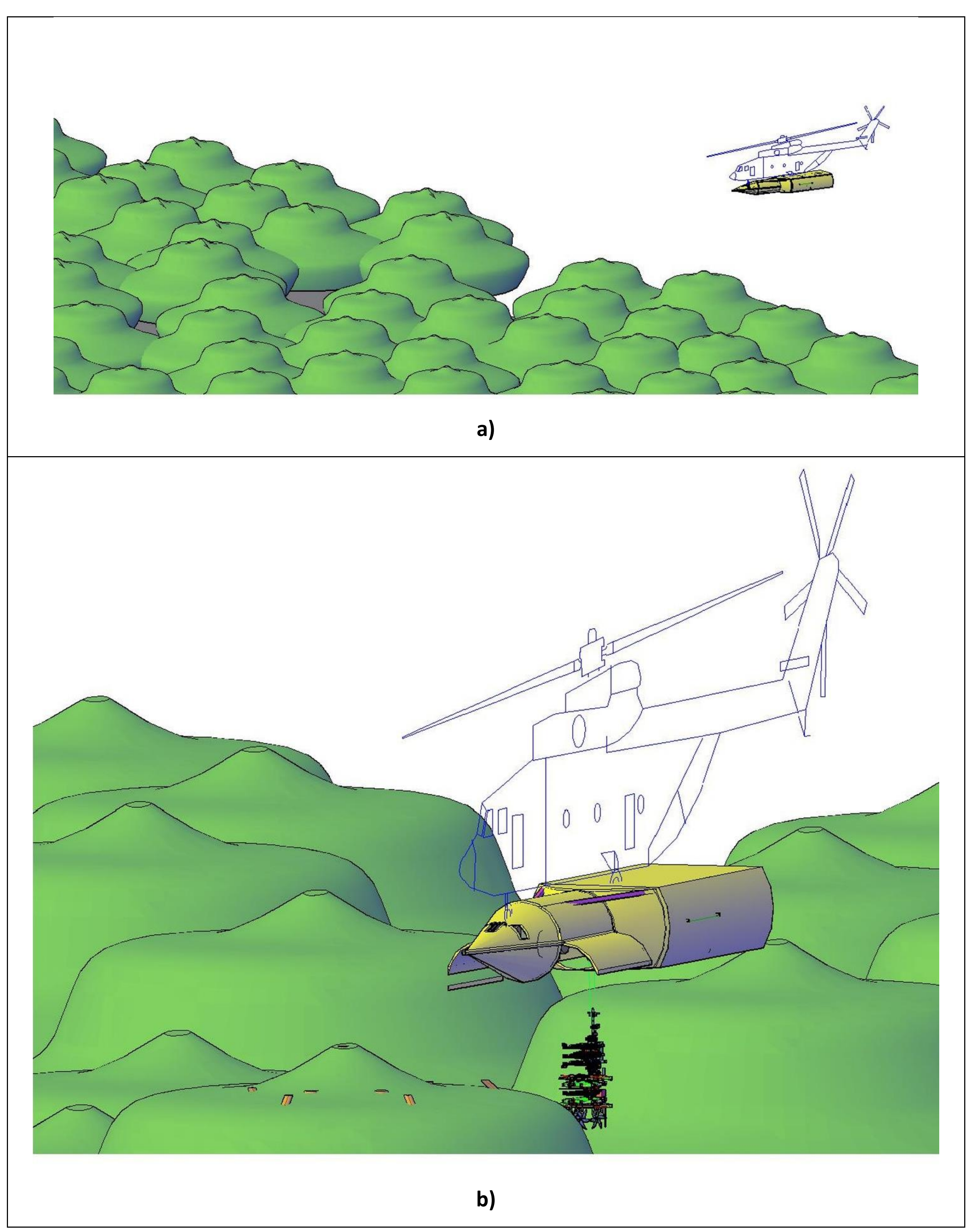

Figure 10. URIEL system arriving at the target area: a) URIEL system entering the forest; b) Location and identification of the target tree.

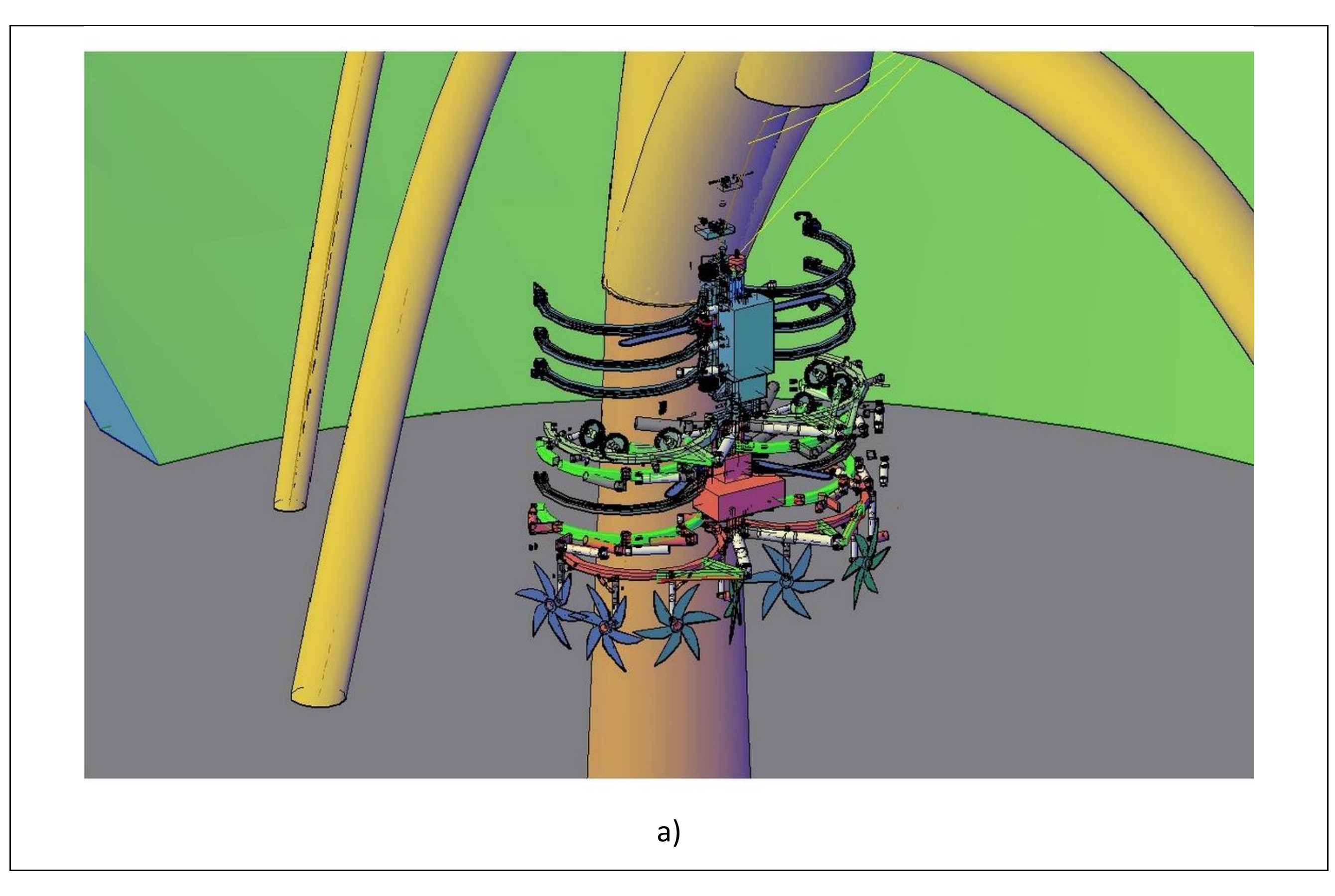

a)

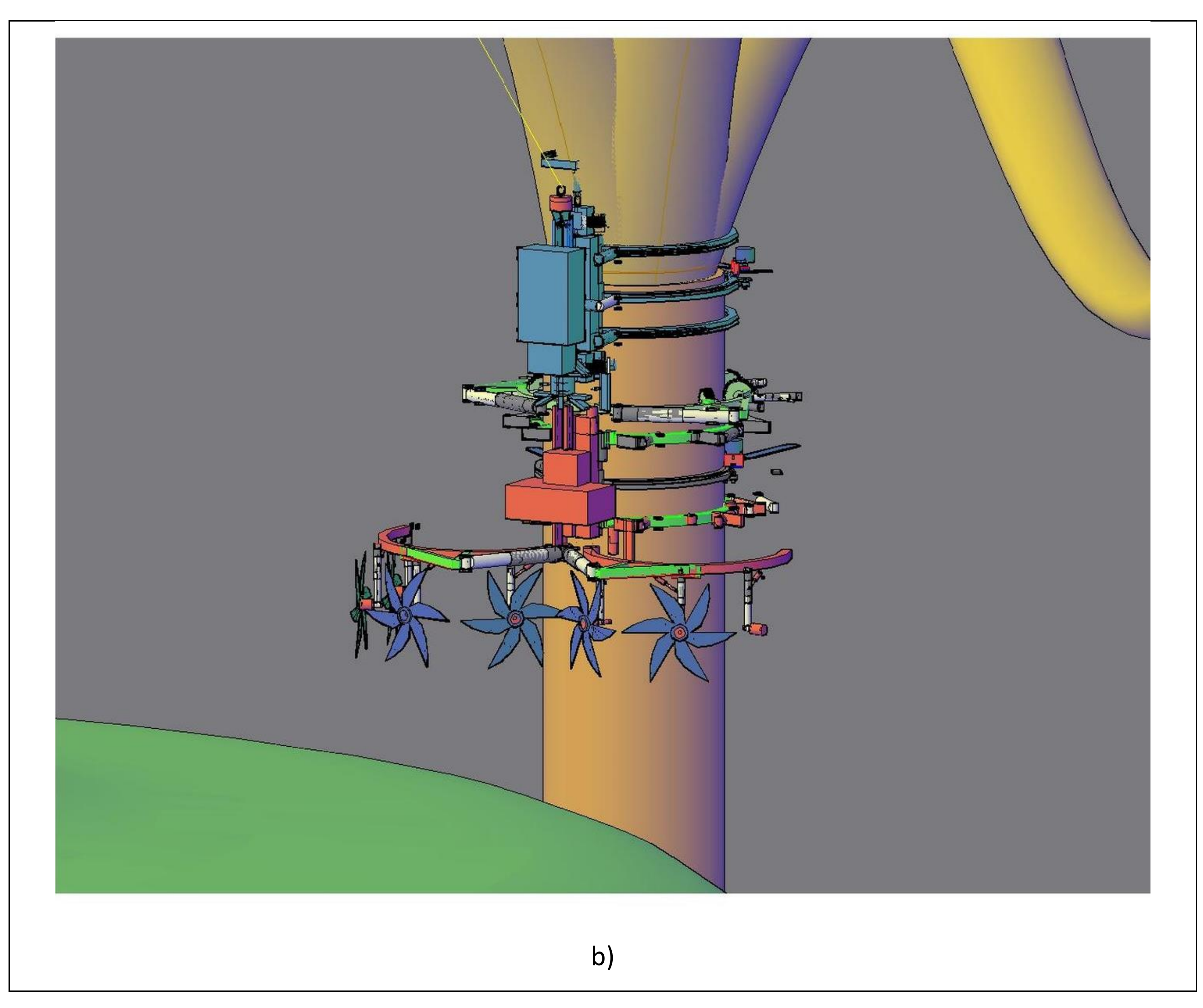

b)

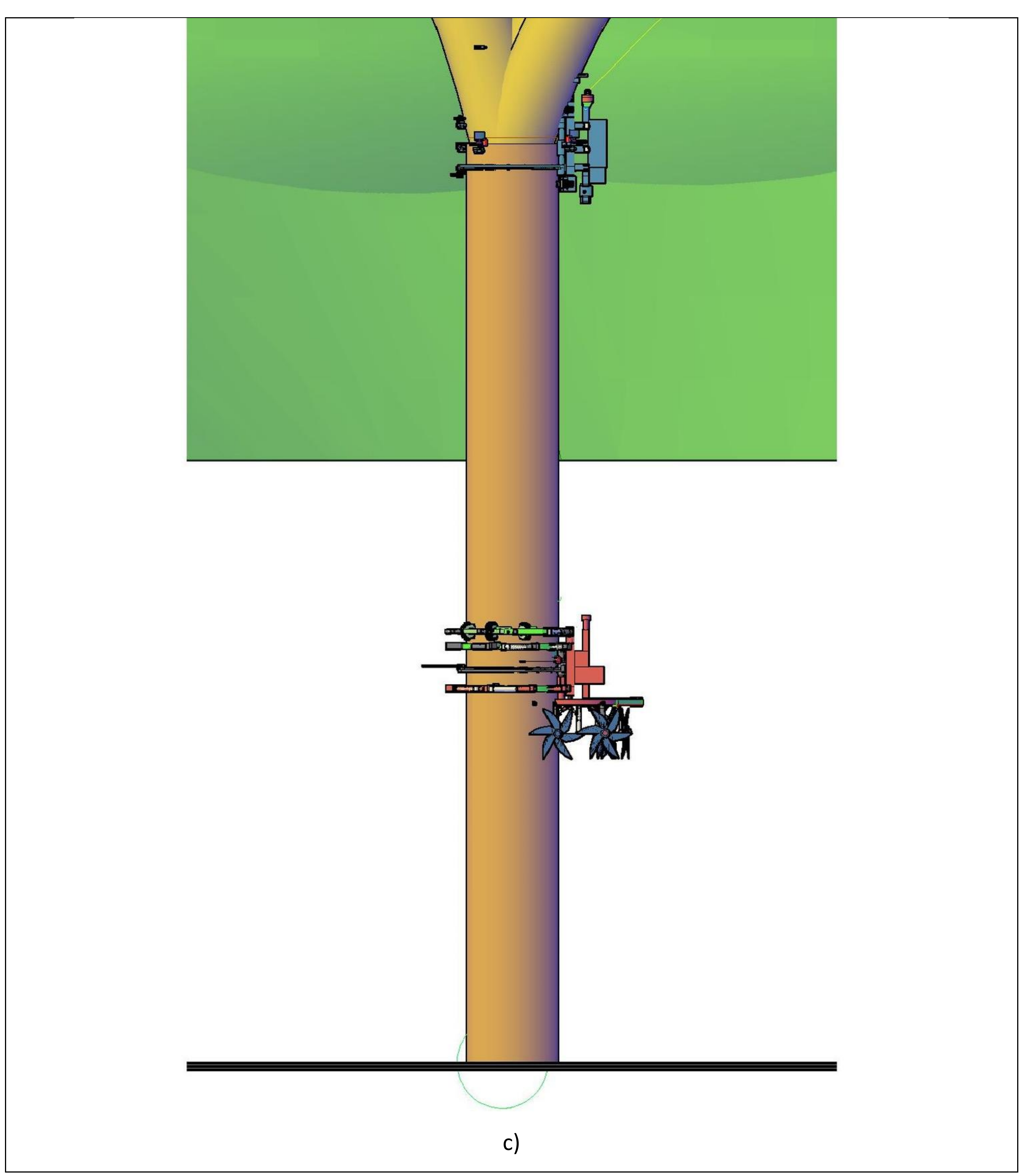

c)

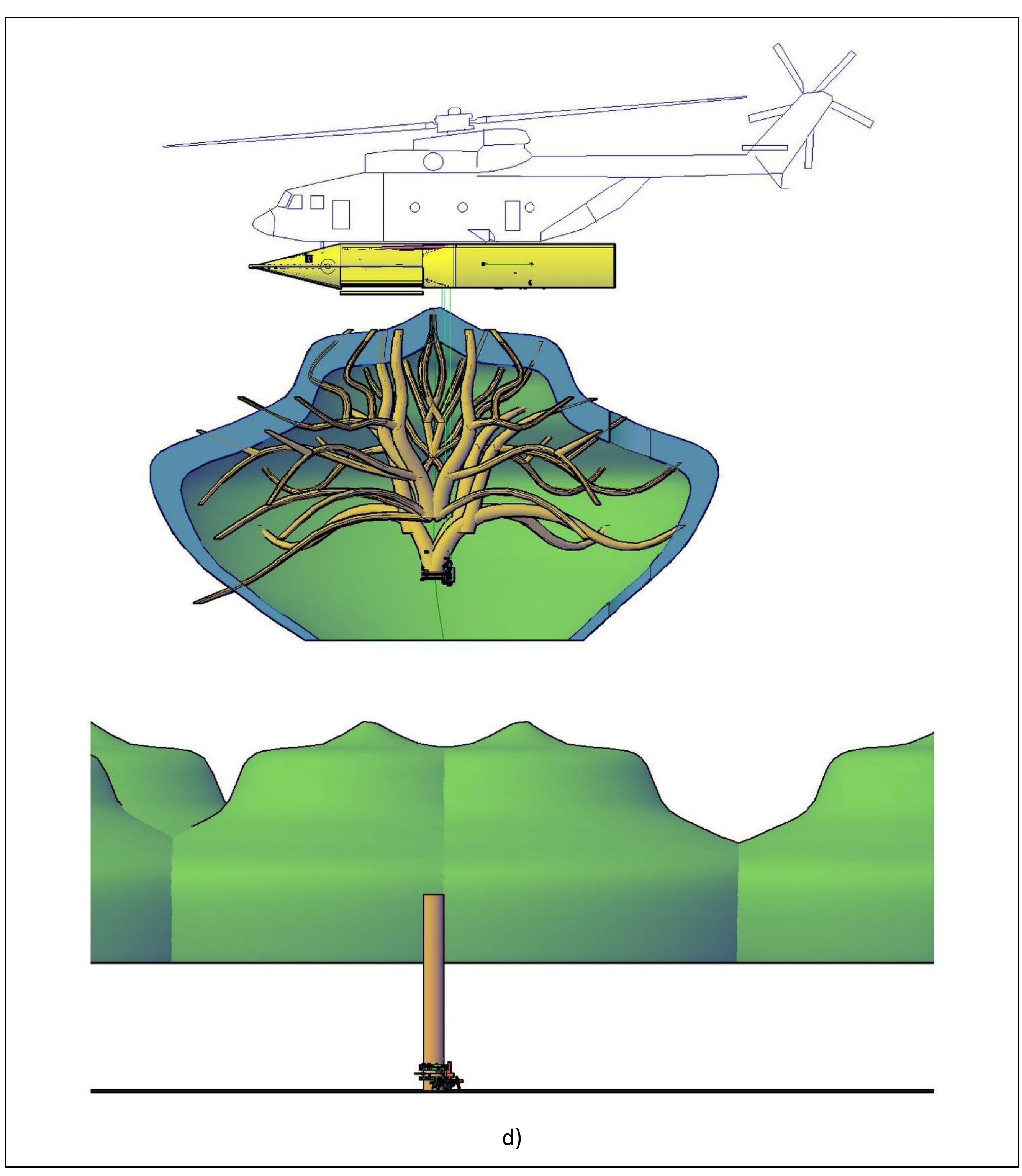

d)

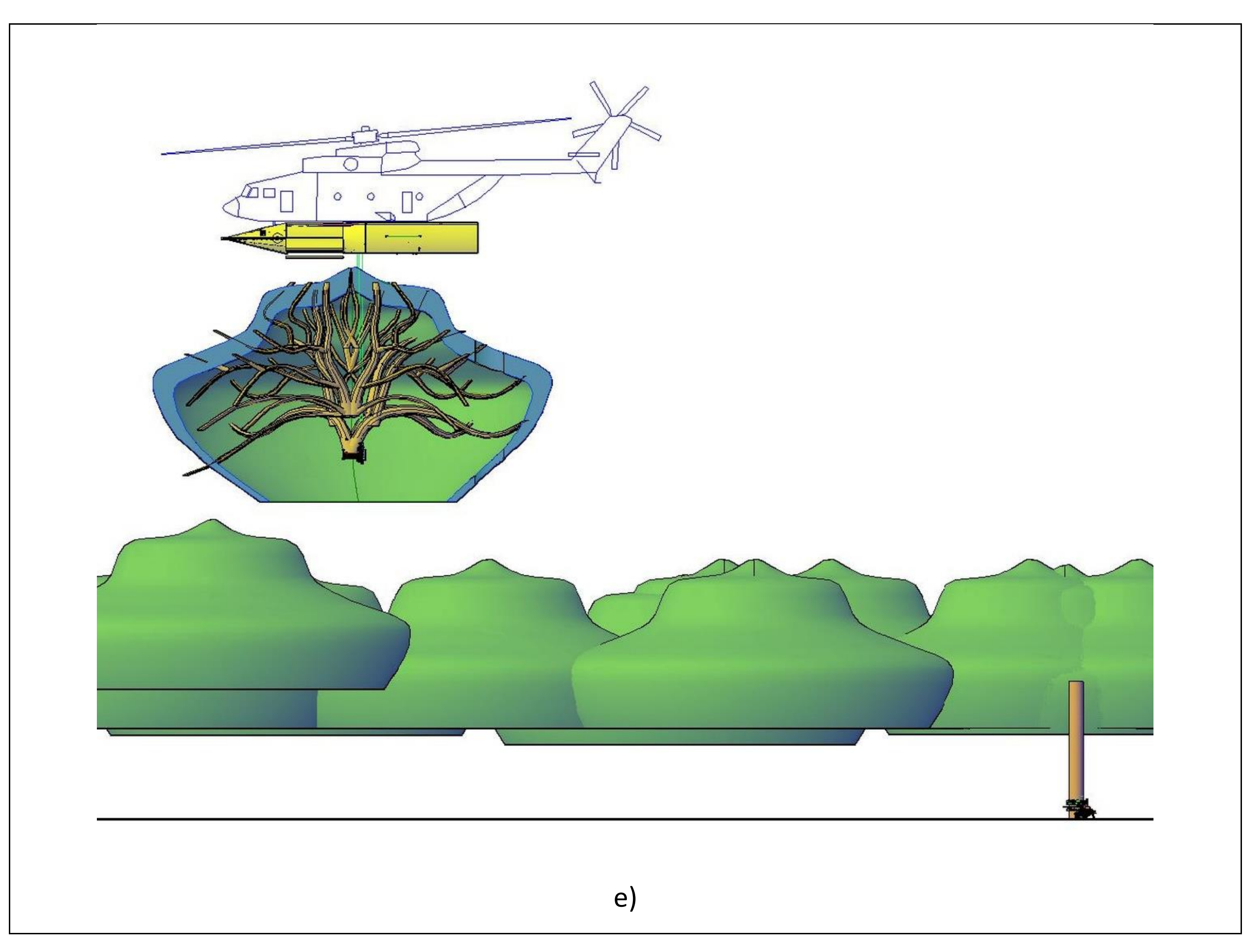

e)

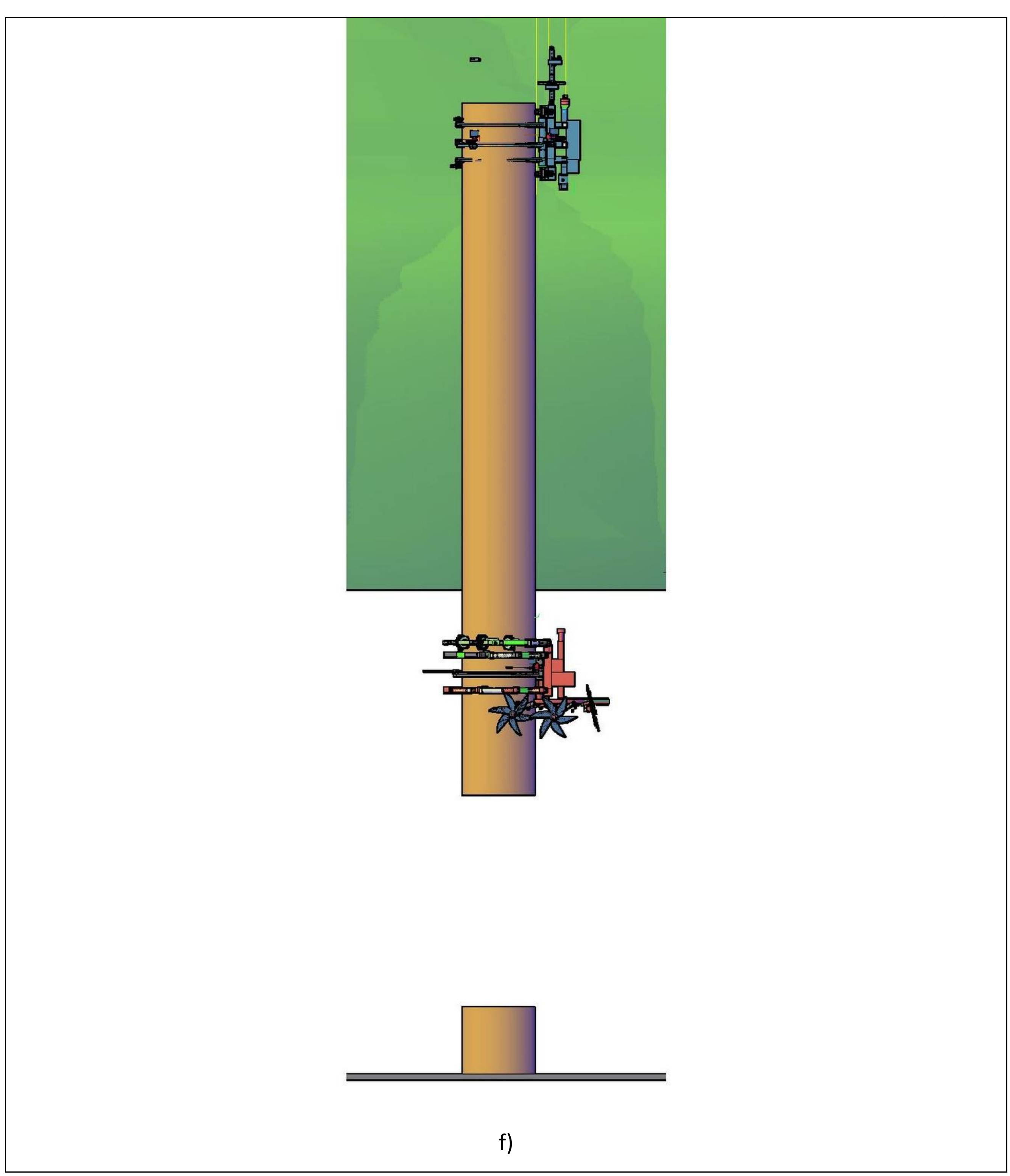

f)

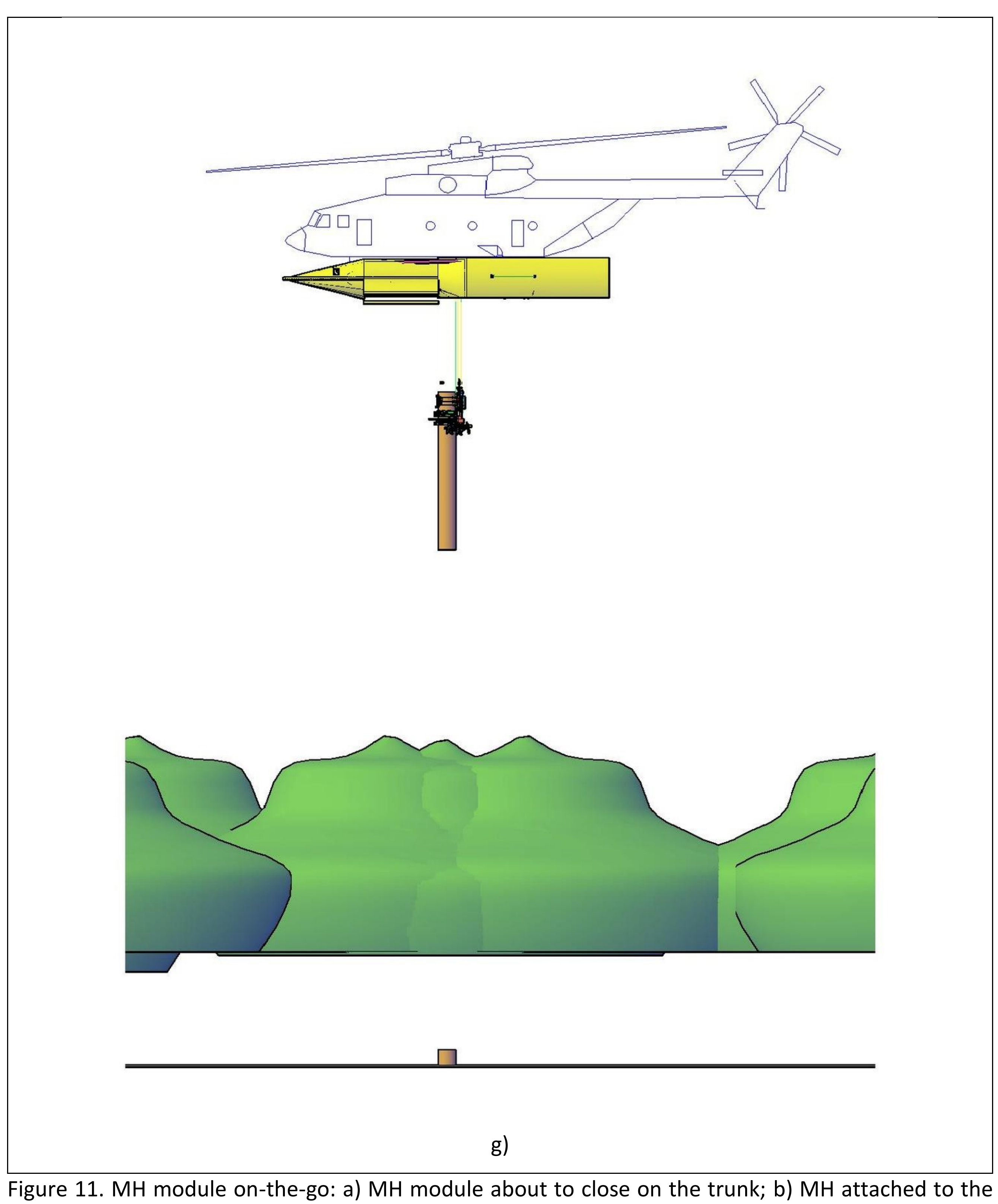

Figure 11. MH module on-the-go: a) MH module about to close on the trunk; b) MH attached to the trunk; c) Decoupling of the decoupling subsystem from the trunk cutting subsystem; d) URIEL system maneuvering to raise the crown and stabilization subsystem performing compensatory force operations to stabilize the cut crown; e) URIEL system moving the crown to the encapsulated crown deposition site on the forest floor; f) Trunk cutting subsystem performed the cut and the stabilization

subsystem raises the trunk by walking upwards from the trunk cutting subsystem; g) Trunk cutting subsystem couples to the decoupling subsystem.

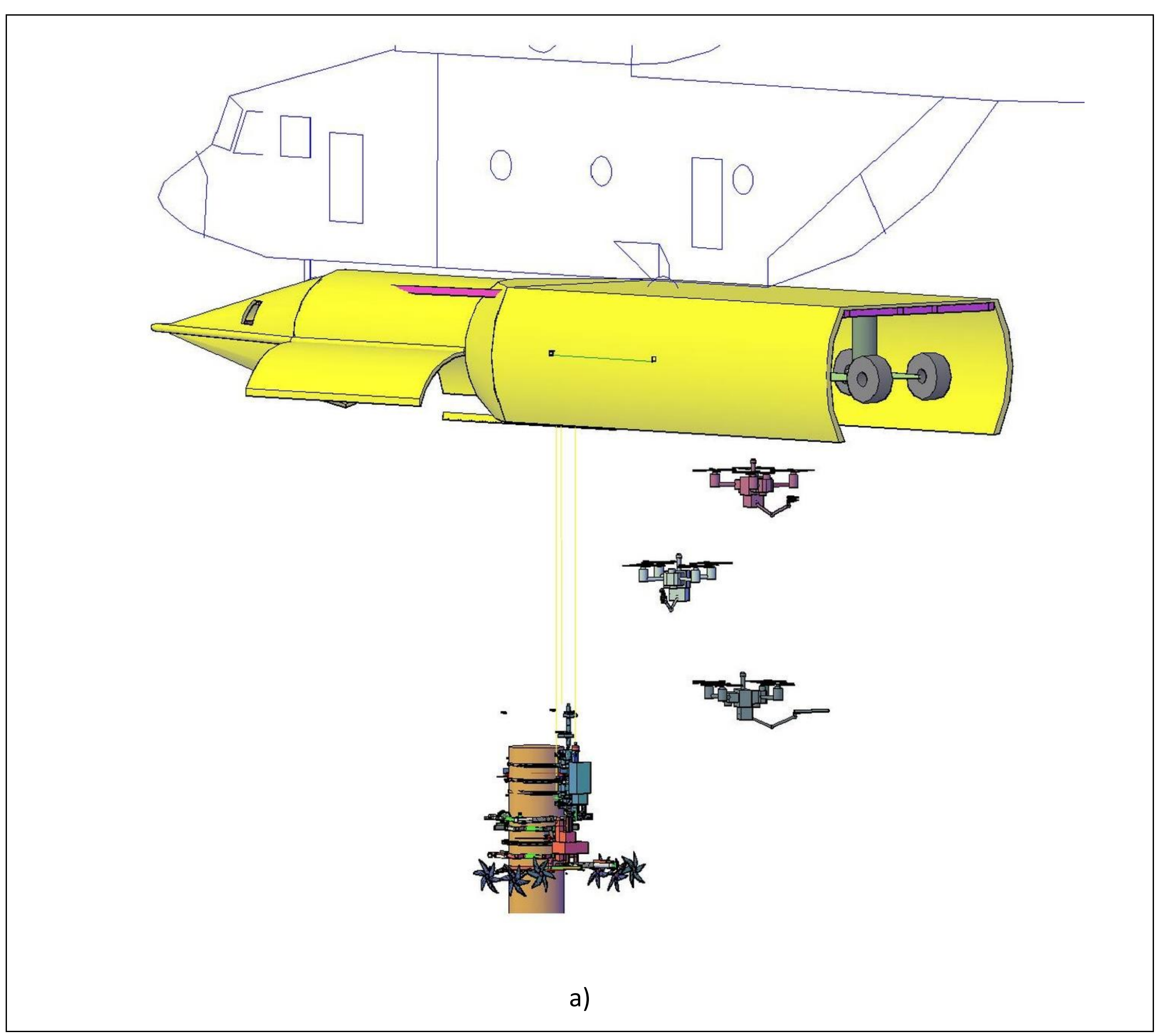

a)

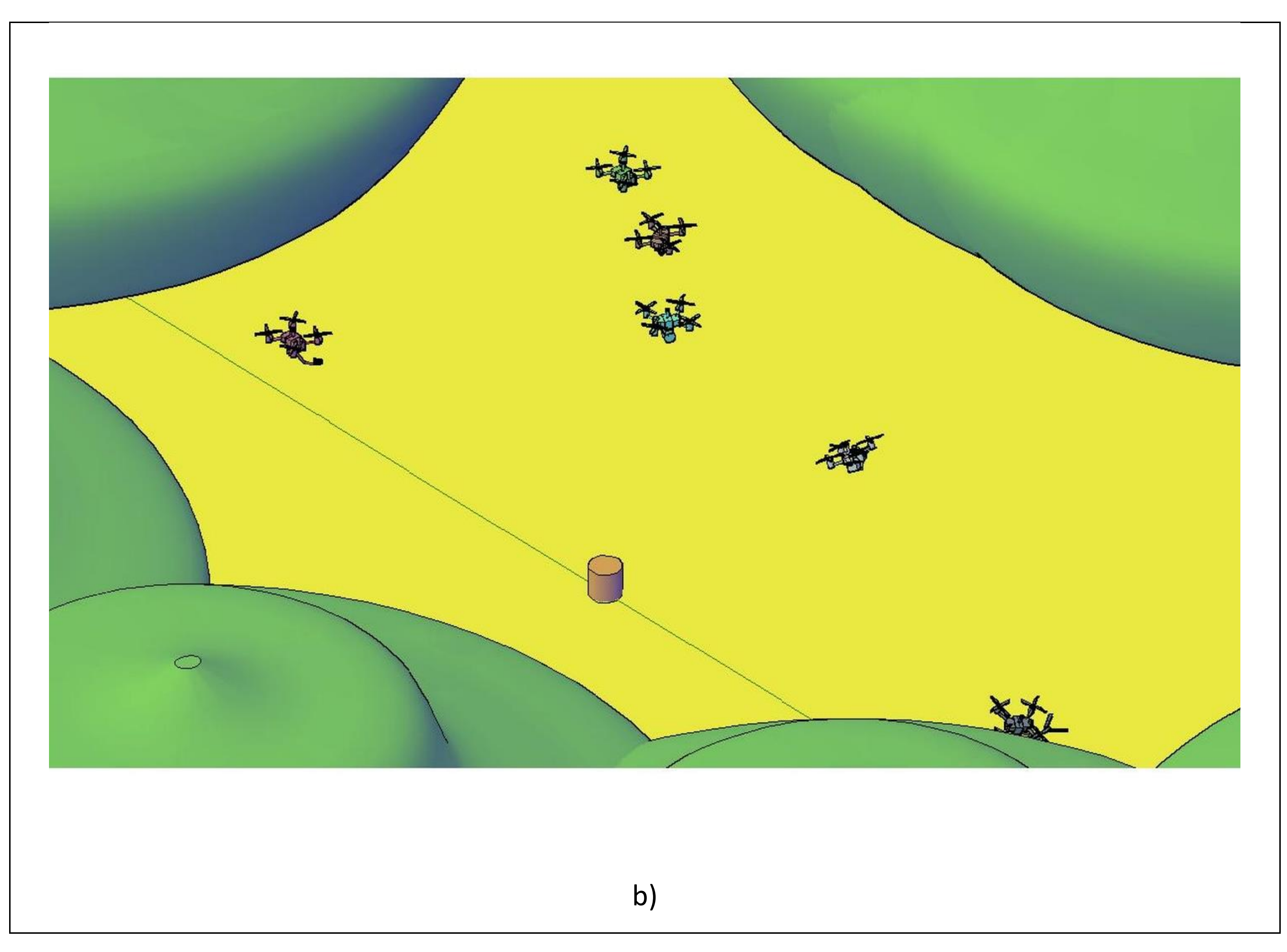

b)

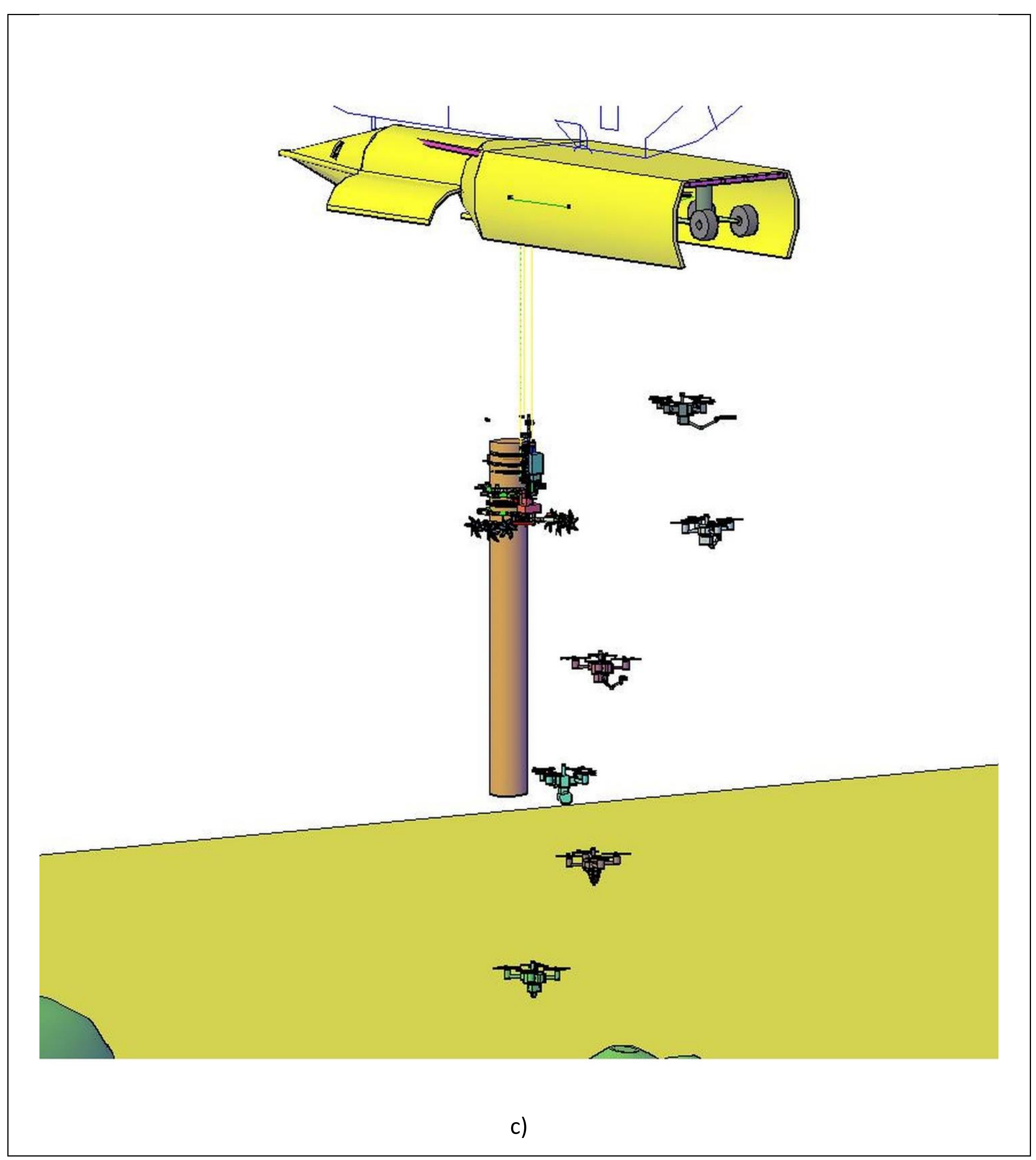

c)

Figure 12. MHST module on-the-go: a) Launch of the drone squadron for area clearing (vine cutting drone, secondary tree cutting drone, and tree girdling drone); b) View from the URIEL Pod of the drone squadrons in operation; c) Perspective view of the drones returning to the URIEL Pod.

Harvesting cycle time (approach, coupling, decoupling, encapsulation, recoupling, stem cutting) and post-harvesting silvicultural treatment cycle time (vine cutting drone, secondary tree cutting drone, tree girdling, digging, planting, irrigation) was digitally simulated and estimated to be 6 minutes and 9 minutes respectively.

### AI implementation

The analysis about AI implementation for operationalizing the URIEL system addresses two active research gaps (Gap B: Multi-Agent Coordination; Gap C: Computer Vision for Tree Species Identification) and one confirmed research void (Gap A), which constitutes one of URIEL's primary novelty claims.

The AI module specifications analyzed in this section correspond to those outlined in Supplementary Tables 16–20, which detail the development costs and proposed architectures for each subsystem. The table below assesses each specified approach against the state of the art identified in the reviewed literature.

Table x. AI modules for URIEL System.

| URIEL Module | AI Specified | Est. Hours | Literature Support | Assessment |
|---|---|---|---|---|
| Hover Stabilization | Reinforcement Learning | 1,168 | No precedent (Gap A void) | Validated — frame as pioneering |
| Delimbing | Deep Learning / YOLO | 13,140 | [83,84] | Appropriate for aerial phase; near-field adaptation required |
| Stem Cutting | Deep Learning / YOLO | 8,240 | [84] — mAP 98.91% | Appropriate for aerial phase; near-field adaptation required |
| 6-Drone Fleet (HSTM) | CNN + RNN | ~58,000 | Partial — no tropical equivalent | Solid choice — novelty claim on heterogeneous fleet |

**Hover Stabilization — Reinforcement Learning.** The RL approach is technically appropriate and unsupported by any existing literature, making it a genuine novelty claim. RL outperforms classical PID controllers in non-stationary environments with wind disturbance and variable payload, consistent with [85]. The DBN component models system-state uncertainty while the RL layer optimizes the control policy — a distinction that should be made explicit in the architecture description to anticipate reviewer questions about the 3T framework.

**Delimbing and Stem Cutting — YOLO.** YOLO-based detection is well-supported for aerial tree identification ( [84,86]: mAP 98.91%;: 93.1%), but both studies operate on open-canopy aerial imagery — a different input domain from close-range robotic operations at sub-metre distance. For near-field branch segmentation required for cutting-plane calculation, a two-stage approach is recommended: YOLO for rapid crown/canopy detection, followed by a lightweight instance segmentation model (e.g., YOLACT or MobileNet-based Mask R-CNN) running on onboard embedded hardware. Full-scale models such as SAM (636M parameters) are not viable on drone-mounted platforms. The design tables should clarify whether YOLO operates on aerial pre-identification imagery or onboard near-field camera feeds.

**Six-Drone Fleet (HSTM) — CNN + RNN.** The CNN + RNN architecture is defensible: CNNs handle spatial perception while RNNs capture temporal flight dynamics. The reviewed corpus contains no equivalent for heterogeneous drone fleets in forested environments — an honest limitation that doubles as a novelty claim. [87] on overlapping coalition formation is the closest reference. A note on whether a behavior tree layer sits above the CNN + RNN for mission-level task switching would strengthen the architecture description

### Gap A: DBN-Based Hover Stabilization

A systematic review of the Scopus database using queries combining 'dynamic bayesian network', 'UAV swarm coordination', and 'marl' with 'forestry', 'logging', and 'aerial robotics' (2022–2026) identified no peer-reviewed contributions addressing DBN-based architectures for hover stabilization of helicopter-suspended harvesting modules, confirming that the 3T AI architecture proposed for URIEL constitutes an original contribution without direct precedents in the scientific literature.

### Gap B — Multi-Agent Drone Coordination

Table xx.Literature on multi-agent UAV coordination relevant to URIEL.

| Authors (year) | Topic | Key Result |
|---|---|---|
| [88] | UAV swarms — monitoring & disease ID | Scalable AI swarm for agricultural monitoring |
| [85] | MARL for UAV fleet control | PPO and MADDPG outperform classical approaches |
| [87] | Overlapping coalition formation | Hypergraph attention improves energy efficiency (+228%) |
| [89] | Cooperative environmental control | Emergent group-level optimization via decentralized RL |

The following concepts are entirely absent from the corpus and each represent a direct novelty claim for URIEL: behavior tree + MARL integration; heterogeneous fleet coordination; consensus-based task allocation for multi-role logging drones; LoRa/low-power UAV communication in multi-agent forestry systems; federated learning for swarm intelligence.

The URIEL design specifies LoRa for inter-drone communication — a choice absent from all 25 reviewed articles, but pragmatically suited to the Amazon's RF environment. An important bandwidth constraint must be noted: LoRa's low bitrate (<50 kbps) precludes real-time video or image sharing between drones, but is adequate for telemetry, task status updates, and coordination commands. Higher-bandwidth tasks (e.g., model updates) would require drone return-to-pod for physical data transfer. This limitation should be acknowledged in the system architecture description.

### Gap C — Computer Vision for Tree Species Identification

Computer vision for tree species classification is the dominant stream of the corpus literature. Hyperspectral/spectral methods are most prevalent, followed by LiDAR/point cloud and YOLO/object detection, top results are summarized below.

Table xx. Computer vision literature for URIEL System.

| Authors (year) | Method | Best Accuracy |
|---|---|---|
| (2023) [90] | SLA-NET — morphological + spatial CNN (HSI) | State-of-the-art HSI classification |
| [83] | YOLO + semantic segmentation | mAP 0.564 — urban street trees |

| [91] | EfficientNet on UAV multispectral (14 classes) | Macro F1 0.61 — temperate forest |
|---|---|---|
| [92] | TASAM — texture-aware self-attention (HSI) | Best-in-class HSI feature extraction |
| [93] | Random Forest + auxiliary data on Sentinel-2 | OA 82.69%, Kappa 0.80 |

**Critical caveat on hyperspectral studies**: [90,94] operate on hyperspectral data (hundreds of spectral bands), whereas URIEL's onboard sensors operate in the RGB–multispectral range (4–5 bands). This distinction should be stated explicitly: classification in URIEL must rely more heavily on spatial and textural features than on spectral signatures, which further justifies the YOLO + segmentation pipeline over HSI-based approaches. The URIEL architecture, which integrates satellite data for pre-mapping and drone data for close-range operations, aligns with emerging 'space-to-tree' monitoring frameworks [95].

**Tropical species — URIEL's strongest Gap C claim:** No article in the corpus addresses species-level identification under Amazonian canopy conditions. All studies use temperate European or North American species (Scots pine, Norway spruce, beech, oak) or East Asian plantation species. The three URIEL target species — *Cedrela fissilis* (cedar), *Handroanthus impetiginosus* (ipê), and *Hymenaea courbaril* (jatobá) — appear in no reviewed dataset. This gap must be stated explicitly in both the Introduction and Discussion, and implies that URIEL will need to construct a proprietary annotated dataset for model training. Given the absence of annotated datasets for the target species, transfer learning from existing temperate-zone datasets (e.g., TreeSatAI, ForInstance) will be necessary, followed by fine-tuning on a newly collected Amazonian corpus. The domain shift between temperate plantations and tropical closed-canopy forests is substantial and should not be underestimated.

**Edge AI — Onboard Inference Feasibility**

All reviewed studies perform inference offline on ground-based servers. None addresses real-time onboard inference on a weight-constrained aerial platform — a critical concern for URIEL's delimbing and cutting drones, where latency between visual detection and mechanical actuation is safety-critical. While real-time onboard inference for drones remains a challenge, the group has previously demonstrated the feasibility of deploying AI-driven alert systems based on smart sensor networks for agricultural applications [96,97], providing a foundational experience for the edge computing requirements of URIEL.

Deployment of deep learning models on resource-constrained UAV platforms requires model compression (structured pruning, INT8 quantization, knowledge distillation). Lightweight variants such as YOLOv8-nano and MobileNet-v3 achieve 30–120 FPS on embedded GPUs (NVIDIA Jetson Orin class) while retaining >90% of full-model accuracy, demonstrating that onboard deep learning inference is operationally feasible within the power envelope of a 6 kg MTOW platform.

**Development costs**

Considering the defined concepts and design assumptions based on the simulated digital proof of concept, it was possible to estimate the costs for developing a prototype of the URIEL system.

These costs were divided into costs related to: structural systems, automation systems, robotic systems, drone-operated systems, metal-mechanical labor; AI programming labor and labor for accessory systems (communications, power, drivetrain, oil-hydraulics, mechanical mechanisms, magnetic mechanisms, electro-electronic mechanisms). The costs for developing a prototype are presented in Table 1. Detailed cost data for each subsystem are presented in the Supplementary Tables.

Table 1. Summary of URIEL System development costs.

| Subsystem | Cost (US$) |
| --- | --- |
| MH Module | 2,791,571.00 |
| MHST Module | 1,394,682.00 |
| URIEL Pod | 932,142.90 |
| Power and Energy | 473,610.70 |

| AI Programming (Labor hours)* | 830,778.60 |
|---|---|
| Development Engineering (Labor hours)* | 310,284.99 |
| **TOTAL** | **6,733,070.00** |

* Based on the value of the postdoctoral fellowship from the São Paulo Research Foundation(FAPESP).

Estimated total weight of the URIEL System (Aluminium alloys 7050 and 7075; Titanium alloys Ti-6Al-4V and Ti-10V-2Fe-3Al, Carbon fiber elements: Hexcel IM7) and **: 2080 kg.** This mass was subtracted from the helicopters' payload.

**Economic viability**

In machine design engineering, it is common practice to consider that the cost of a prototype is 5 to 6 times the value of a production machine (already considering the profit margin of the venture) [66,67,75,98]. Thus, considering the value for the development of a prototype of the URIEL System, a sales price of a system in series production of around US$ 1,000,000.00 can be considered feasible. This value, added to the costs related to the use of helicopters (flight hours, pilots' working hours, URIEL operator's working hours, depreciation, investment value), and total costs of sawmill gives the total cost value.

The revenue value comes from the value related to the sale of wood in the international market. Aiming for the best cost-benefit ratio, the sales value of sawn timber in planks was considered. In this configuration, it was necessary to calculate the cost of this sawn timber, which refers to the investment in a large-scale sawmill, as well as the fixed and variable costs related to the operation of such a sawmill. The investment value to set up a large-scale sawmill is US$ 94,340.00 [99], and the total cost to process one cubic meter of sawwood is US$ 149.00 [100].

By obtaining all these values, it was possible to carry out the economic feasibility analysis for a period of 15 years based on the parameters of Net Present Value (NPV), Internal Rate of Return (IRR) and Payback. In this project, a maximum payback period of 5 years was considered. Then, from these, the Internal Rate of Return (IRR), equation 5, will be calculated and compared with the Minimum Attractive Rate of Return (MARR), considered as the SELIC rate (Special Settlement and Custody System) of the Brazilian Central Bank + 8% [101], since in Brazil in 2025 the SELIC is 10%, so the MARR for this study is 18%.

The predefined values for the economic feasibility calculation are: payload helicopters; investment costs; flight hour costs (including pilot and URIEL operator working hours); price of the timber considered; Cubic measurements of the timber considered in two scenarios (scenario 1: DBH of 0.8 m and 25 m of stem height; scenario 2: DBH of 1 m and 30 meters of stem height). Detailed revenue and cost data are presented in the Supplementary Tables.

The spreadsheet with all the data is available in the REDU-UNICAMP repository (https://doi.org/10.25824/redu/WUCHFJ).

Table 2. Economic viability assessment for CH47 in cedar harvesting under the two scenarios.

| Scenario* | Helicopter Condition | Investment Helicopter+ URIEL+Sawmill (US$) | Revenue 100days/8h 15 years (US$) | Distance to Target (km) | NPV (US$) | IRR** (%) | Payback*** (years) |
|---|---|---|---|---|---|---|---|
| 1 | New | 40,094,340.00 | 7,680,400.00 | 10 | $-37.50 \times 10^6$ | - | - |
| 1 | Used-new | 21,094,340.00 | 7,680,400.00 | 10 | $-18.50 \times 10^6$ | - | - |
| 1 | Used-old | 11,094,340.00 | 7,680,400.00 | 10 | $-8.50 \times 10^6$ | - | - |
| 2 | New | 40,094,340.00 | 13,904,800.00 | 10 | $-5.97 \times 10^6$ | - | - |
| 2 | Used-new | 21,094,340.00 | 13,904,800.00 | 10 | $13.02 \times 10^6$ | 40 | 4 |
| 2 | Used-old | 11,094,340.00 | 13,904,800.00 | 10 | $23.02 \times 10^6$ | 76 | 3 |
| 1 | Used-old | 11,094,340.00 | 7,680,400.00 | 50 | $-24.29 \times 10^6$ | - | - |
| 2 | Used-old | 11,094,340.00 | 8,342,880.00 | 50 | $4.42 \times 10^6$ | 25 | 8 |
| 2 | Used-old | 11,094,340.00 | 5,561,920.00 | 100 | $-9.65 \times 10^6$ | - | - |

*Scenario 1; DBH=0.8 m and 25 m height; Scenario 2; DBH=1 m and 30 m height. **MARR=18%.*** Maximum payback of 5 years.

Table 3. Economic viability assessment for CH47 in Ipê tree harvesting under the two scenarios.

| Scenario* | Helicopter Condition | Investment Helicopter+ URIEL+Sawmill (US$) | Revenue 100days/8h 15 years (US$) | Distance to Target (km) | NPV (US$) | IRR** (%) | Payback*** (years) |
|---|---|---|---|---|---|---|---|
| 1 | New | 40,094,340.00 | 16,212,500.00 | 10 | $5.71 \times 10^6$ | 26 | 5 |
| 1 | Used-new | 21,094,340.00 | 16,212,500.00 | 10 | $24.71 \times 10^6$ | 51 | 3 |
| 1 | Used-old | 11,094,340.00 | 16,212,500.00 | 10 | $34.71 \times 10^6$ | 97 | 2 |
| 2 | New | 40,094,340.00 | 16,212,500.00 | 10 | $5.71 \times 10^6$ | 26 | 5 |
| 2 | Used-new | 21,094,340.00 | 16,212,500.00 | 10 | $24.71 \times 10^6$ | 51 | 3 |
| 2 | Used-old | 11,094,340.00 | 16,212,500.00 | 10 | $34.71 \times 10^6$ | 97 | 2 |
| 1 | Used-old | 11,094,340.00 | 9,727,500.00 | 50 | $1.36 \times 10^6$ | 38 | 5 |
| 1 | Used-old | 11,094,340.00 | 6,485,000.00 | 100 | $-15.03 \times 10^6$ | - | - |
| 2 | Used-old | 11,094,340.00 | 16,212,500.00 | 50 | $1.36 \times 10^6$ | 38 | 5 |
| 2 | Used-old | 11,094,340.00 | 16,212,500.00 | 100 | $-15.03 \times 10^6$ | - | - |

*Scenario 1; DBH=0.8 m and 25 m height; Scenario 2; DBH=1 m and 30 m height. **MARR=18%.*** Maximum payback of 5 years.

Table 4. Economic viability assessment for CH47 in Jatobá harvesting under the two scenarios.

| Scenario* | Helicopter Condition | Investment Helicopter+ URIEL+Sawmill (US$) | Revenue 100days/8h 15 years (US$) | Distance to Target (km) | NPV (US$) | IRR** (%) | Payback*** (years) |
|---|---|---|---|---|---|---|---|
| 1 | New | 40,094,340.00 | 8,987,620.00 | 10 | $-21.87 \times 10^6$ | - | - |
| 1 | Used-new | 21,094,340.00 | 8,987,620.00 | 10 | $-2.87 \times 10^6$ | - | - |
| 1 | Used-old | 11,094,340.00 | 8,987,620.00 | 10 | $7.12 \times 10^6$ | 32 | 3 |
| 2 | New | 40,094,340.00 | 8,987,620.00 | 10 | $-21.87 \times 10^6$ | - | - |
| 2 | Used-new | 21,094,340.00 | 8,987,620.00 | 10 | $-2.87 \times 10^6$ | - | - |
| 2 | Used-old | 11,094,340.00 | 8,987,620.00 | 10 | $7.12 \times 10^6$ | 32 | 3 |
| 1 | Used-old | 11,094,340.00 | 5,392,572.00 | 50 | $-11.36 \times 10^6$ | - | - |
| 1 | Used-old | 11,094,340.00 | 3,595,048.00 | 100 | $-11.36 \times 10^6$ | - | - |
| 2 | Used-old | 11,094,340.00 | 8,987,620.00 | 50 | $-21.87 \times 10^6$ | - | - |
| 2 | Used-old | 11,094,340.00 | 8,987,620.00 | 100 | $-2.7 \times 10^6$ | - | - |

*Scenario 1; DBH=0.8 m and 25 m height; Scenario 2; DBH=1 m and 30 m height. **MARR=18%.*** Maximum payback of 5 years.

Table 5. Economic viability assessment for CH53 in timber harvesting under the two scenarios.

| Scenario* | Helicopter Condition | Investment Helicopter+ URIEL+Sawmill (US$) | Revenue 100days/8h 15 years (US$) | Distance to Target (km) | NPV (US$) | IRR** (%) | Payback*** (years) |
|---|---|---|---|---|---|---|---|
| 1 | Cedro | 88,094,340.00 | 7,680,400.00 | 10 | -190.31 × $10^6$ | - | - |
| 1 | Ipê | 88,094,340.00 | 16,212,500.00 | 10 | -120.64 × $10^6$ | - | - |
| 1 | Jatobá | 88,094,340.00 | 8,987,620.00 | 10 | -168.80 × $10^6$ | - | - |
| 2 | Cedro | 88,094,340.00 | 13,904,800.00 | 10 | -159.19 × $10^6$ | - | - |
| 2 | Ipê | 88,094,340.00 | 21,616,900.00 | 10 | -120.64 × $10^6$ | - | - |
| 2 | Jatobá | 88,094,340.00 | 11,983,760.00 | 10 | -167.80 × $10^6$ | - | - |

*Scenario 1; DBH=0.8 m and 25 m height; Scenario 2; DBH=1 m and 30 m height. **MARR=18%.*** Maximum payback of 5 years.

Table 6. Economic feasibility assessment for MI26 in timber harvesting under the two scenarios.

| Scenario* | Helicopter Condition | Investment Helicopter+ URIEL+Sawmill (US$) | Revenue 100days/8h 15 years (US$) | Distance to Target (km) | NPV (US$) | IRR** (%) | Payback*** (years) |
|---|---|---|---|---|---|---|---|
| 1 | Cedro | 26,094,340.00 | 7,296,000.00 | 10 | -49.20 × $10^6$ | - | - |
| 2 | Cedro | 26,094,340.00 | 7,296,000.00 | 10 | -18.41 × $10^6$ | - | - |
| 1 | Ipê | 26,094,340.00 | 25,670,064.00 | 10 | 47.52 × $10^6$ | 55 | 2 |
| 2 | Ipê | 26,094,340.00 | 25,670,064.00 | 10 | 46.45 × $10^6$ | 55 | 2 |
| 1 | Ipê | 26,094,340.00 | 8,916,875.00 | 50 | -10.73 × $10^6$ | - | - |
| 2 | Ipê | 26,094,340.00 | 14,861,616.00 | 50 | -11.33 × $10^6$ | - | - |
| 1 | Jatobá | 26,094,340.00 | 8,537,099.00 | 10 | -36.62 × $10^6$ | - | - |
| 2 | Jatobá | 26,094,340.00 | 14,230,715.00 | 10 | -13.10 × $10^6$ | - | - |

*Scenario 1; DBH=0.8 m and 25 m height; Scenario 2; DBH=1 m and 30 m height. **MARR=18%.*** Maximum payback of 5 years.

Regarding the chosen woods, considering a harvesting area delimited in the RESEX-TA as a circle with a radius of 10 km, there is a total area of 31,415.92 ha. Considering an average frequency of 4 cedars, 3 ipês, and 3 jatobás per ha, there is an estimate of 125,663 cedars, 94,247 ipês, and 94,247 jatobás. According to [65], the spatial and DBH distribution of trees in a native forest follows a lognormal distribution [102].

As an estimate, a standard deviation of 0.5 was considered for the frequency distribution based on studies by [103], considering that the dimensions of the trees in the scenarios (DBH of 0.8 and 1 m) have a cumulative distribution of about 96% of the sample, it is found that the dimensionless DBH factor (x-axis) in a lognormal distribution is around 2, thus there is a probability (y-axis) of finding trees of these dimensions in a native forest of 12.5%. Therefore, in the first estimate, there are 15,707 cedars, 11,780 ipês and 11,780 jatobás in the delimited area with DBH between 0.8 and 1 m.

Considering the operational envelope of the selected helicopters, up to 20 rides per day are possible. Each ride assumes one tree harvested. To complete the harvest of cedars in a 10 km radius circle, an aircraft would take 785 days; for ipês and jatobás, 589 days each. Therefore, harvesting all three species in this area would take approximately 5 years. Considering that the RESEX-TA has 647,610 ha, there are 20 harvesting modules of 31,415 ha each. If each module requires 5 years for complete harvesting, the return to the first area occurs after approximately 100 years.

## Discussion

Based on the results of the economic feasibility assessment presented in Tables 2, 3, 4, and 6, it is clear that the **hypothesis of this paper has been proven**. Therefore, a URIEL System, considering the current prior art in forest harvesting systems, load leveling in cranes, and specific autonomous drones, together with AI-integrated robotic systems**, is economically viable** in various harvesting scenarios for the selected timbers for various operational envelopes of the chosen helicopters. The innovation of this proposal lies **in combining these already mastered technologies in the same system** through robotic integration operated by AI algorithms for forest harvesting with minimal damage.

It is noteworthy that the Ipê+CH47 combination performed excellently economically in various configurations, even reaching an IRR of 97% for the Ipê + target at 10 km from the base + used old CH47 combination (US$ 10,000,000.00). On the other hand, the CH53 did not show economic viability in any configuration. The MI26 proved viable for the Ipê + target combination 10 km from the base, with a superb IRR of 55%.

On the other hand, in contrast to Ipê, Cedar never achieves economic viability under Scenario 1 in any helicopter configuration. This is a significant negative result, due to its lower timber density and lower market price relative to Ipê and Jatobá.

It is important to consider that there are caveats regarding the operational availability of a used CH47, i.e., perhaps such a machine cannot fly 100 days a year for 8 hours a day. The availability of a Chinook at the end of its service life is around 60% [104], even in this situation an IRR of 39% is obtained, which is quite good. Considering a new CH47 (US$ 39,000,000.00), despite the high investment value, the feasibility analysis proved positive with a very interesting IRR of around 26% in the case of Ipê, which is well above the MARR of 18%.

It is also necessary to consider that in the same harvesting area there may be all three species, in the case of CH47 with targets within a 10 km radius, there are interesting economic viability for several harvesting rides in the same region, with IRR varying from 97% (Ipê, scenarios 1 and 2) to 25% (Cedar, scenario 2), passing through 32% (Jatobá, scenarios 1 and 2).

All these analyses always consider a characteristic MARR of 18% for Brazil in 2025, when considering tropical forests in other countries the results may be even more interesting, for example in Madagascar the MARR is 12% [105], in Indonesia it is 4.5% [106].

It is important to emphasize that the choice of Cedar (Cedrela fissilis), Ipê (Handroanthus impetiginosus) and Jatobá (Hymenaea courbaril) woods was primarily due to their existence in the Tapajós-Arapiuns Extractive Reserve (RESEX-TA), but also to provide a scale for comparison of commercial value in the international market for noble woods, where cedar has a medium commercial value, ipê has a high value, and jatobá has a low commercial value. However, Brazil has many other native timber species of high commercial value, so other RESEXs may have other even more valuable woods, such as Brazilian Mahogany (Swietenia macrophylla), where the average price per cubic meter of sawn timber reaches US$ 3,000.00 [107], thus a URIEL System, depending on its target wood, can have even better results in economic terms.

Considering only the economic aspect, it becomes clear that this proposed new logging method, the URIEL Method operated by the URIEL System, is very interesting, but there is another valuable component in this study: the Ecosystem Services component in the context of Sustainable Forest Management. The URIEL System allows for the minimization of damage to the forest, almost to the point of eliminating damage.

The URIEL Method, by combining RIL techniques with Heli-logging (HL), eliminates damage due to logging. This can extend cutting cycles beyond that recommended by [12], resulting in trees with larger DBH and height, which will increase economic returns. One proposal from the authors of this paper is to **extend the return to the harvest cycle to 100 years**. This is possible considering silvicultural intervention techniques as advocated by [18,24], which with a URIEL System will have their efficiency raised to a level never seen before, as the use of drones for HST operations enables extremely efficient and rapid action.

The action of the squadron of six drones with their specialized terminal tools will enable the excellent results indicated by [25] that guaranteed a 60% increase in the growth of Future Crop Trees (FCTs). This, combined with the innovation of the technique of encapsulating the cut tree canopy in another region close to the origin, but distant enough to allow for opened canopy gaps in the forest that offers optimal microclimatic conditions for the establishment of the seedling [25].

With enrichment through planting together with reduced competition and maintenance of favorable growing conditions for the seedling [26] MHST will enable an unprecedented gain in the reestablishment of these harvested species. [26] advise that silvicultural treatments should not begin long after logging actions, as gaps are quickly closed by surrounding trees, pioneer species and lianas, with the use of URIEL there is no such time gap, as immediately after cutting, the drones go on-the-go.

The proposal of this new URIEL system completely avoids soil impacts from conventional machine skid trails and eliminates residual stand damage by maximizing directional feller and liana cutting actions. Encased optimizes post-harvest silvicultural treatments by encapsulating the canopy in an area close to harvest, enabling post-harvest silvicultural treatments in the open harvest gap. In this context, the use of a dedicated URIEL System solely for floristic restoration of native forests where commercial timber species have become extinct is not considered impossible.

A pertinent question is whether logger operators will adapt to the technologies presented, because although they are prior art in industrial terms, they are disruptive in the forestry context. On this issue, the authors [108] state that there is great interest from loggers in robotics applied to forestry operations, as there is currently a lack of personnel willing to work in operations, in addition to increased costs due to regulatory compliance.

In this paper, a discussion regarding the **Political Economy** in which the new URIEL Method is embedded is inevitable, since the interactions between the potential producing industry of the URIEL System, the timber companies that are clients of this industry, and the native inhabitants of the RESEX are essentially the basis on which the concrete effectiveness of this new logging method rests, which strongly requires the background of national political governments.

There is a guiding structural issue here that refers to the hegemony of capitalism in the face of the emergence of environmentalism, since the contacts between both ideologies generate tensions even considering that technically and economically the method is a good bet for preserving tropical forests. According to [109], the political system and the economic environment mutually influence each other, including how profit and economic development affect natural resources. In this environmental-political-economic-social context, it is understood that for the URIEL Method to become established, a synergy of actors is needed, which unfortunately is not trivial.

First, it is necessary for some high-tech manufacturing industry (for example, the aeronautical industry) to be interested in conducting the development of the URIEL System, which, although the vast majority of its subsystems can be considered prior art (proven technologies made commercially available to the public), has a non-negligible engineering development component, which will require many hours of work.

Secondly, in a project of this type, there is a risk of technological innovation that an industry is unlikely to undertake without the guarantee of a certain consumer market. Therefore, it is necessary to have non-reimbursable funding resources from some government or funding institution so that the management of the manufacturing industry can convince its investors of the possibility of a foreseen consumer market. Even with these resources, acceptance is not guaranteed because the strategic planning of this hypothetical industry will certainly not consider the URIEL System as its core business; therefore, there must be at least a managerial, perhaps even institutional, rearrangement.

The third actor in this equation is the timber companies; these companies generally do not have high internal rates of return (IRR) (at least not the legally registered and certified timber companies) [110,111], and although the economic analysis is very interesting, such companies will certainly have difficulty accessing large, high-tech helicopters like those indicated in this paper; this will require political will to open lines of credit for such companies. This is without considering the enormous challenge of hiring highly qualified professionals in helicopter piloting, robotics, AI, and complex aircraft maintenance.

And lastly, and obviously most importantly, there are the native inhabitants of the RESEX (Extractive Reserve). Of all the actors, they are the most vulnerable, the least served by governments, those with the least bargaining power, and those ignored by banks and developers. In short, the true owners of the valuable timber existing in their RESEX. Assuming, in a very optimistic scenario, that the other three actors manage to integrate and operationalize the implementation of a URIEL Method, it is imperative that, first and foremost, at the negotiating table with the native inhabitants, their wishes, their desires, their culture, and above all, that they are fairly compensated with the dividends derived from their timber.

In a preliminary analysis lacking the necessary depth, it appears that the URIEL Method will foster an **ontological conflict among all actors** involved in two opposing and extremist directions: a capitalist direction with a purely economic bias and an environmentalist direction with a purely conservationist bias. Within what is discussed [112], a false opposition is created between conventional conservation and concern for humanity, when the real conflict lies in the modern organization of society as a capital accumulation machine that internalizes benefits and displaces socio-environmental costs.

This debate is by no means simple and is beyond the scope of this paper, but it is necessary for a good accommodation of the diverse interests of the possible stakeholders using the URIEL Method. As a possible reconciliation between all those involved (both internally and externally), a middle way could be suggested. It is clear that in today's world, if there is no profit in a venture, it is doomed to failure; the only variable is the time it takes for bankruptcy to occur. On the other hand, the indiscriminate destruction of tropical forests for the sake of immediate and unwise profit is also another path to ruin, especially when considering the increasingly intense effects of climate change [1,113–115].

Ultimately, it is a matter of governance; there needs to be a sharing of decision-making power. On the one hand, respect for the desires, needs, and possibilities of native inhabitants; on the other, the essential need for profit for the companies involved. Would it be utopian to seek a just (fair) price [116]?

As a final word in this discussion, it is necessary to write about the context of illegal logging, which is omnipresent in tropical forests. It is not considered dangerous that a URIEL System could increase the piracy of valuable timber in the world's tropical forests, because the operation of the Uriel Pod is only compatible with the operation of helicopters of the size described; such aircraft require ground support and maintenance support only available at medium-sized airports.

In these specific aviation environments, characterized by state and federal oversight, the presence of heavy-lift aircraft—such as the Mi-26 Letayushchaya Korova, the CH-53K King Stallion, or the CH-47 Chinook—inherently invites scrutiny. Given the immense operational costs and logistical complexities associated with such 'aerial titans,' any administrative justification that fails to align with the significant time and capital investment required for their deployment would likely trigger suspicion from even a baseline regulatory observer.

Even if a Uriel Pod is hidden in some clearing in the forest, it will need to be transported by river to the hideout, considering that it has the appearance and size of an ICBM (InterContinental Ballistic Missile), it is deduced that it would hardly go unnoticed, even if it were covered up. And even if such a feat occurs, assembling the robotic subsystems of a URIEL in the middle of the jungle is an undertaking completely beyond the capabilities of any criminal logging group.

## Conclusion

The proposed new URIEL method, activated by prior-art technologies in forest harvesting and crane load leveling, integrated with prior-art robotics and AI technologies, presented very interesting economic viability in various helicopter+timber+scenario configurations, proving the hypothesis of this paper.

The use of the URIEL System will make it possible to optimize the extraction of noble woods from tropical forests, minimizing damage due to logging actions. The use of intensive and optimized post-harvest silvicultural treatment by autonomous drones may enable the recovery of the harvested timber population, with a real possibility of increasing the abundance of these trees in the forest.

The biggest challenge regarding the URIEL method is the integration of user stakeholders; this integration is essential for the realization of the ecosystem potential presented in this paper. A possibility of success is envisioned if there are shared governance actions among all parties.

There is hope that this new logging method, winged like the angel of vegetation, will protect Earth's paradisiacal tropical forests with the flaming sword of robotics, enabling the sustainable and intelligent use of their resources, preserving them and their beautiful birds for future generations.

## Data Availability

The economic data and spreadsheet are available in the repository. UNICAMP-REDU: Albiero, Daniel, 2026, "Spreadsheet of data and calculations for URIEL system feasibility study", https://doi.org/10.25824/redu/WUCHFJ, Repositório de Dados de Pesquisa da Unicamp, V1, UNF:6:yPRlggp

## Acknowledgements

The first author of this paper thanks the National Council for Scientific and Technological Development (CNPq) and Coordination for the Improvement of Higher Education Personnel (CAPES) for financial support. The authors of this paper thank FEAGRI-UNICAMP for the infrastructure provided for carrying out this work. The first author of this paper thanks Boeing Inc. for permission to use the image of the CH-47, especially to Ms. Heather Anderson for her mediation.

## Author contributions

D.A: Conceptualization, Methodology, Formal analysis, Validation, Economics analysis, Drawing, Resources, Writing— original draft, Supervision, Funding acquisition, Project administration; Supervision; G.F.M: Study of tree species; D.H.: RESEX-TA study; F.R.F.G: Software, Data curation; A.V.A.S.: Software, Robotic Systems; W.L.A: Structural calculus, Simulations; A.M.F: AI studies; C.K.U.: Data curation, Visualization; M.P.: Forestry studies; F.T.: Validation AI; A.I.R: Ecosystem services; A.J.S.F: Power electronics; A.F.D.N: Data curation, Formal analysis; A.P.G.: Validation, Writing—review & editing, Project administration.

## Funding

This research received external funding from the National Council for Scientific Research (CNPq-Brazil), process 303331/2021-6.

## Competing interests

The authors declare that there is no any conflict of interest.

## SupplCompeting interests

## Supplementary Materials

Supplementary materials relating to all economic calculation tables (SupplementaryTables) and all figures detailing the systems (SupplementaryFigures) are available in the online version (contains supplementary material available at on-line in Scientific Report Supplementary Materials.

## Supplementary Tables

Considering the defined concepts and design premises based on the simulated digital proof of concept, it was possible to estimate the costs for developing a prototype of the URIEL system.

These costs were divided into costs related to: structural systems, automation systems, robotic systems, drone-operated systems, metal-mechanical labor; AI programming labor and labor for accessory systems (communications, power, drivetrain, oil-hydraulics, mechanical mechanisms, magnetic mechanisms, electro-electronic mechanisms). The costs for developing a prototype are presented below.

Table 6. Structural costs of the stabilization subsystem.

| Item | Componente | Estimated mass (kg) | Estimated cost (US$) |
|---|---|---|---|
| 1 | Rotation pivot | 1.8 | 180.00 |
| 3 | Support rods (4 units) | 0.3 | 30.00 |
| 4 | Structural frame | 9 | 900.00 |
| 5 | Longitudinal and transverse rails | 3 | 300.00 |
| 6 | Overhead crane | 5 | 500.00 |
| 7 | Wheel set (3 systems) | 4 | 400.00 |
| 9 | Stabilizing pulleys (3 units) | 6 | 600.00 |
| 10 | Stabilizing cables (3 units) | 100 | 178.57 |
|  | **Total** | **129.1** | **3,088.57** |

Obs.: Aluminium alloys 7050 and 7075; Titanium alloys Ti-6Al-4V and Ti-10V-2Fe-3Al, Carbon fiber elements: Hexcel IM7.

Table 7. Structural costs of the decoupling subset.

| Item | Component | Estimated Mass (kg) | Cost (US$) |
|---|---|---|---|
| 1 | Hydraulic rotator | 18 | 1,800.00 |
| 2 | Gripper support chassis | 6 | 600.00 |
| 3 | Battery box and electro-electronic systems | 1 | 100.00 |
| 4 | Hydraulic pump box and accessories | 6 | 600.00 |
| 5 | Coupling mast chassis | 4 | 400.00 |
| 7 | Coupling mast | 2 | 200.00 |
| 8 | Right beam of the main rigid chassis | 6.5 | 650.00 |
| 9 | Left beam of the main rigid chassis | 6.5 | 650.00 |
| 10 | Left pulley – upper handle clamping cable | 4 | 400.00 |
| 11 | Upper handle cable guide | 1.2 | 120.00 |
| 12 | Left pulley – lower handle clamping cable | 4 | 400.00 |
| 13 | Lower handle cable guide | 1.2 | 120.00 |
| 14 | Right pulley – lower hook clamping cable | 4 | 400.00 |
| 15 | Right pulley – upper hook clamping cable | 4 | 400.00 |
| 16 | Main magnetic cannon | 2 | 200.00 |
| 17 | Magnetic cannon support flange | 1 | 100.00 |
| 18 | Magnetic cannon cable pulley | 2 | 200.00 |
| 19 | Motor/gearbox box – upper cables | 3 | 300.00 |
| 20 | Motor/gearbox box – lower cables | 3 | 300.00 |
|  | **Total** | **79.4** | **7,940.00** |

Obs.: Aluminium alloys 7050 and 7075; Titanium alloys Ti-6Al-4V and Ti-10V-2Fe-3Al, Carbon fiber elements: Hexcel IM7.

Table 8. Structural costs of the claw subassembly.

| Item | Componente | Massa estimada (kg) | Cost (US$) |
|---|---|---|---|
| 1 | Clamping cable coupling handle | 0.3 | 30.00 |
| 2 | Upper claw arm | 4 | 400.00 |
| 3 | Clamping cable coupling hook | 0.4 | 40.00 |
| 4 | Pivoting claw chassis | 8 | 800.00 |
| 6 | Claw pivot bearing | 3 | 300.00 |
| 7 | Chainsaw bar and chain (upper) | 10 | 133.92 |
| 9 | Lower cable coupling handle | 2 | 200.00 |

| 10 | Rail-travel trolley system | 6 | 600.00 |
|---|---|---|---|
| 11 | Lower claw arm | 6.5 | 650.00 |
| 12 | Hook transport trolley | 6 | 600.00 |
| 13 | Handle wire-rope trunnion | 1 | 100.00 |
| 14 | Handle trolley chassis | 2 | 200.00 |
| 15 | Trolley wheels with in-hub motors (handle) | 6 | 600.00 |
| 17 | Hook wire-rope trunnion | 1.5 | 150.00 |
| 19 | Hook trolley chassis | 6 | 600.00 |
| 20 | Trolley wheels with in-hub motors (hook) | 6 | 600.00 |
| 22 | Chainsaw transport trolley | 5.5 | 550.00 |
| 23 | Machined trolley rail on intermediate claw | 10 | 1,000.00 |
| 24 | Chainsaw bar (intermediate) | 10 | 133.92 |
|  | **Total** | **94.2** | **7,687.85** |

Obs.: Aluminium alloys 7050 and 7075; Titanium alloys Ti-6Al-4V and Ti-10V-2Fe-3Al, Carbon fiber elements: Hexcel IM7

Table 9. Structural costs of the load alignment subset.

| Item | Component | Estimated Mass (kg) | Estimated Cost (R$) |
|---|---|---|---|
| 9 | Rigid chain/tube locking system | 4 | 400.00 |
| 10 | Coupling handle of the load/alignment subassembly | 3 | 300.00 |
| 11 | Aligning steel cable pulley | 2 | 200.00 |
| 12 | Housing for electric motor, gearbox, and drive system | 15 | 8,928.57 |
| 13 | Aligning steel cable | 2 | 200.00 |
| 14 | Upper supporting carriage | 5 | 500.00 |
| 15 | Chain gear and locking system | 4 | 1,785.71 |
|  | **Total** | **35** | **12,314.28** |

Obs.: Aluminium alloys 7050 and 7075; Titanium alloys Ti-6Al-4V and Ti-10V-2Fe-3Al, Carbon fiber elements: Hexcel IM7.

Table 10. Structural costs of the shaft traverse subset.

| Item | Component | Estimated Mass (kg) | Cost (US$) |
|---|---|---|---|
| 3 | Widia sprocket | 6 | 892.85 |

| Item | Component | Estimated Mass (kg) | Cost (US$) |
|---|---|---|---|
| 6 | Main chassis of the subassembly | 8 | 800.00 |
| 7 | Torque support bracket | 5 | 500.00 |
| 8 | Coupling mast with the decapper system | 4 | 400.00 |
| 9 | Electro-electronic systems housing | 2 | 200.00 |
| 10 | Battery and accessories housing | 2 | 200.00 |
| 11 | Tenaz clamp | 2 | 200.00 |
| | **Total** | **29** | **3,192.85** |

Obs.: Aluminium alloys 7050 and 7075; Titanium alloys Ti-6Al-4V and Ti-10V-2Fe-3Al, Carbon fiber elements: Hexcel IM7.

Table 11. Structural costs of the delimbing subset.

| Item | Component | Function (units) | Cost (US$) |
|---|---|---|---|
| 1 | Pressing claw of the delimbing system | 5 | 500.00 |
| 3 | Delimbing blades | 6 | 600.00 |
| 8 | Torque bracket for linear actuator | 0.5 | 50.00 |
| 9 | Right bar of the main chassis | 6 | 600.00 |
| 10 | Left bar of the main chassis | 6 | 600.00 |
| 11 | Structural supports for linear actuators | 2 | 200.00 |
| | **Total** | **25.5** | **2,550.00** |

Obs.: Aluminium alloys 7050 and 7075; Titanium alloys Ti-6Al-4V and Ti-10V-2Fe-3Al, Carbon fiber elements: Hexcel IM7.

Table 12. Structural costs of the shaft cutting subassembly.

| Item | Component | Estimated Mass (kg) | Cost (US$) |
|---|---|---|---|
| 1 | Chainsaw bar | 1.5 | 150.00 |
| 3 | Chainsaw transport trolley | 5 | 500.00 |
| 4 | Trolley wheels with electric hub motors | 4 | 400.00 |
| 5 | Trolley locomotion rail | 3.5 | 350.00 |
| 7 | Chain tensioning system | 1 | 100.00 |
| | **Total** | **15** | **1,500.00** |

Obs.: Aluminium alloys 7050 and 7075; Titanium alloys Ti-6Al-4V and Ti-10V-2Fe-3Al, Carbon fiber elements: Hexcel IM7.

Table 13. Structural costs of the MHST subset.

| Item | Component | Estimated Mass (kg) | Cost (US$) |
|---|---|---|---|
| 1 | Rotation pivot | 1.8 | 180.00 |
| 3 | Support stringers (6 units) | 0.3 | 300.00 |
| 4 | Structural frame | 9 | 900.00 |
| 5 | Overhead crane | 5 | 500.00 |
| 6 | Wheel set (6 systems) | 4 | 800.00 |
| 7 | Stabilizing pulleys (12 units) | 6 | 595.71 |
| 8 | Hydraulic rotator | 1 | 1,928.57 |
| | **Total** | **25** | **3,088.57** |

Obs.: Aluminium alloys 7050 and 7075; Titanium alloys Ti-6Al-4V and Ti-10V-2Fe-3Al, Carbon fiber elements: Hexcel IM7.

Table 14. Structural costs of the Uriel Pod subset.

| Item | Component | Estimated Mass (kg) | Cost (US$) |
|---|---|---|---|
| 1 | Structural longitudinal-frame chassis | 120 | 12,000.00 |
| 2 | Structural crossmembers of the chassis | 45 | 4,500.00 |
| 3 | Aerodynamic fuselage sheet metal – MH section | 95 | 9,500.00 |
| 4 | Aerodynamic fuselage sheet metal – MHST section | 70 | 7,000.00 |
| 5 | Movable ventral doors of the MH section | 35 | 3,500.00 |
| 6 | Door sealing system | 8 | 800.00 |
| 7 | URIEL System control station | 40 | 5,357.14 |
| 9 | Front pylons for helicopter coupling | 25 | 2,500.00 |
| 10 | Intermediate coupling pylons | 20 | 2,000.00 |
| 11 | Rear coupling pylons | 20 | 2,000.00 |
| 12 | Front tandem landing gear assembly | 30 | 8,928.57 |
| 13 | Rear tandem landing gear assembly | 30 | 8,928.57 |
| 14 | Wheels (x8) | 400 | 17,857.14 |
| 15 | Tandem steering system | 35 | 2,678.57 |
| | **Total** | **973** | **87,550.00** |

Obs.: Aluminium alloys 7050 and 7075; Titanium alloys Ti-6Al-4V and Ti-10V-2Fe-3Al, Carbon fiber elements: Hexcel IM7.

Estimated total structural weight of the URIEL System: 2080 kg

Table 15. Hardware development costs for the stabilization subsystem.

| Item | Description | Quantity | Cost (US$) |
|---|---|---|---|
| 1 | Servo Driver DC 72V | 6 | 21,428.57 |
| 2 | 1:40 Gearbox | 6 | 5,357.14 |
| 3 | Limit Switch with Roller | 12 | 750.00 |
| 4 | DC 72V Servo Driver for overhead crane with locking | 4 | 71,428.57 |
| 5 | Torsion-resistant steel mechanical structure | 1 | 76,785.71 |
| 7 | Inertial Sensor | 1 | 8,928.57 |
| 5 | Digital Controller | 1 | 8,928.57 |
| 6 | Industrial Planning Computer | 1 | 12,500.00 |
| 7 | Power electronics + CAN (Inertial Sensor, Limit Switch with Roller, Inductive Sensor, GPS Module, Lidar Sensor) | 1 | 75,000.00 |
| 8 | Engineering development man-hours | 3,840 | 54,857.14 |
| **Total** | | **327,035.71** | |

Table 16. Hardware development costs for the decoupling subsystem.

| Item | Description | Quantity | Cost (US$) |
|---|---|---|---|
| 1 | 72V 400Ah Lithium Battery | 1 | 5,357.14 |
| 2 | Hydraulic System (Motor + Pump Assembly) | 1 | 17,857.14 |
| 3 | Hydraulic Motor for Rotator | 1 | 1,785.71 |
| 4 | Power Inverter | 1 | 3,571.43 |
| 5 | Digital Controller | 1 | 8,928.57 |
| 6 | Industrial Planning Computer | 1 | 12,500.00 |
| 7 | Computer Vision System | 1 | 17,857.14 |
| 8 | Inertial Sensor | 1 | 8,928.57 |
| 9 | GPS Module | 1 | 5,357.14 |
| 10 | Lidar Sensor | 1 | 16,071.43 |

| Item | Description | Quantity | Cost (US$) |
|---|---|---|---|
| 11 | Inductive Sensor | 8 | 428.57 |
| 12 | High-Torque Linear Actuator | 6 | 75,000.00 |
| 13 | Capacitive Sensor | 2 | 107.14 |
| 14 | High-Torque, High-Speed Electric Motor | 2 | 7,142.86 |
| 15 | Power Electronics + CAN (Inertial Sensor, Roller Limit Switch Sensor, Inductive Sensor, GPS Module, Lidar Sensor) | 1 | 75,000.00 |
| 16 | Development Engineering Work-Hours | 3840 | 54,857.14 |
| **Total** | | **310,750.00** | |

Table 17. Hardware development costs for the shaft cutting subsystem.

| Item | Description | Quantity | Cost (US$) |
|---|---|---|---|
| 1 | 72V 400Ah Lithium Battery | 1 | 5,357.00 |
| 2 | Hydraulic System (Motor + Pump Assembly) | 1 | 17,857.00 |
| 3 | Hydraulic Motor for Rotator | 1 | 1,786.00 |
| 4 | Power Inverter | 1 | 3,571.00 |
| 5 | Digital Controller | 1 | 8,929.00 |
| 6 | Industrial Planning Computer | 1 | 12,500.00 |
| 7 | Computer Vision System | 1 | 17,857.00 |
| 8 | Inertial Sensor | 1 | 8,929.00 |
| 9 | GPS Module | 1 | 5,357.00 |
| 10 | LoRa Communication Module | 1 | 179.00 |
| 10 | Lidar Sensor | 1 | 16,071.00 |
| 11 | Inductive Sensor | 8 | 429.00 |
| 12 | High-Torque Linear Actuator | 4 | 50,000.00 |
| 13 | Linear Actuator | 16 | 28,571.00 |
| 14 | Linear Actuator | 12 | 10,714.00 |
| 15 | High-Torque, High-Speed Electric Motor | 2 | 14,286.00 |
| 16 | Medium-Torque, Medium-Speed Electric Motor | 6 | 21,429.00 |
| 17 | Electric Motor Speed Controller | 8 | 14,286.00 |
| 18 | Power Electronics + CAN (Inertial Sensor, Roller Limit Switch, Inductive Sensor, GPS Module, Lidar Sensor) | 1 | 75,000.00 |
| 19 | Development Engineering Work-Hours | 1,920 | 27,428.57 |
| **Total** | | **340,535.57** | |

Table 18. Hardware development costs for the MHST Module.

| Item | Description | Quantity | Cost (US$) |
|---|---|---|---|
| 1 | DJI Agras 70P Planting Drone | 1 | 53,571.42 |
| 2 | DJI Agras 70P Hole-Digging Drone | 1 | 53,571.42 |
| 3 | DJI Agras 70P Vine-Cutting Drone | 1 | 53,571.42 |
| 4 | DJI Agras 70P Girdling Drone | 1 | 53,571.42 |
| 5 | DJI Agras 70P Secondary Tree-Cutting Drone | 1 | 53,571.42 |
| 6 | DJI Agras 70P Irrigation Drone | 1 | 53,571.42 |
| 7 | 9-DOF Robotic Arm | 6 | 21,428.71 |
| 8 | Power Electronics + CAN (Inertial Sensor, Roller Limit Switch Sensor, Inductive Sensor, GPS Module, Lidar Sensor) | 1 | 75,000.00 |
| 9 | Development Engineering Work-Hours | 3,840 | 54,857.14 |
| **Total** | | **665,571.42** | |

Table 19. Structural and hardware development costs of the URIEL Pod.

| Item | Description | Cost (US$) |
|---|---|---|
| 1 | Aeronautical structure | 285,714.30 |
| 2 | Aerodynamic fuselage | 164,285.70 |
| 3 | Movable doors | 75,000.00 |
| 4 | Motorized landing gear | 196,428.60 |
| 5 | Control station | 135,714.30 |
| 6 | Power electronics + CAN (Inertial Sensor, Limit Switch with Roller, Inductive Sensor, GPS Module, Lidar Sensor) | 75,000.00 |
| 7 | Engineering development man-hours | 18,285.66 |
| | **Total** | **950,428.56** |

Table 20. AI Programming Development Costs.

| Module | Subsystem | Function | Objective of Applicable AI | Types of Applicable AI | Database | Type of Annotation (Database) | Work | Dataset Acquisition (hours) | Annotations (hours) | Model Programming and Training (hours) | Estimated Time (hours) | Cost (US$) |
|---|---|---|---|---|---|---|---|---|---|---|---|---|
| HM | 1 | Stabilization | Self-regulation of balance | Reinforcement Learning | Stabilization data | – | Model creation, training hours (reinforcement learning) and safety tests (simulations and refined learning in the module itself) | – | – | 1,168 | 1,168 | 16,685.71 |
| | 2 | Delimbing | Identification of the end of the crown (cutting point) and identification of the presence of cuttable branches (distinguishing them from non-cuttable ones by the module) | Deep Learning model for computer vision (e.g., YOLO) | Images – Trees and crowns, preferably of target species, from several angles and positions (preferably similar to what the drone will see) | Location of the end of the crown / start of the trunk | Creation of the database, annotation + programming of the classification model and training hours + programming of the flight autonomy model + training hours | 1,971 | 7,884 | 3,285 | 13,140 | 187,714.30 |
| | 3 | Stem cutting | Identification of the ideal place for the lower cut of the trunk | Deep Learning model for computer vision (e.g., YOLO) | Images – Trees, preferably native, from various angles and distances | Cutting location at the base of the trunk | Creation of the database, annotation + programming of the model and training hours + | 1,071 | 5,884 | 1,285 | 8,240 | 117,714.30 |

| Module | Subsystem | Function | Objective of Applicable AI | Types of Applicable AI | Database | Type of Annotation (Database) | Work | Dataset Acquisition (hours) | Annotations (hours) | Model Programming and Training (hours) | Estimated Time (hours) | Cost (US$) |
|---|---|---|---|---|---|---|---|---|---|---|---|---|
| | | | | | (preferably similar to what the drone will see) | | programming of the flight autonomy model + training hours | | | | | |
| HSTM | 1 | Planting drone | 1- Flight autonomy: identification of path and obstacles during flight. 2- Task execution: identification of planting site and route creation to the site. | 1 – CNN and RNN. 2 – CNN for identification of the holes made by the hole-drilling drone (HSTM – 2 subsystem) and RNN to draw the route | Images – Various forest terrains, dirty and clean, with holes dug similar to those made by the hole-drilling drone | Holes dug by the hole-drilling drone | Creation of the database, annotation + programming of the model and training hours + programming of the flight autonomy model + training hours | 971 | 884 | 1,285 | 3,140 | 44,857.14 |
| | 2 | Hole-drilling drone | 1- Flight autonomy: identification of path and obstacles during flight. 2- Task execution: identification of hole-drilling sites and creation of the route to the site. | 1 – CNN and RNN. 2 – CNN for identification of hole-drilling locations and RNN to draw the route and select the best distributions | Images – Various forest terrains, dirty and clean, with indications of possible hole-drilling sites | Possible hole-drilling areas on the soil | Creation of the database, annotation + programming of the model and training hours + programming of the flight autonomy model + training hours | 971 | 884 | 1,285 | 3,140 | 44,857.14 |

| Module | Subsystem | Function | Objective of Applicable AI | Types of Applicable AI | Database | Type of Annotation (Database) | Work | Dataset Acquisition (hours) | Annotations (hours) | Model Programming and Training (hours) | Estimated Time (hours) | Cost (US$) |
| --- | --- | --- | --- | --- | --- | --- | --- | --- | --- | --- | --- | --- |
| | 3 | Liana-cutting drone | 1- Flight autonomy: identification of path and obstacles during flight. 2- Task execution: Identification of lianas, classification of cuttability and identification of cutting point | 1 – CNN and RNN. 2 – CNN for identification of lianas, classification (cuttable or not depending on how much they are in the way) and identification of the cutting point, and RNN to draw the routes from the clearing to each cutting point | Images – Lianas of several species (mainly local), and various sizes and angles with indication of cutting points | Identification of lianas and indication of cutting points | Creation of the database, annotation + programming of the model and training hours + programming of the flight autonomy model + training hours | 1,628 | 8,512 | 2,380 | 12,523 | 178,900.00 |
| | 4 | Girdling drone | 1- Flight autonomy: identification of path and obstacles during flight. 2- Task execution: Identification of trees suitable for girdling and identification of the girdling | 1 – CNN and RNN. 2 – CNN for identification of trees and the girdling location and RNN to draw the routes through the clearing and to each target | Images – Trees, preferably native, from various angles and distances | Creation of the annotated database (various local species with indication of girdling location) | Creation of the database, annotation + programming of the model and training hours + programming of the flight autonomy model + training hours | 314 | 1,256 | 590 | 2,160 | 30,857.14 |

| Module | Subsystem | Function | Objective of Applicable AI | Types of Applicable AI | Database | Type of Annotation (Database) | Work | Dataset Acquisition (hours) | Annotations (hours) | Model Programming and Training (hours) | Estimated Time (hours) | Cost (US$) |
|---|---|---|---|---|---|---|---|---|---|---|---|---|
| | | | point on the tree | | | | | | | | | |
| | 5 | Secondary-tree-cutting drone | 1- Flight autonomy: identification of path and obstacles during flight. 2- Task execution: classification of surrounding trees regarding the need for cutting and identification of the cutting point on the tree and route creation to the site. | 1 – CNN and RNN. 2 – CNN for identification of surrounding trees and classification regarding the need for cutting, and RNN to draw the routes through the clearing to the target | Images – Trees, preferably native, from various angles and distances, with and without overlapping lianas | Identification of trees and classification of the level of need for cutting | Creation of the database, annotation + programming of the model and training hours + programming of the flight autonomy model + training hours | 1,628 | 8,512 | 2,380 | 12,523 | 178,900.00 |
| | 6 | Irrigation drone | 1- Flight autonomy: identification of path and obstacles during flight. 2- Task execution: Identification of sites requiring irrigation | 1 – CNN and RNN. 2 – CNN for identification and RNN to draw the route | Images – Various forest terrains, dirty and clean | Identifications of planting sites | Creation of the database, annotation + programming of the model and training hours + programming of the flight autonomy model + training hours | 371 | 884 | 528 | 1,783 | 25,471.43 |

| Module | Subsystem | Function | Objective of Applicable AI | Types of Applicable AI | Database | Type of Annotation (Database) | Work | Dataset Acquisition (hours) | Annotations (hours) | Model Programming and Training (hours) | Estimated Time (hours) | Cost (US$) |
|---|---|---|---|---|---|---|---|---|---|---|---|---|
| | | | (receiving coordinates from the planting drone) with route creation to the site and identification of the planting location | | | | | | | | | |
| | | | | | | | **Total** | | | | | **830,778.60** |

Note: Considering the hourly rate for the FAPESP postdoctoral fellowship: US$ 2,285.71 per month.

**Economic Feasibility**

In machine design engineering, it is common practice to consider that the cost of a prototype is 5 to 6 times the value of a production machine (already considering the profit margin of the venture) [127,129,142,143].

Thus, considering the value for the development of a prototype of the URIEL System, a sales price of a system in series production of around US$ 1,000,000.00 can be considered feasible. This value, added to the costs related to the use of helicopters (flight hours, pilot working hours, URIEL operator working hours, depreciation, investment value), gives the total cost value.

The revenue value comes from the value related to the sale of wood in the international market. Aiming for the best cost-benefit ratio, the sales value of sawn timber in planks was considered. In this configuration, it was necessary to calculate the cost of this sawn timber, which refers to the investment in a large-scale sawmill, as well as the fixed and variable costs related to the operation of such a sawmill. The investment value to set up a large-scale sawmill is US$ 94,340.00 [144], and the total cost to process one cubic meter of wood is US$ 149.00 [145].

By obtaining all these values, it was possible to carry out the economic feasibility analysis for a period of 15 years based on the parameters of Net Present Value (NPV), Internal Rate of Return (IRR) and Payback.

The predefined values for the economic feasibility calculation are: payload helicopters; investment costs; flight hour costs (including pilot and URIEL operator working hours); price of the timber considered; Volume of timber considered in two scenarios (scenario 1: DBH of 0.8 m and 25 m trunk height; scenario 2: DBH of 1 m and 30 m trunk height).

Table 22. Characteristics of the selected helicopters.

| HELICOPTER | PAYLOAD (kg) | MISSION RADIUS (km) | CRUISE SPEED (km/h) | FLIGHT HOUR PRICE (US$) | INVESTMENT (US$) |
|---|---|---|---|---|---|
| CH-47 (new) | 12,565 | 306 | 291 | 6,705.00 | 39,000,000.00 |
| CH-47 (used – newer) | 12,565 | 306 | 291 | 6,705.00 | 20,000,000.00 |
| CH-47 (used – older) | 12,565 | 306 | 291 | 6,705.00 | 10,000,000.00 |

| HELICOPTER | PAYLOAD (kg) | MISSION RADIUS (km) | CRUISE SPEED (km/h) | FLIGHT HOUR PRICE (US$) | INVESTMENT (US$) |
|---|---|---|---|---|---|
| CH-53 | 16,329 | 200 | 270 | 34,497.00 | 87,000,000.00 |
| MI-26 | 20,000 | 400 | 255 | 15,000.00 | 25,000,000.00 |

Considering that each flight harvests one barrel in a harvesting time of 15 minutes per helicopter, the barrel unloading time is negligible. With 8 hours of work per day, 100 days per year, the operational envelope for various target distances is presented in the following table.

Table 23. Operational envelope of helicopters.

| Helicopter | Target Distance (km) | Flight Time to Target (minutes) | Rides per Day | Rides per Day* |
|---|---|---|---|---|
| CH47 | 10 | 4.12 | 19.89 | 20 |
| CH47 | 50 | 20.61 | 11.81 | 12 |
| CH47 | 100 | 41.23 | 7.83 | 8 |
| CH47 | 150 | 61.85 | 5.86 | 6 |
| CH47 | 200 | 82.47 | 4.68 | 5 |
| CH47 | 300 | 123.71 | 3.34 | 3 |
| CH53 | 10 | 4.44 | 19.63 | 20 |
| CH53 | 50 | 22.22 | 11.36 | 11 |
| CH53 | 100 | 44.44 | 7.44 | 7 |
| CH53 | 150 | 66.66 | 5.53 | 5 |
| CH53 | 200 | 88.88 | 4.40 | 4 |
| MI26 | 10 | 4.70 | 19.42 | 19 |
| MI26 | 50 | 23.52 | 11.02 | 11 |
| MI26 | 100 | 47.05 | 7.15 | 7 |
| MI26 | 150 | 70.58 | 5.29 | 5 |
| MI26 | 200 | 94.11 | 4.20 | 4 |
| MI26 | 300 | 141.17 | 2.97 | 3 |
| MI26 | 400 | 188.23 | 2.30 | 2 |

Note: *Disregarding the fractional part, in the roundings up the working days were slightly longer than 8 hours, in the roundings down the working days were slightly shorter than 8 hours.

Table 24. Total operational cost of helicopters.

| HELICOPTER | PAYLOAD (excluding URIEL) (kg) | Price per Flight Hour (US$) | Total Cost for 100 Working Days (US$) |
|---|---|---|---|
| CH-47 | 10,485 | 53,640.00 | 5,364,000.00 |
| CH-53 | 14,249 | 275,976.00 | 27,597,600.00 |
| MI-26 | 17,920 | 120,000.00 | 12,000,000.00 |

Tabela 25. Financial timber revenue per log (scenario 1).

| Wood | m³/Log | US$/m³ | Green Weight* (kg) | Sawmill Cost (US$) | Gross Revenue (US$) | Boards Billing (US$) |
|---|---|---|---|---|---|---|
| Cedar | 4.22 | 1,059.00 | 4,473 | 628.78 | 4,468.98 | 3,840.20 |
| Ipe | 10.50 | 1,446.00 | 20,160 | 1,564.50 | 15,183.00 | 13,618.50 |
| Jatoba | 7.11 | 868.00 | 13,651 | 1,059.39 | 6,171.48 | 5,112.09 |

Note: * Reduction need.

Due to the maximum payload of the CH-47 being 12,565 kg, considering the weight of the URIEL System of 2080 kg. For Ipê and Jatobá trees, it is necessary to divide the trunk into several sections of 10,000 kg, which represents several harvesting flights for the same tree. Thus, a reduction factor was added to the calculation of Gross Revenue for harvesting Ipê and Jatobá trees. The same logic applies to the CH-53 (payload of 16,329 kg) with regard to Ipê trees.

Table 26. Financial timber revenue per log (scenario 1 with reduction factor).

| Wood species | Helicopter | Reduction factor | Sawmill cost (US$) | Gross revenue (US$) | Boards billing (US$) |
|---|---|---|---|---|---|
| Ipê | CH47 | 0.595 | 931.25 | 9,037.50 | 8,106.25 |
| Jatobá | CH47 | 0.879 | 931.26 | 5,425.07 | 4,493.81 |
| Ipê | CH53 | 0.793 | 1,241.66 | 12,050.00 | 10,808.33 |

Table 27. Financial timber revenue per log (scenario 2).

| Timber | $m^3$/log | US$/$m^3$ | Green Weight* (kg) | Sawmill Cost (US$) | Gross Revenue (US$) | Boards Billing (US$) |
|---|---|---|---|---|---|---|
| Cedar | 7.64 | 1,059 | 8,098 | 1,138.36 | 8,090.76 | 6,952.40 |
| Ipê | 19.47 | 1,446 | 37,382 | 2,901.03 | 28,153.62 | 25,252.59 |
| Jatobá | 11.44 | 868 | 21,964 | 1,704.56 | 9,929.92 | 8,225.36 |

Note: * Reduction need.

In scenario 2, all helicopters need to have reduction values based on the green masses of the ipê and jatobá trees. Considering the URIEL System mass of 2080 kg, the useful transport mass of the fuselage for the CH-47 is 10,485 kg, for the CH-53 = 14,249 kg, and for the MI-26 = 17,920 kg.

Tabela 28. Financial timber revenue per log (scenario 2 with reduction factor).

| Wood species | Helicopter | Reduction factor | Sawmill cost (US$) | Gross revenue (US$) | Boards billing (US$) |
|---|---|---|---|---|---|
| Ipê | CH47 | 0.321 | 931.25 | 9,037.59 | 8,106.33 |
| Jatobá | CH47 | 0.546 | 931.28 | 5,425.19 | 4,493.91 |
| Ipê | CH53 | 0.428 | 1,241.67 | 12,050.12 | 10,808.45 |
| Jatobá | CH53 | 0.728 | 1,241.71 | 7,233.59 | 5,991.88 |
| Ipê | MI26 | 0.535 | 1,552.09 | 15,062.66 | 13,510.56 |
| Jatobá | MI26 | 0.910 | 1,552.13 | 9,041.99 | 7,489.85 |

Tabela 29. Financial timber revenue for 100 days with 8 h, logging scenario 1 (DAP=0,8 m e H=25 m ).

| Helicopter | Distance from Target (km) | Wood species | Billing per log (US$) | Number of trips | Total revenue 100 days/8 h (US$) |
|---|---|---|---|---|---|
| CH47 | 10 | Cedar | 3,840.20 | 20 | 7,680,400.00 |
| | | Ipê | 8,106.25 | 20 | 16,212,500.00 |
| | | Jatobá | 4,493.81 | 20 | 8,987,620.00 |
| | 50 | Cedar | 3,840.20 | 12 | 4,608,240.00 |
| | | Ipê | 8,106.25 | 12 | 9,727,500.00 |
| | | Jatobá | 4,493.81 | 12 | 5,392,572.00 |
| | 100 | Cedar | 3,840.20 | 8 | 3,072,160.00 |
| | | Ipê | 8,106.25 | 8 | 6,485,000.00 |

| Helicopter | Distance from Target (km) | Wood species | Billing per log (US$) | Number of trips | Total revenue 100 days/8 h (US$) |
|---|---|---|---|---|---|
| | | Jatobá | 4,493.81 | 8 | 3,595,048.00 |
| CH53 | 10 | Cedar | 3,840.20 | 20 | 7,680,400.00 |
| | | Ipê | 8,106.25 | 20 | 16,212,500.00 |
| | | Jatobá | 4,493.81 | 20 | 8,987,620.00 |
| | 50 | Cedar | 3,840.20 | 11 | 4,224,220.00 |
| | | Ipê | 8,106.25 | 11 | 8,916,875.00 |
| | | Jatobá | 4,493.81 | 11 | 4,943,191.00 |
| | 100 | Cedar | 3,840.20 | 7 | 2,688,140.00 |
| | | Ipê | 8,106.25 | 7 | 5,674,375.00 |
| | | Jatobá | 4,493.81 | 7 | 3,145,667.00 |
| MI26 | 10 | Cedar | 3,840.20 | 19 | 7,296,000.00 |
| | | Ipê | 8,106.25 | 19 | 15,401,875.00 |
| | | Jatobá | 4,493.81 | 19 | 8,537,099.00 |
| | 50 | Cedar | 3,840.20 | 11 | 4,224,220.00 |
| | | Ipê | 8,106.25 | 11 | 8,916,875.00 |
| | | Jatobá | 4,493.81 | 11 | 4,943,191.00 |
| | 100 | Cedar | 3,840.20 | 7 | 2,688,140.00 |
| | | Ipê | 8,106.25 | 7 | 5,674,375.00 |
| | | Jatobá | 4,493.81 | 7 | 3,145,667.00 |

Scenario 1; DBH=0.8 m and 25 m height; Scenario 2; DBH=1 m and 30 m height.

Table 30. Financial timber revenue for 100 days with 8 h, logging scenario 2 (DAP=1 m and H=30 m).

| Helicopter | Distance to Target (km) | Wood Species | Billing per Log (US$) | Rides per 100 Days (8 h/day) | Total Revenue for 100 Days (US$) |
|---|---|---|---|---|---|
| CH47 | 10 | Cedar | 6,952.40 | 20 | 13,904,800.00 |
| CH47 | 10 | Ipe | 8,106.25 | 20 | 16,212,500.00 |
| CH47 | 10 | Jatoba | 4,493.81 | 20 | 8,987,620.00 |
| CH47 | 50 | Cedar | 6,952.40 | 12 | 8,342,880.00 |
| CH47 | 50 | Ipe | 8,106.25 | 12 | 16,212,500.00 |
| CH47 | 50 | Jatoba | 4,493.81 | 12 | 8,987,620.00 |
| CH47 | 100 | Cedar | 6,952.40 | 8 | 5,561,920.00 |
| CH47 | 100 | Ipe | 8,106.25 | 8 | 16,212,500.00 |
| CH47 | 100 | Jatoba | 4,493.81 | 8 | 8,987,620.00 |

| Helicopter | Distance to Target (km) | Wood Species | Billing per Log (US$) | Rides per 100 Days (8 h/day) | Total Revenue for 100 Days (US$) |
|---|---|---|---|---|---|
| CH53 | 10 | Cedar | 6,952.40 | 20 | 13,904,800.00 |
| CH53 | 10 | Ipe | 10,808.45 | 20 | 21,616,900.00 |
| CH53 | 10 | Jatoba | 5,991.88 | 20 | 11,983,760.00 |
| CH53 | 50 | Cedar | 6,952.40 | 11 | 7,647,640.00 |
| CH53 | 50 | Ipe | 10,808.45 | 11 | 11,889,295.00 |
| CH53 | 50 | Jatoba | 5,991.88 | 11 | 4,943,191.00 |
| CH53 | 100 | Cedar | 6,952.40 | 7 | 4,866,680.00 |
| CH53 | 100 | Ipe | 10,808.45 | 7 | 7,565,915.00 |
| CH53 | 100 | Jatoba | 5,991.88 | 7 | 4,194,316.00 |
| MI26 | 10 | Cedar | 6,952.40 | 19 | 13,209,560.00 |
| MI26 | 10 | Ipe | 13,510.56 | 19 | 25,670,064.00 |
| MI26 | 10 | Jatoba | 7,489.85 | 19 | 14,230,715.00 |
| MI26 | 50 | Cedar | 6,952.40 | 11 | 7,647,640.00 |
| MI26 | 50 | Ipe | 13,510.56 | 11 | 14,861,616.00 |
| MI26 | 50 | Jatoba | 7,489.85 | 11 | 8,238,835.00 |
| MI26 | 100 | Cedar | 6,952.40 | 7 | 4,866,680.00 |
| MI26 | 100 | Ipe | 13,510.56 | 7 | 9,457,392.00 |
| MI26 | 100 | Jatoba | 7,489.85 | 7 | 5,242,895.00 |

## Supplementary Figures

### Harvesting Module (HM)

Considering the process flowchart, Figure 11, and the morphological matrix and solution synthesis methodologies described by [102,103] and used by [126–130], it was possible to find optimal solution paths and synthesize the concept of the Harvesting Module (HM).

Figure 11. Flowchart of the harvesting process with a URIEL.

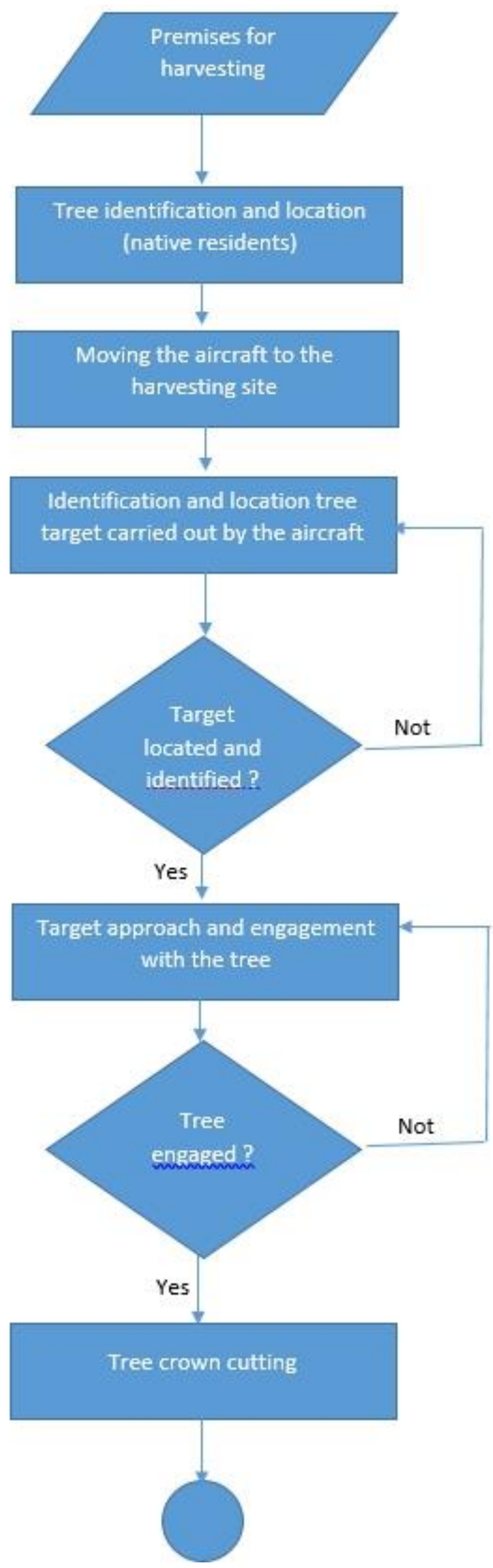
Premises for harvesting
Tree identification and location (native residents)
Moving the aircraft to the harvesting site
Identification and location tree target carried out by the aircraft
Target located and identified ?
Not
Yes
Target approach and engagement with the tree
Tree engaged ?
Not
Yes
Tree crown cutting

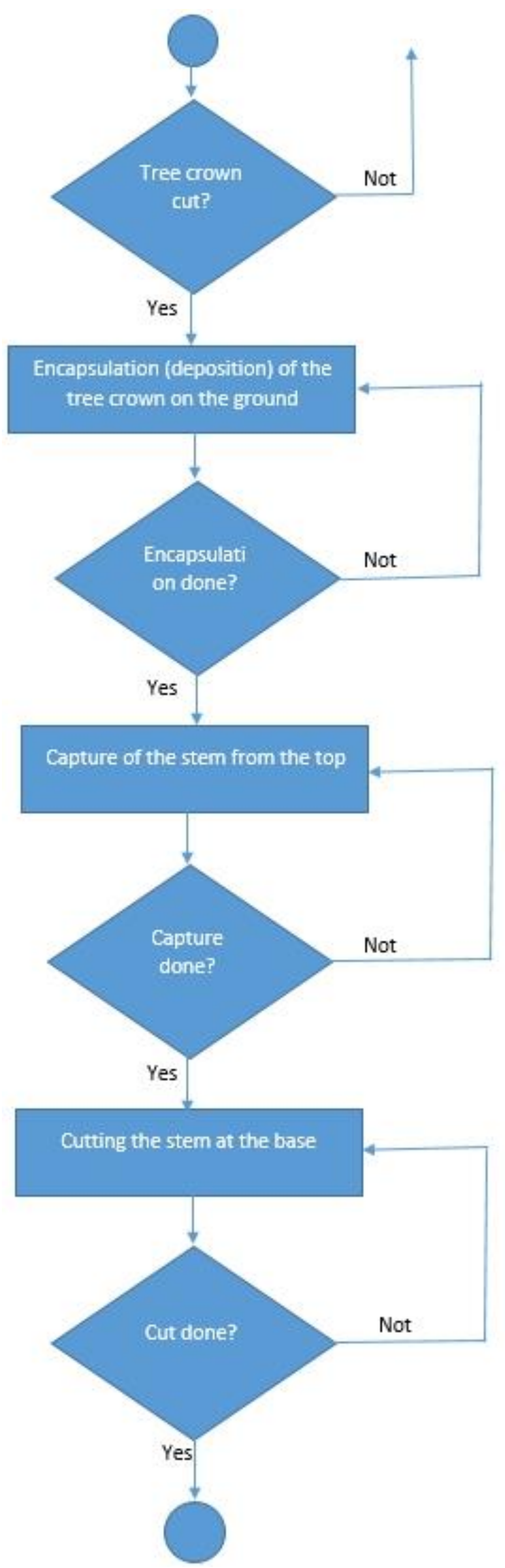
Tree crown cut?
Not
Yes
Encapsulation (deposition) of the tree crown on the ground
Encapsulati on done?
Not
Yes
Capture of the stem from the top
Capture done?
Not
Yes
Cutting the stem at the base
Cut done?
Not
Yes

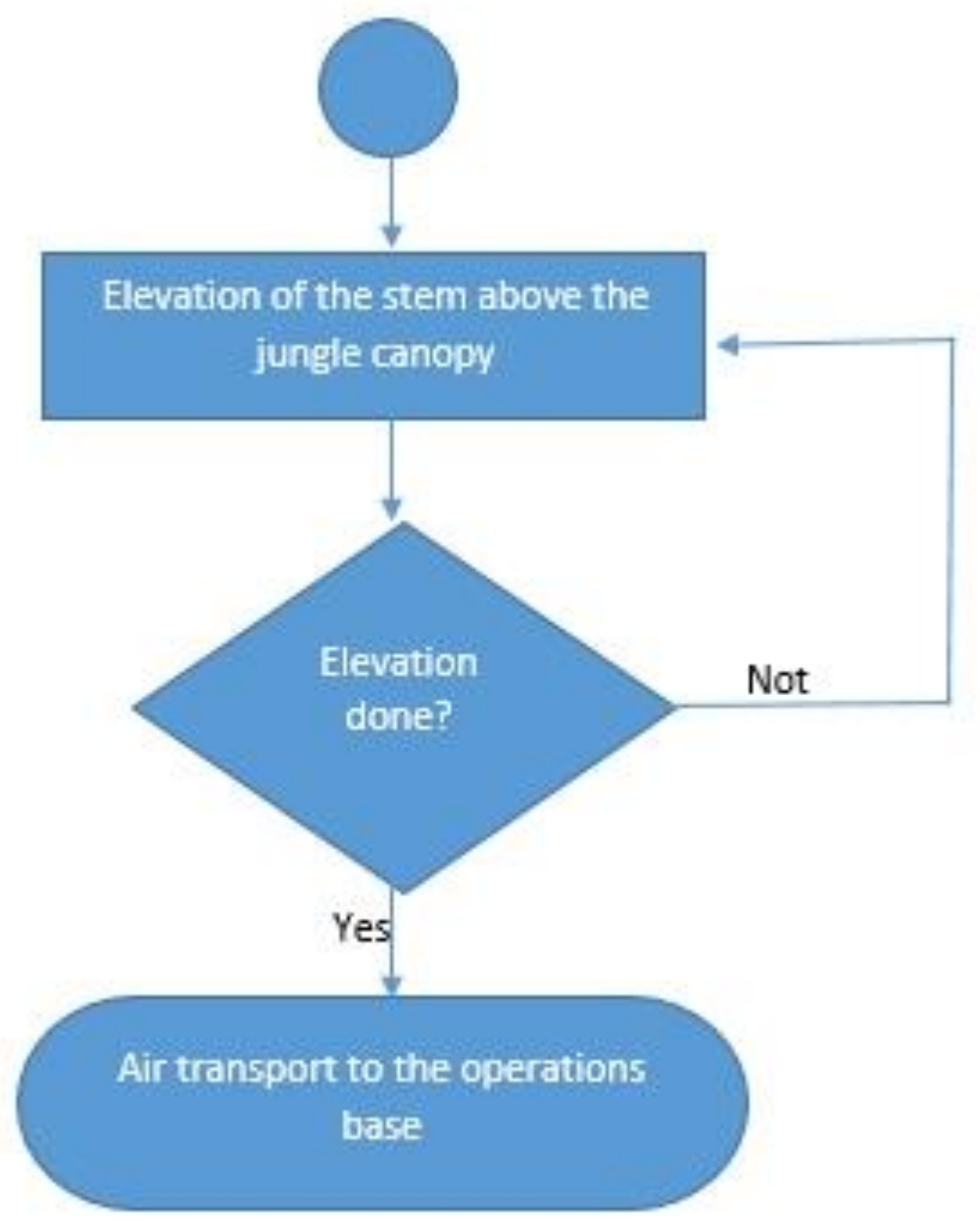

The Harvesting Module (HM) consists of three subsystems, Figure 12: stabilization subsystem (1), decoupling subsystem (2) and stem cutting subsystem (3). All these subsystems are integrated and operated with an electrical power source from lithium batteries, the operation is autonomously controlled by advanced AI in a 3T architecture divided into three algorithmic cores in the planning layer, one for each specific subsystem, each acting separately, but configured in a heterogeneous collaborative system whose dominance is the stabilization subsystem. These three subsystems follow the state of the art in: stabilization subsystem  crane leveling and control systems; decoupling subsystem Feller-Bunchers type forest harvesters and stem cutting system  Harvesters type forest harvesters.

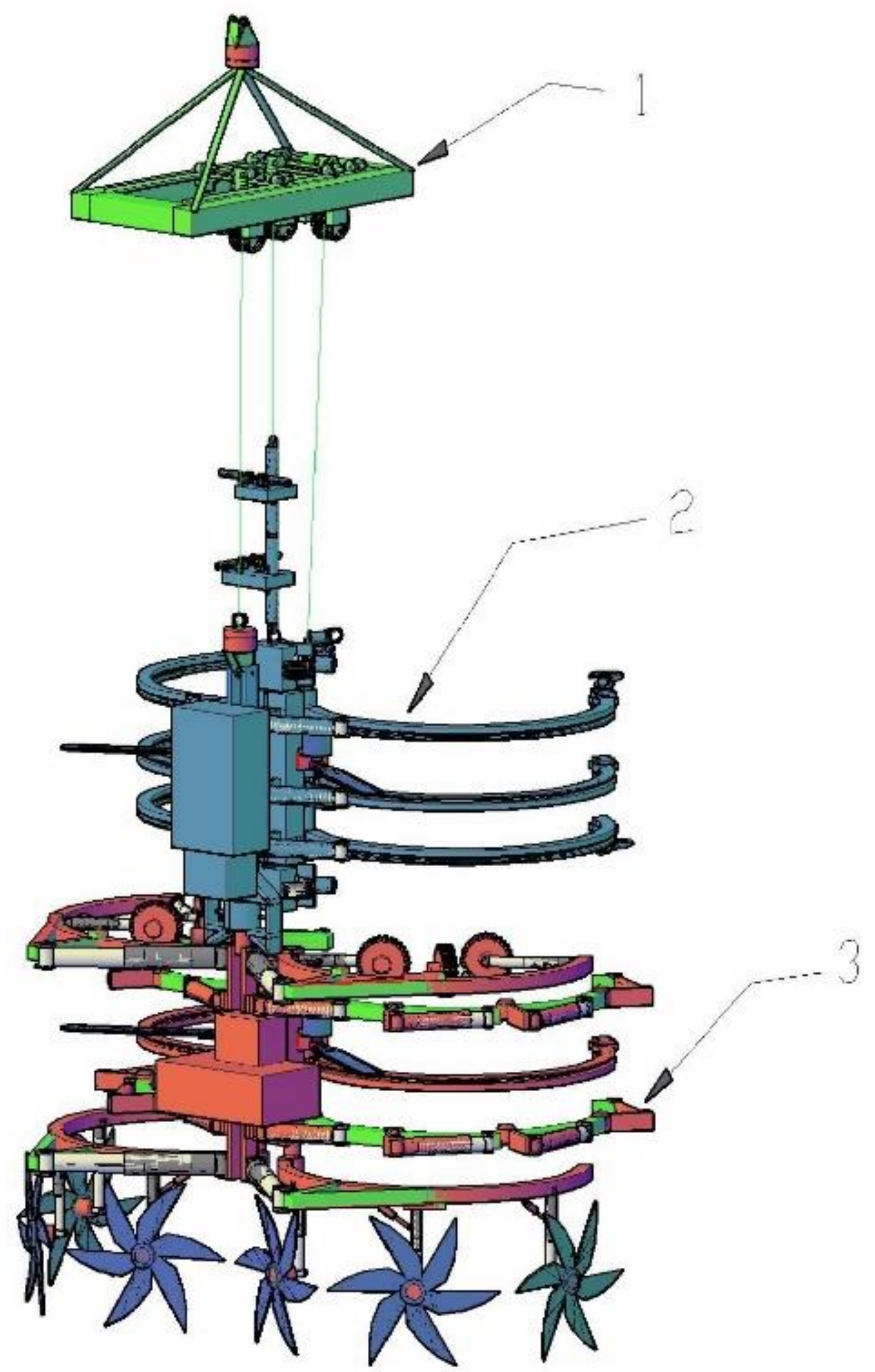

Figure 12. Harvesting Module: 1-stabilization subsystem; 2-decoupling subsystem; 3-stem cutting subsystem.

The stabilization subsystem, Figures 13 a) and b). consists of a rectangular frame with longitudinal and transverse rails that support three locomotion systems with metal wheels driven by electric motors. This frame is supported by four steel bars that converge at a vertex, the top of which has a rotator with one degree of freedom (Z). Each locomotion system has two degrees of freedom (X, Y) and carries a set of pulleys for steel cables. These pulleys are driven by high-power electric motors. The steel cables are coupled to the decoupling system in such a way that the vector movement of these cables makes it possible to control the orientation of the MH, control the deposition of the crown on the ground, and also control the positioning and stabilize the wood load. An AI-powered stabilization control system is responsible for activating the locomotion systems and pulleys in such a way that the steel cables are manipulated to operate the motorhome and stabilize the load.

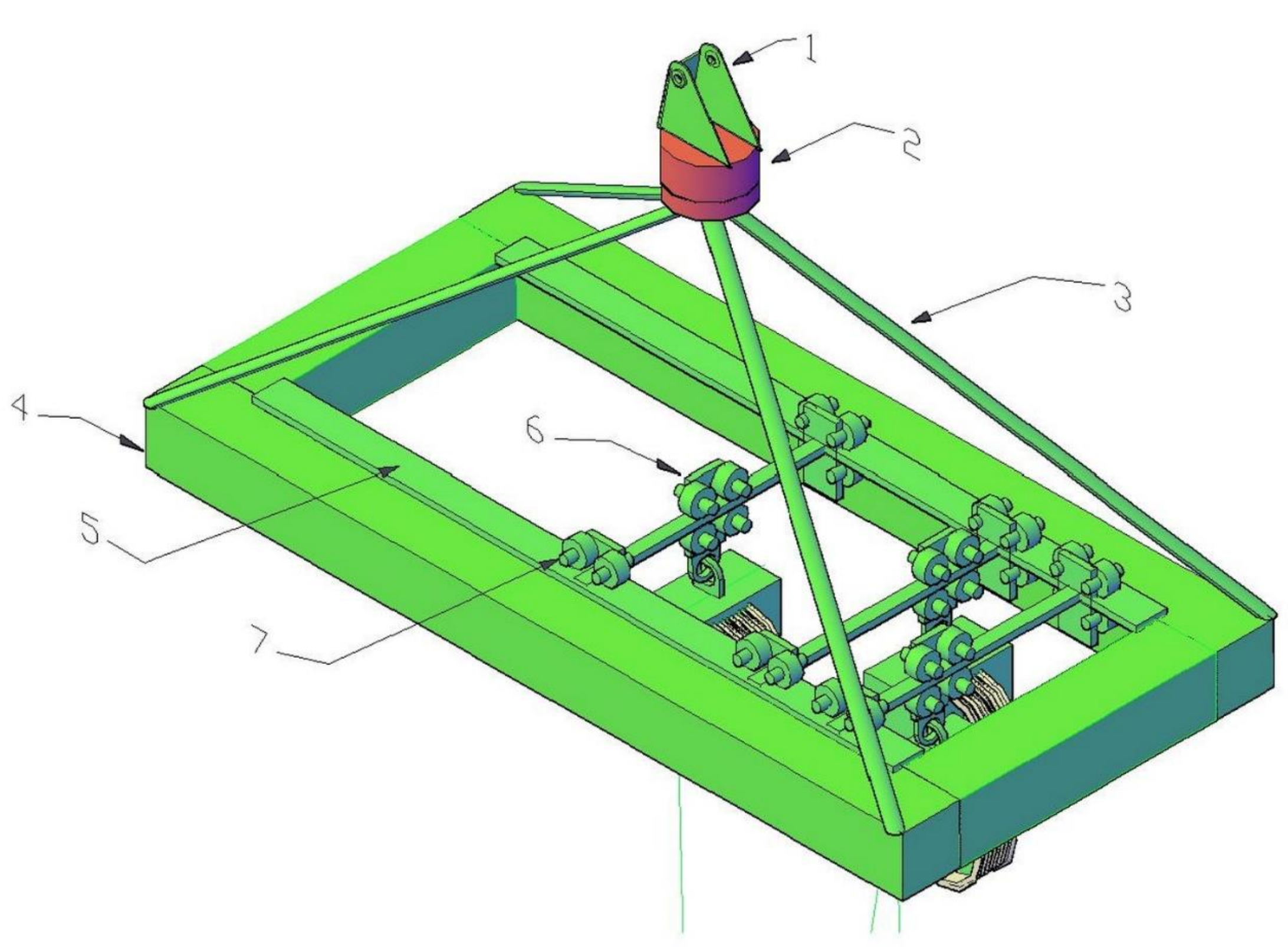

a)

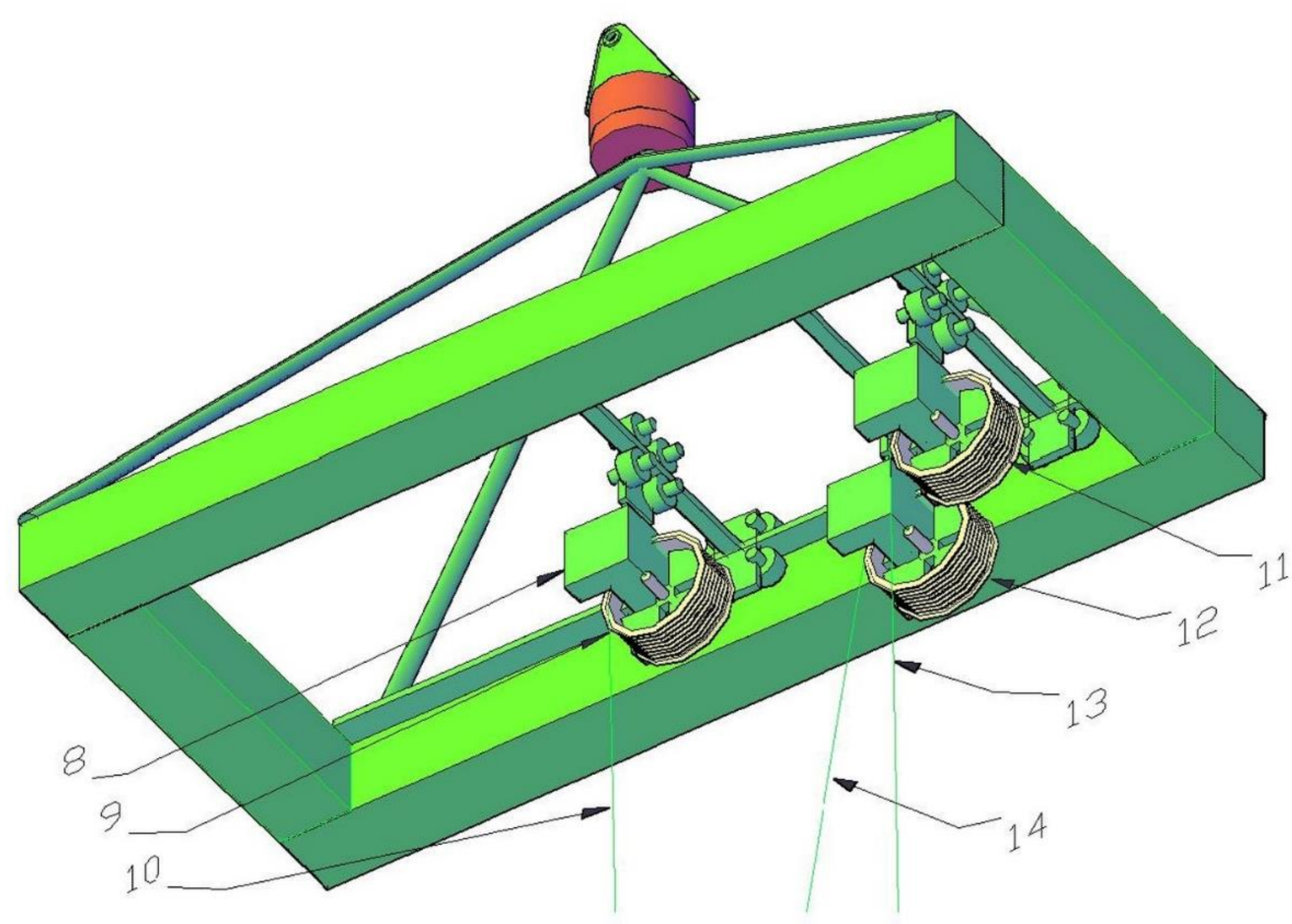

b)

Figure 13. Stabilization subsystem. a) Top view: 1-Rotation pivot for transport position; 2-Electric rotator; 3-Support rods; 4-Frame; 5-Pulley bridge movement rail; 6-Overhead crane; 7-Longitudinal displacement wheel assembly. b) Bottom view: 8-Battery chassis, reducer and electric motor; 9-Stabilizing pulley 1; 10-Stabilizing cable 1; 11-Stabilizing pulley 3; 12-Stabilizing pulley 2; 13-Stabilizing cable 2; 14-Stabilizing cable 3.

The decoupling subsystem is formed by three subassemblies, Figure 14: (A)-Grapple subassembly; (B)-Alignment and load subassembly and (C)-Rotation and support subassembly. The rotation and support subassembly, Figure 15 a), consists of a rigid main chassis, Figure 16, where, on the left side of this chassis, a high-capacity lithium battery is housed, and the hydraulic system for a rotator, the power inverter systems, the digital controllers, the industrial computer of the planning layer, and the computer vision system are also coupled. The entire subsystem is supported by a hydraulic rotator from a harvester head located on the upper part of the chassis. This rotator is operated by an oil flow from a hydraulic pump driven by an electric motor. The rotator is coupled to the main support steel cable of the MH. At the bottom of the chassis, there is the coupling mast for the stem cutting subsystem.

On the right side of the decoupling subsystem chassis, there are bearings and shafts and a main beam that support three claws, each composed of two grippers, Figure 15 b) and c). Each gripper is actuated by a high-force electric linear actuator. The upper and lower claws have an automatic hook and loop system at their tips, Figure 15 d). These hooks and loops are coupled to steel cables, which are attached to reel systems located next to the main support shaft of the claws and are driven by high-power electric motors. Two chainsaw cutting systems driven by electric motors are coupled to the intermediate claw, and in the center of the support mast there is a magnetic cannon loaded with a metal arrow.

At the top of the main beam of the claws, there is a system consisting of a high-capacity load chain, Figure 17 a) and b), where two magnetic cannon support carriages are coupled. Each carriage has two magnetic cannons oriented diametrically to each other. These vehicles move using chain-driven gears coupled to a chain and powered by electric motors. The magnetic cannons are loaded with metal arrows that, at the rear after the steering fins, are attached to small-diameter steel cables. These magnetic cannons have an automatic laser aiming system to guide them to target the center of the tree trunk at the height of the canopy.

The stabilization and decoupling subsystems communicate with each other and with the control center on board the helicopter via a CAN wired network.

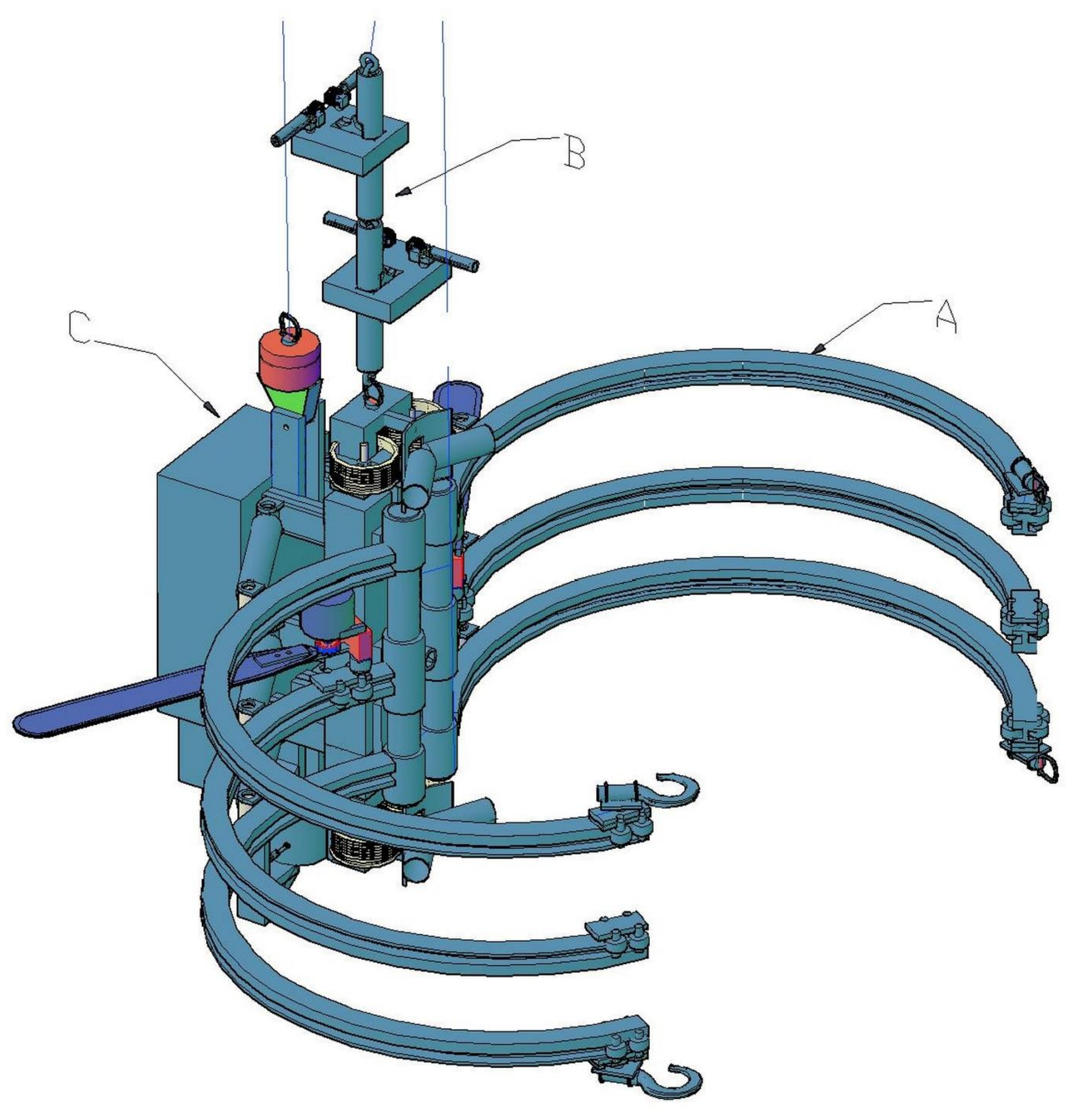

Figure 14. Decoupling subsystem: A; Gripper subassembly; B-Load/alignment subassembly; C-Rotation and support subassembly.

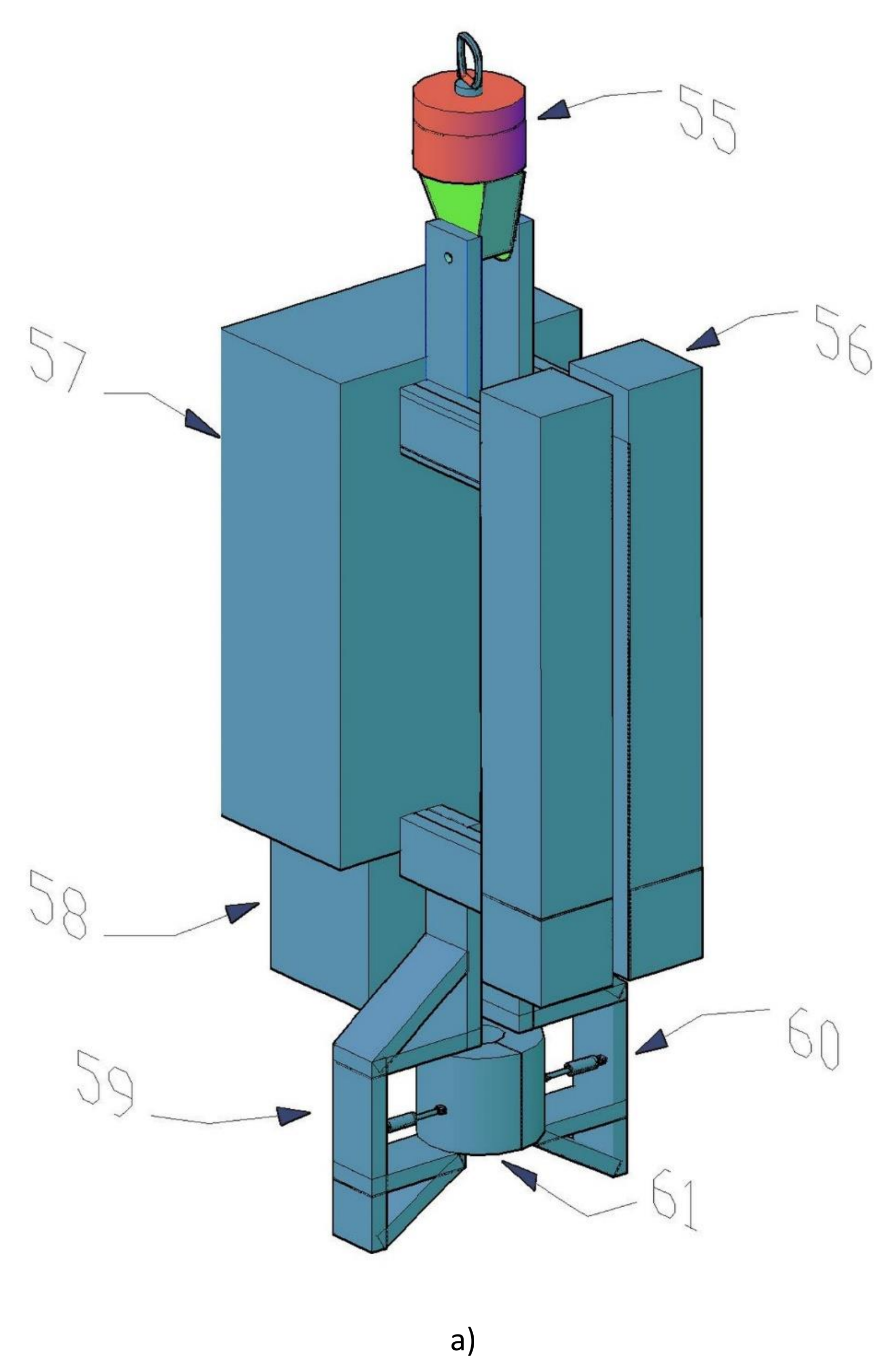

a)

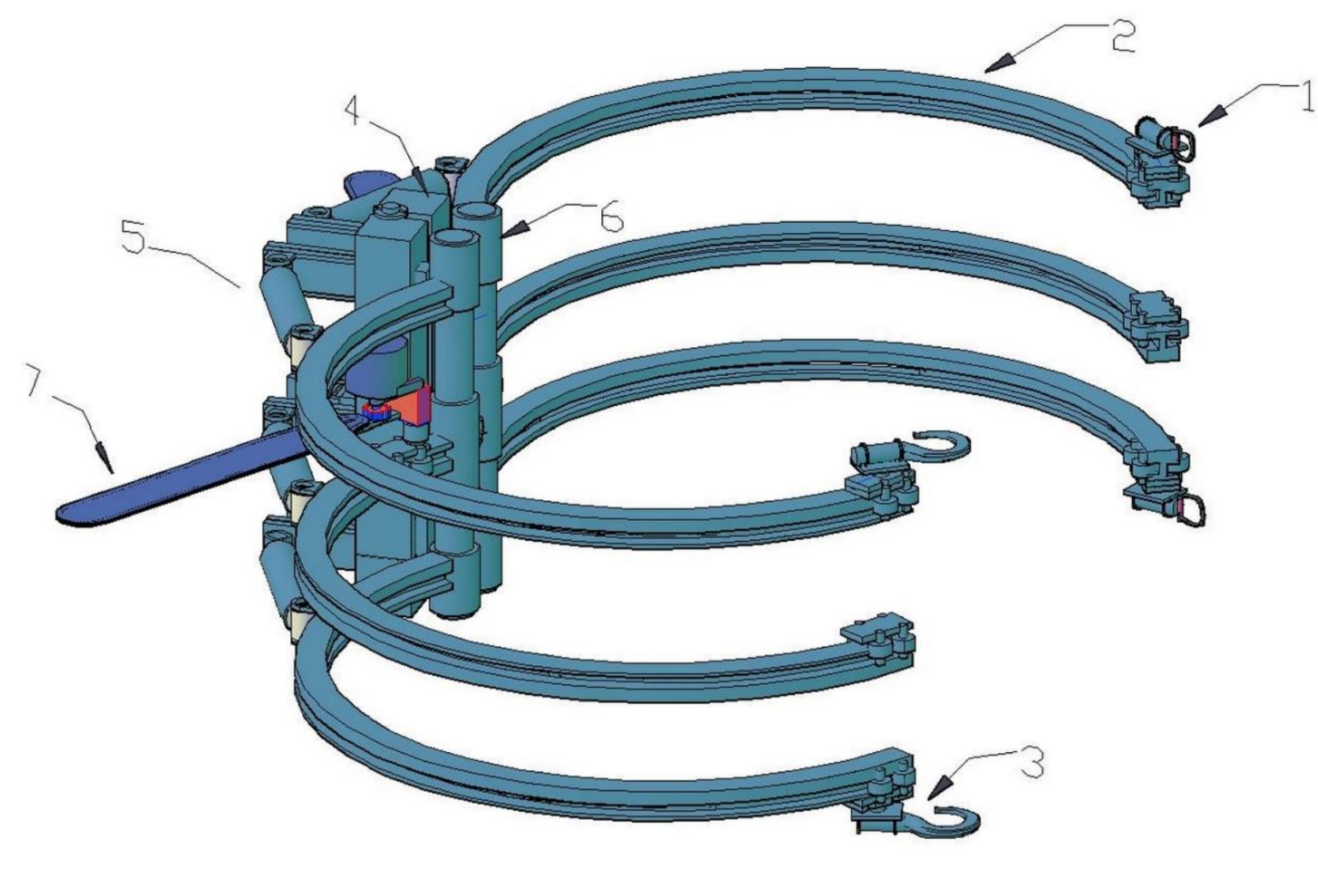

b)

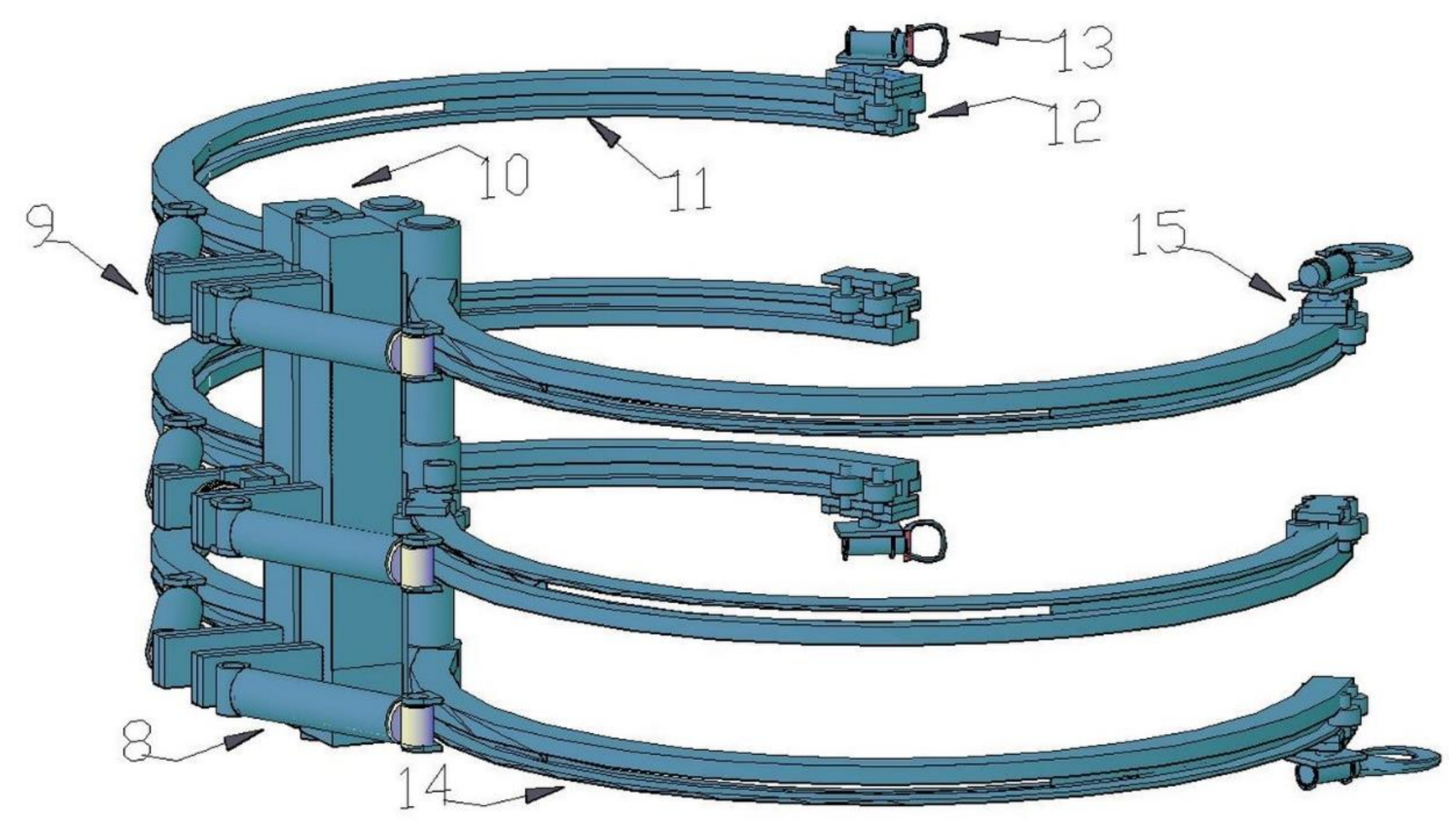

c)

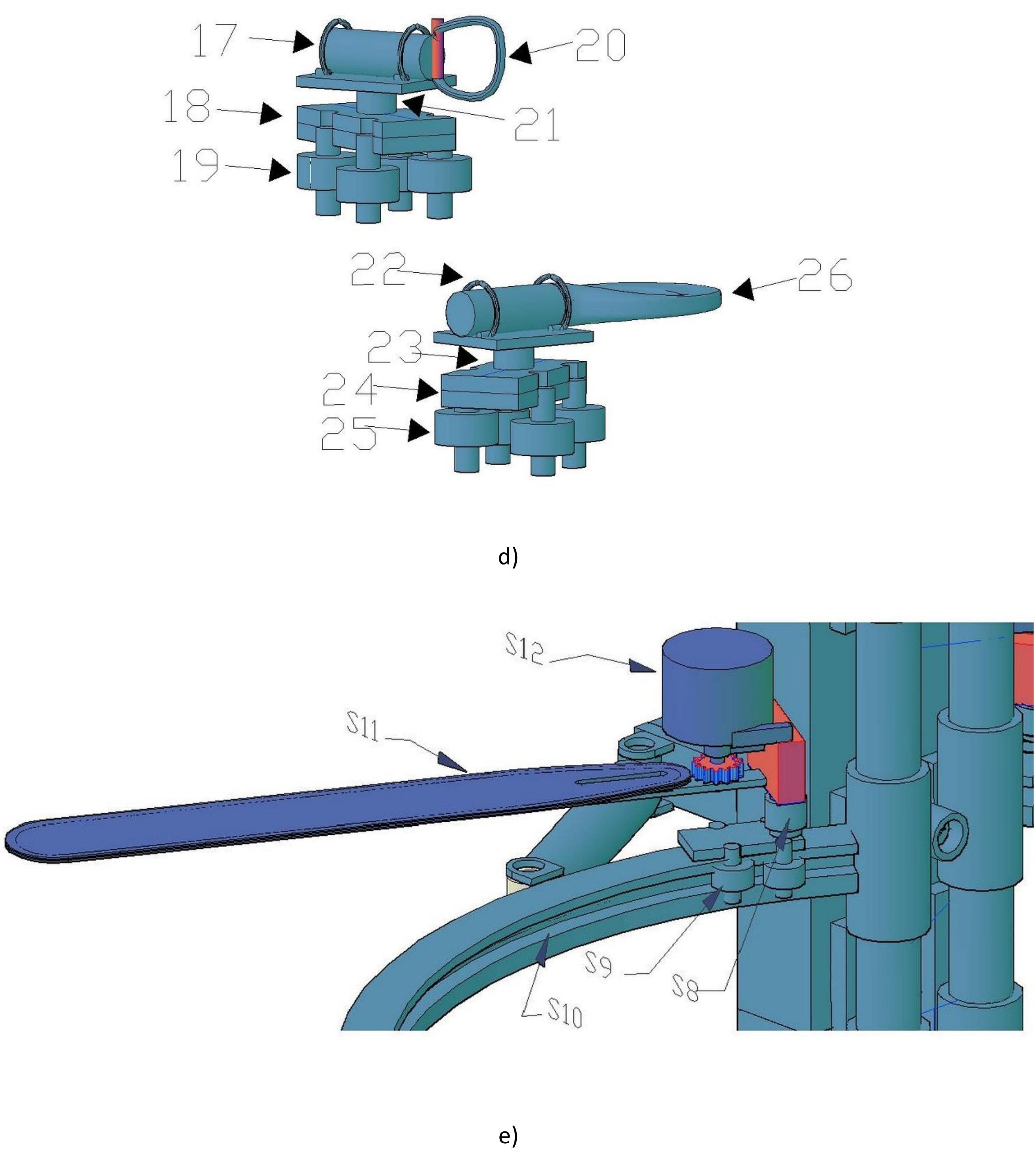

Figure 15. Subassemblies of the decoupage system. a) Rotation and support subassembly: 55- Hydraulic rotator; 56- Claw support chassis; 57- Battery box and electro-electronic systems; 58- Hydraulic pump box and accessories; 59- Coupling mast chassis; 60- Mast linear actuator; 61- Coupling mast. b) Claw subassembly, front view: 1- Coupling handle for the crimping cable; 2- Upper claw arm; 3- Coupling hook for the crimping cable; 4- Pivoting claw chassis; 5- Upper claw linear actuator; 6- Claw pivoting bearing; 7- Chainsaw bar with cutting chain. c) Claw subassembly, rear

view: 8- Lower claw linear actuator; 13- Cable coupling handle; 13- Rail-mounted displacement trolley system; 14-Lower claw arm; 15-Hook transport trolley. d) Cable clamping system: 17-Handle steel cable trunnion; 18-Handle trolley chassis; 19-Trolley wheels with electric motors in the hubs; 20-Cable coupling handle; 21-Circular actuator for handle rotation; 22-Hook steel cable trunnion; 23-Hook circular actuator; 24-Hook trolley chassis; 25-Trolley wheels with electric motors in the hubs. e) Cutting system detail: S8-Rotary actuator supporting the chainsaw chassis; S9-Chainsaw transport trolley; S10-Transport trolley rail machined into the intermediate claw; S11-Chainsaw bar; S12-High torque and rotation electric motor.

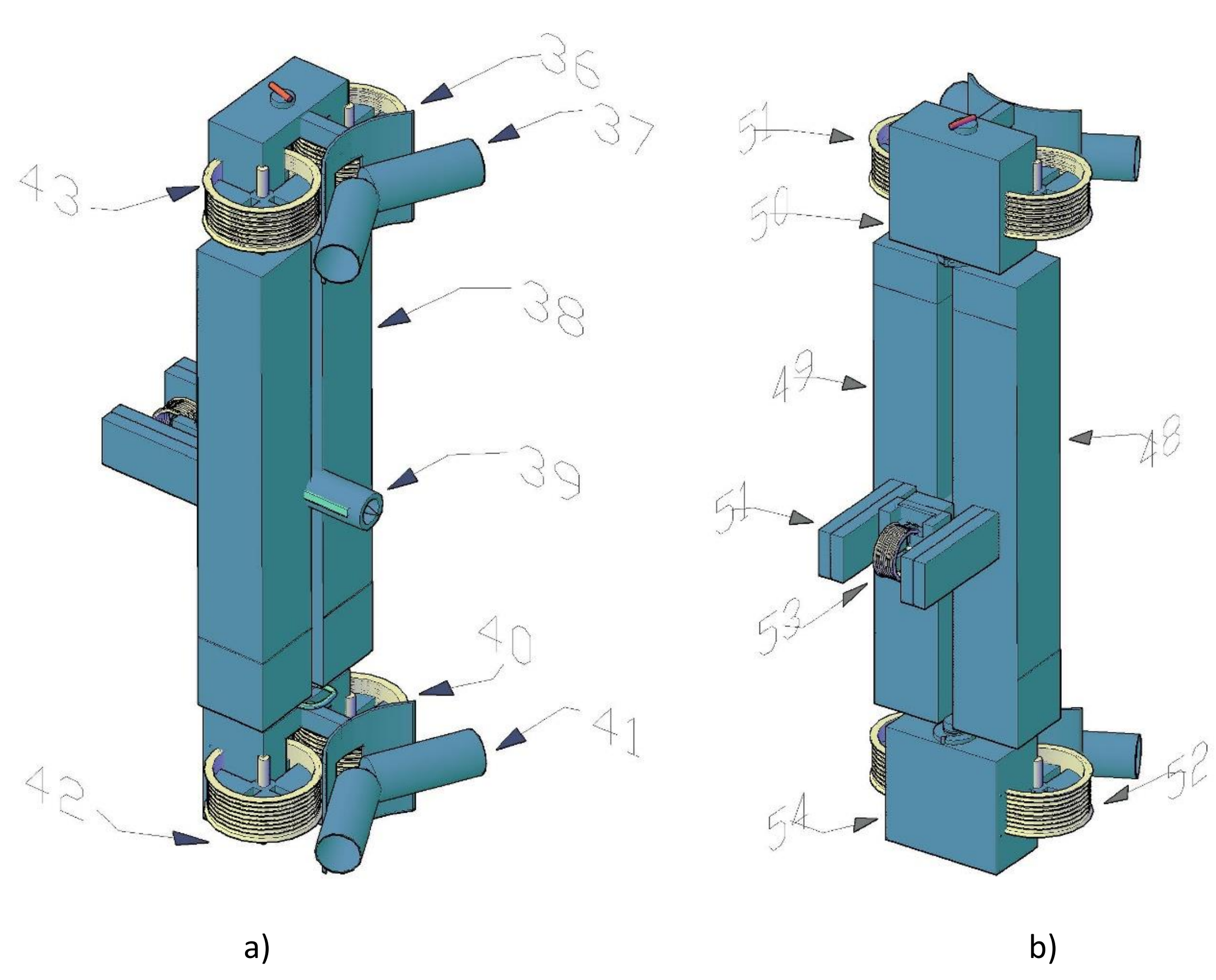

Figure 16. Rigid main chassis. a) Front view: 36-Left pulley of the upper handle clamping cable; 37-Upper handle clamping cable conductor; 38-Rigid main chassis; 39-Main magnetic cannon; 40-Left pulley of the lower handle clamping cable; 41-Lower handle clamping cable conductor; 42-Right pulley of the lower hook clamping cable; 43-Right pulley of the upper hook clamping cable. b) Rear view: 48-Right beam of the rigid main chassis; 49-Left beam of the rigid main chassis; 50-Electric motor housing,

reducers and drives for the upper clamping cable pulleys; 52-Support flange of the main magnetic cannon; 53-Pulley of the main magnetic cannon clamping cable; 50-Electric motor housing, reducers and pulley drives for lower clamping cables.

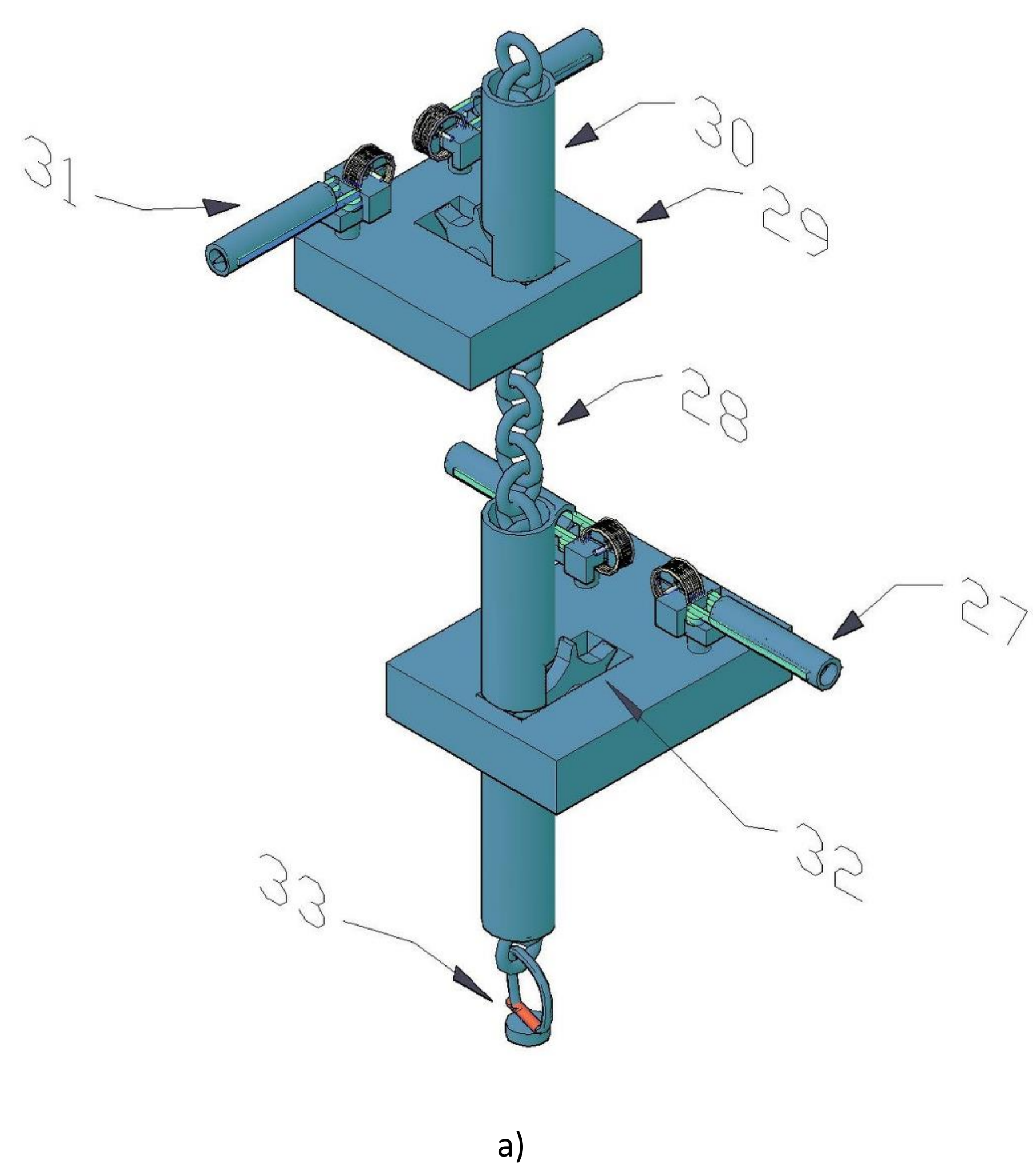

a)

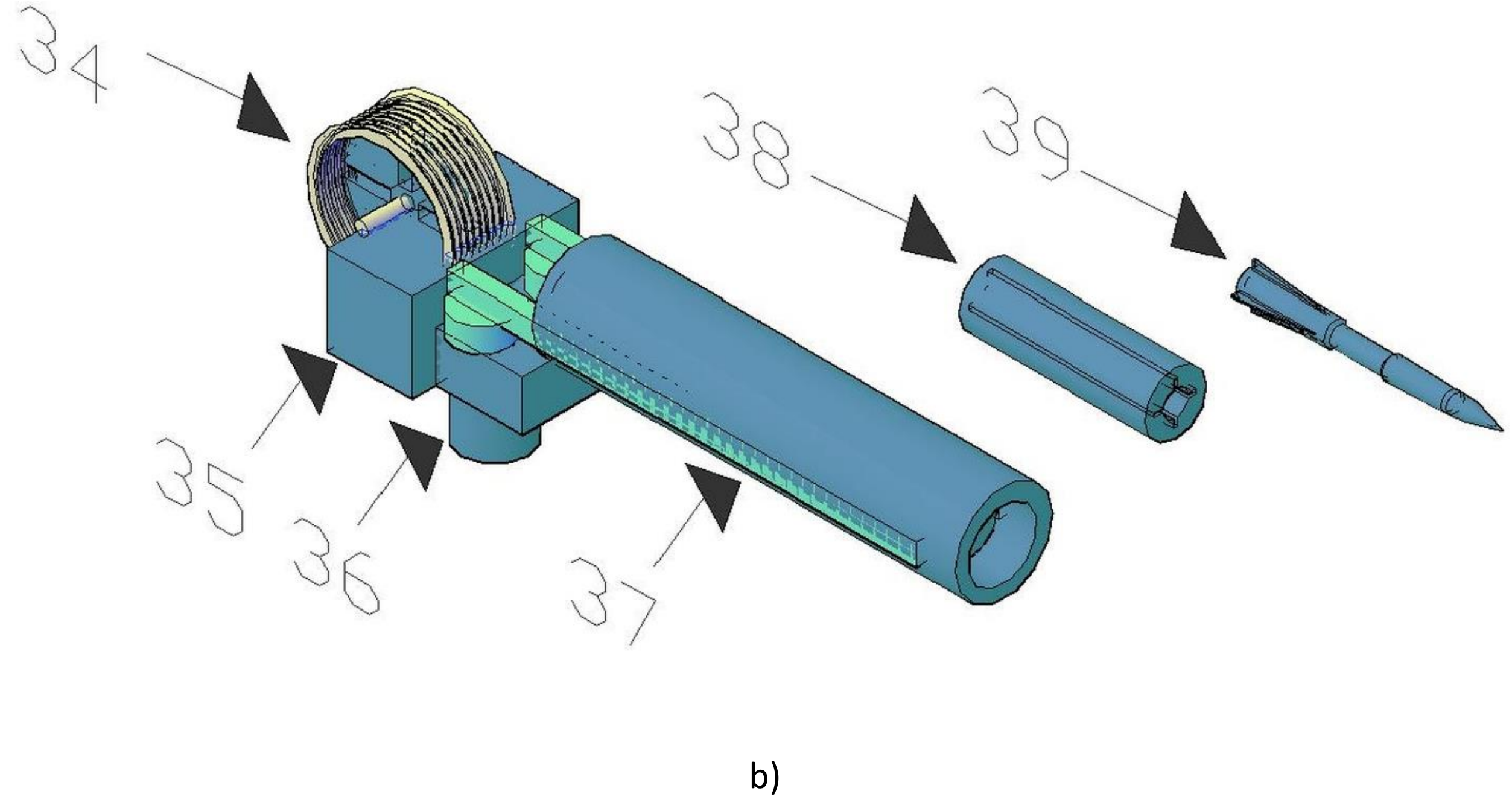

b)

Figure 17. A) Load/Alignment Subassembly: 27-Port/Starboard Magnetic Cannons; 28-High-Capacity Chain; 29-Upper Support Carriage; 30-Chain Rigidity Tube; 31-Bow/Stern Magnetic Cannons; 32-Chain Gear for Carriage Lifting and Rigid Chain/Rigidity Tube Flexion Locking; 33-Loop Coupling of the Load/Alignment Subassembly with the Main Rigid Subassembly. b) Magnetic Cannon: 34-Alignment Steel Cable Pulley; 35-Electric Motor Housing, Reducer and Alignment Cable Drive; 36-High-Power Shielded Magnetic Unit; 37-Magnetic Barrel with Magnetic Induction Rail; 38-Raked Magnetic Polarizer; 39-Ferrimagnetic Steel Dart.

The stem cutting subsystem, Figure 18, consists of a stem traversing subset (A), a cutting subset (B), and a delimbing subset (C). A main chassis, Figure 19, houses a high-capacity lithium battery on the left side, and also connects the canopy clearing system, the stem traversing system, the power inverter systems, the digital controllers, the industrial planning layer computer, and the computer vision and LoRa communication systems.

The shaft walking system, Figure 20, consists of a gripper located at the top of the system composed of two tongs, each tong is actuated by a high-force electric linear actuator, in each tong there are three metal wheels with sharp prismatic tungsten carbide tips each driven by high-power electric motors, these wheels and motors are located at the ends of a telescopic system actuated by a high-force electric linear actuator, each wheel/motor/actuator assembly is deflected at an angle oriented radially to the arc of the gripper by electric linear actuators.

The delimbing system, Figure 21, is located at the bottom of the chassis and consists of a claw composed of two pincers, each pincer is driven by an electric linear actuator, in each pincer there are three high-speed circular cutting systems with sharp carbide blades oriented longitudinally to the shaft, these blades are driven by high-speed electric motors, and are supported by a telescopic system driven by an electric linear actuator and also has a deflection angle controlled by another electric linear actuator. At the top of the chassis there is the coupling claw to the coupling mast of the delimbing subsystem.

On the right side of the shaft walking subsystem chassis there are bearings and shafts and a main beam that support three claws composed of two pincers each. The intermediate pincer is driven by a high-force electric linear actuator. This claw has two chainsaw cutting systems coupled to it, Figure 22, driven by electric motors. The upper and lower claws are segmented into four sections, each section is actuated by high-force electric linear actuators.

The shaft cutting subsystem communicates with the stabilization and decoupling subsystems and with the control center on board the helicopter via a LoRa wireless network.

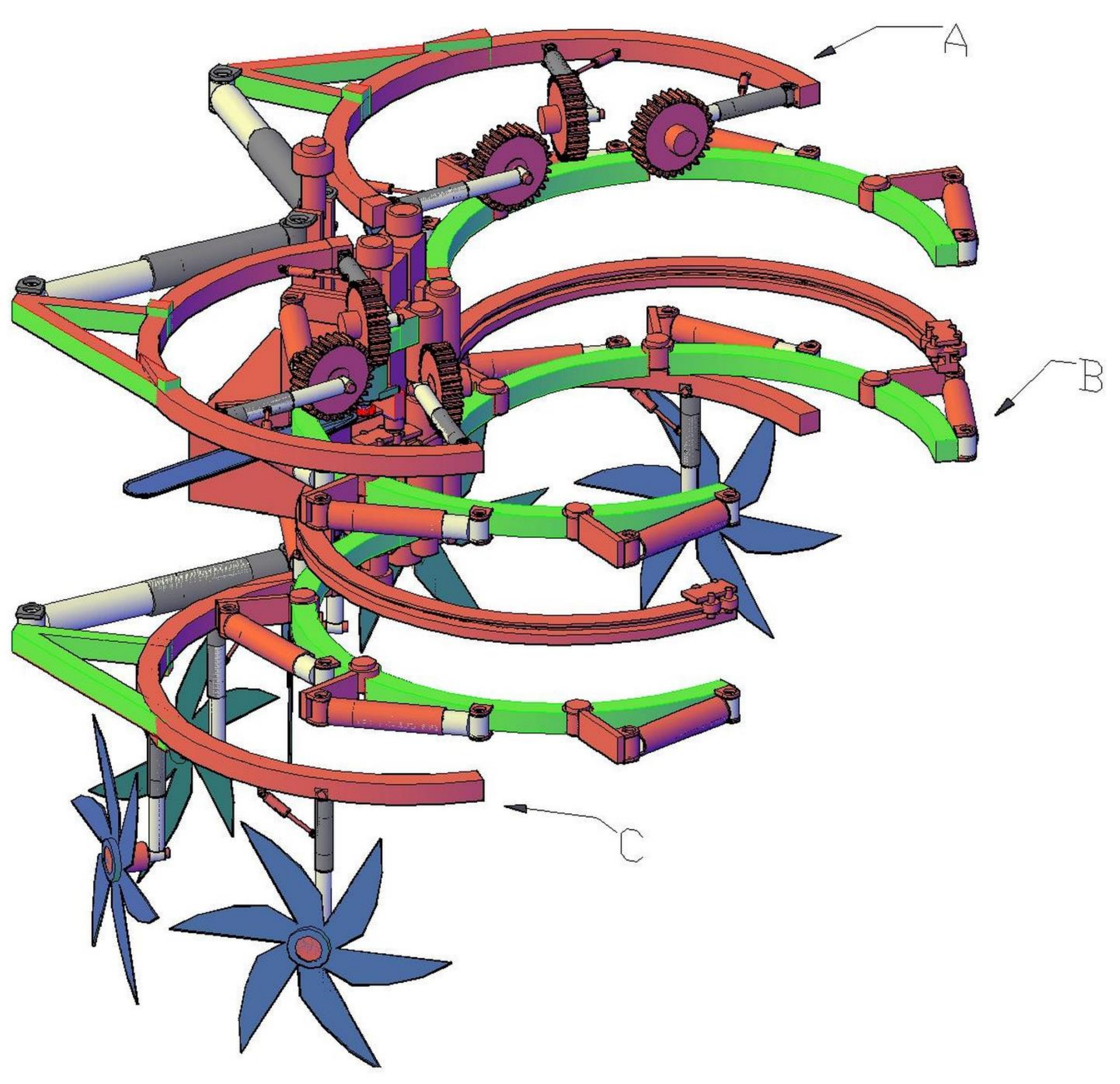

Figure 18. Stem cutting subsystem: A) Stem walking subassembly; B: cutting subassembly and C) delimbing subassembly.

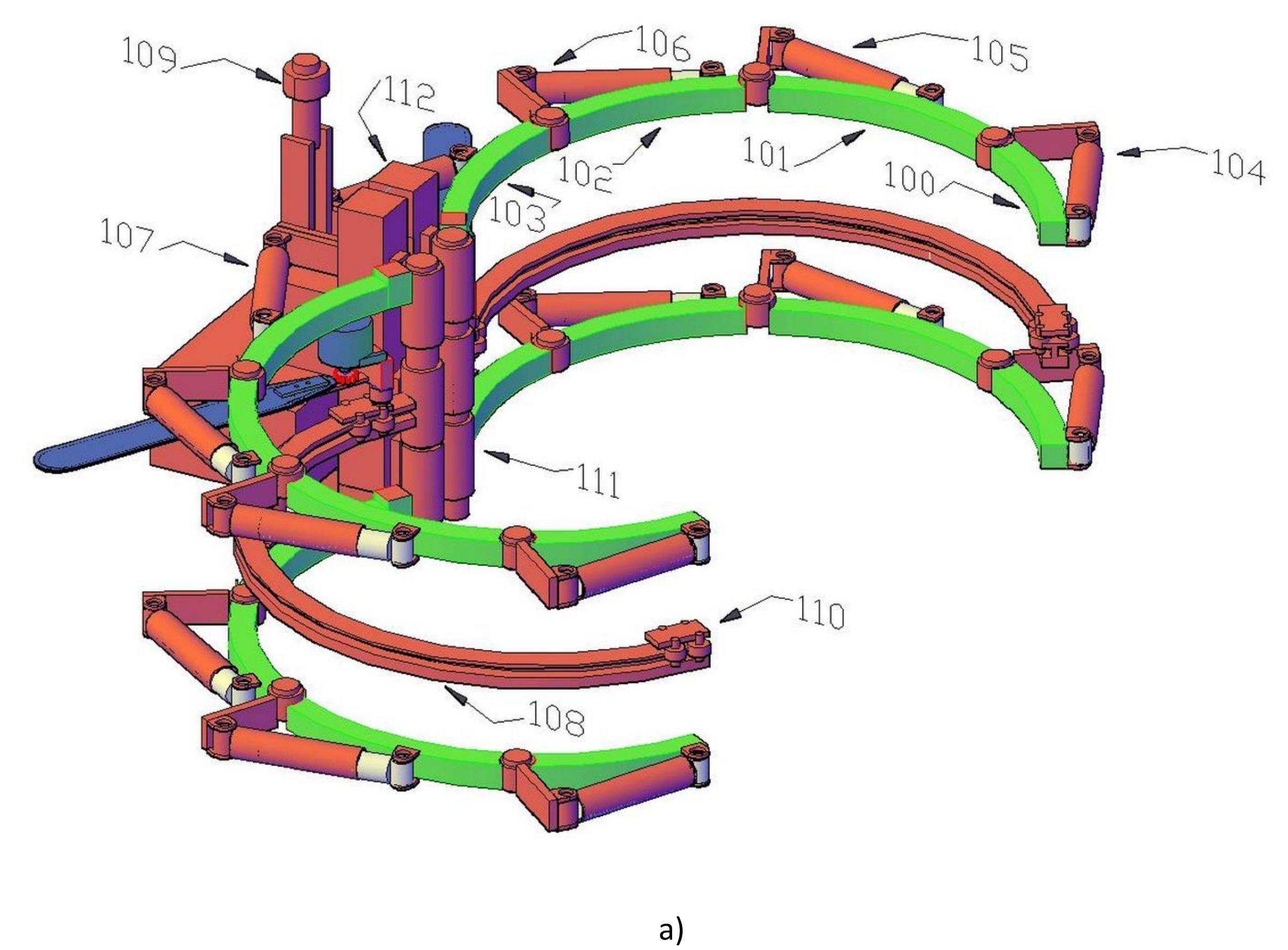
109
112
106
105
102
101
104
103
100
107
111
110
108

a)

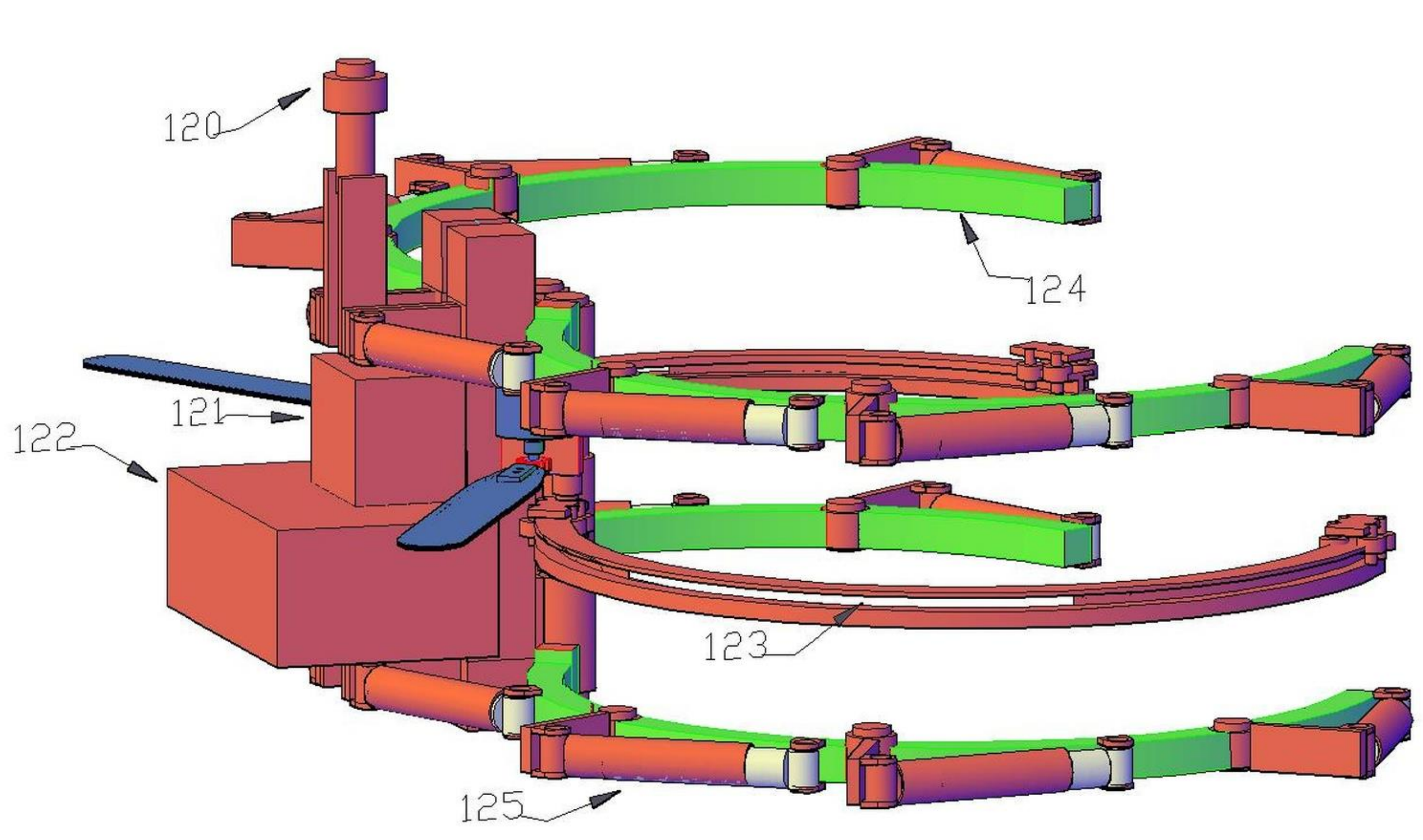
120
124
121
122
123
125

b)

Figure 19. Main chassis: Front view a) 100-Articulated claw front section; 101-Articulated claw intermediate section 2; 102-Articulated claw intermediate section 1; 103-Articulated claw base section; 104-Linear actuator of the front section; 105-Linear actuator of the intermediate section 2; 106-Linear actuator of the intermediate section 1; 107-Linear actuator of the base section; 108-Intermediate claw; 109-Coupling mast; 110-Chainsaw transport trolley; 111-Claw pivot bearings; 112-Main chassis. Rear view b) 120-Head of the coupling mast; 121-Electro-electronic control, actuation and sensing systems box; 122-Battery and accessories box; 123-Track of the chainsaw transport trolley; 124-Front section of the articulated grapple; 125-Linear actuator of the base section of the articulated grapple.

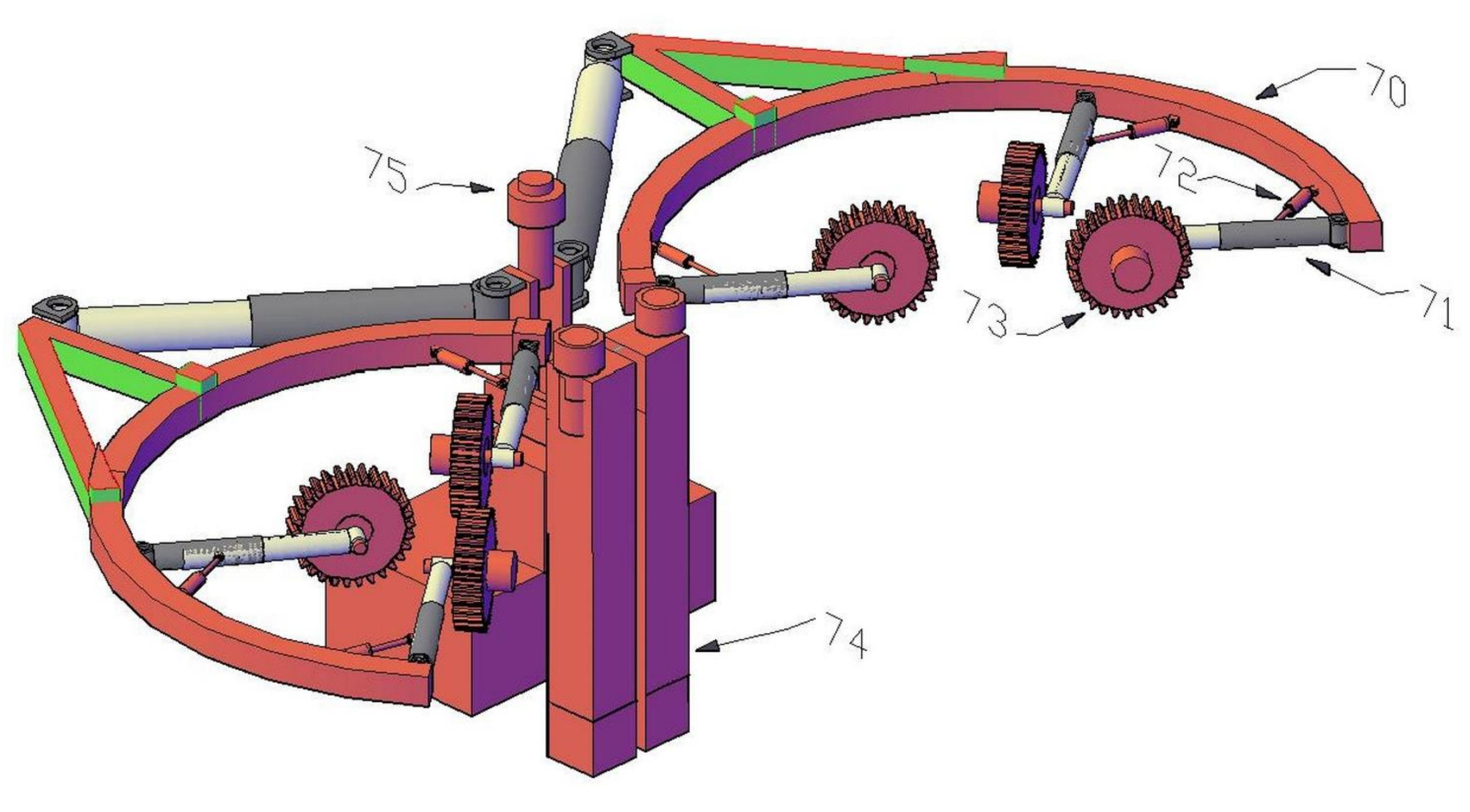

a)

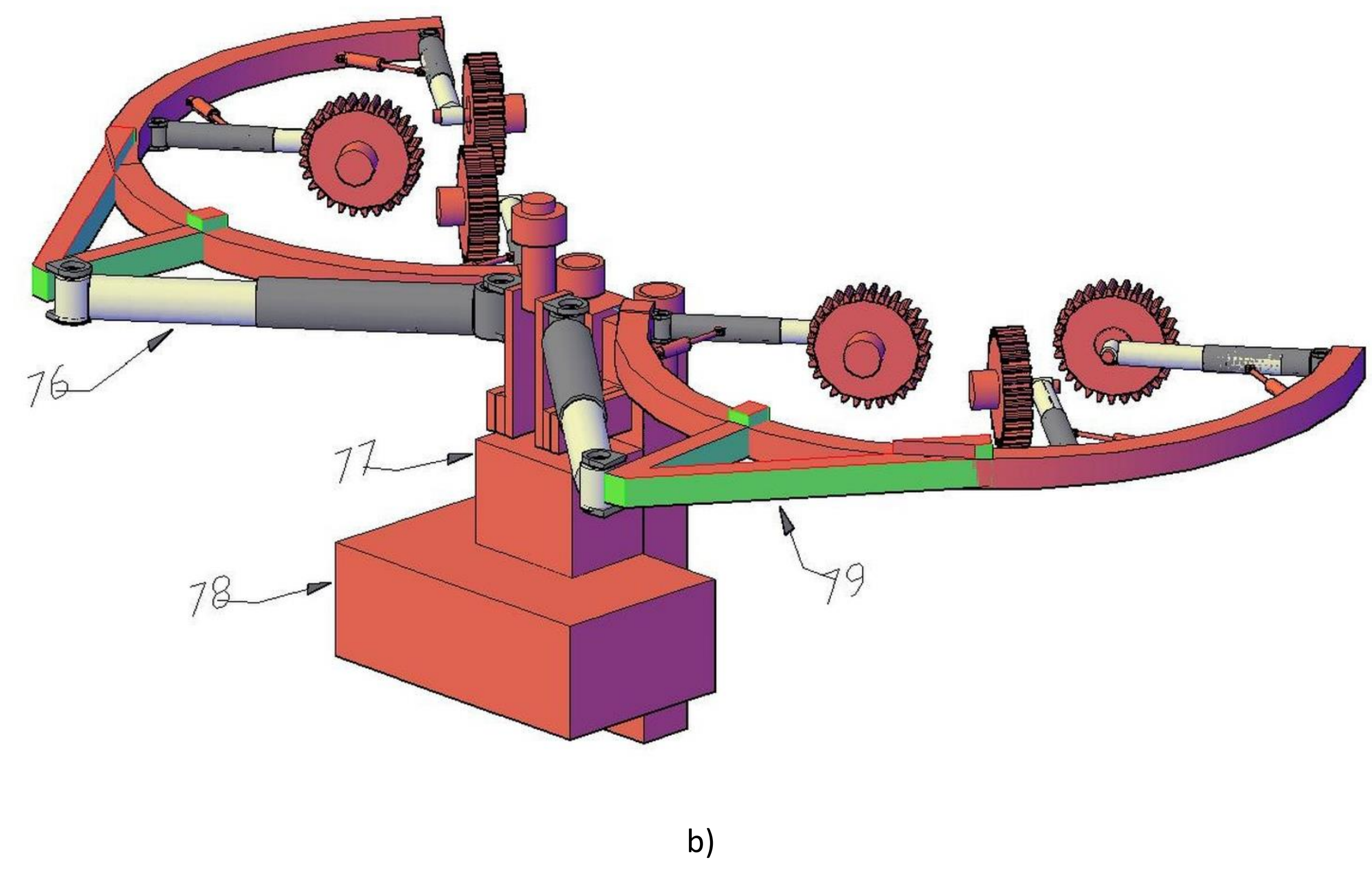

b)

Figure 20. Shaft travel subsystem. Front view a): 70-Tense gripper; 71-High-force linear actuator supporting the right front sprocket; 72-Linear actuator for angling the sprocket system; 73-Wydia sprockets; 74-Main chassis; 75-Coupling mast with the decoupage system. Rear view b): 76-High-force linear actuator for shaft gripping; 77-Electro-electronic control, drive and sensing systems box; 78-Battery and accessories box; 79-Torque support bracket for the high-force linear actuator.

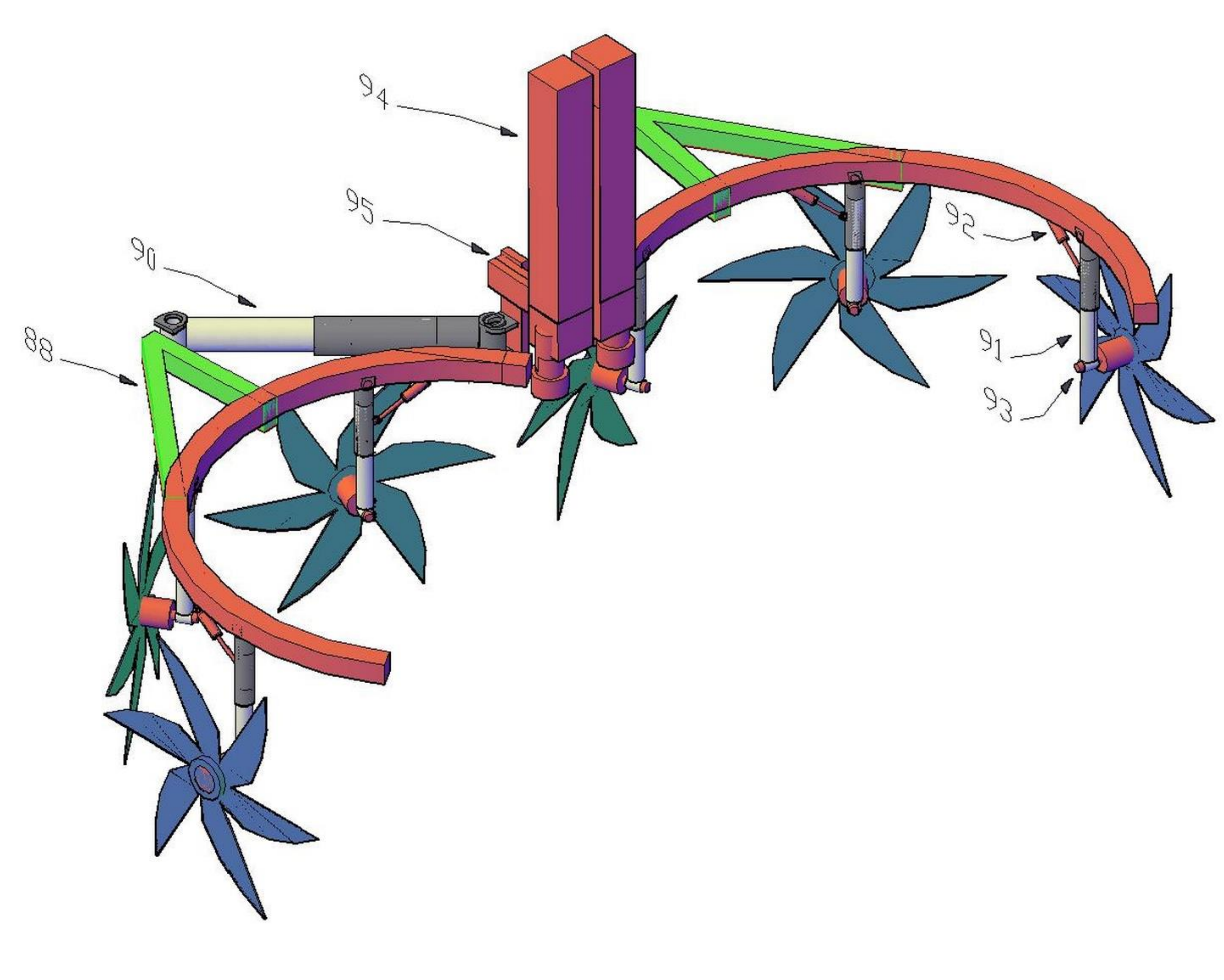

a)

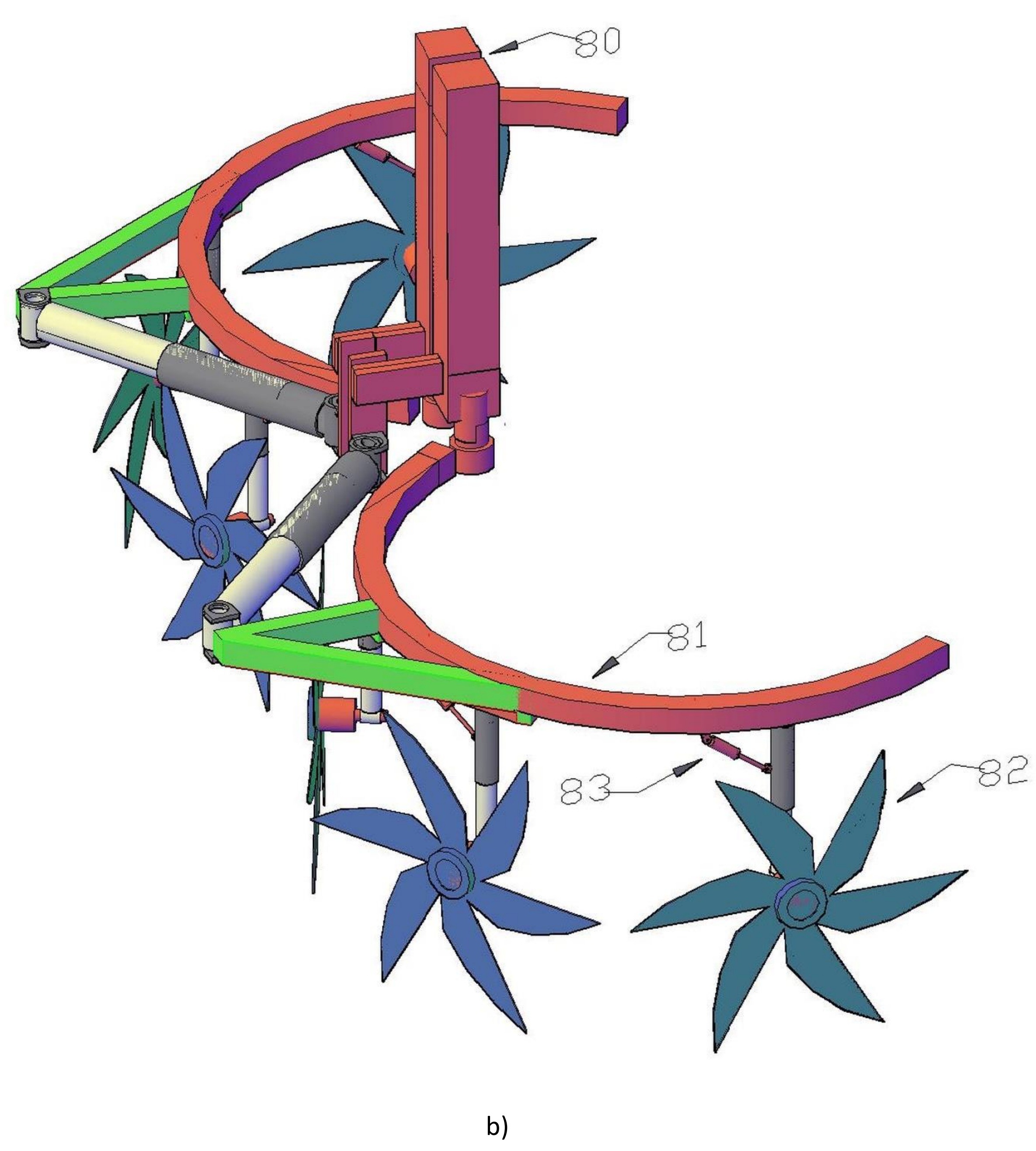

b)

Figure 21. Branch removal system. Front view a): 88-Torque bracket for linear actuator; 90-High-force linear actuator; 91-Linear actuator supporting the branch removal blade; 92-Linear actuator for angling the branch removal blade; 93-High-torque, high-speed electric motor; 94-Right main chassis bar; 95-Structural supports for linear actuators. Rear view: 80-Left main chassis bar; 81-Pressing gripper; 82-Bullet removal blades; 83-Linear actuator for angling the branch removal blades.

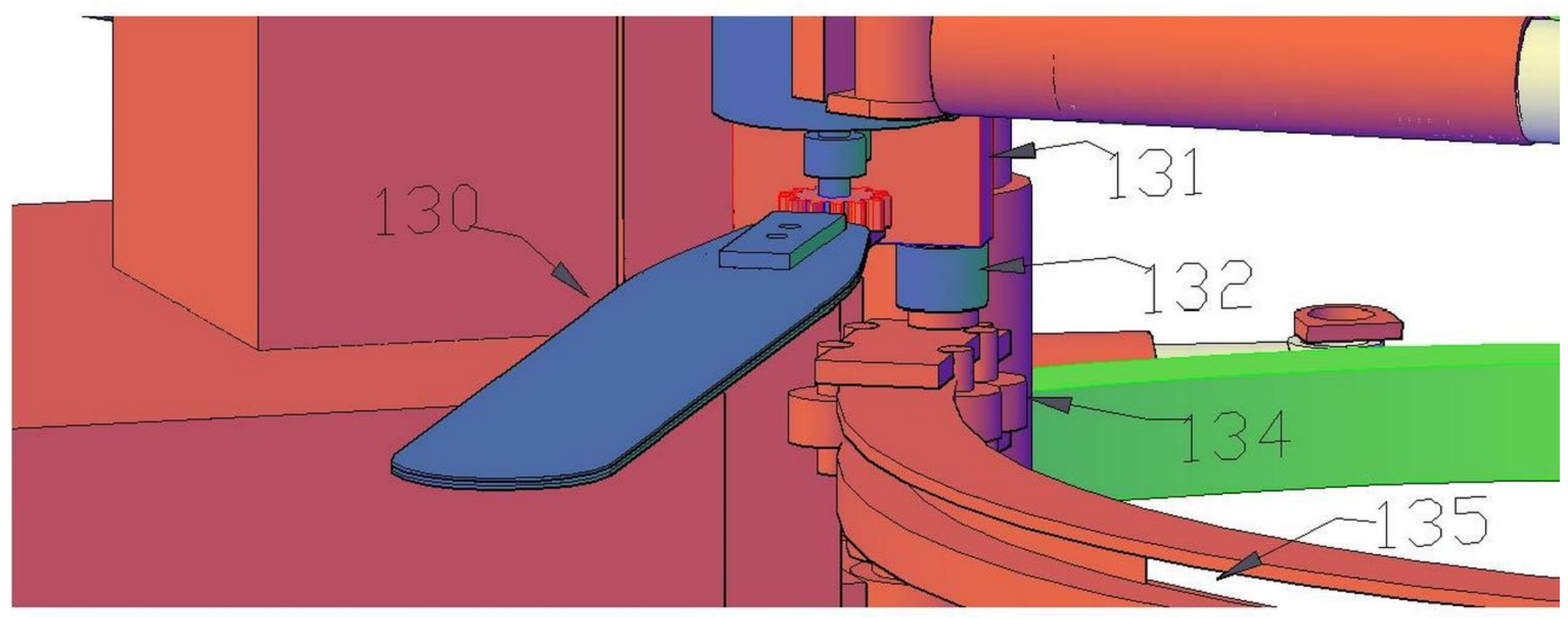

Figure 22. Stem cutting system: 130-Chainsaw bar; 131-High torque and high rotation electric motor; 132-Chainsaw transport trolley; 134-Wheels with electric motors in the hubs; 135-Trolley locomotion rail.

**Harvesting Silvicultural Treatments Module**

Considering the process flowchart, Figure 23, and the morphological matrix and solution synthesis methodologies described by [102,103] used by [126–130], it was possible to find optimal solution paths and synthesize the concept of the HST Module (MHST).

Figure 23. Flowchart of the HST process with a URIEL.

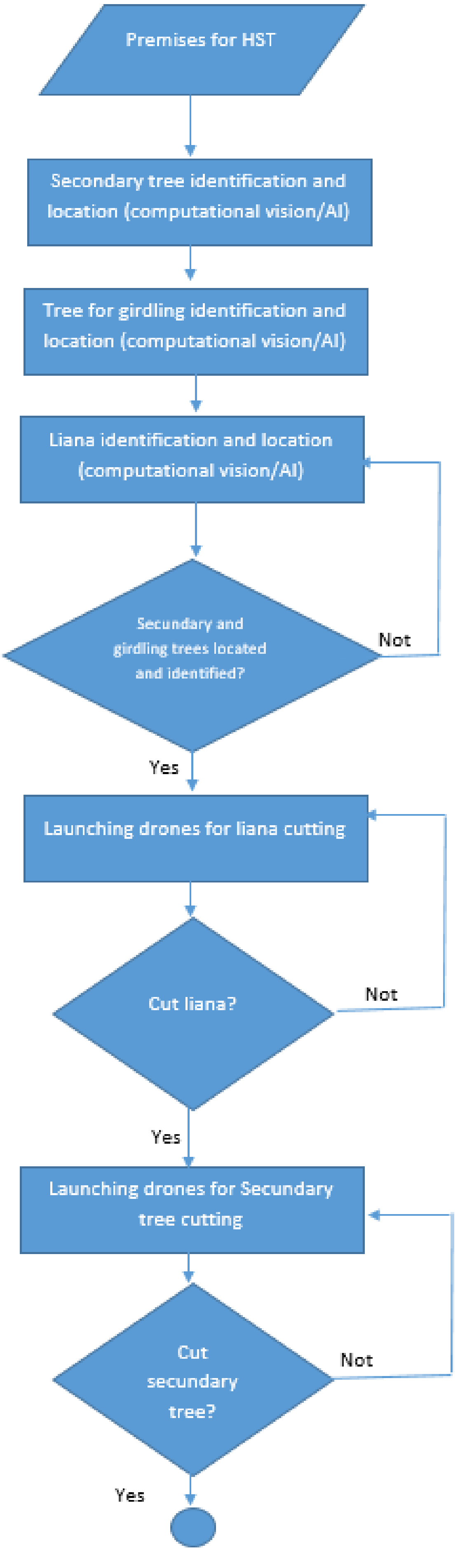
Premises for HST
Secondary tree identification and location (computational vision/AI)
Tree for girdling identification and location (computational vision/AI)
Liana identification and location (computational vision/AI)
Secundary and girdling trees located and identified?
Not
Yes
Launching drones for liana cutting
Cut liana?
Not
Yes
Launching drones for Secundary tree cutting
Cut secundary tree?
Not
Yes

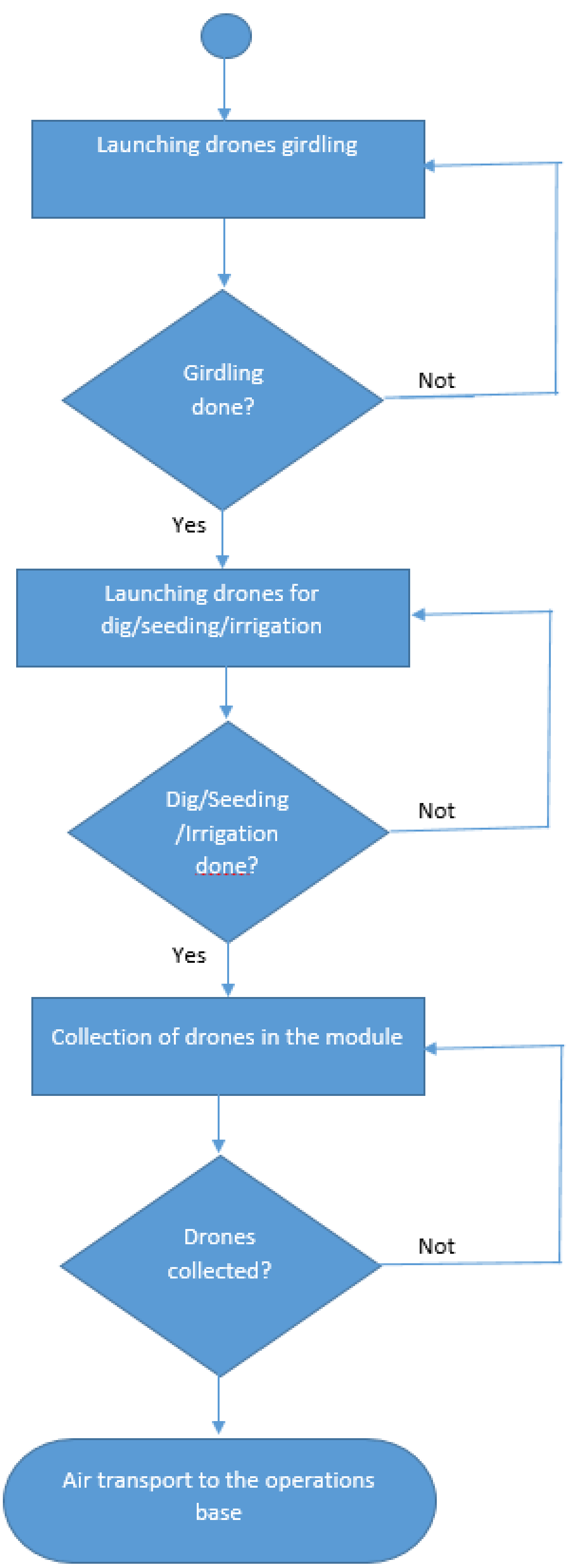
Launching drones girdling
Girdling done?
Not
Yes
Launching drones for dig/seeding/irrigation
Dig/Seeding /Irrigation done?
Not
Yes
Collection of drones in the module
Drones collected?
Not
Air transport to the operations base

The concept of the HST module is based on a drone-carrying system, Figure 24, which transports specific drones responsible for each of the main missions of a Harvest Silvicultural Treatment system: Cutting of lianas; Cutting of small secondary trees; girdling of large secondary trees; preparation of planting pits; irrigation and planting of seedlings.

It is important to emphasize that there are already drones that airlift commercial (TRL9) or advanced-stage (TRL7) systems for several of these proposed HST actions: Drone for cutting lianas [136]; cutting of small secondary trees [137]; making pits [138]; irrigation [139]; planting/sowing [140].

The drone carrier system consists of a hexagonal star-shaped chassis. Each arm of this star has a rail where a locomotion system, consisting of a hitch and wheels, attaches to the star arm. The power source is electric. This hitch supports a coupling mast consisting of a female gripper actuated by linear actuators. Each drone has a male pivot that is suitable for gripping by the female gripper; this pivot is passive and connects to the drone's central structure.

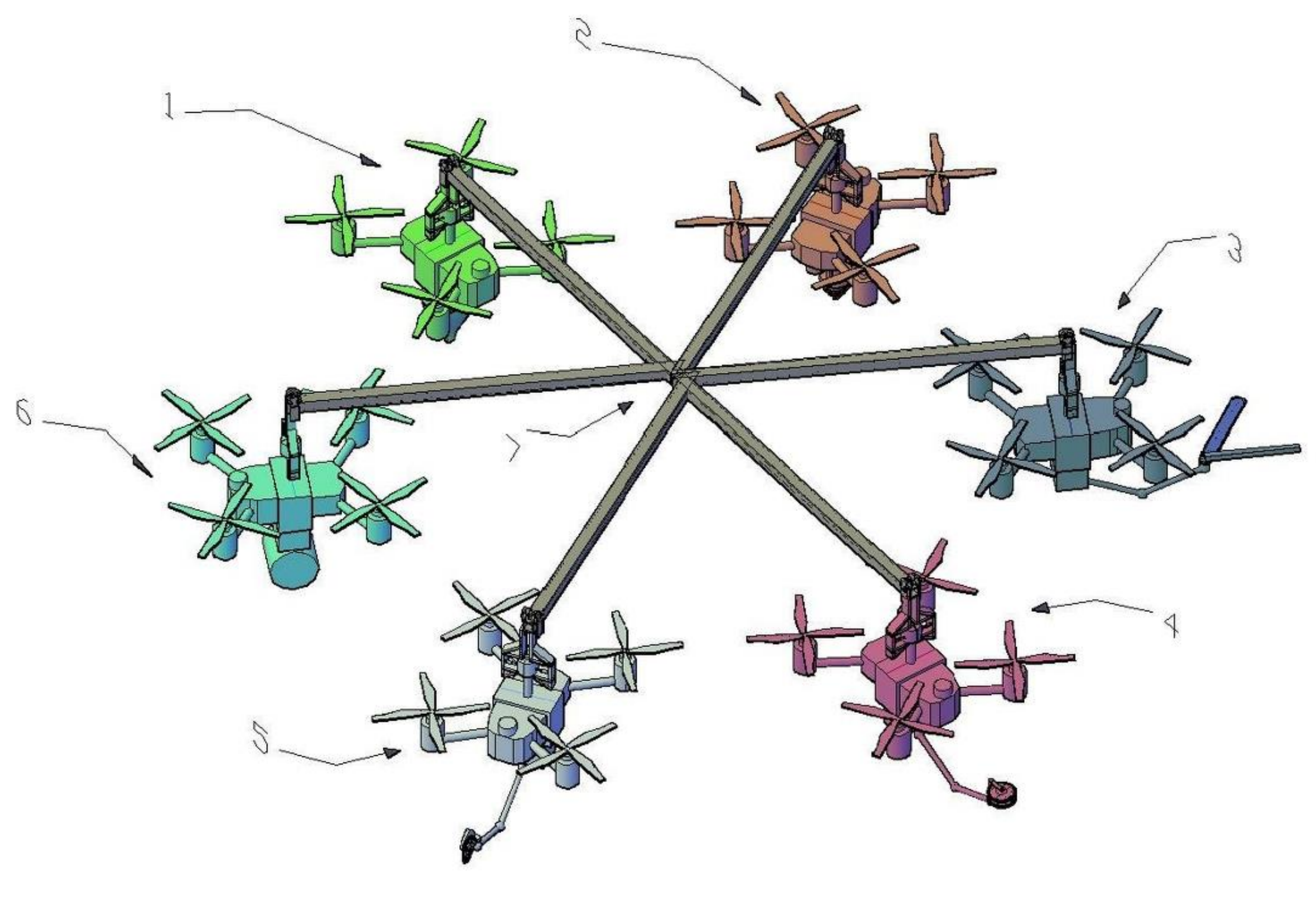

a)

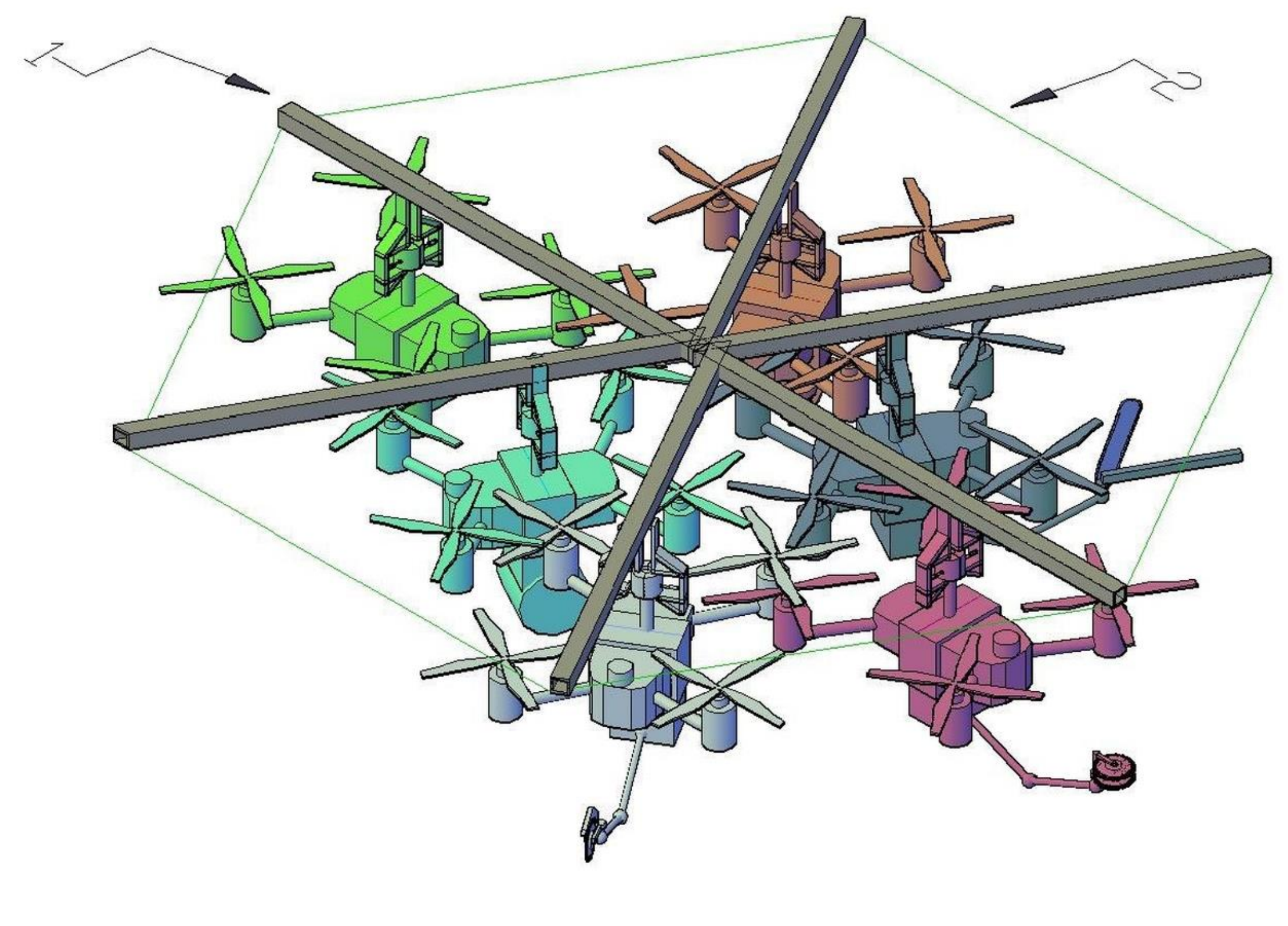

b)

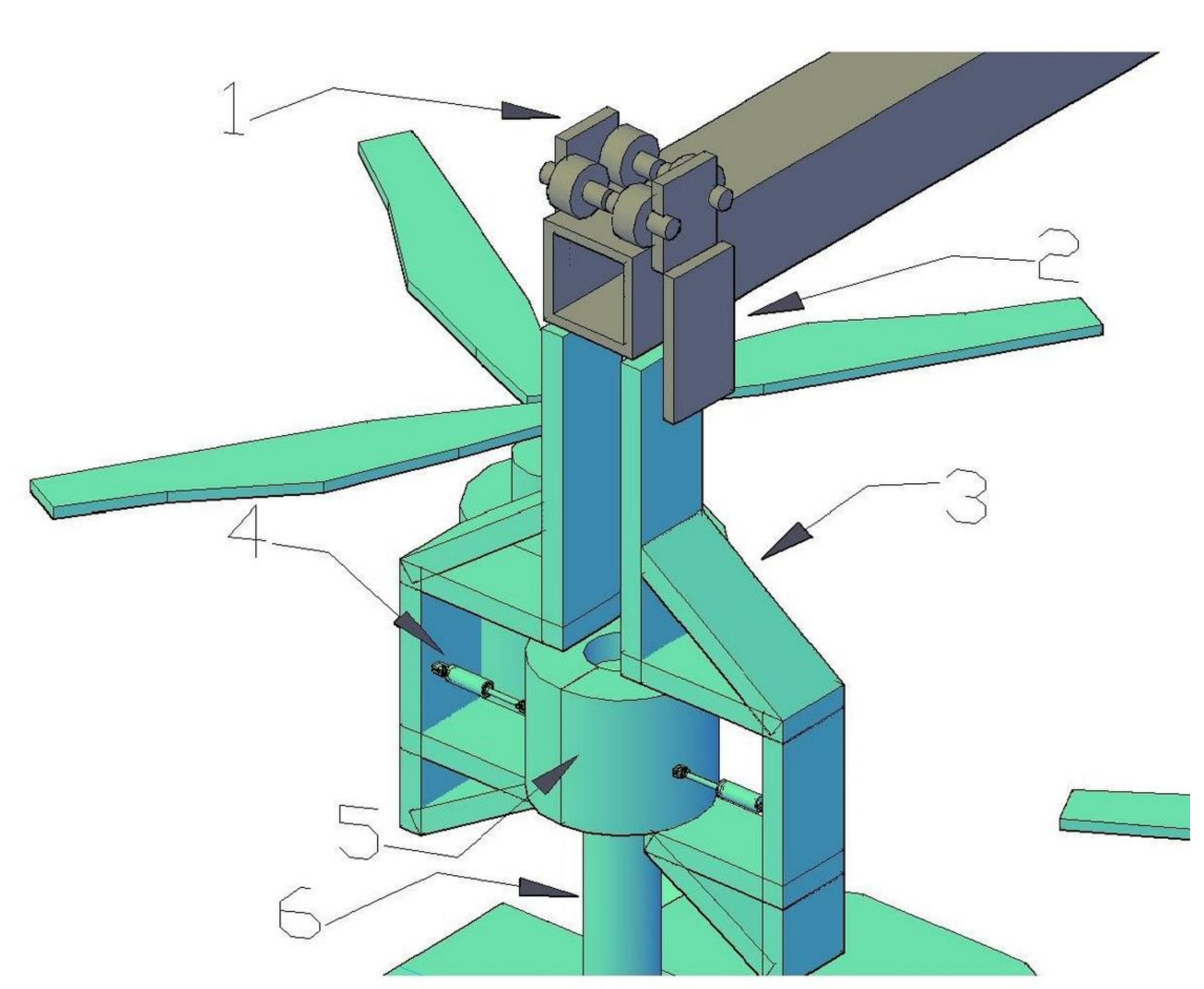

c)

Figure 24. Drone carrier system. Open star a): 1-Planting drone; 2-Filling drone; 3-Liana cutting drone; 4-Girdling drone; 5-Secondary tree cutting drone; 6-Irrigation drone; 7-Hexagonal drone carrier chassis. Closed star b): 1-Hexagonal star chassis; 2-Perimeter of the open star. c) Drone movement system between closed and open positions and vice versa: 1-Wheels with electric motors in the hubs; 2-Movement trolley; 3-Coupling mast chassis; 4-Linear actuator for opening and closing the female coupling head; 5-Female coupling head; 6-Coupling mast fixed to the drone.

The drones are all of the same model (for example, the DJI Agras 70P), specified as a quadcopter with conventional computer vision systems for drones consisting of RGB cameras, radars, and LIDAR, and a GPS-RTK navigation system (the RTK antenna is mounted on the helicopter). The payload is 70 kg and the flight autonomy is 15 minutes. Each drone can have a robotic arm with 9 degrees of freedom diagrammed as two links (arm and forearm) and three joints (rotations in x, y, and z). For each HST mission there is a different system attached to the drone (robotic arm or structure on the fuselage) namely:

a) Liana cutting drone (DCL), Figure 25: there is a terminal tool attached to the robotic arm consisting of a chainsaw bar and its respective cutting chain, this bar mounted on a cylindrical support that houses a high-speed electric motor that drives a gear that moves the cutting chain. On the same support there is a counter-blade consisting of a rigid rectangular support located geometrically just above the bar. There is a servomotor coupled between the saber/gear/motor system and the cylindrical support in such a way that, under command, controlled rotation occurs in terms of angle and angular velocity of the saber on the cylinder axis, generating a mixed cutting action (cutting chain plus counter-knife). This action is important considering the specificities of lianas entangled between trees and between other lianas. The combined cutting action causes a bending moment between the biological elements in the form of a "rope" and the cutting system, thus enabling the cutting of loose elements without a support base, such as lianas suspended in trees.

b) Drone for cutting small secondary trees (DCA), Figure 26: there is a terminal tool coupled to the robotic arm consisting of a circular saw with Wydia teeth and a support cylindrical saw. This circular saw/support cylindrical saw assembly is mounted on a cylindrical support. The rotation of the circular saw occurs due to a high-speed motor coupled to the hub of the circular saw and locked onto the fork of the circular saw shaft. In the cylindrical support there is a servomotor coupled in such a way that, under command, controlled rotation occurs in terms of angle and angular speed of the circular saw/support hammer assembly.

c) Drone for ringing large secondary trees (DAN), Figure 27: there is a terminal tool coupled to the robotic arm consisting of two circular saws with Wydia teeth arranged in parallel and between them a movable debarking blade, this whole assembly is mounted on a cylindrical support. The rotation of the circular saws occurs due to a high-speed motor coupled to the hub of the circular saws and locked onto the fork of the circular saw shaft. In the cylindrical support there is a servomotor coupled in such a way that, under command, controlled rotation occurs in terms of angle and angular speed of the assembly. The debarking blade is positioned so that its cutting edge aligns 5 millimeters with the base of the circular saw teeth, ensuring that when the saws perform a double cut, the blade immediately penetrates the resulting bark strip. A high-torque servomotor mounted on the saw shaft block generates a penetration action of this blade into the doubly sectioned bark. In this action, the robotic arm maneuvers, in conjunction with the drone's positioning and orientation, to circle the tree trunk, thus performing the ring strip cut by the saws and immediately on-the-go debarking of this ring by the debarking blade. Of all the operations planned for the MHST, this is by far the most complex and the only one for which no system exists in the available literature.

d) Hole-digging drone (DCO), Figure 28: the hole-digging drone does not have a robotic arm. The hole-digging system consists of a treated alloy steel helical conical drill bit that is coupled to a shaft connected to a geared motor system driven by an electric motor. The geared motor assembly generates high torque with the medium/low rotations necessary for making holes in hard soil. The entire assembly is mounted on a highly shear- and bend-resistant, yet lightweight, aluminum alloy cylinder tube. This structure is attached to the drone via a lower annular support and four support rods that are fixed to a chassis attached to the drone's fuselage.

e) Planting drone (DPL), Figure 29: the planting drone does not have a robotic arm. The planting system is a commercial system developed by the company AirSeeds Technologies and consists of an encapsulated seed ejector module. These seeds are stored in a reservoir attached to the ejector module, and this assembly is attached to the drone via rods that connect the module to the drone's support chassis.

f) Irrigation Drone (DIR), Figure 30: the irrigation drone does not have a robotic arm. The irrigation system consists of a water storage tank (40 liters) that has a conductor tube where at the end there is an interchangeable liquid fragmentation/jet launch assembly, thus there is the choice of spraying water in droplets through a conventional atomizer system or launching a continuous jet of water. The hydraulic pump is housed in a box above the tank which, in addition to the pump, also carries lithium batteries. This supplementary power module is necessary due to the energy consumption of the pump and the irrigation system, since if only the drone's batteries were used, its autonomy would be greatly reduced. The entire assembly is attached to the drone's fuselage by rectangular brackets and bolted to the drone's chassis beams, given the assembly's considerable weight and its dynamic actions during flight maneuvers.

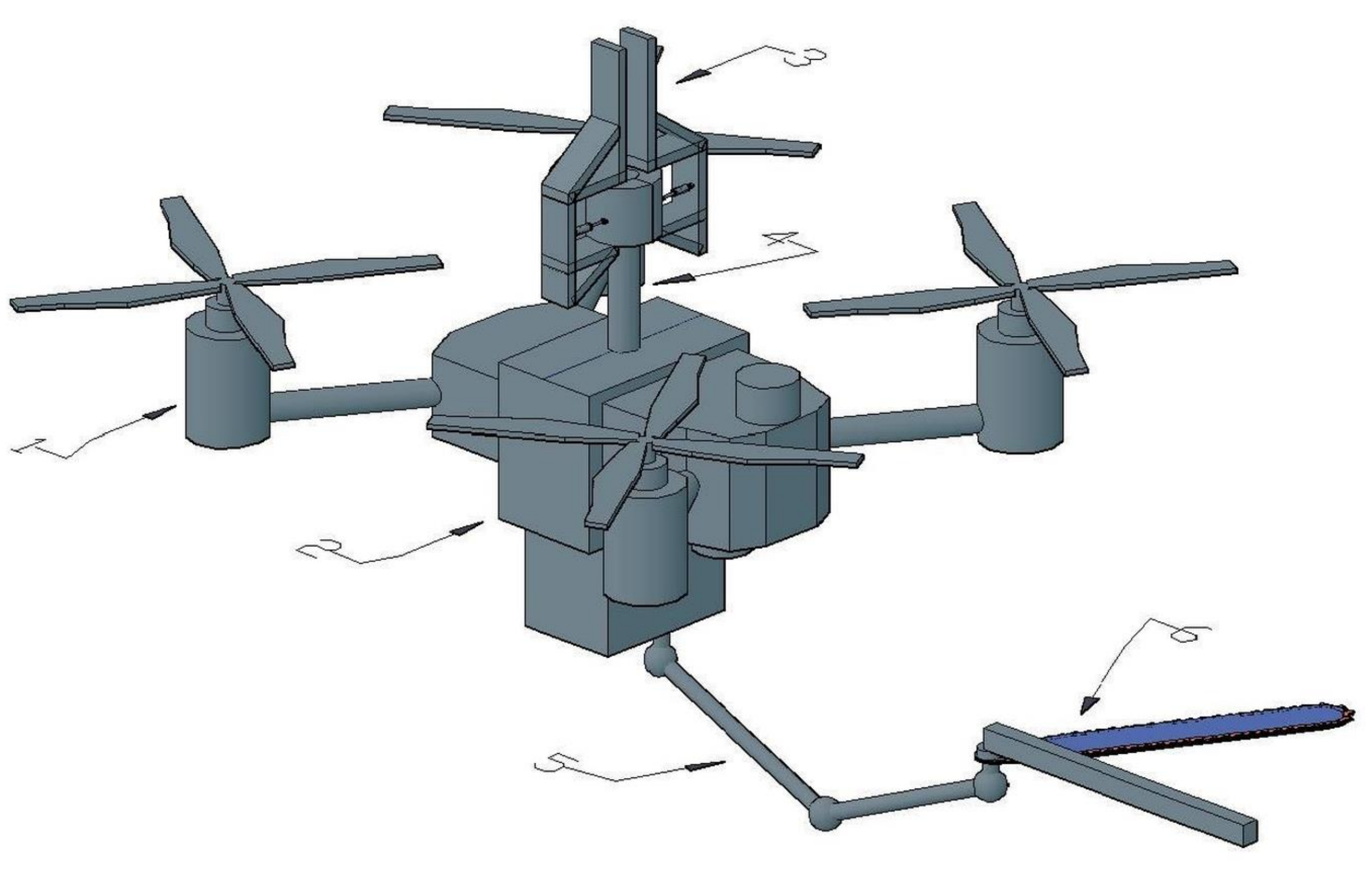

a)

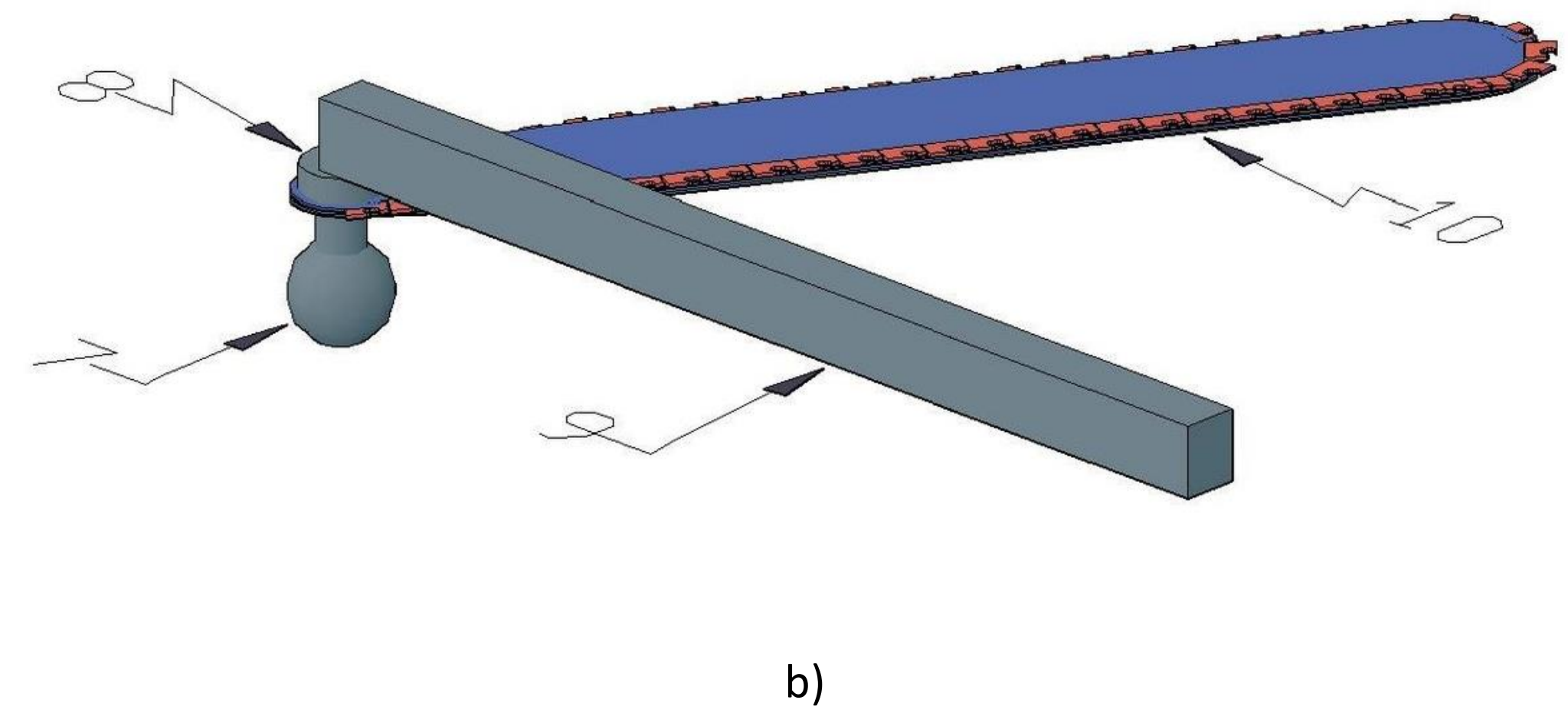

b)

Figure 25. Vine cutting drone. Front view a): 1-Electric motor; 2-Drone; 3-Coupling mast; 4-Coupling mast fixed to the drone; 5-9-axis robotic arm; 6-Vine cutting system by chainsaw. Detail b): 7-Robotic arm ball joint 3; 8-Circular electric actuator for blade deflection; 9-Vine cutting counter-blade; 10-Chainsaw; 11-High torque, high rotation electric motor driving the cutting chain.

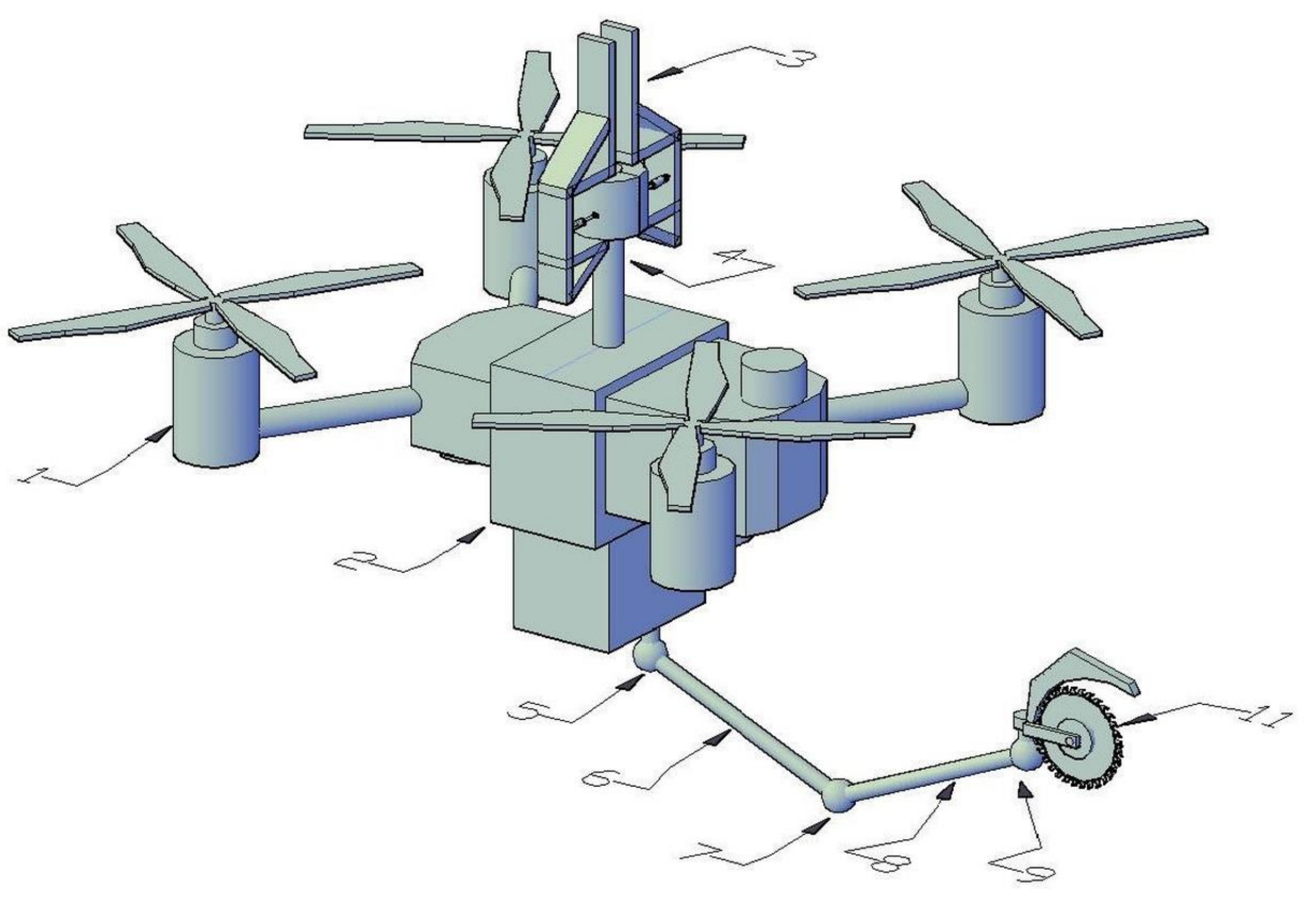

a)

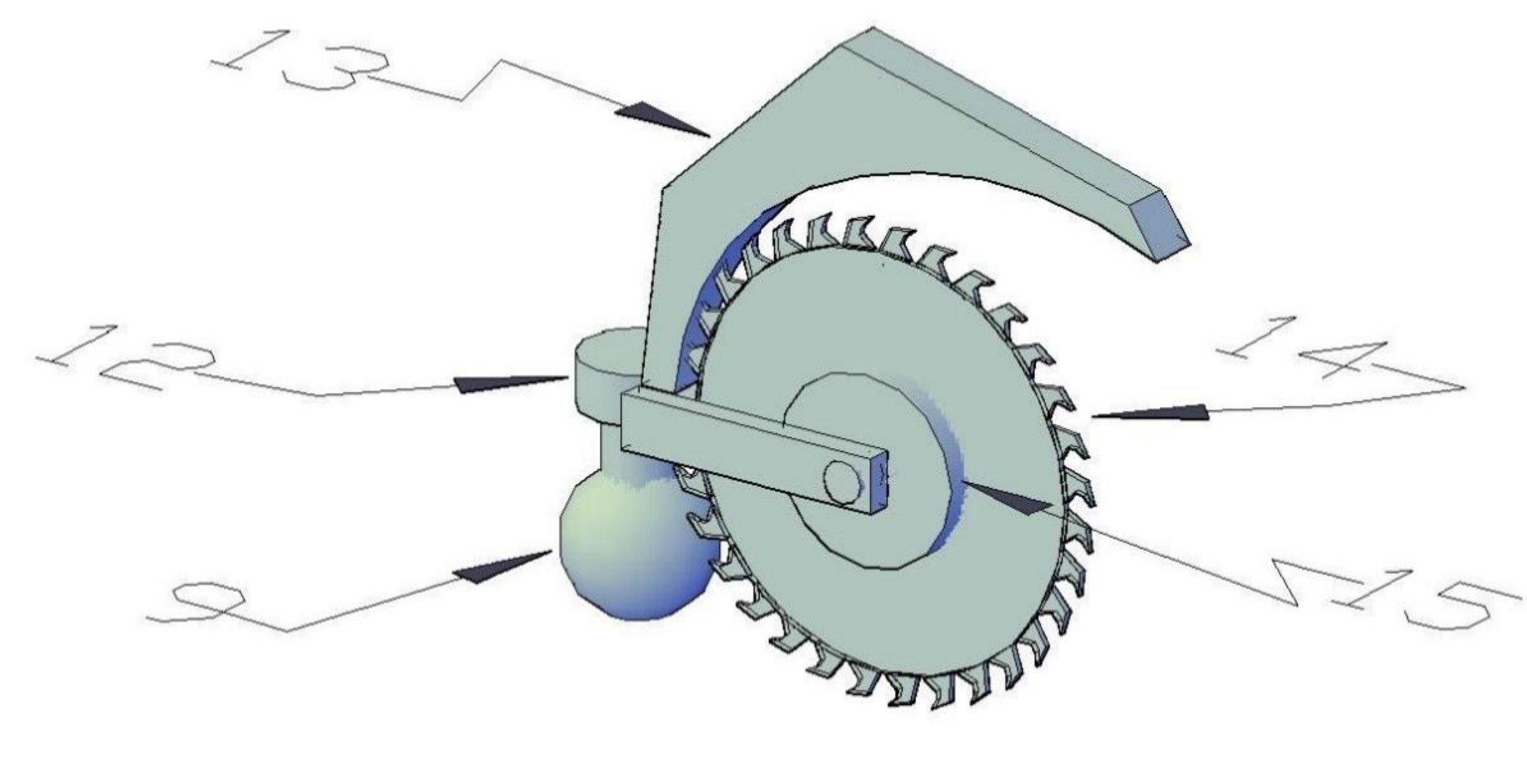

b)

Figure 26. Secondary tree cutting drone. Front view a): 1-Electric motor; 2-Drone; 3-Coupling mast; 4-Coupling mast fixed to the drone; 5-Rotary arm ball joint 1 (9-axis); 6-First link of the robotic arm; 7-Rotary arm ball joint 2; 8-Second link of the robotic arm; 9-Third ball joint of the robotic arm; 11-Secondary tree cutting system using a circular saw. Cutting system detail b): 9-Third ball joint of the robotic arm; 12-Electric circular actuator for circular saw deflection; 13-Cutting hammer; 14-Wydia circular saw; 15-High torque and high rotation electric motor integrated into the circular saw hub.

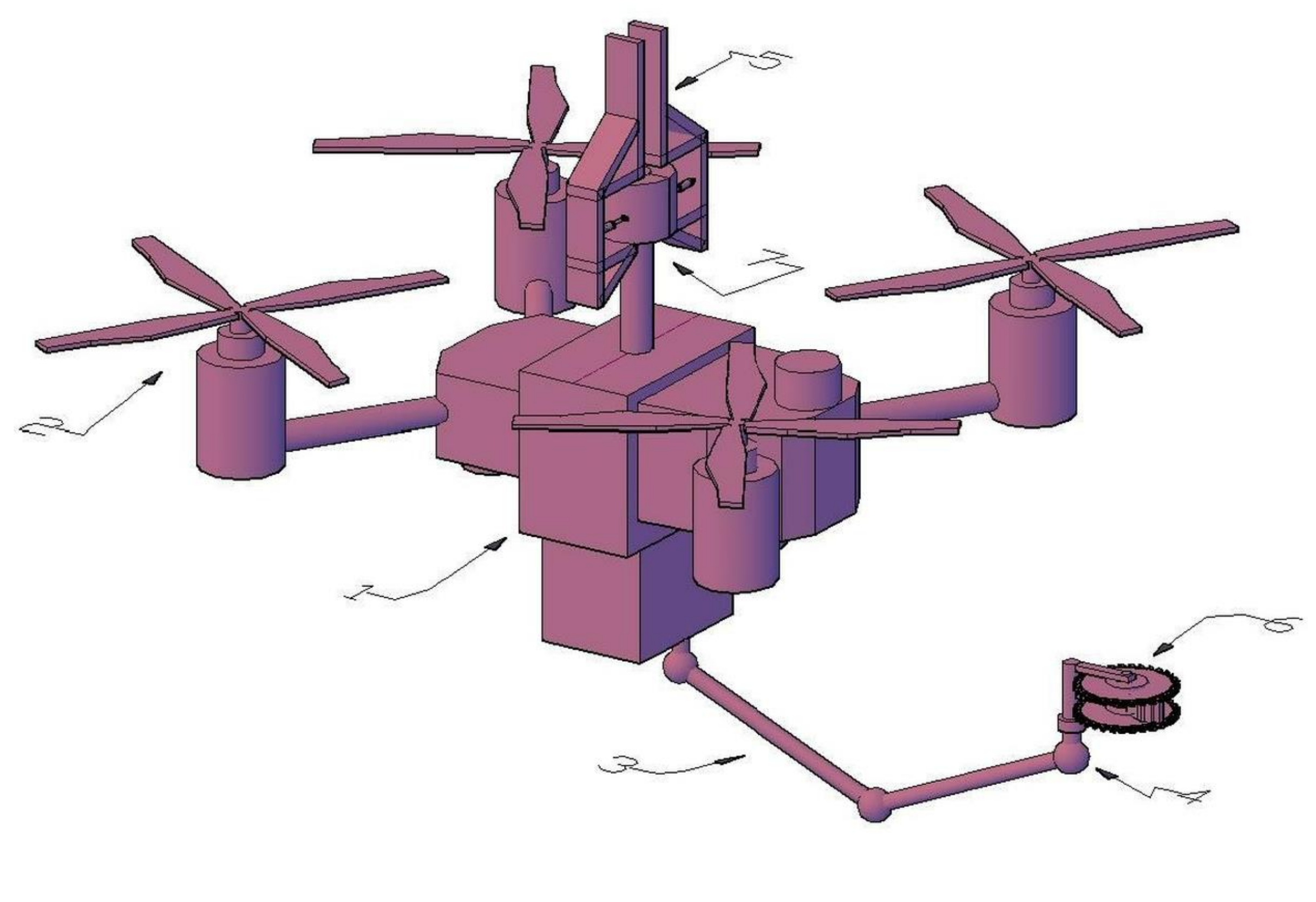

a)

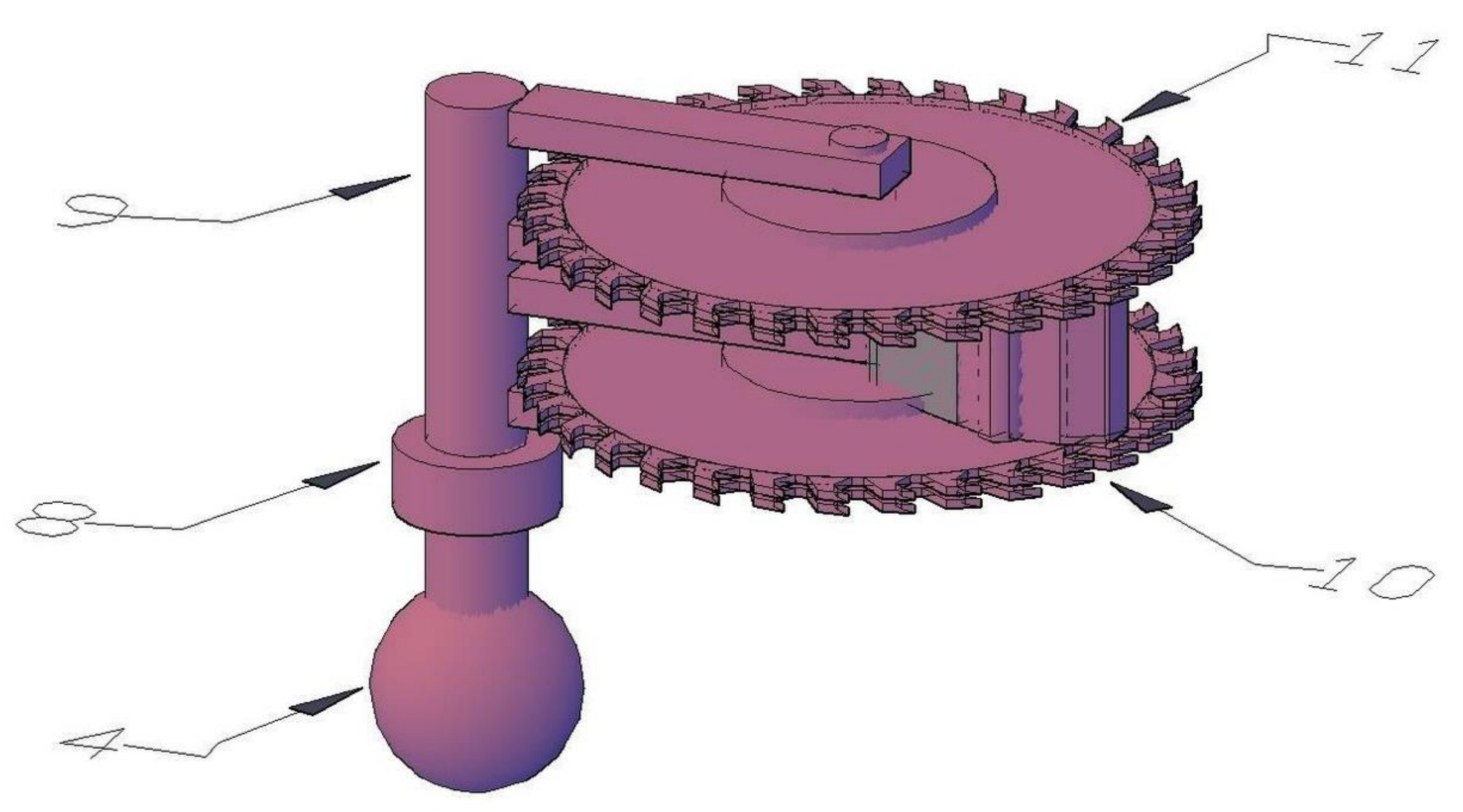

b)

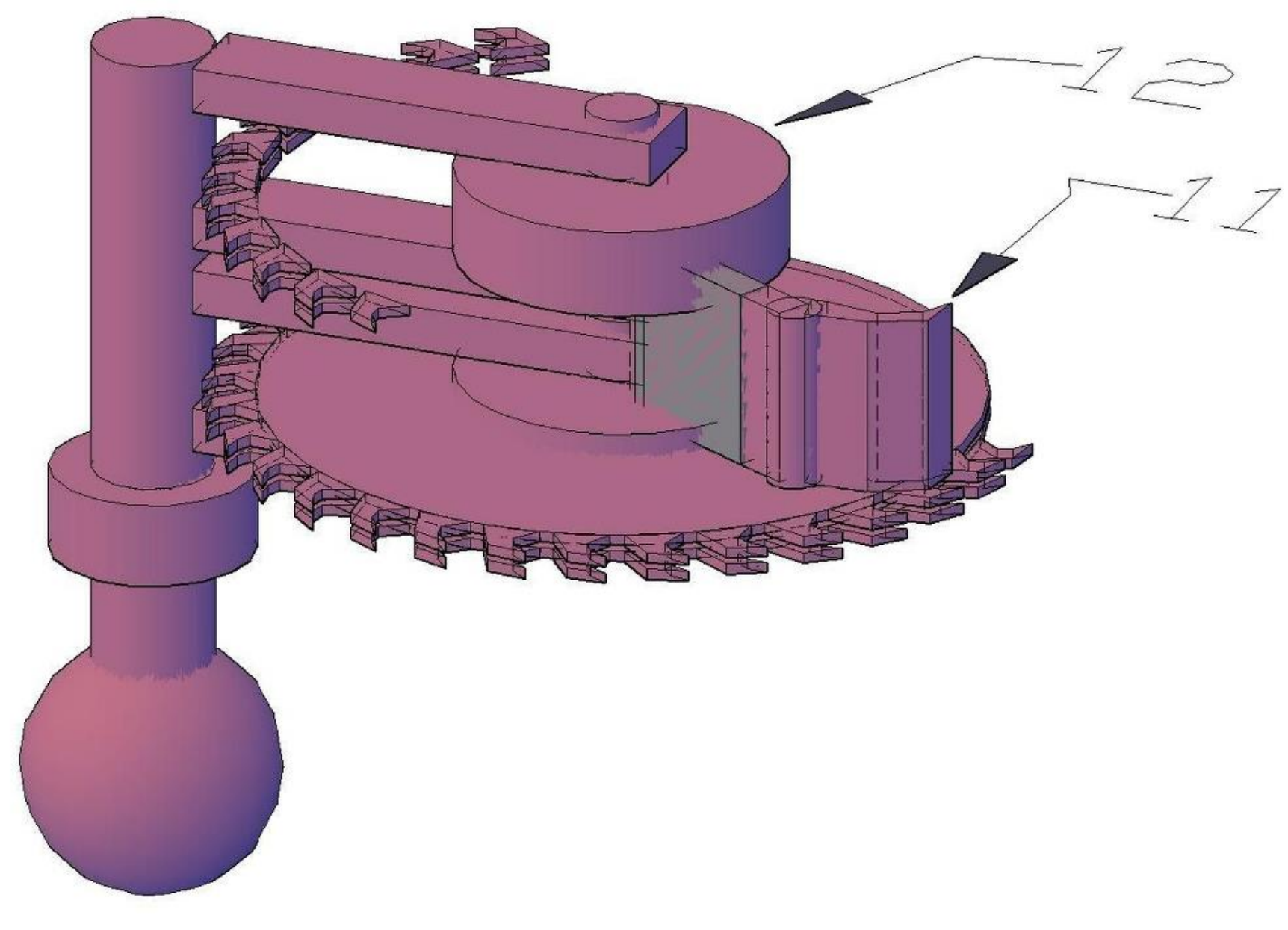

c)

Figure 27. Girdling (ringing) drone. Front view a): 1-Electric motor; 2-Drone; 3-9-axis robotic arm; 4-Third ball joint of the robotic arm; 5-Ringing system; 6-Coupling mast; 7-Coupling mast fixed to the drone. Detail of the ringing system b): 4-Third ball joint of the robotic arm; 8-Circular electric actuator for deflecting the ringing saws; 9-Support chassis for the ringing saws; 10-Lower ringing saw; 11-Upper ringing saw. Cutaway view of the ringing system c): 11-Ringing stripping blade; 12-Electric motor integrated into the hub of the high-torque, high-speed ringing saws.

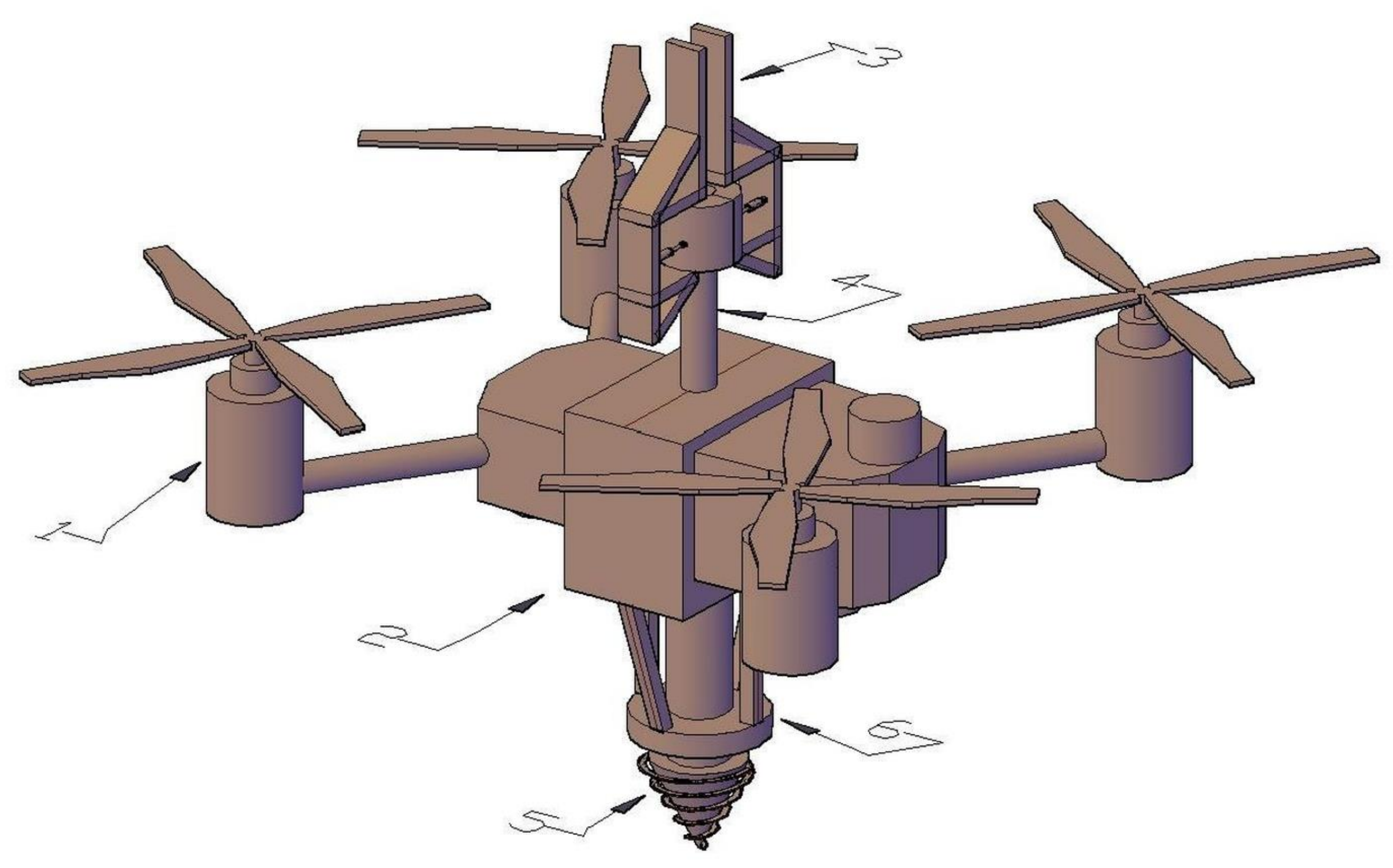

Figure 28. Burrowing drone. Front view a): 1-Electric motor; 2-Drone; 3-Coupling mast; 4-Coupling mast fixed to the drone; 5-Helical borehole drill; 6-Borrowing chassis.

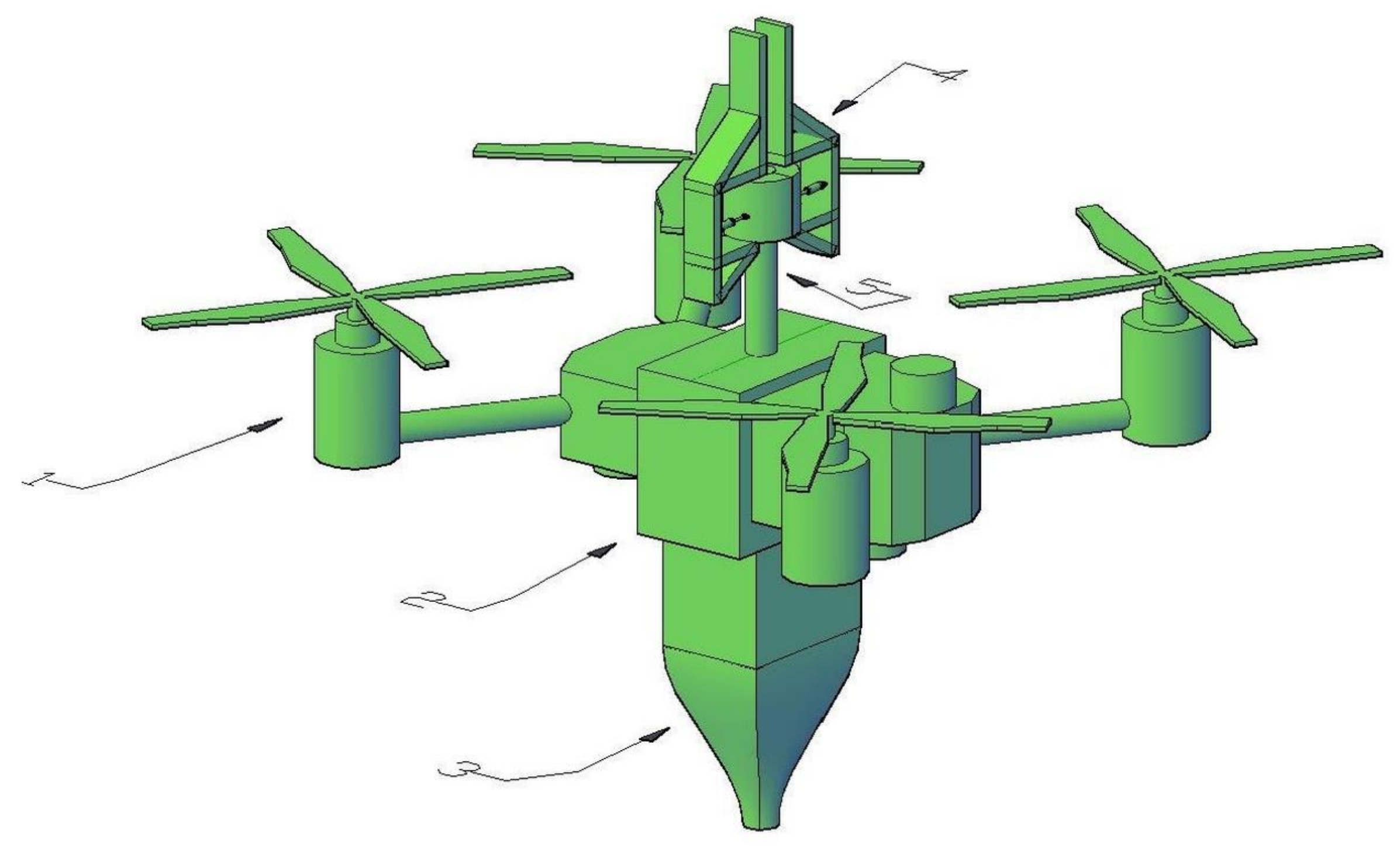

Figure 29. Planting drone. Front view a): 1-Electric motor; 2-Drone; 3-Planting module; 4-Coupling mast; 5-Coupling mast fixed to the drone.

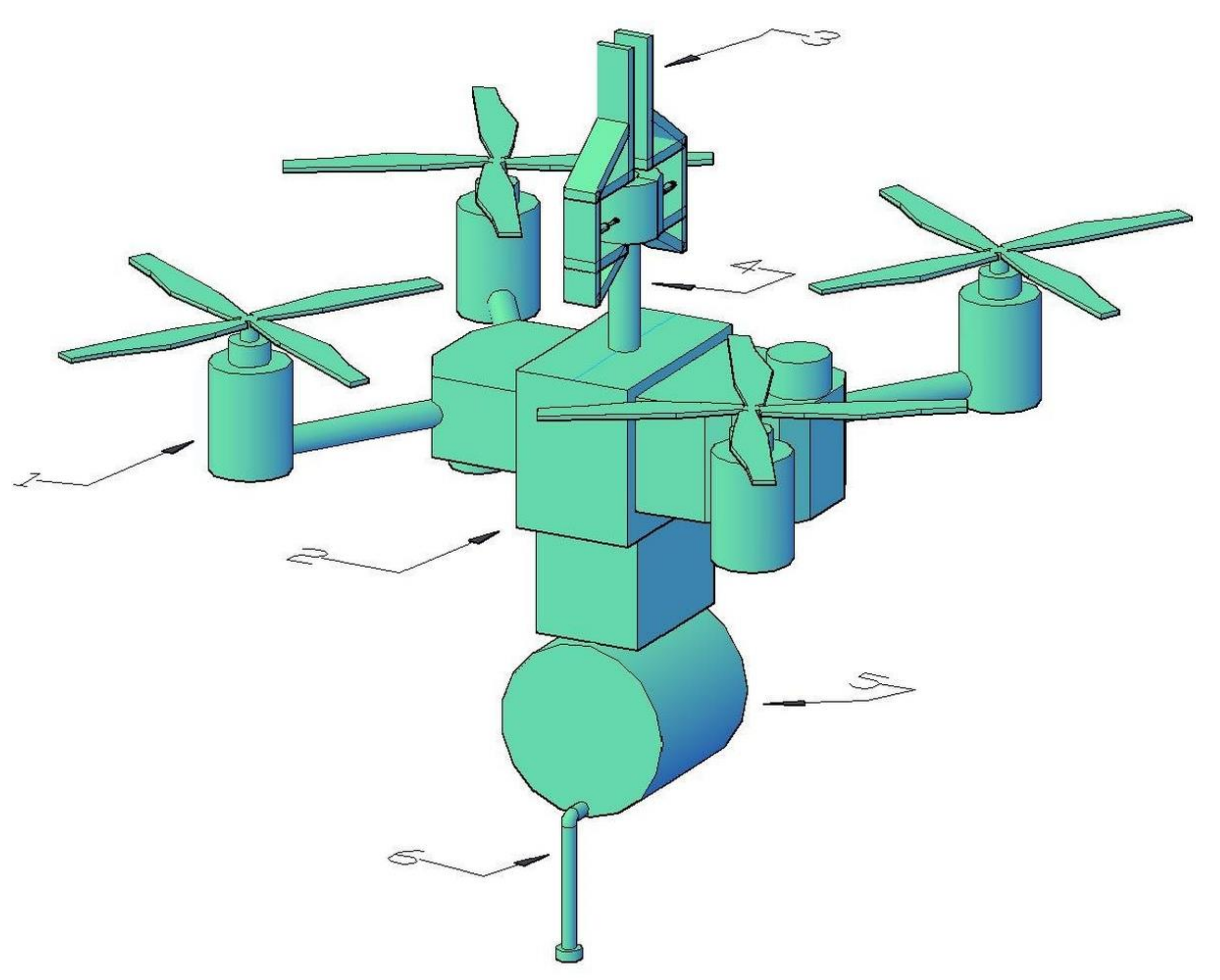

Figure 30. Irrigation drone. Front view a): 1-Electric motor; 2-Drone; 3-Coupling mast; 4-Coupling mast fixed to the drone; 5-Water tank; 6-Spray tube or solid jet.

**Pod URIEL**

Harvesting modules (MH) and silvicultural treatment modules (MHST) are large pieces of equipment with elements that can be damaged if they impact tree branches, birds, or even due to the aerodynamic drag force from air displacement in helicopters flying at more than 200 km/h. Thus, it is necessary to design a transport pod that must be fitted ventrally to the helicopter fuselage.

As inspiration for aircraft coupling, aerodynamic geometry, and structural characteristics, the AN/ALQ-131 ECM (Electronic Counter-Measures) pod used by the McDonnell Douglas RF-4C Phantom II reconnaissance fighter was used [141]. The transport pod, Figure 31, was structured with a chassis of stringers in aeronautical aluminum beams, and its fuselage follows an aerodynamic geometry to avoid excessive drag. It consists of two continuous sections with different dimensions, one for the MH system and the other for the MHST. The section that houses the MH system contains the Uriel System command post, Figure 34, which is operated by a person who manages all the activities of the two modules, the ground movement operations of the Pod, and also the final approach of the helicopter to the target. This section is closed by ventral and movable sealing hatches. The MHST section follows right behind the MH section and is ventrally open.

Due to the size of the MH and MHST modules and their longitudinal arrangement within the transport Pod, it was necessary to design a landing gear system, Figure 32, because the transport Pod must be coupled and uncoupled to the helicopter while it is hovering over the Pod. This is because its dimensions prevent this coupling with aircraft on the ground. This system consists of two sets, one front (next to the MH module) and one rear (next to the MHST module). Each set is formed by a tandem system that supports two axles, each axle with 2 wheels. Each set is directional and has electric traction motors coupled to the wheel hubs. The power source for moving the Pod by the wheels and also for steering the tandems is external, consisting of a lithium battery pack that is towed at the rear of the Pod.

The URIEL System operator: When the Pod is positioned at the coupling point with the helicopter and the helicopter approaches in hovering flight over the Pod, the ground crew couples the Pod to the helicopter by the front, intermediate, and rear pylons. At this moment, the Pod's electrical system is coupled to the helicopter's electrical power source, and the ground crew disconnects the external source and removes it from the flight path.

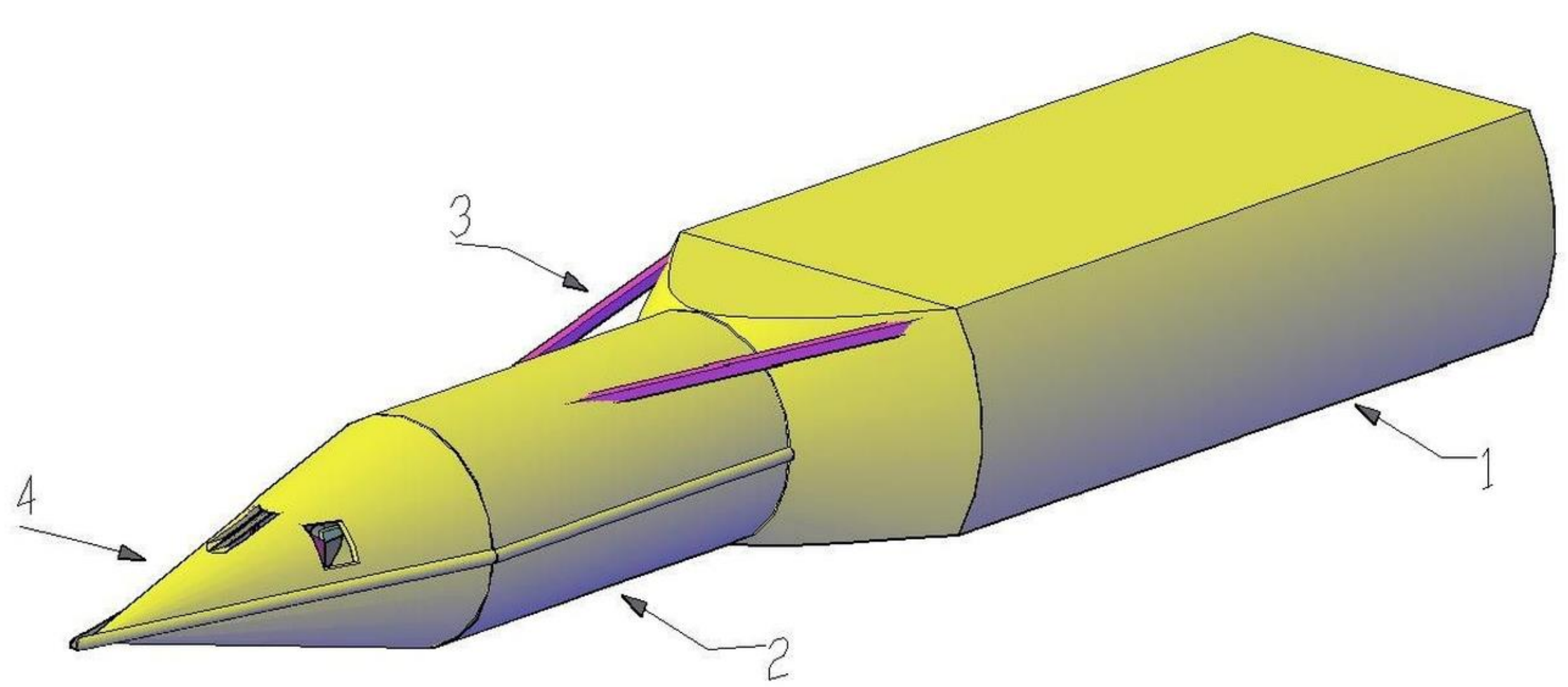

a)

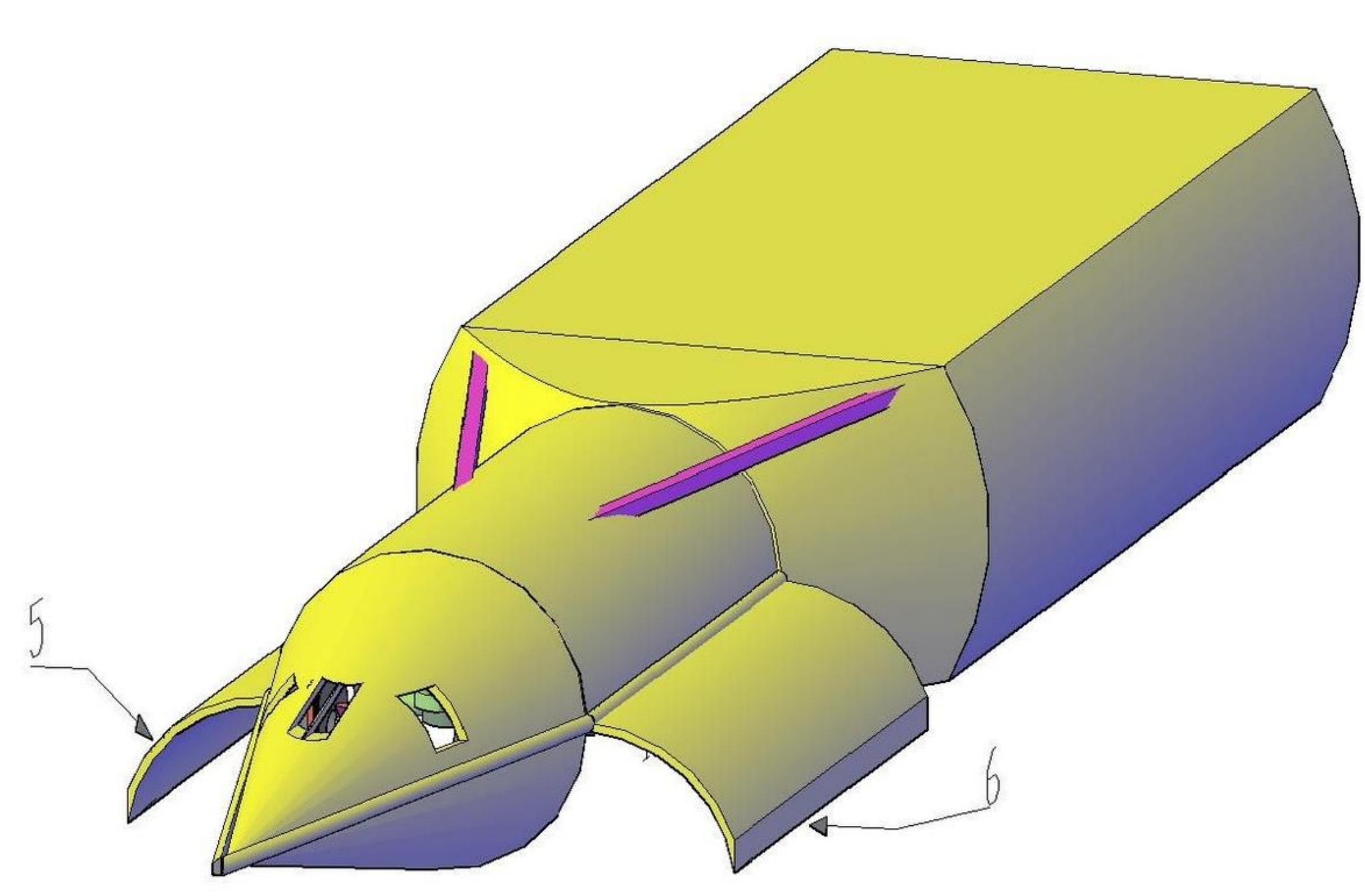

b)

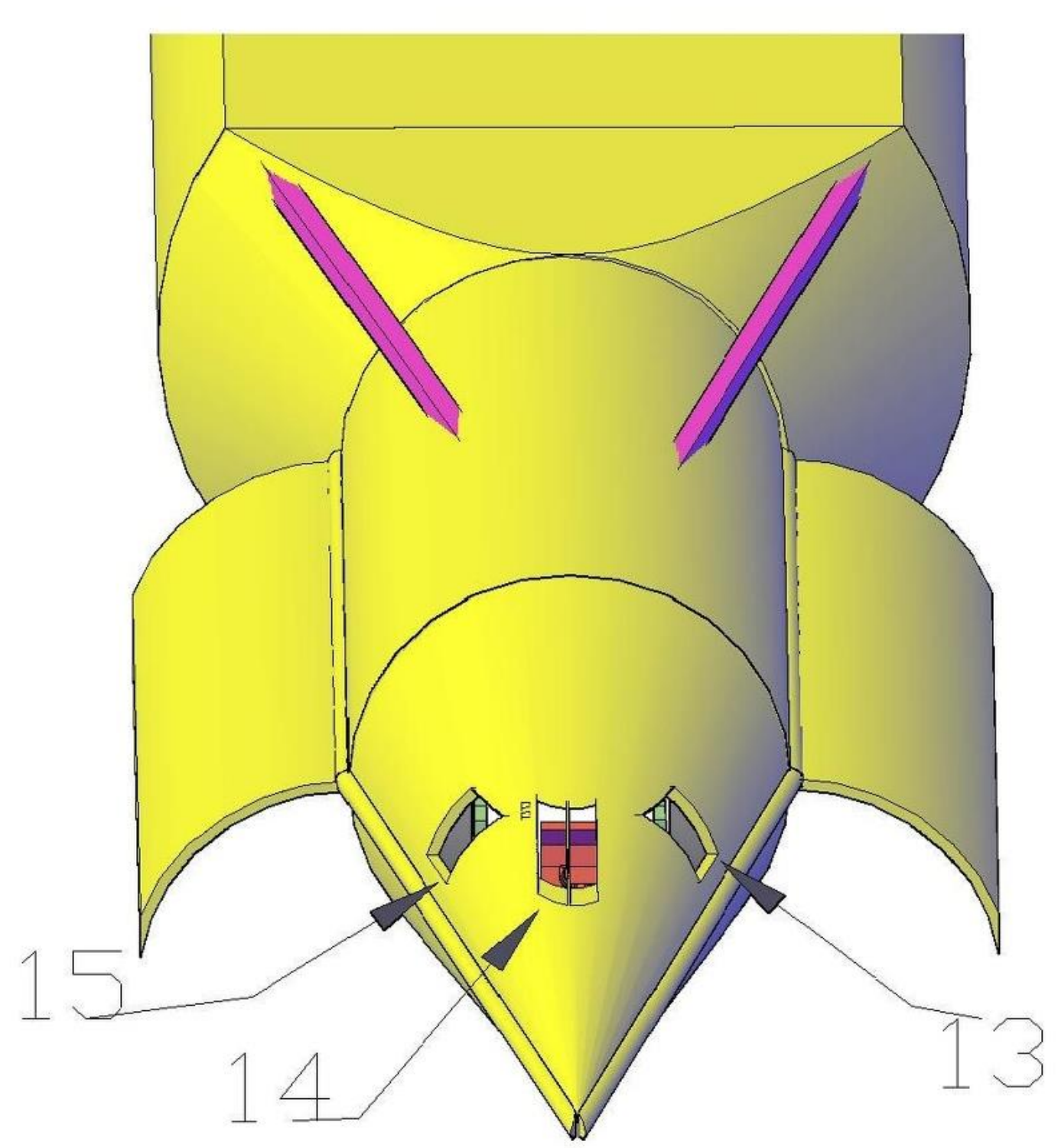

c)

Figure 31. URIEL transport pod. Front view a): 1-MHST transport section; 2-MH transport section; 3-Structural chassis stringer; 4-Aerodynamic cone. Sealing gates b): 5-Right gate; 6-Left gate. Cone view c): 13-Left window; 14-Front window; 15-Right window.

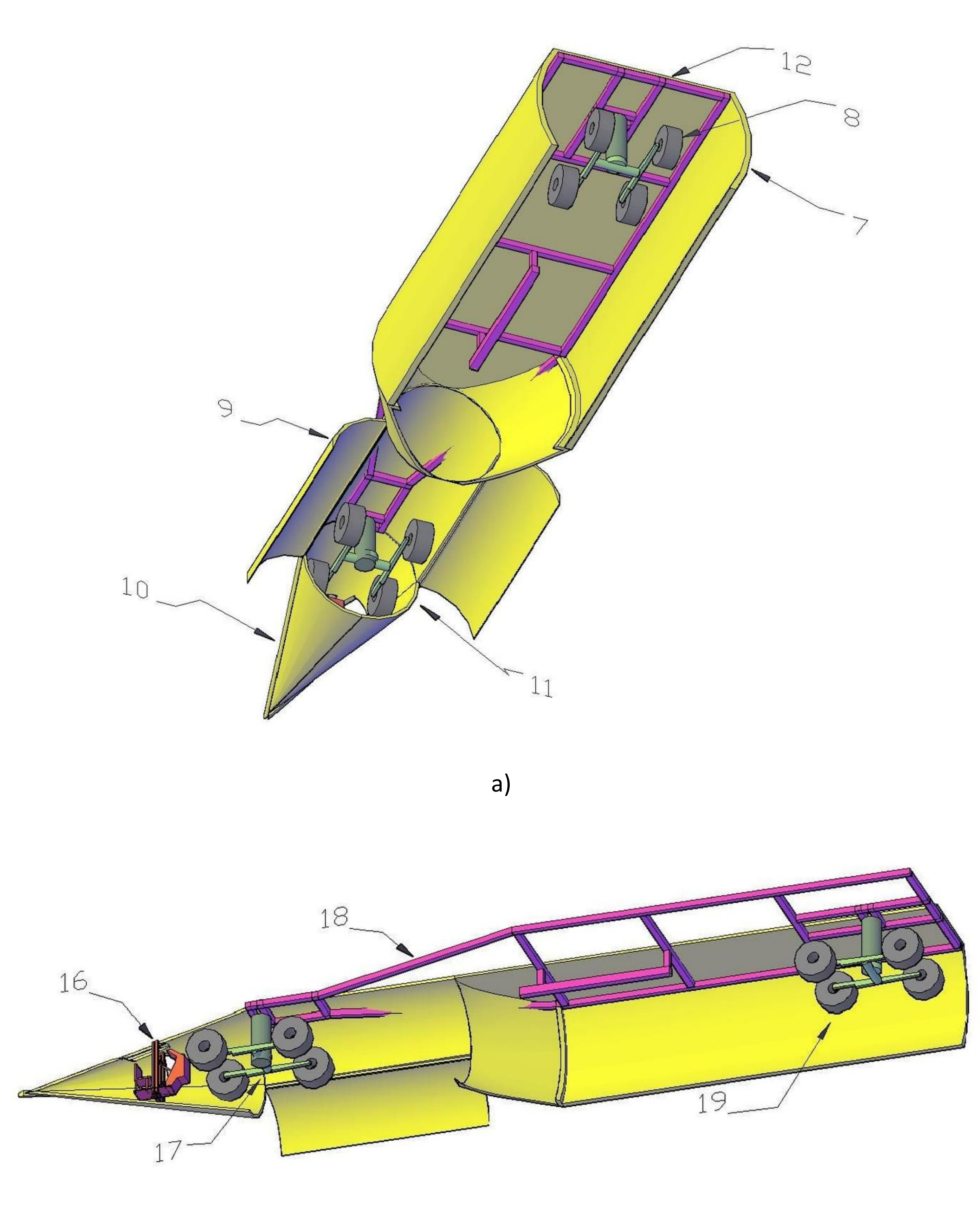

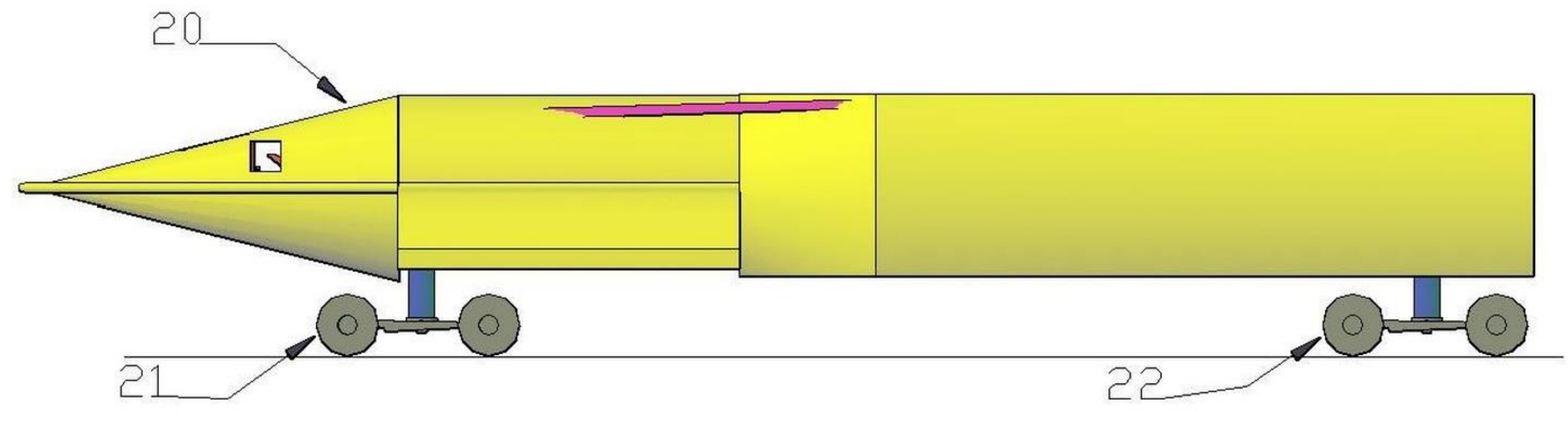

c)

Figure 32. URIEL transport pod. Ventral view a): 7-Open fuselage, rear section of the transport pod; 8-Rear landing gear; 9-Right hatch; 10-Aerodynamic cone; 11-Front landing gear; 12-Chassis, rear section. Structural detail of the pod b): 16-URIEL system command post, note that it is oriented towards the modules; 17-Front landing gear retracted; 18-Chassis spar; 19-Rear landing gear retracted. URIEL landing configuration c): 20-URIEL pod fuselage; 21-Front landing gear extended; 22-Rear landing gear extended.

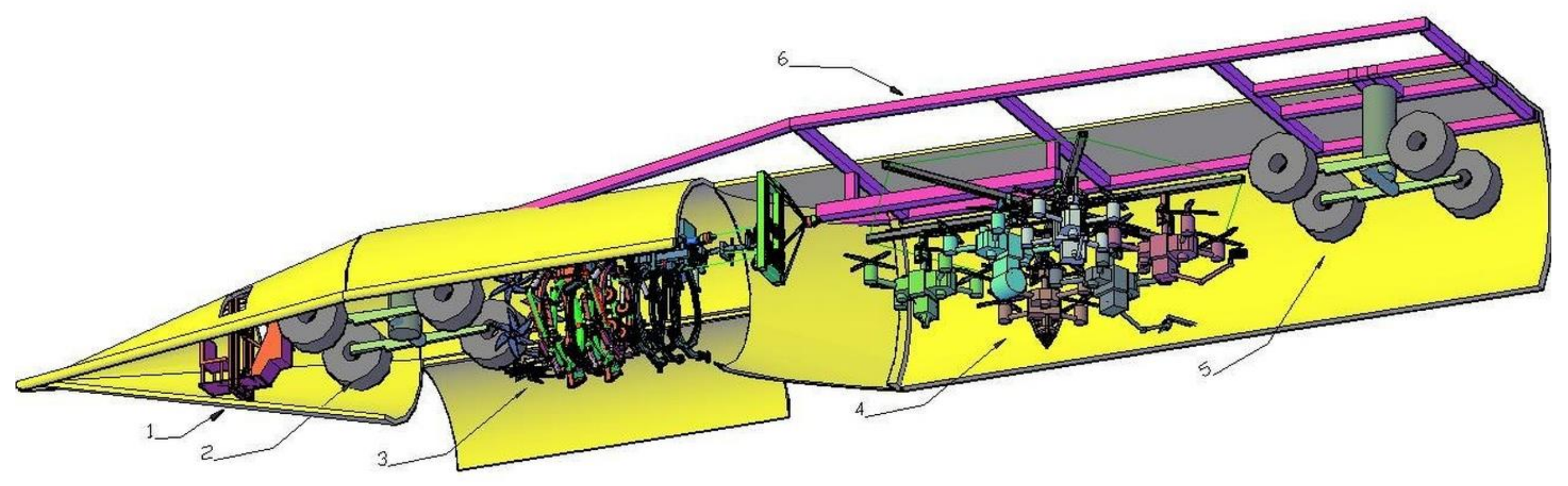

a)

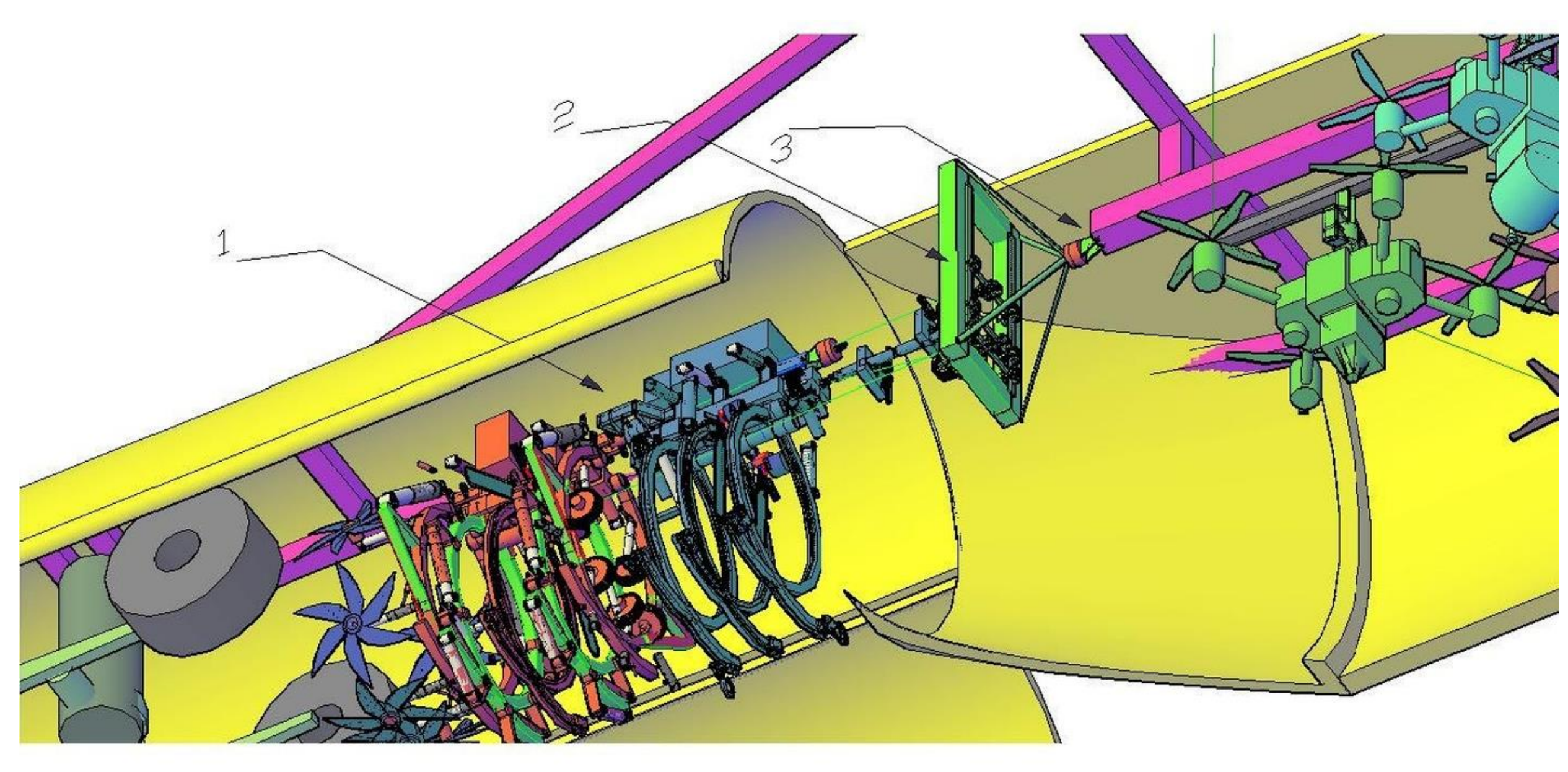

b)

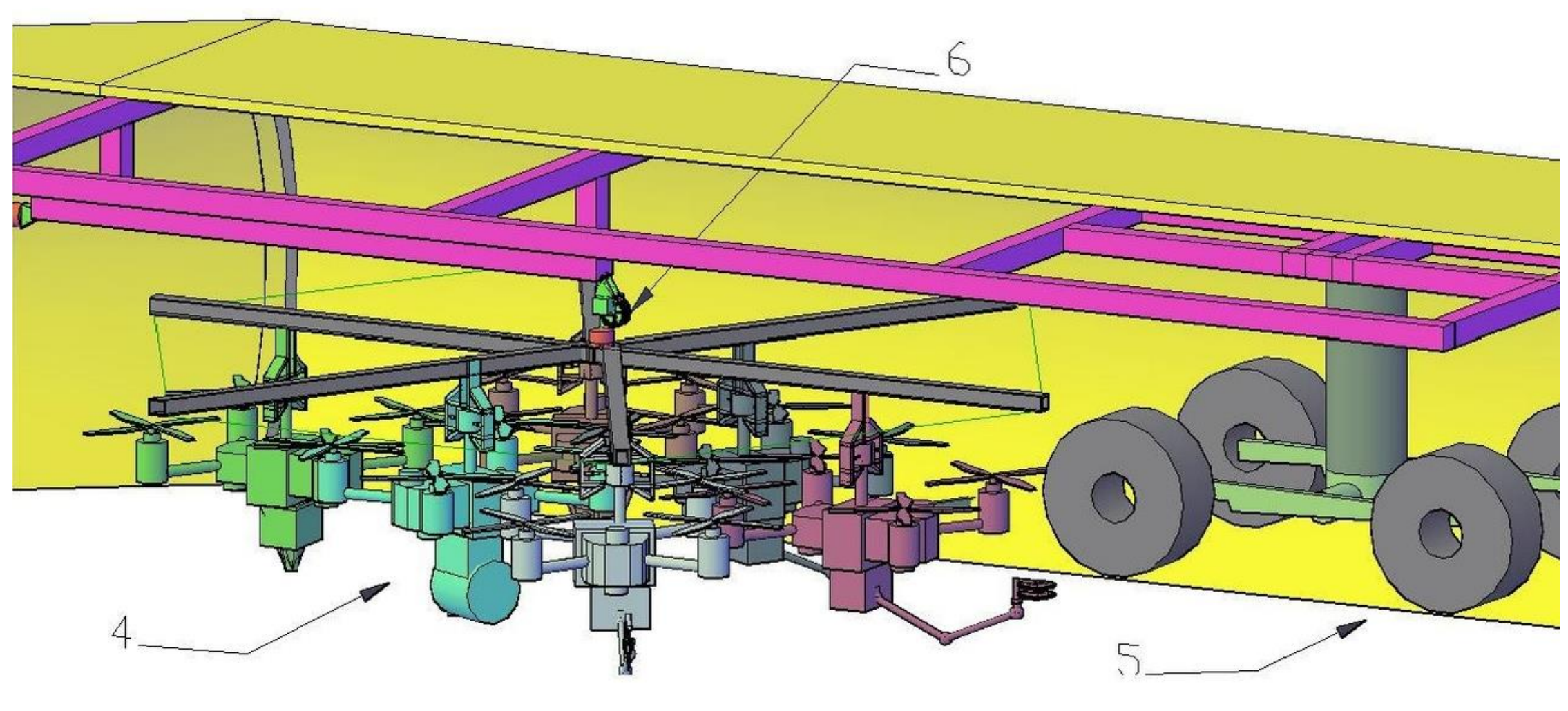

c)

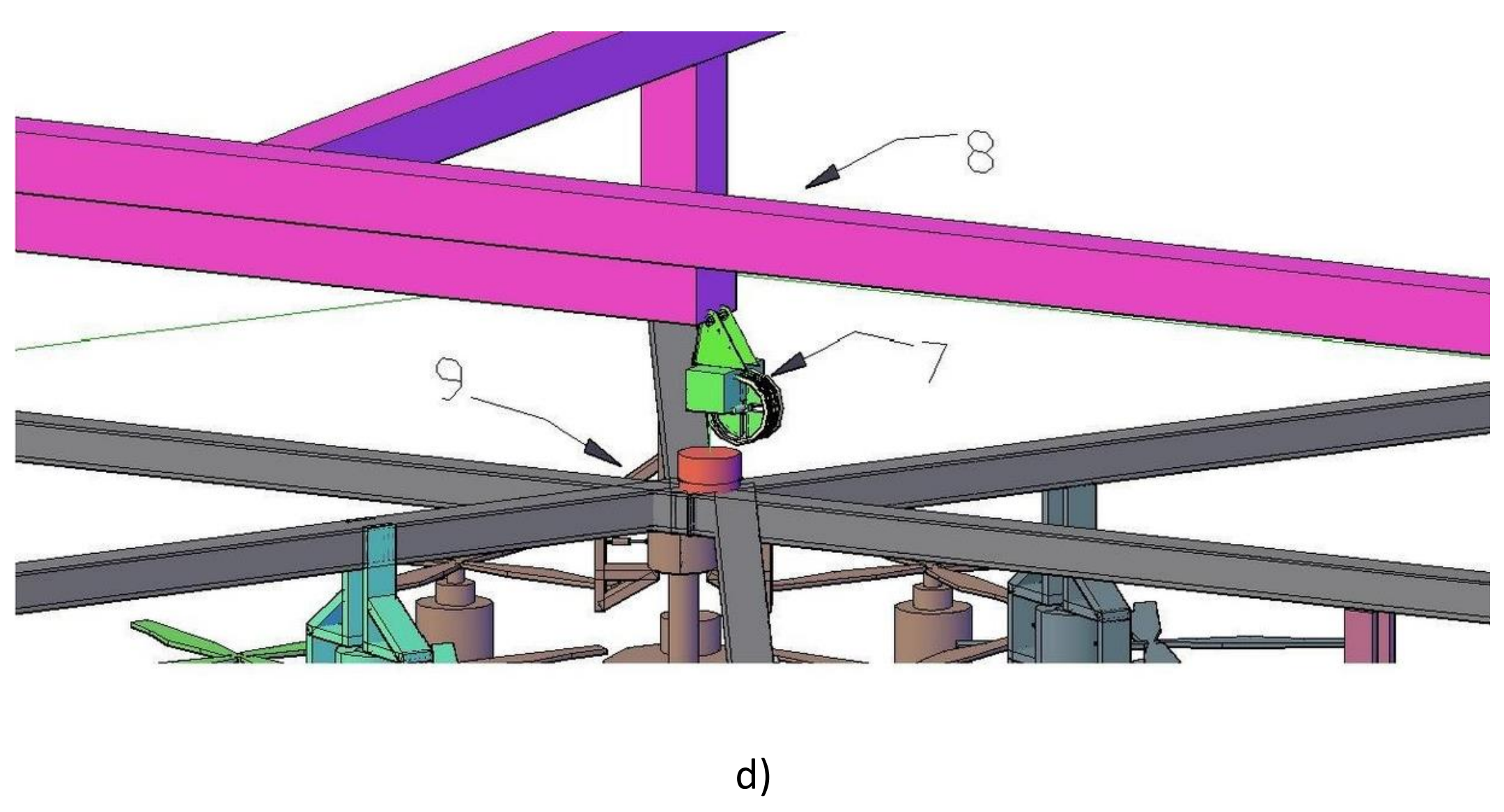

d)

Figure 33. Complete configuration of the URIEL System. Ventral view of complete assembly a): 1-URIEL control station; 2-Front landing gear retracted; 3-Harvest module (MH); 4-Silvicultural treatment module (MHST); 5-Rear landing gear retracted; 6-Main chassis stringer. Detail of Pod/MH assembly b): 1-Harvest module (MH); 2-Stabilization system in horizontal configuration; 3-Support pivot in horizontal configuration. Detail of Pod/MHST assembly c): 4-Silvicultural treatment module (MHST); 5-Rear landing gear retracted; 6-Vertical movement crane of the drone support star. Detail of MHST coupling to the Pod chassis d): 7-Vertical movement crane; 8-Crane/chassis coupling position; 9-Rotator of the drone support star.

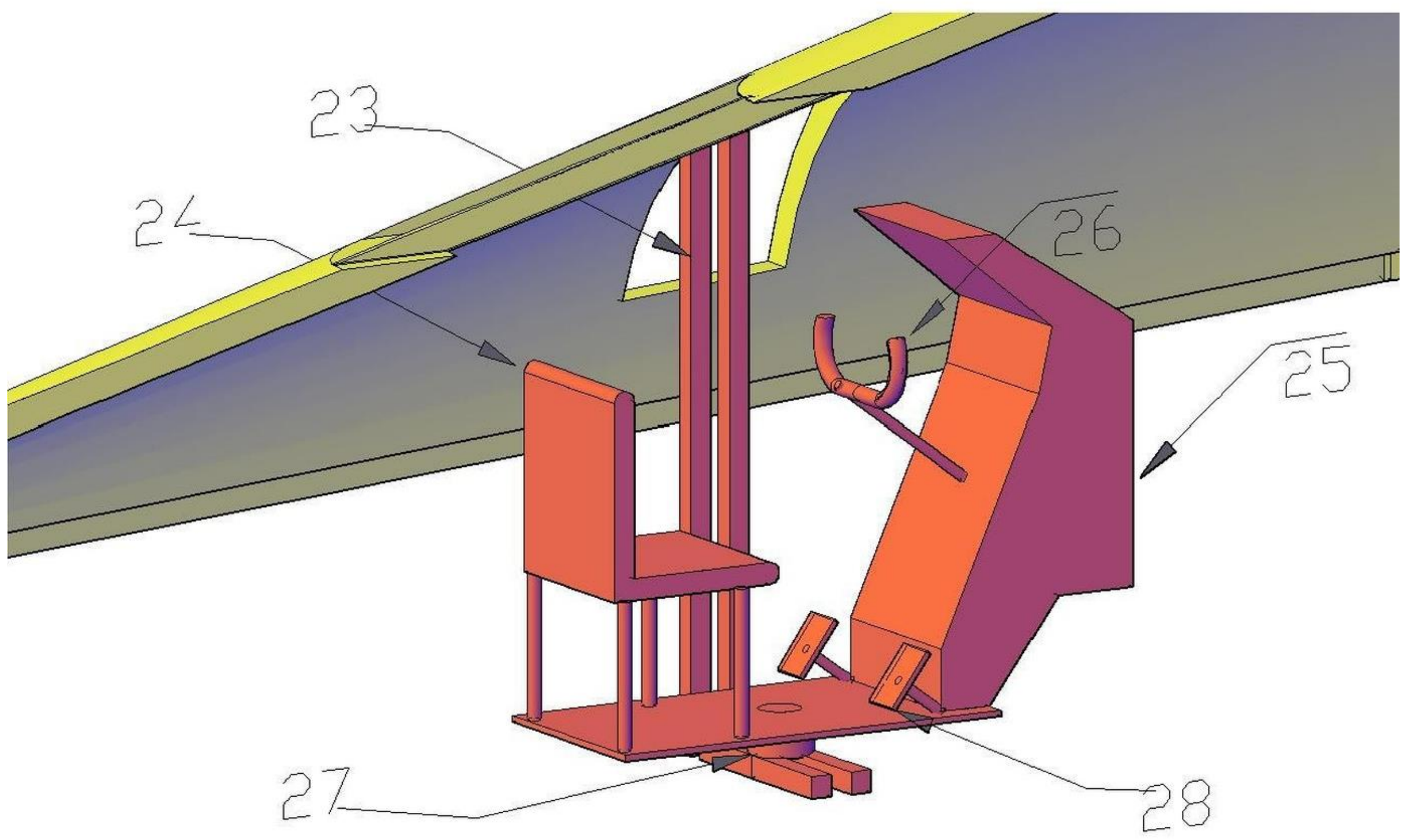

Figure 34. URIEL Command Post: 23-Command center stabilizer support; 24-Command chair; 25-Command console; 26-Remote helicopter control stick (pull stick = ascend vertically; push stick = descend vertically); tilt stick transversely (tilt forward = move forward and tilt backward = move backward); rotate stick on axis (lateral tilt of helicopter); 27-Command post rotation pivot, 720° range; 28-Rudder pedals (rotation of helicopter on the vertical axis and also steering of the selected landing gear (front only, rear only or both).

**Simulations**

Based on the detailed technical drawings, a digital proof of concept (POC) of the URIEL System was developed using a digital model of the system; additionally, a three-dimensional forest environment was modeled. With the digital model and the 3D forest environment, it was possible to perform simulations of the URIEL System's operation.

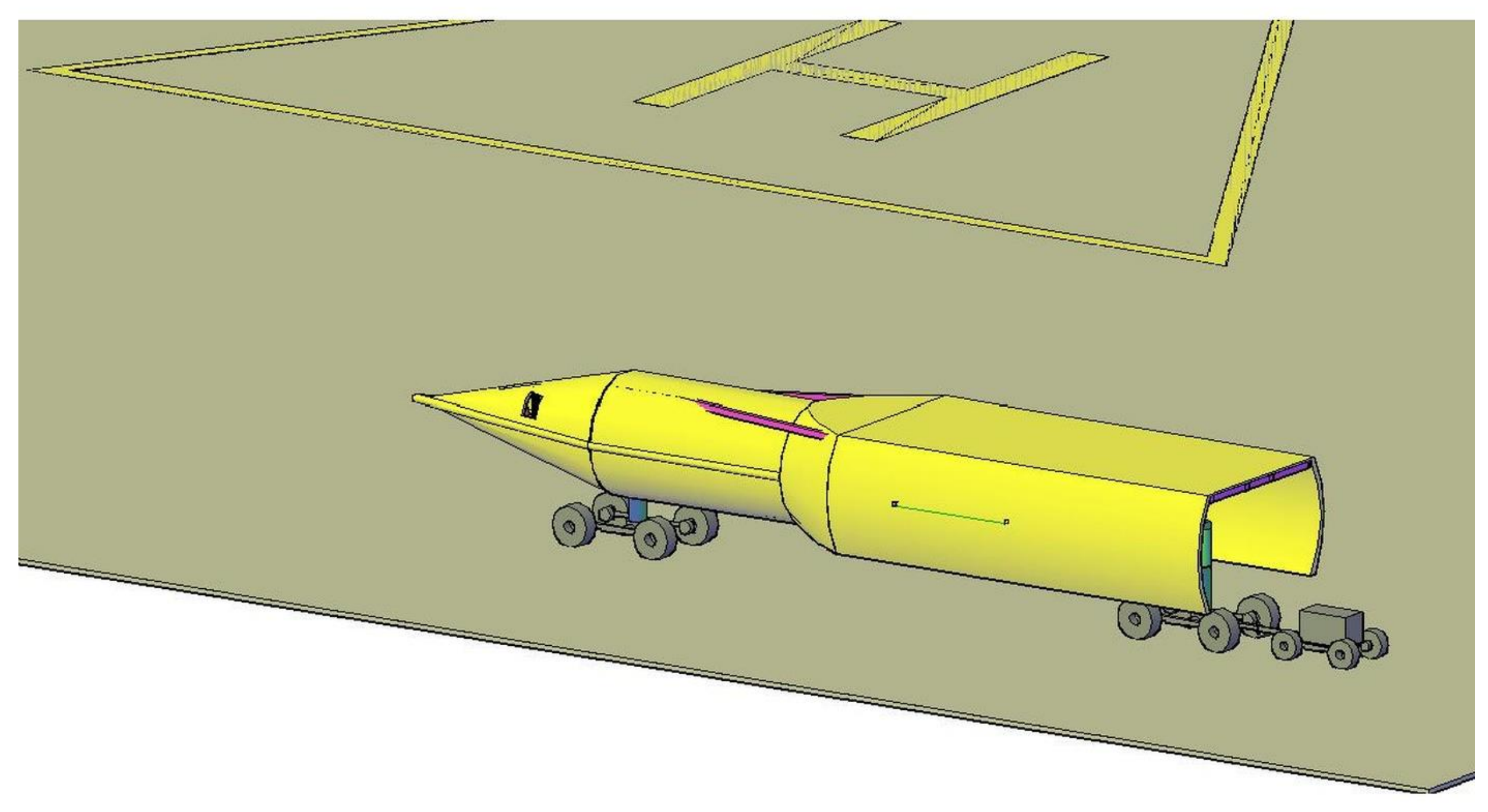

a)

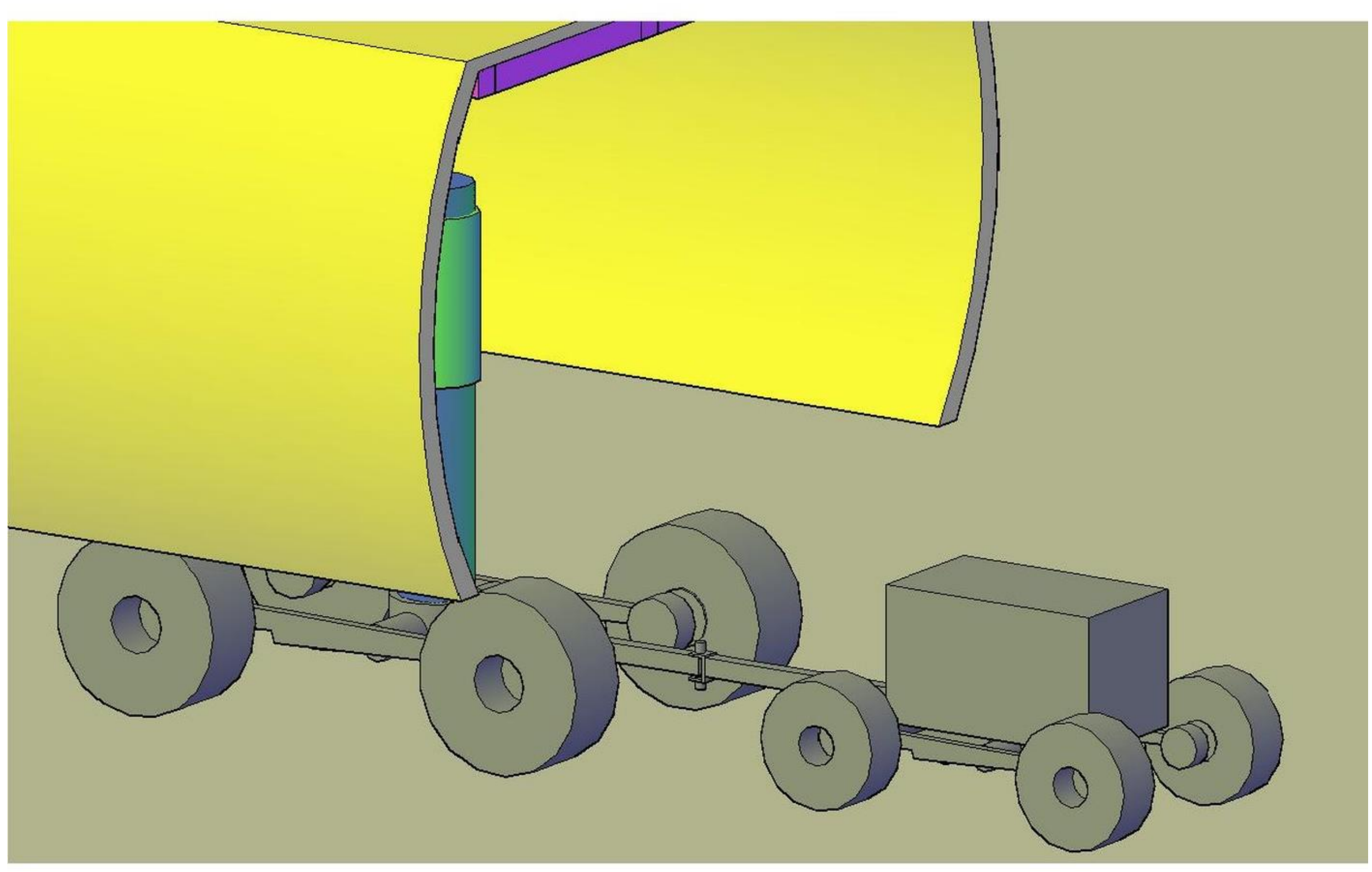

b)

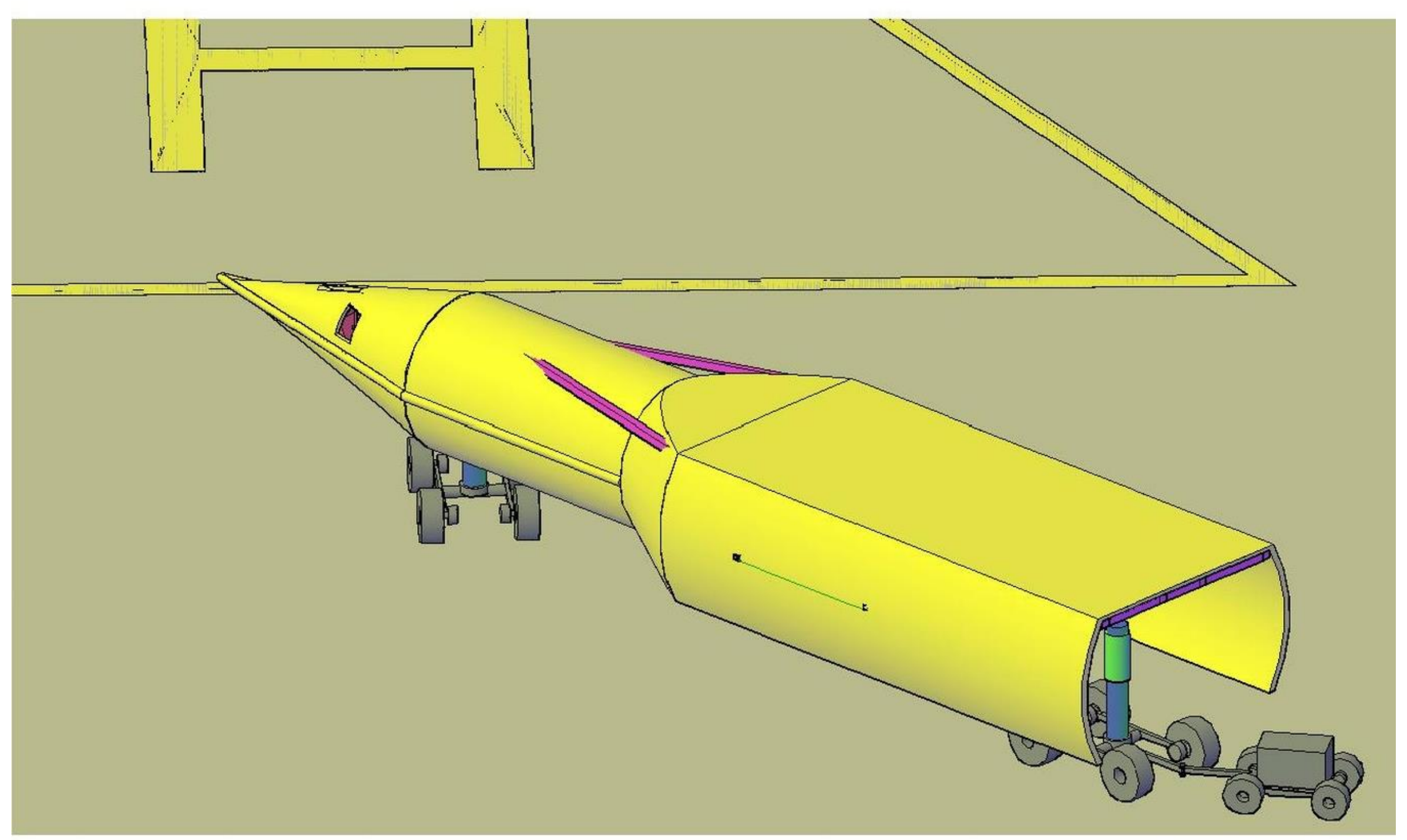

c)

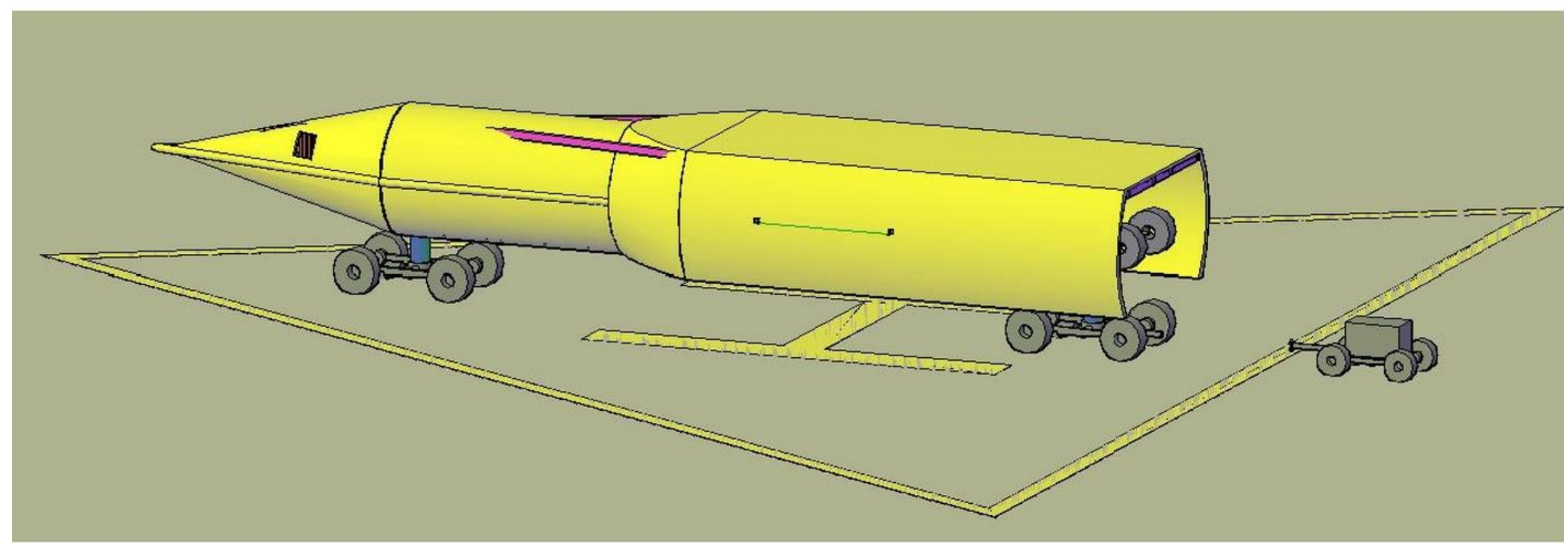

d)

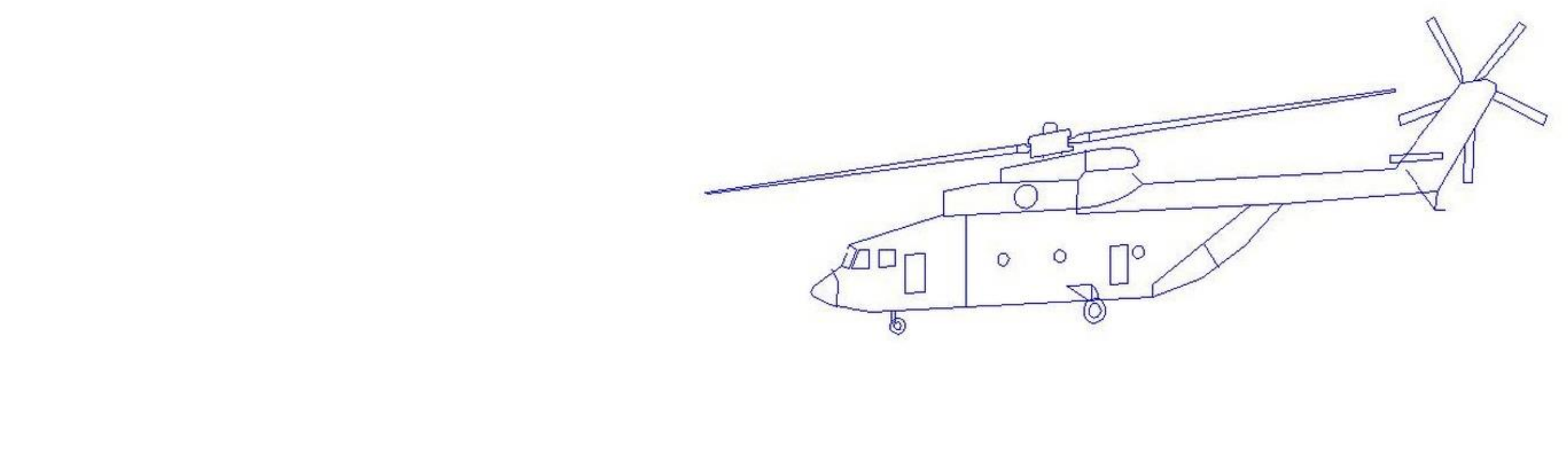

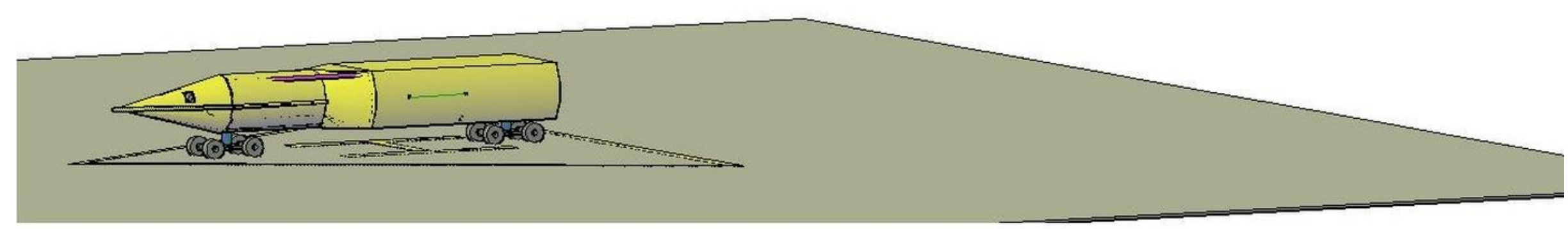

e)

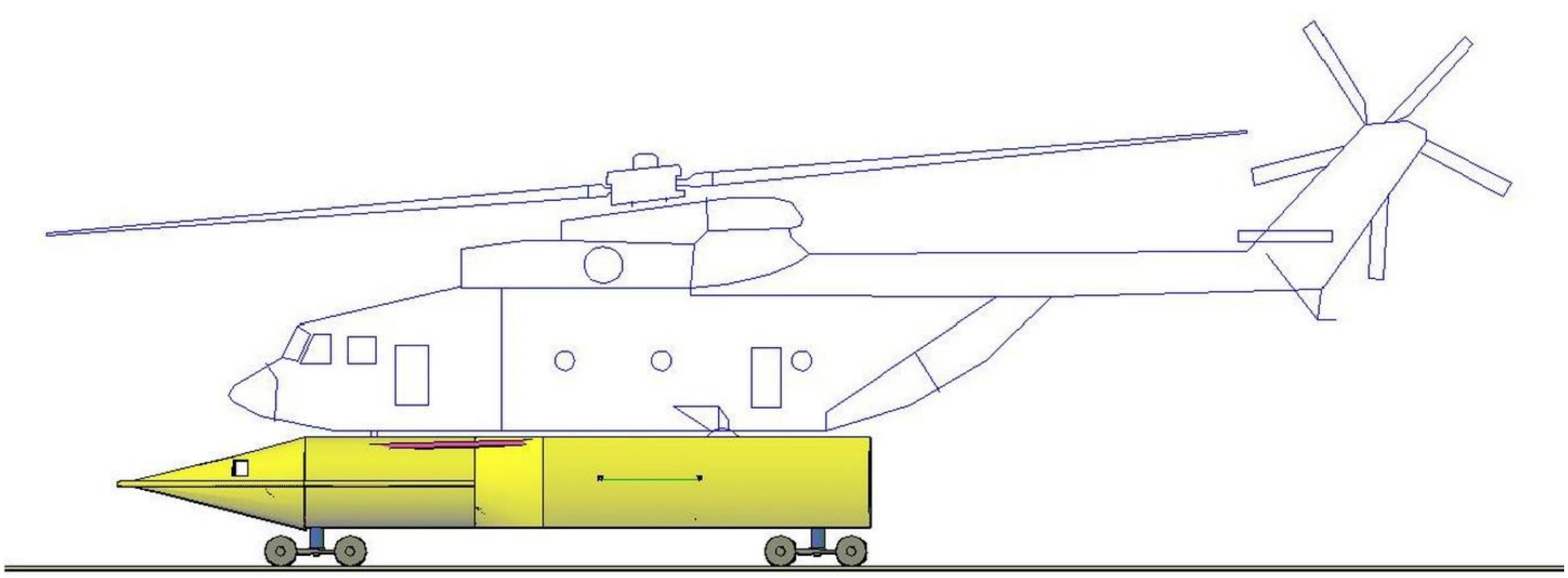

f)

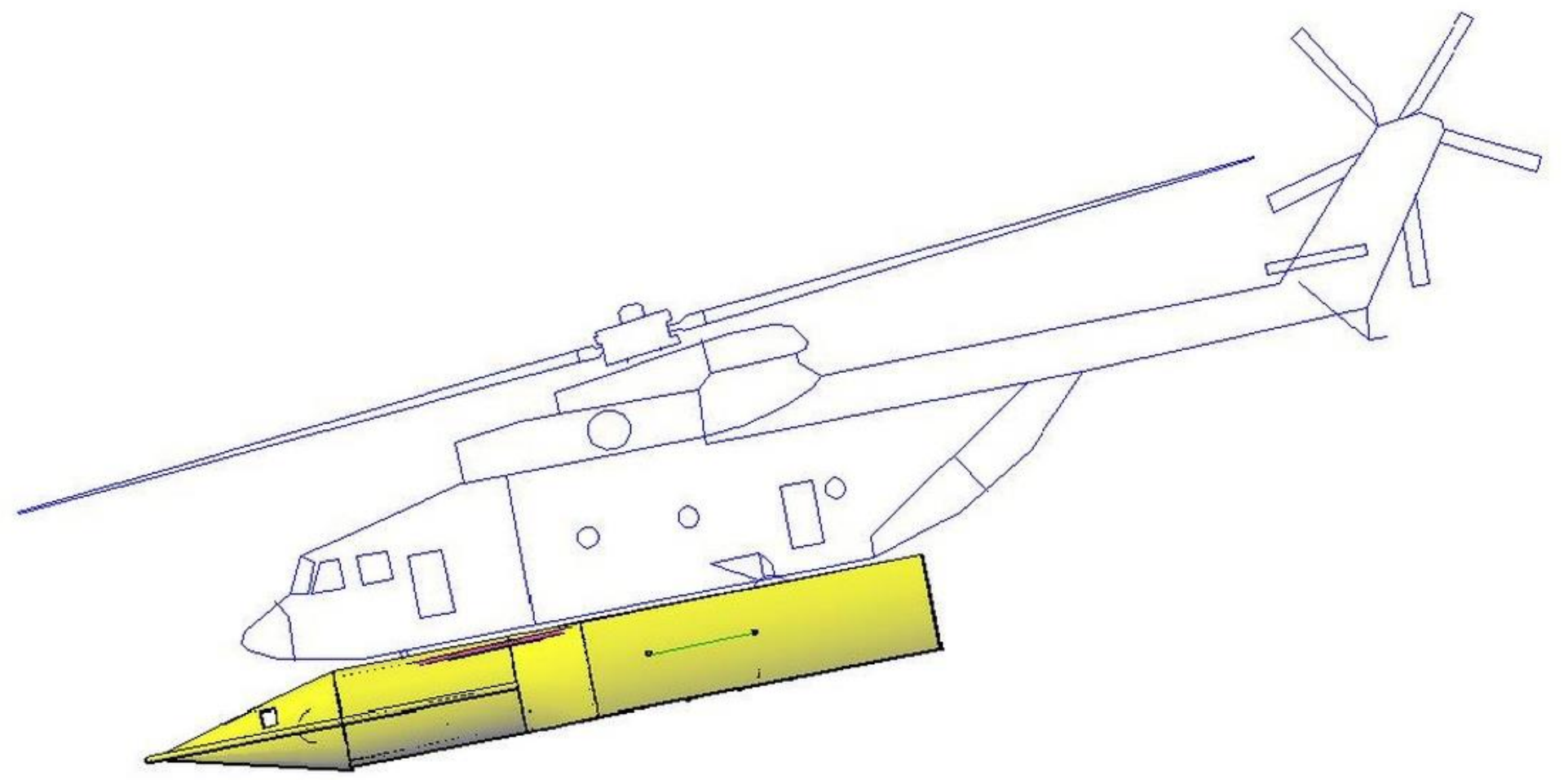

g)

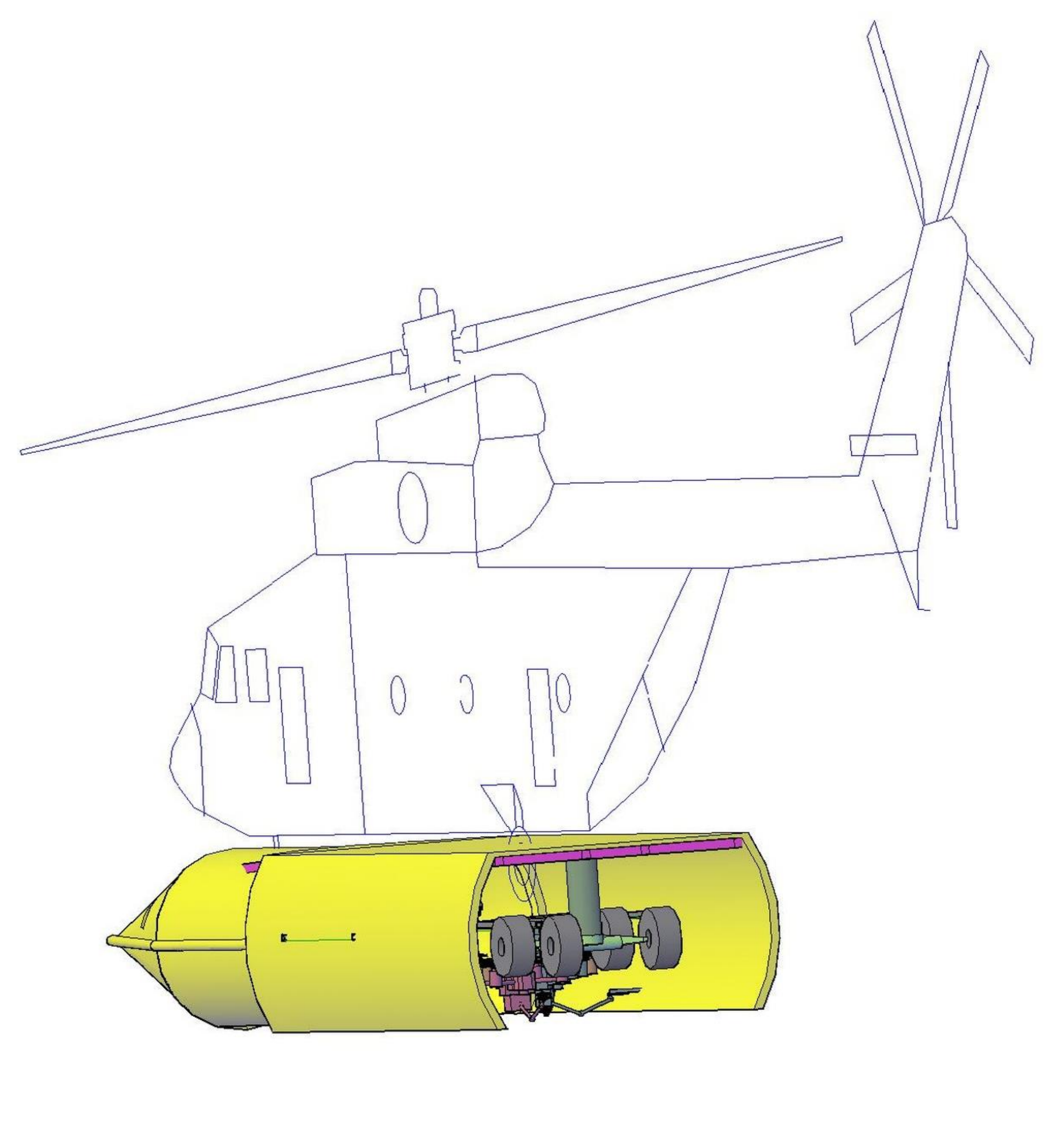

h)

Figure 35. Ground movement simulation of the URIEL Pod docking with a Mi-26: a) URIEL Pod moving from the hangar to the helipad; b) Detail of the battery pack attached to the URIEL Pod powering the systems and traction sources; c) URIEL Pod maneuvering to align with the helipad, note the steered nose landing gear; d) URIEL Pod aligned in flight position on the helipad, note the power pack being removed from the docking area; e) Slow-flying approach of a Mil Mi-26; f) Mi-26 hovering and docking with the URIEL Pod; g) URIEL system in flight towards the target area; h) Rear view of the URIEL system getting underway.

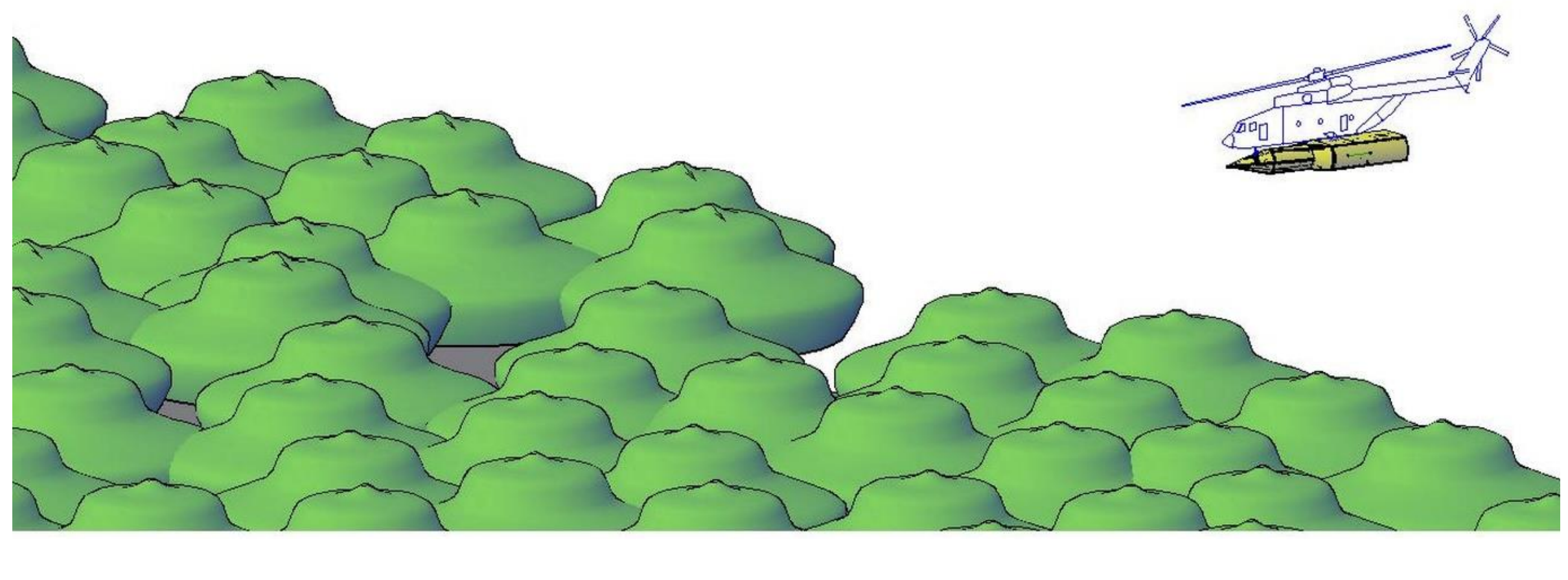

a)

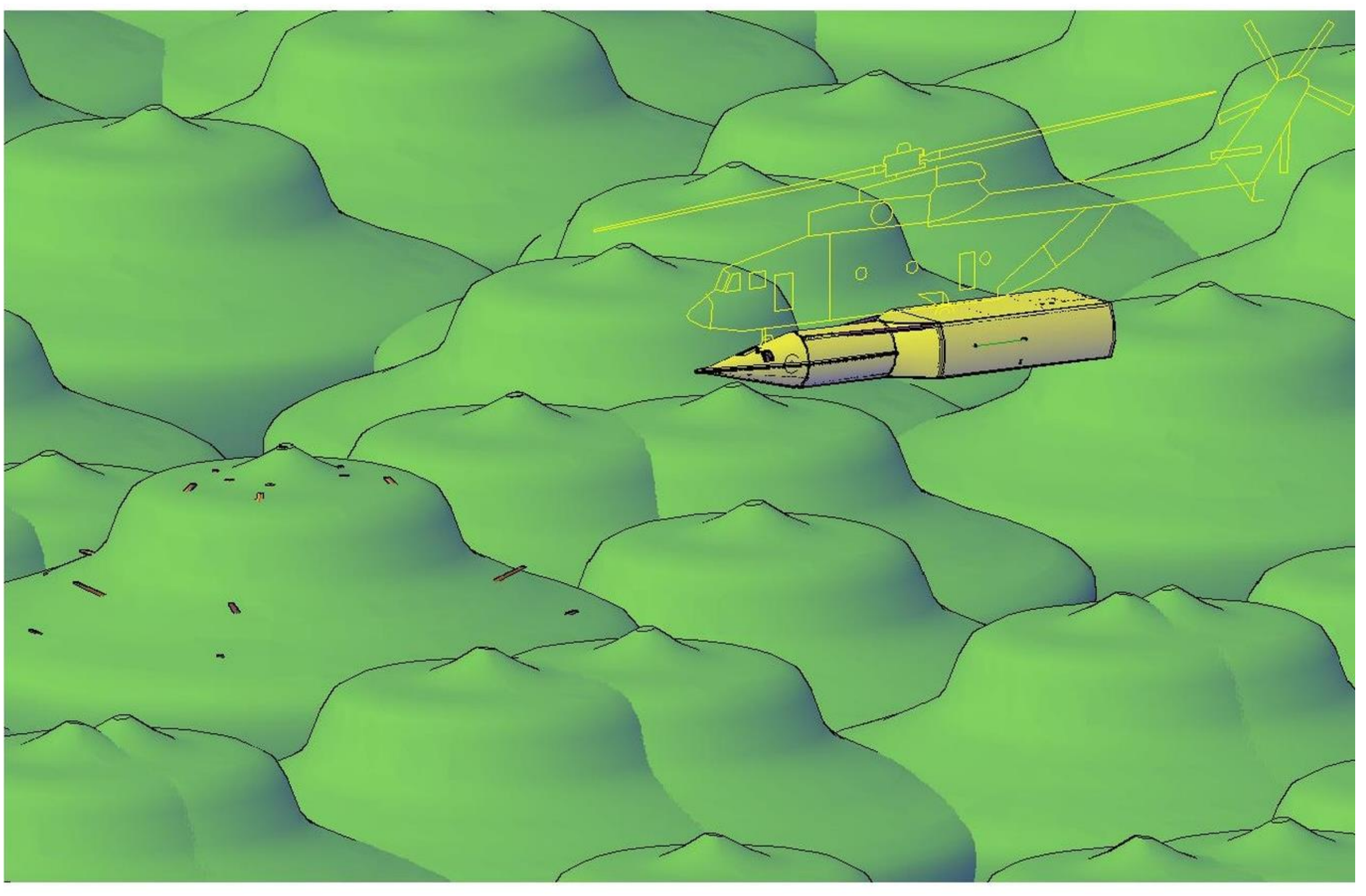

b)

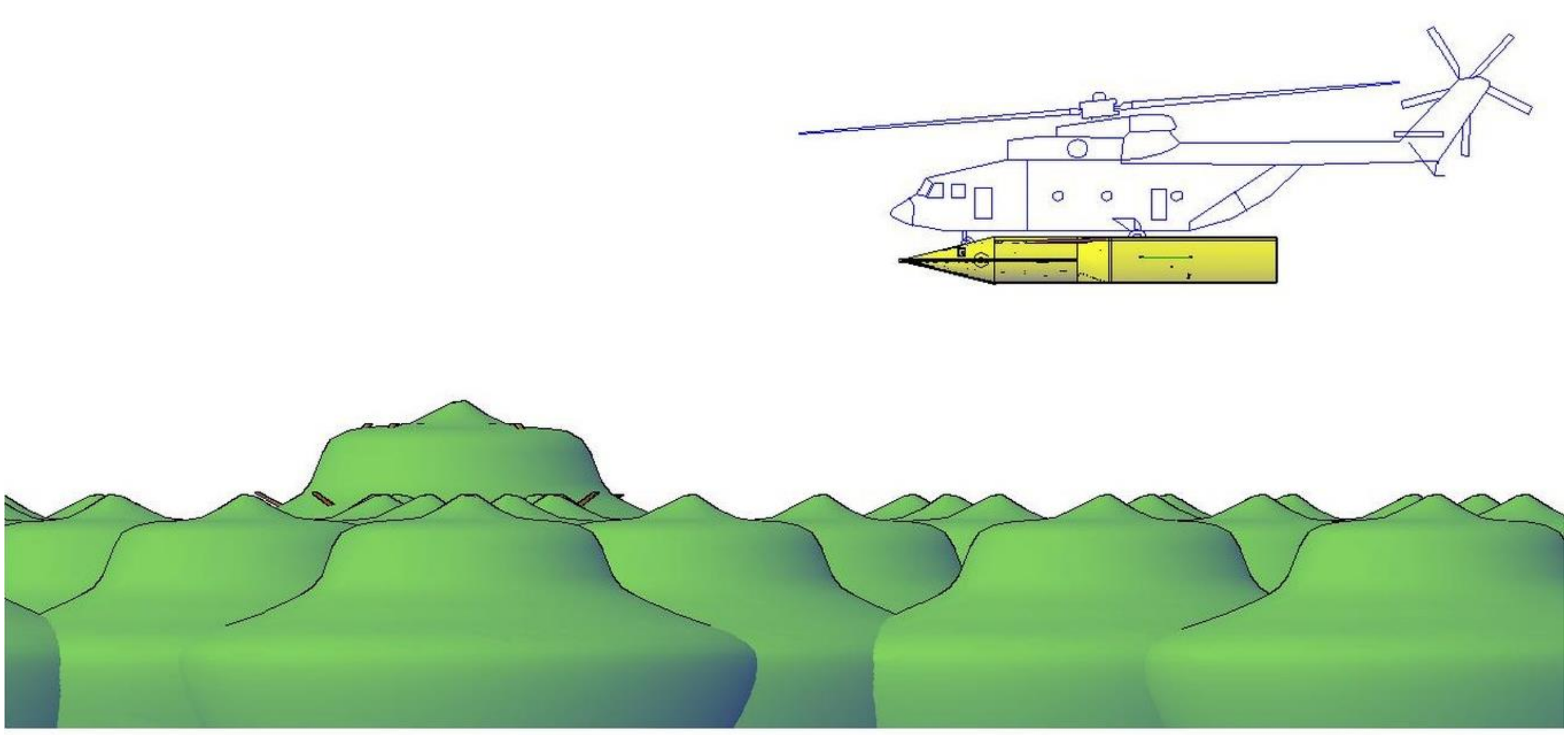

c)

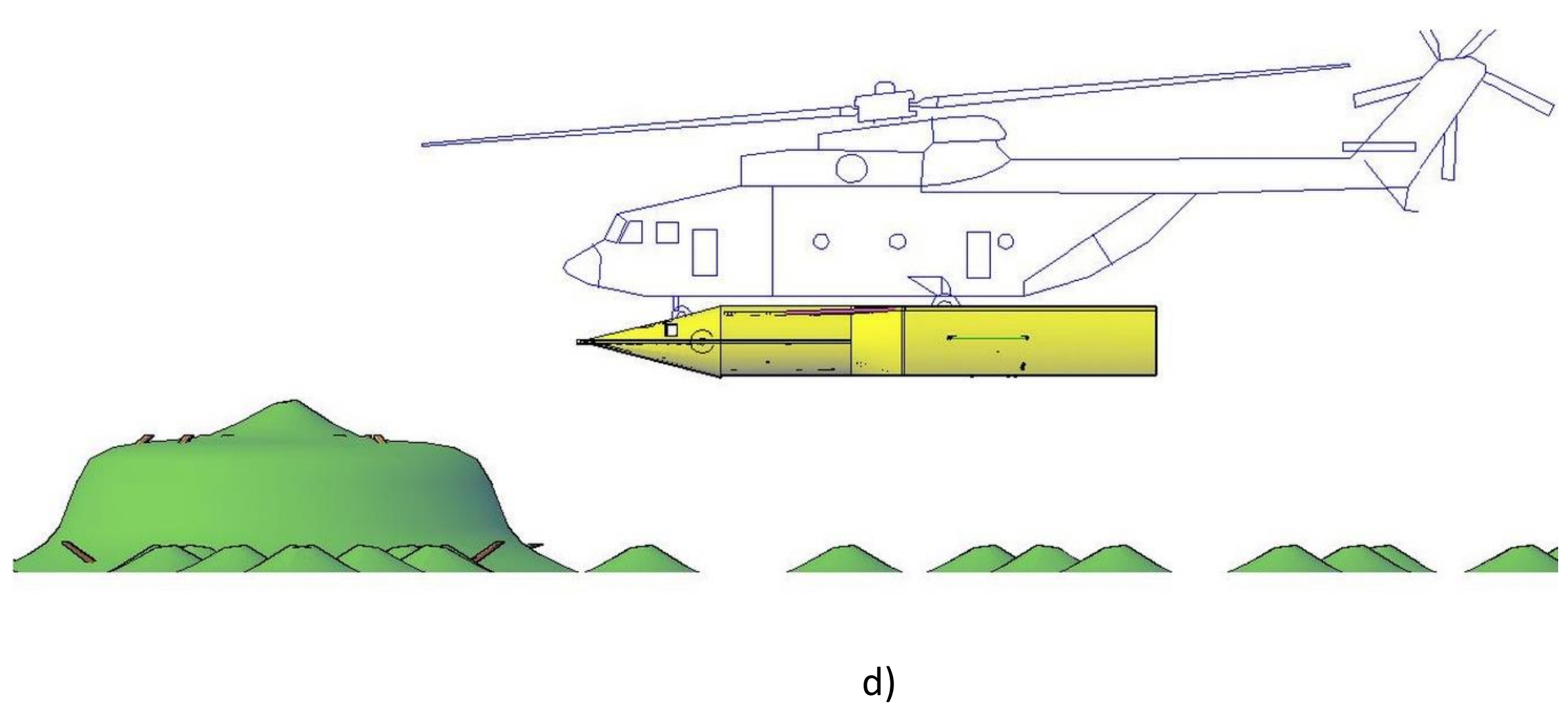

d)

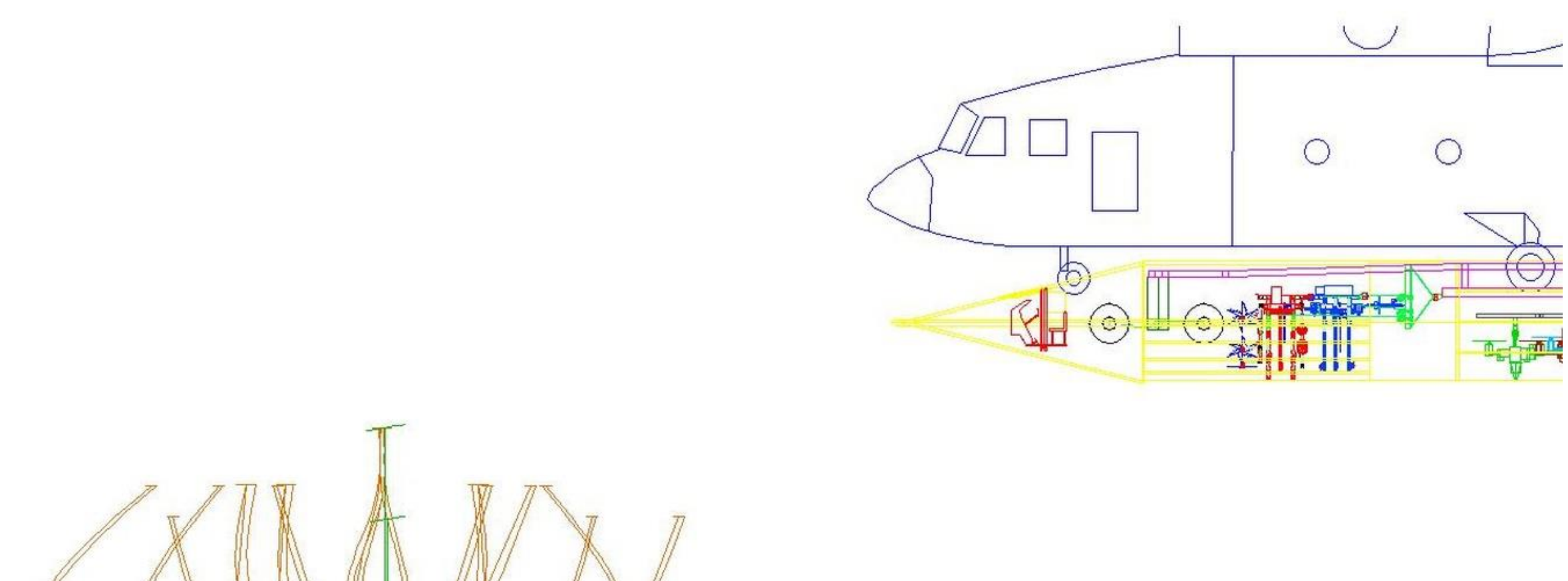

e)

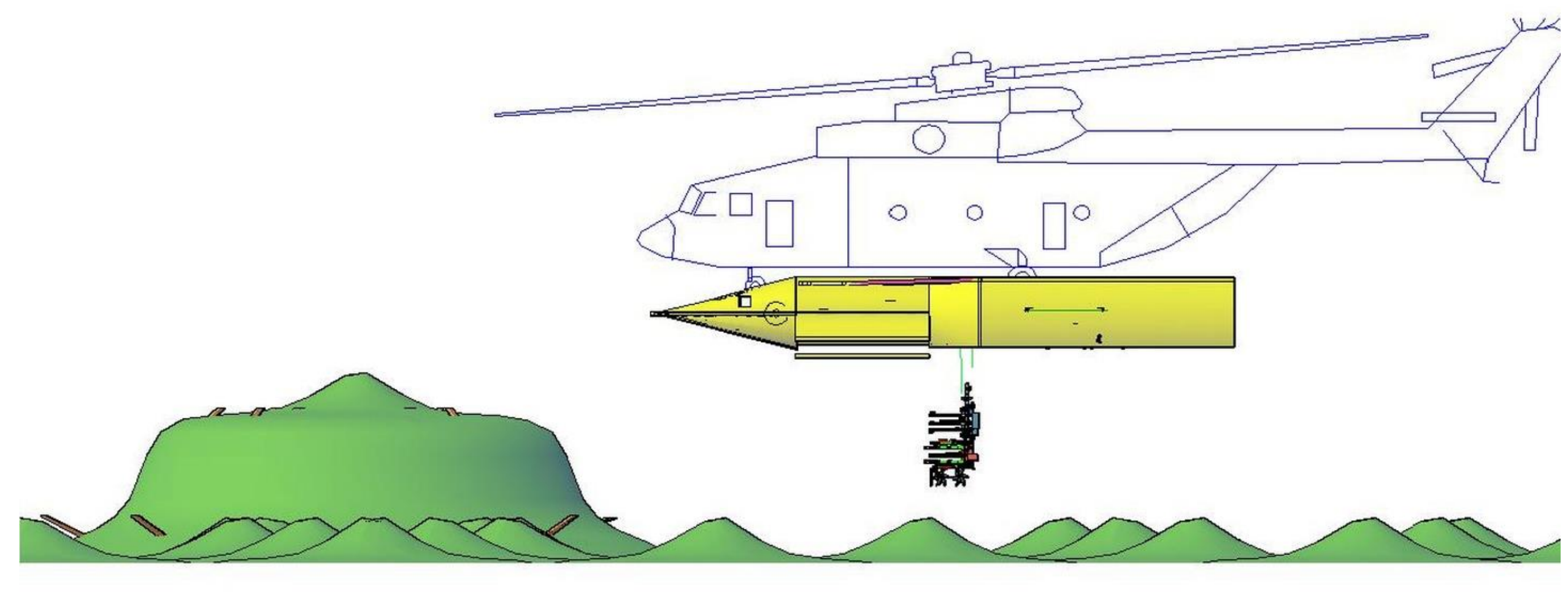

f)

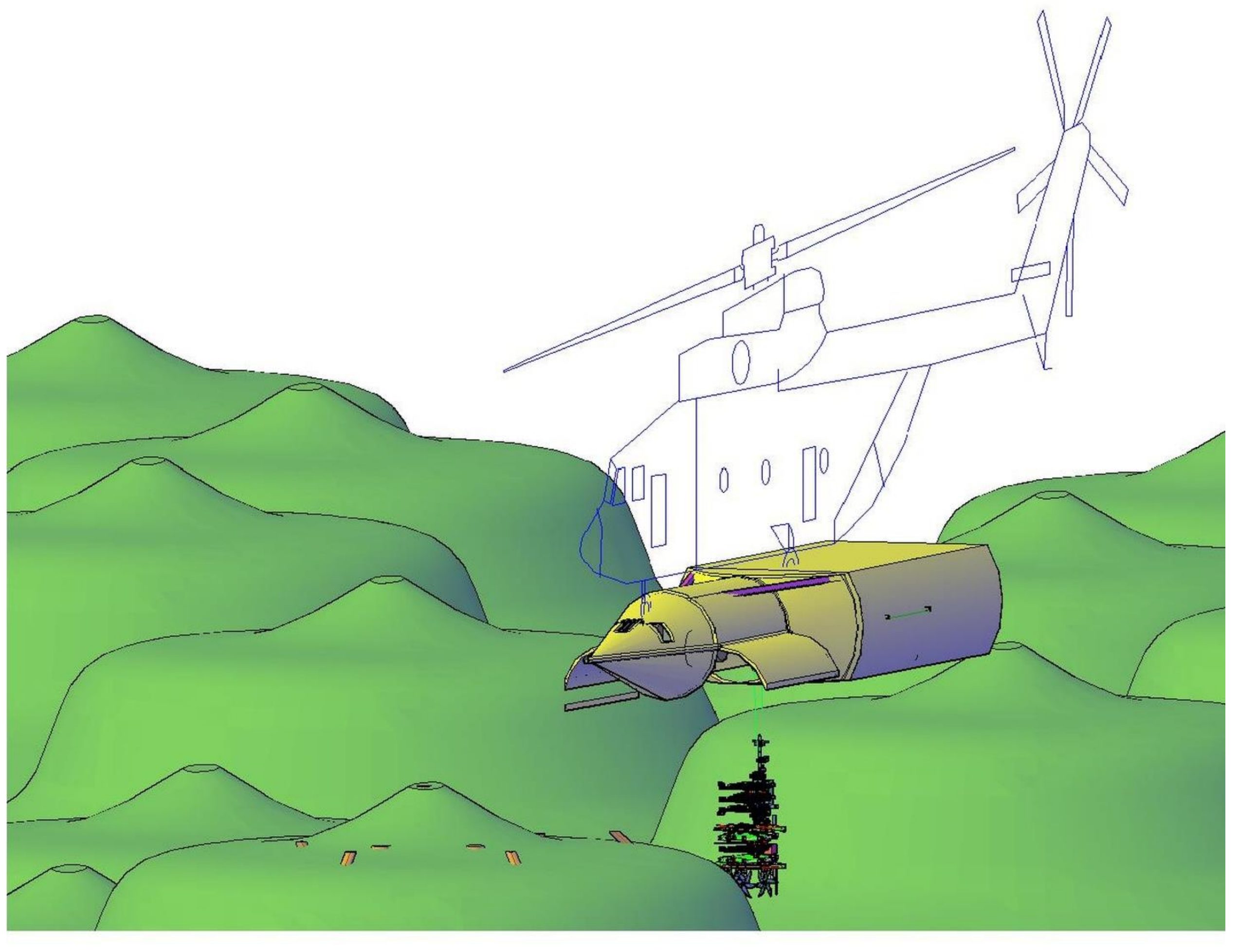

g)

Figure 36. URIEL system arriving at the target area: a) URIEL system entering the forest; b) Location and identification of the target tree; c) Slow and level approach (helicopter command was given to the URIEL operator); d) Precision approach; e) X-ray view of the MH module in resting position; f) MH module active and hoisted for operation; g) Front view of the URIEL system with the MH module hoisted with claws open and ready to begin operation, note the pod hatches open.

.

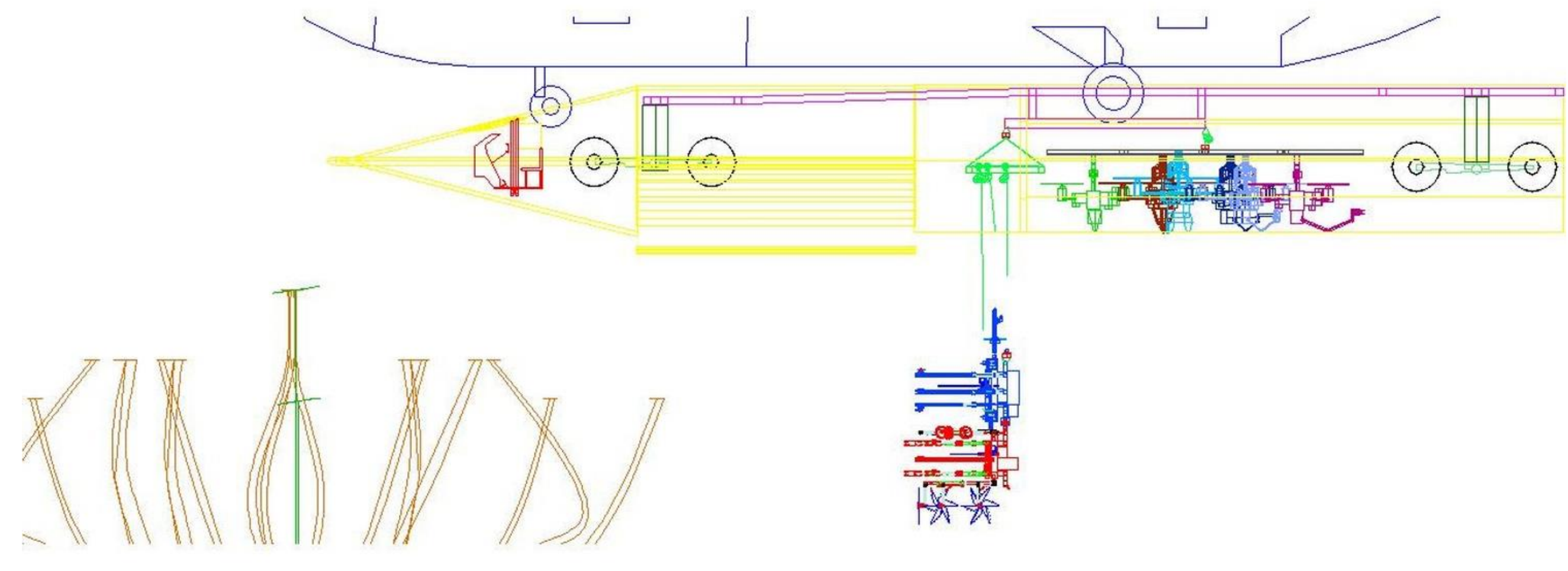

a)

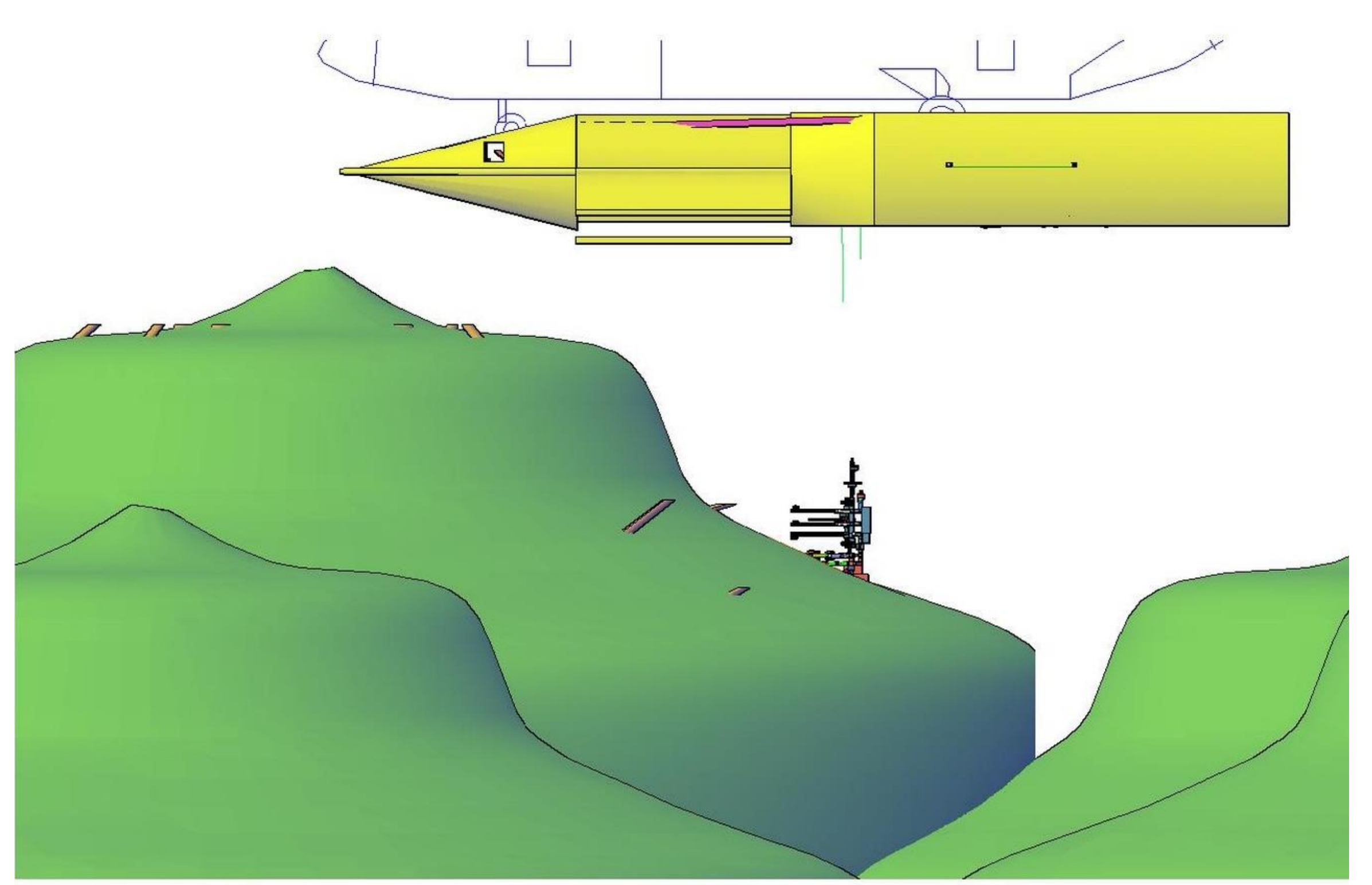

b)

d)

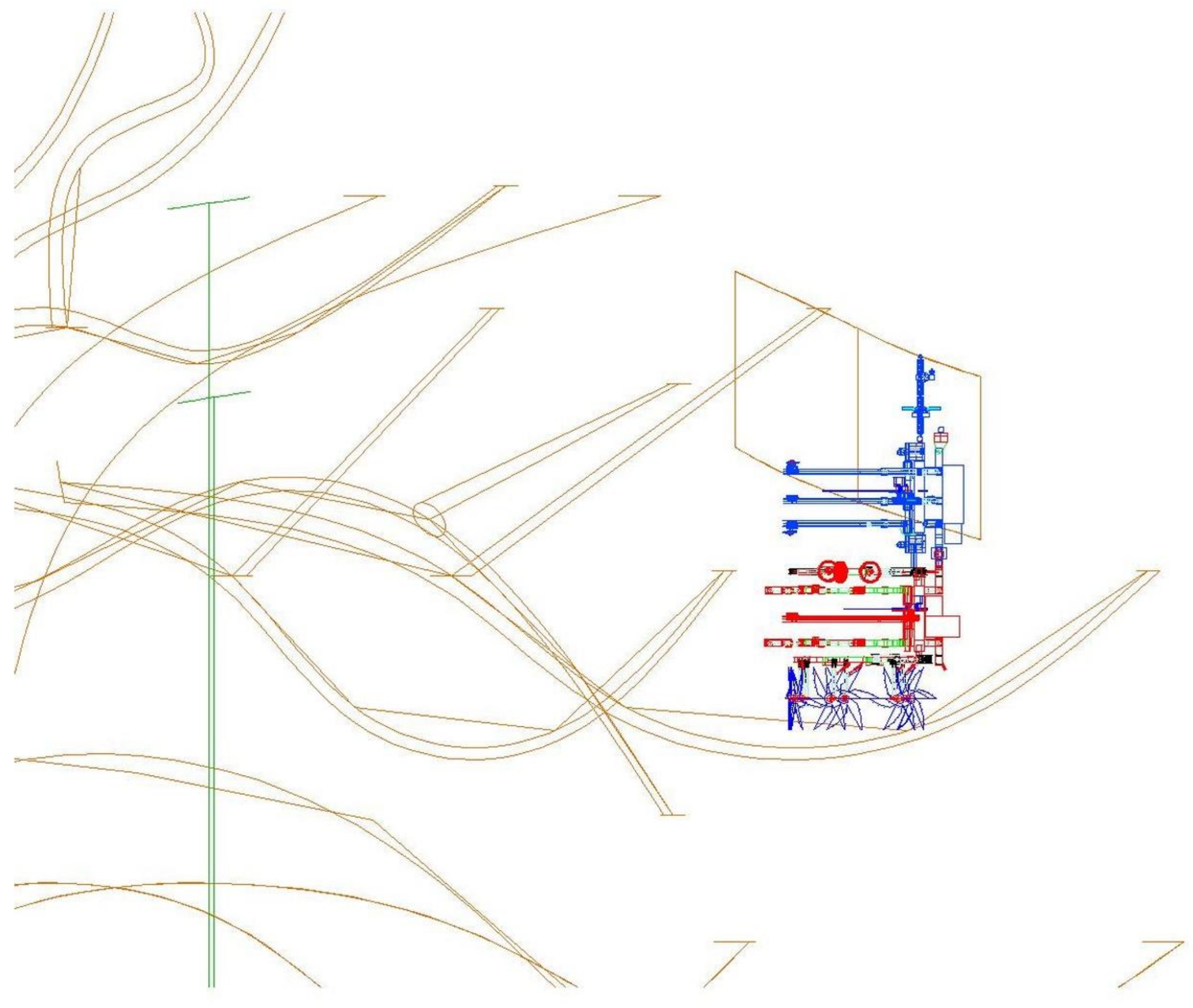

c)

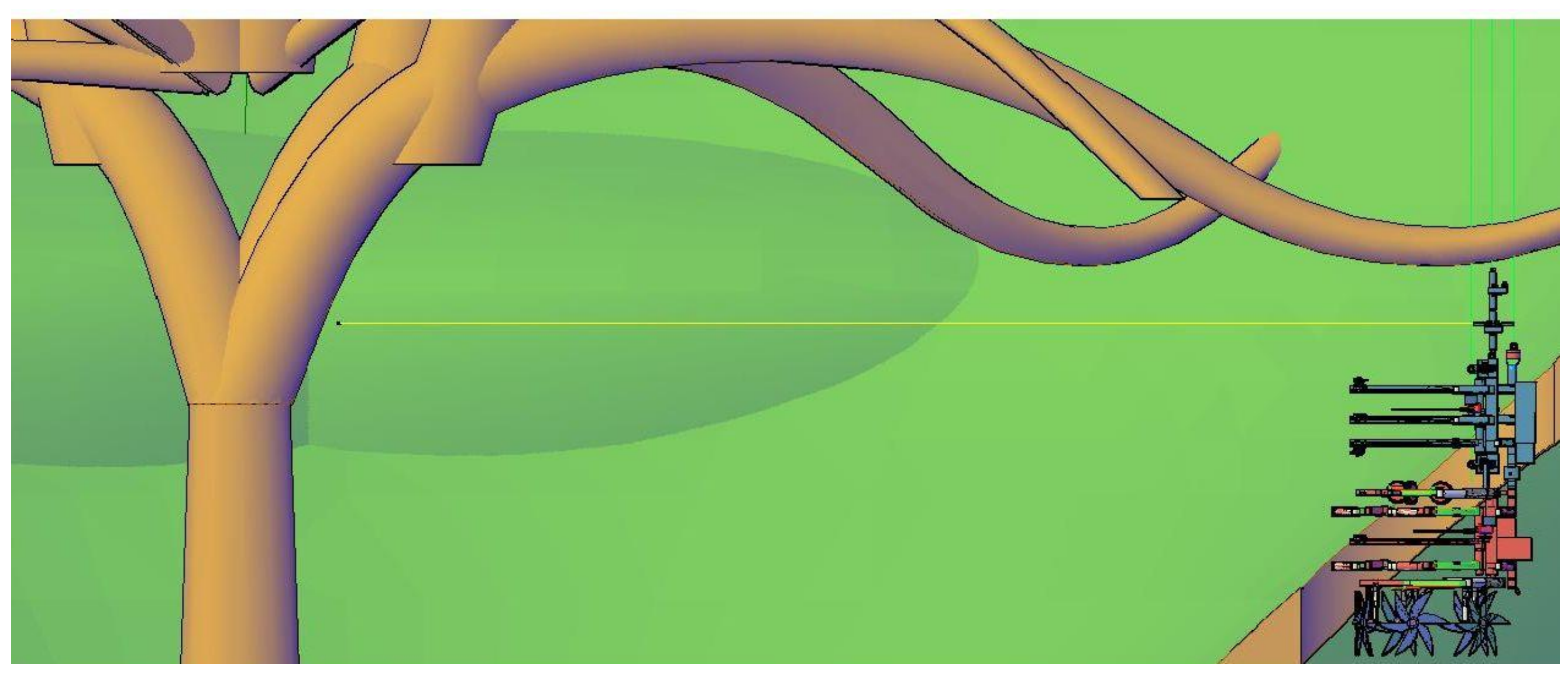

d)

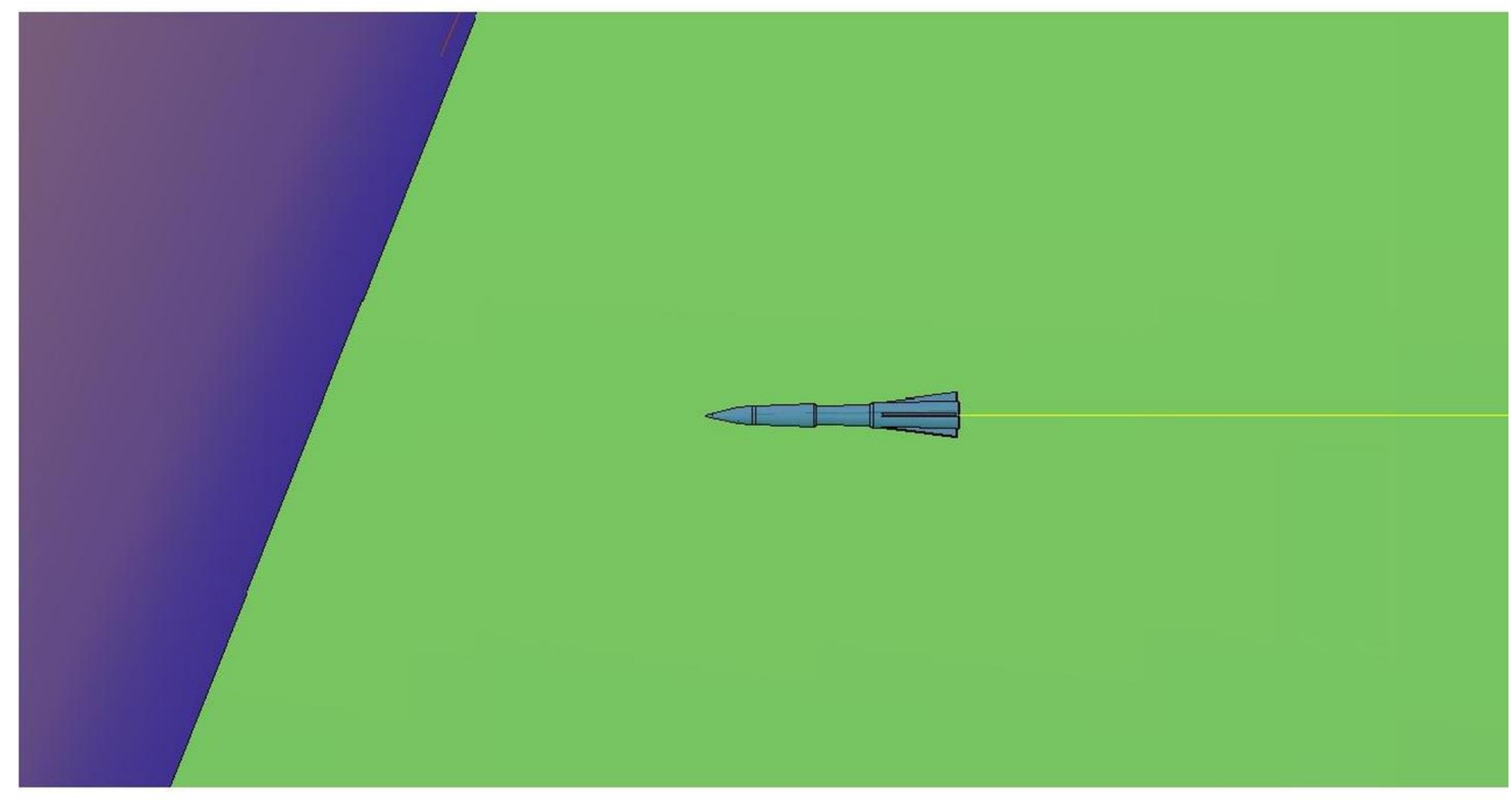

e)

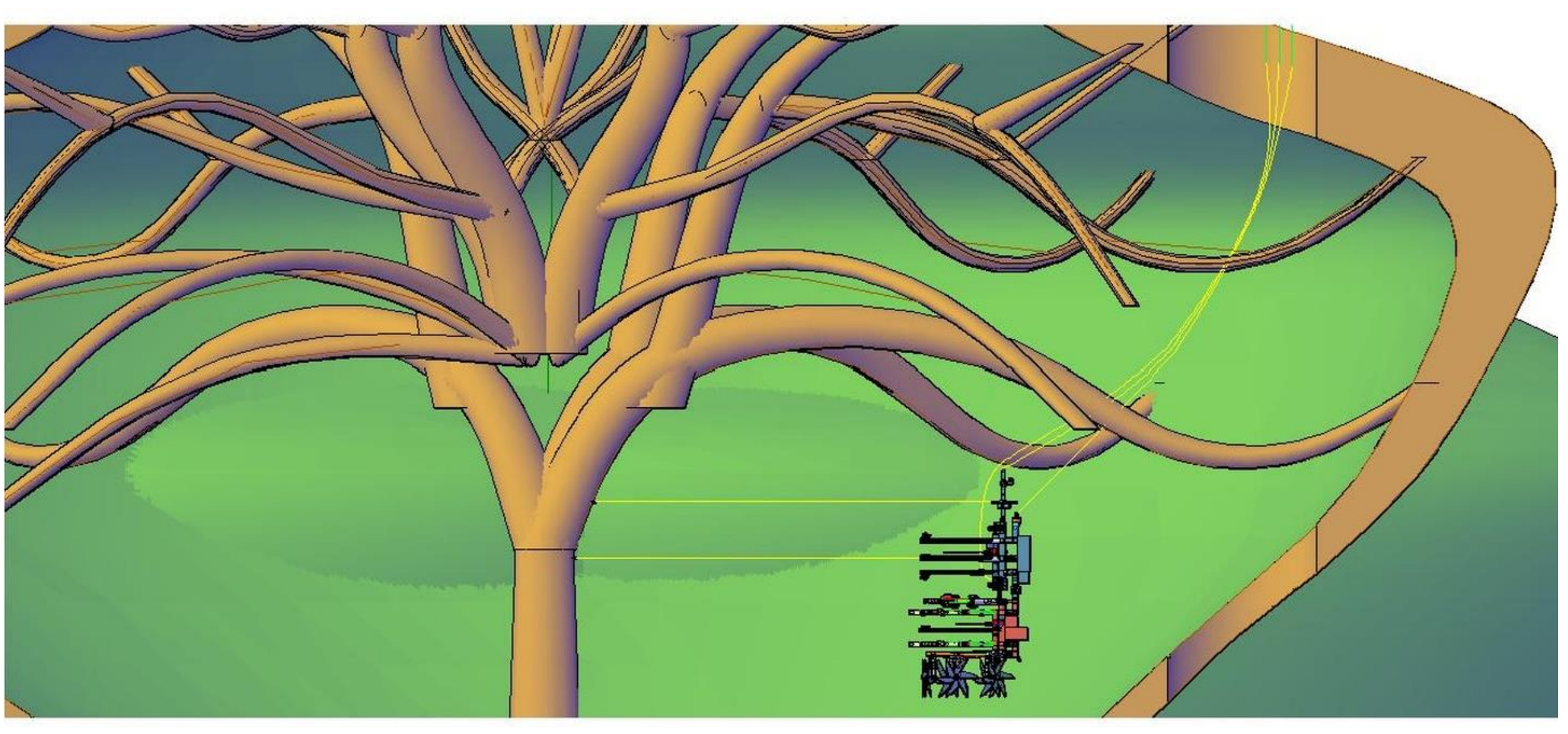

f)

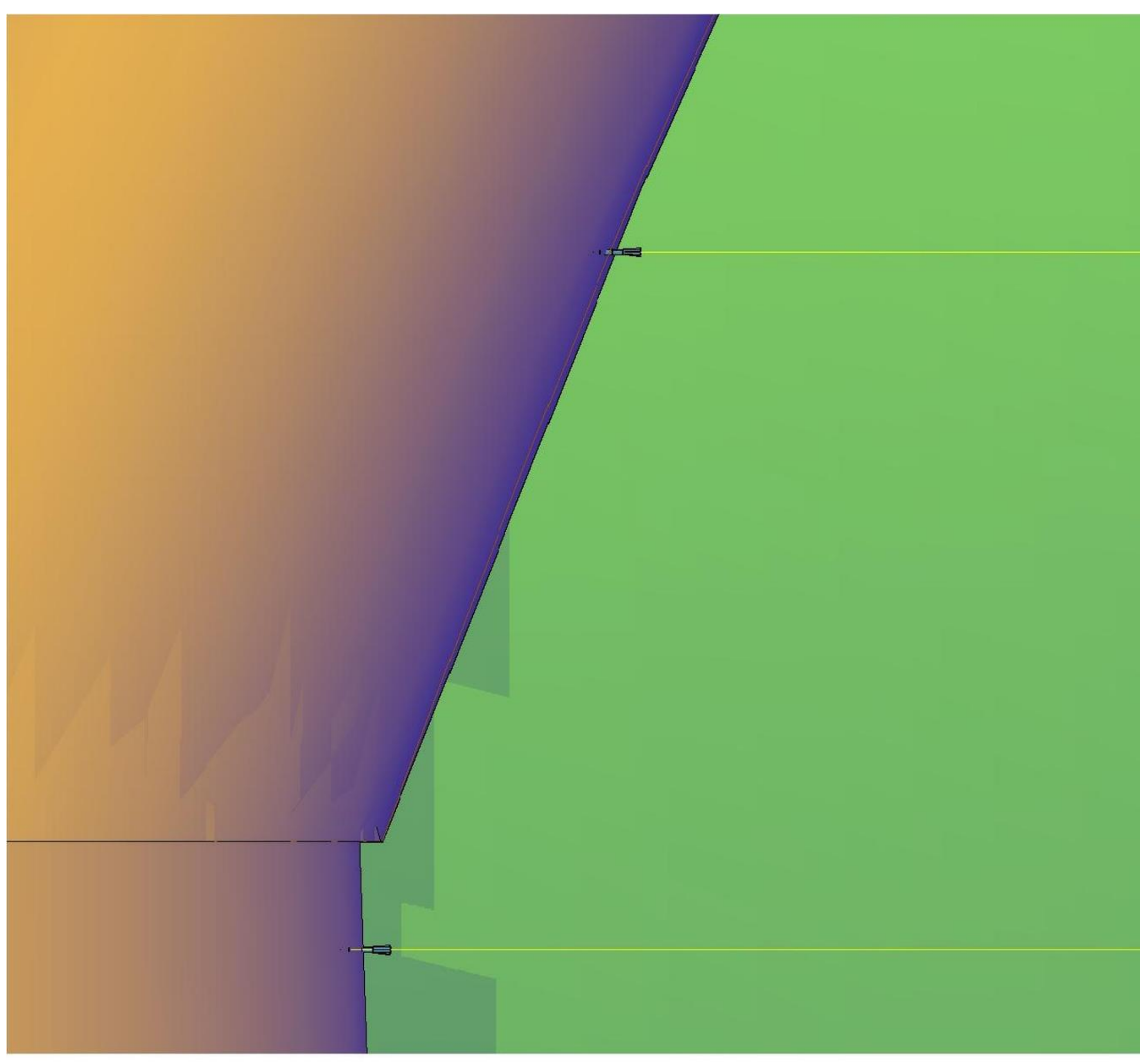

g)

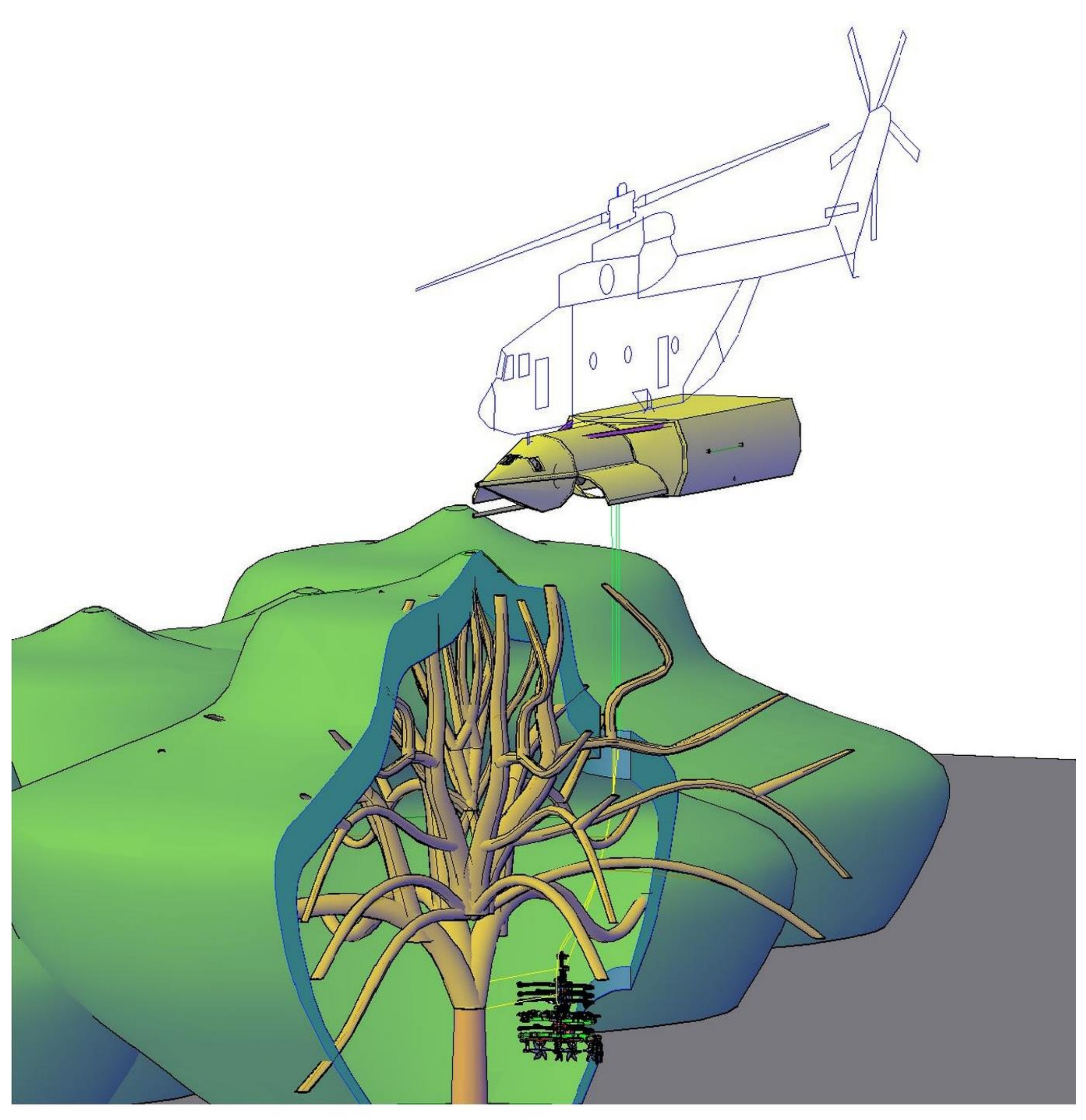

h)

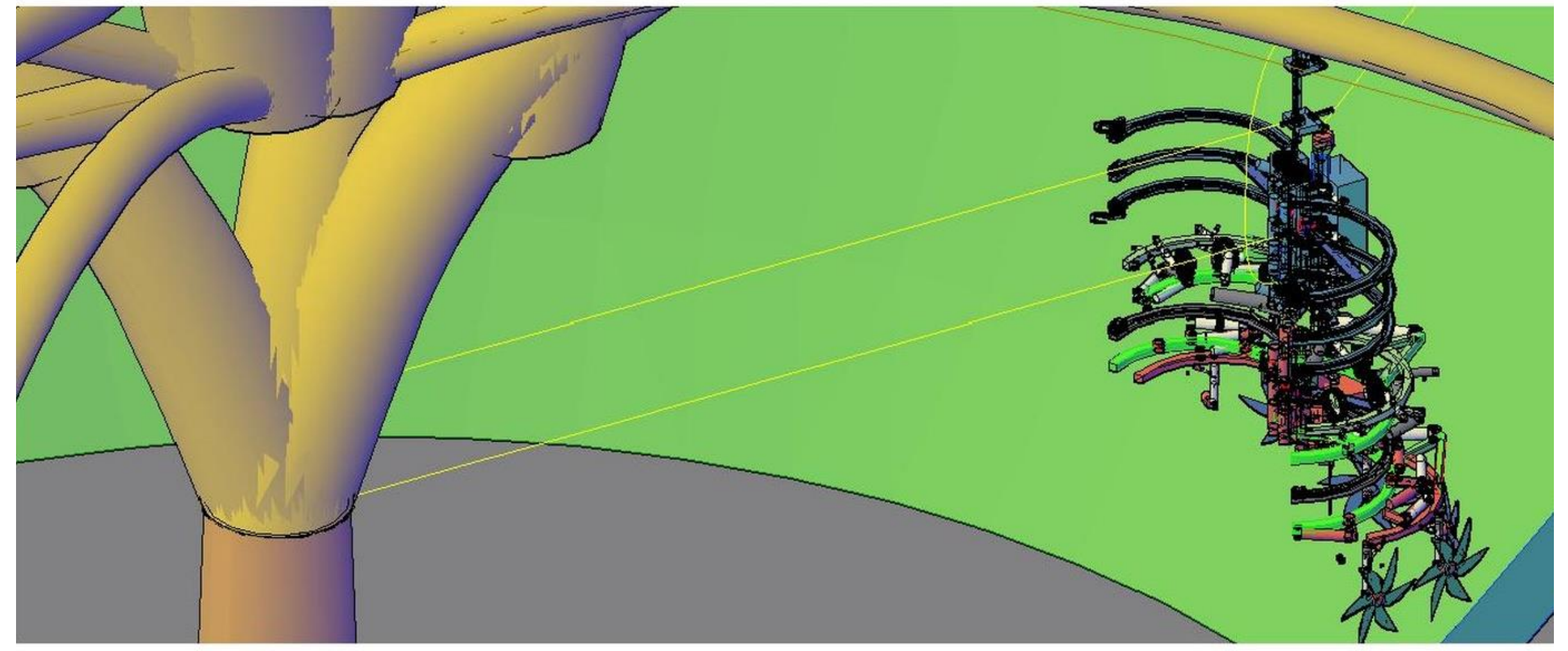

i)

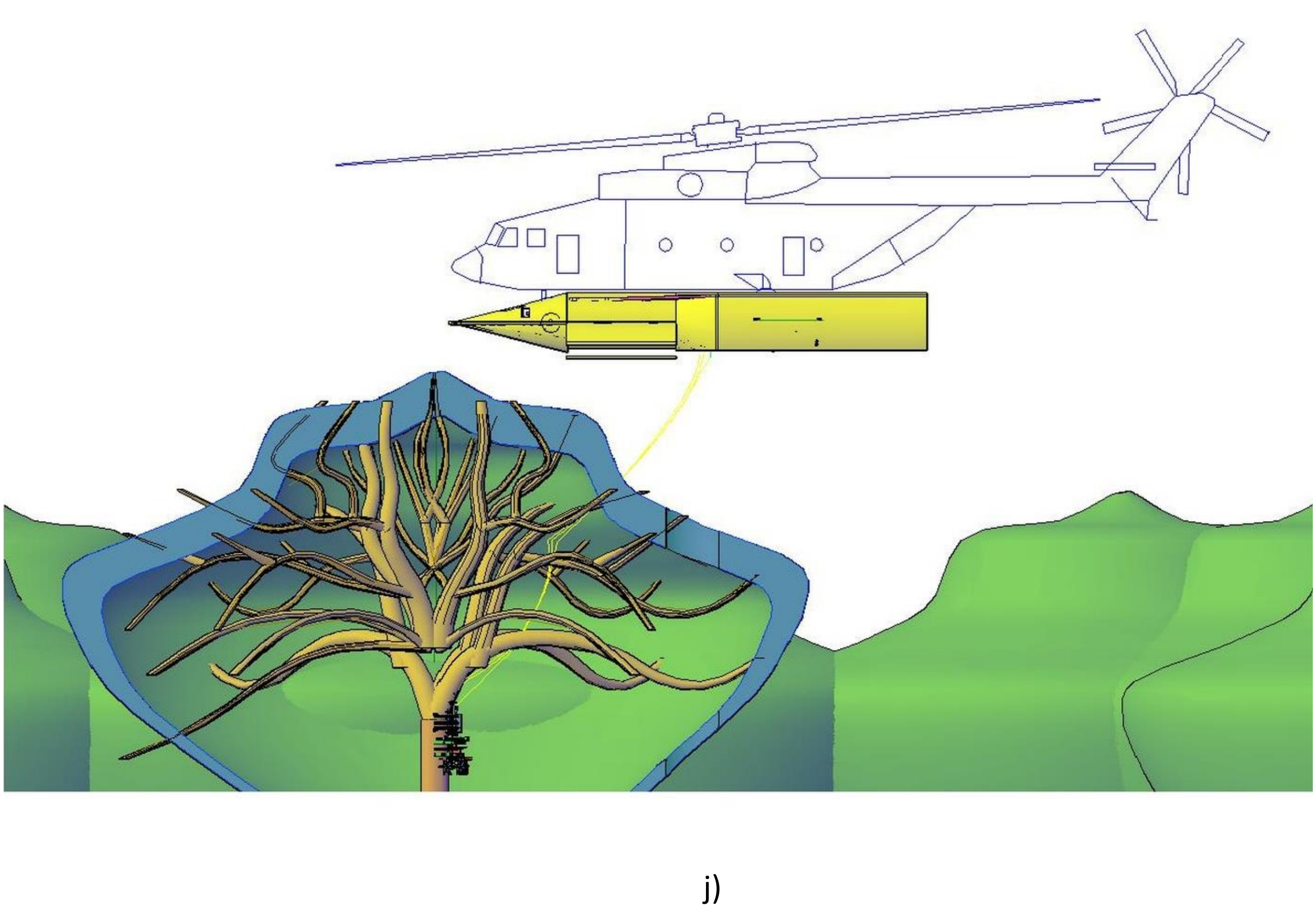

j)

Figure 37. Penetration operation in the canopy and MH approach operation to the trunk: a) X-ray view of the URIEL pod, note the command post oriented towards the target tree; b) MH module performing penetration into the upper canopy of the target; c) X-ray view of the upper canopy of the target being penetrated by the delimbing blades, note cut branches falling below the MH; d) MH module launching a steel dart for trunk approach operation; e) Detail of the steel dart approaching the trunk, note the steel cable attached to the rear of the dart; f) Launching of the second steel dart to stabilize the trunk approach by the MH; g) Detail of the darts penetrating the trunk and the stretched steel cables; h) Front view of the MH being pulled towards the trunk by the steel cables attached to the darts; i) Detail of the MH fully open approaching the trunk; j) MH leaning against the trunk.

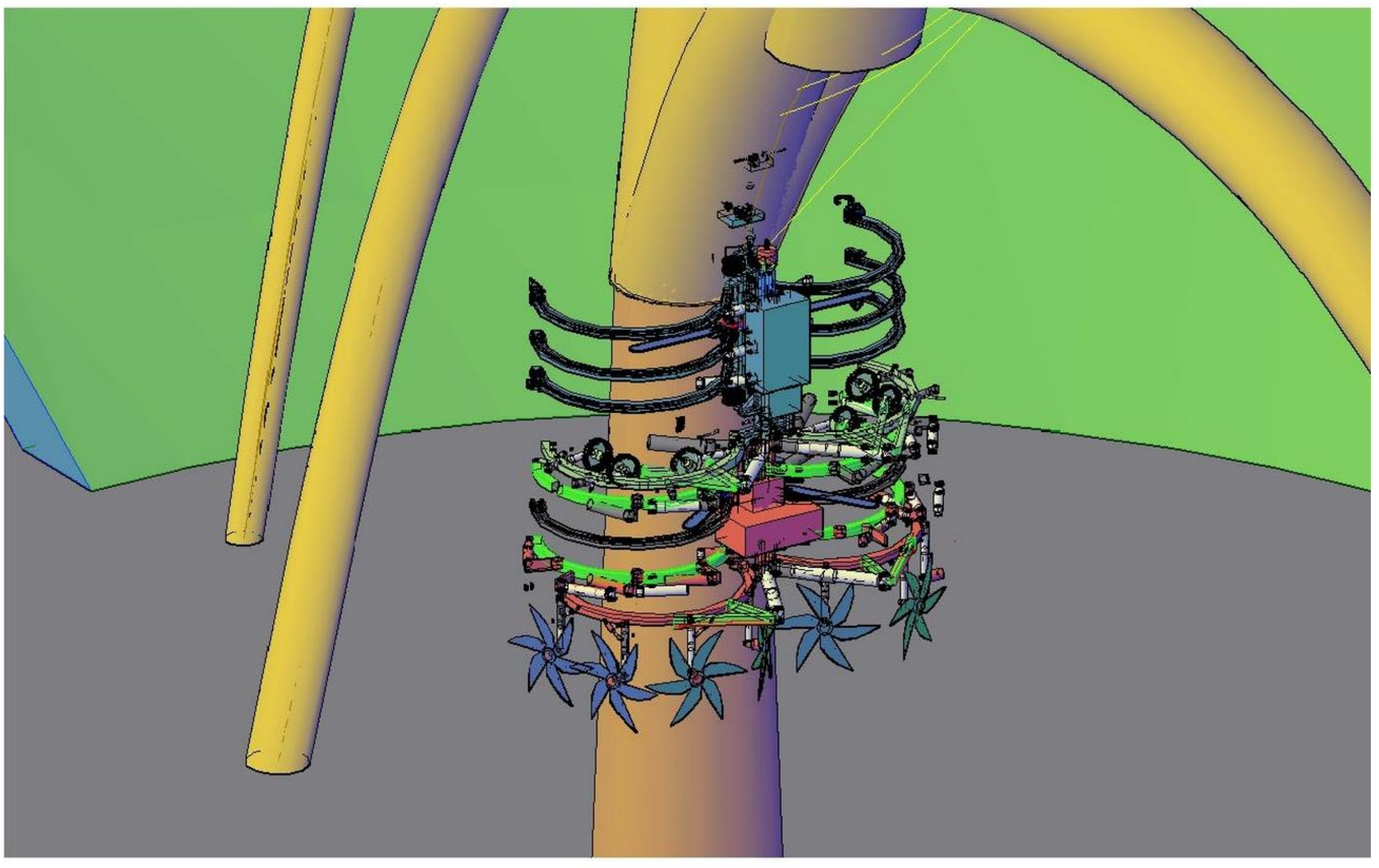

a)

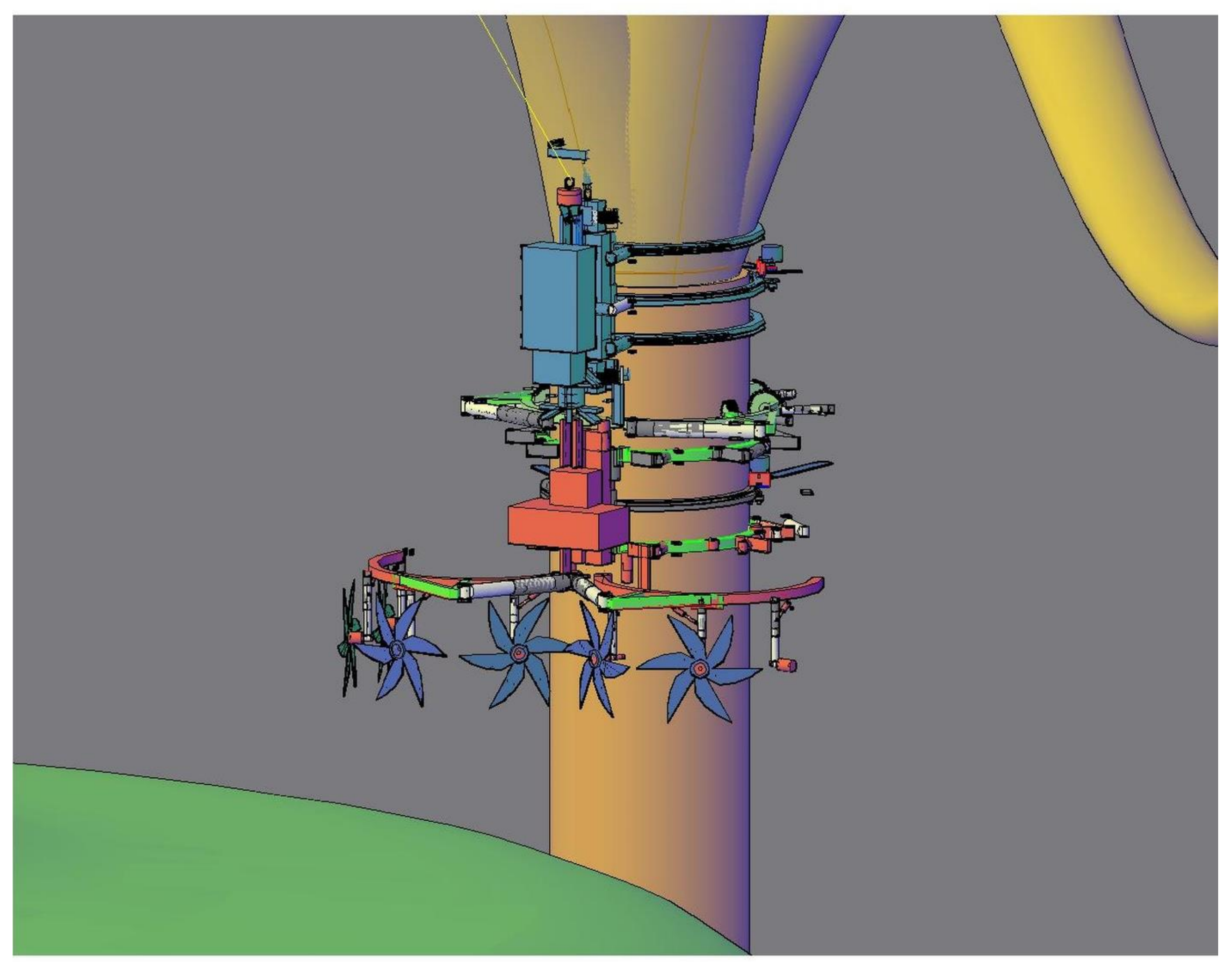

b)

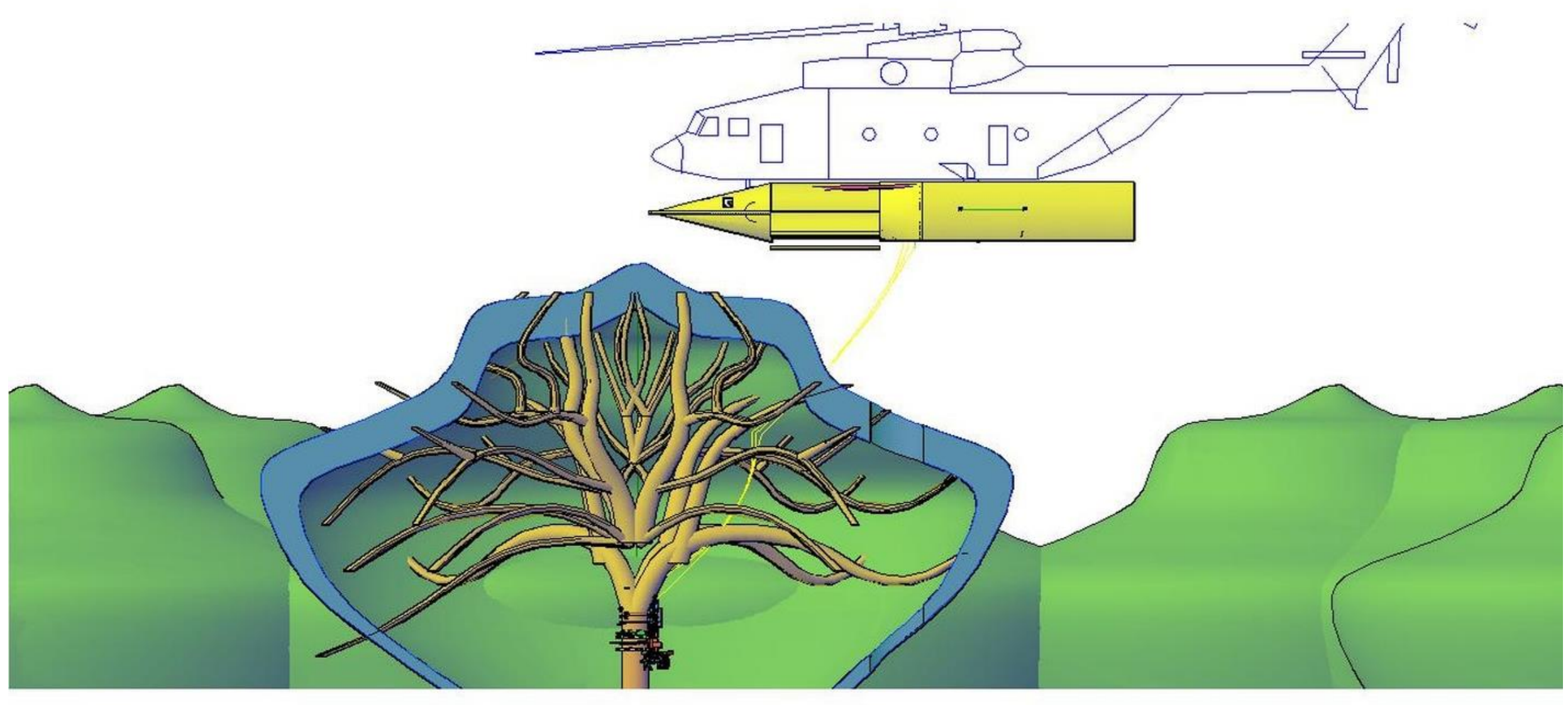

c)

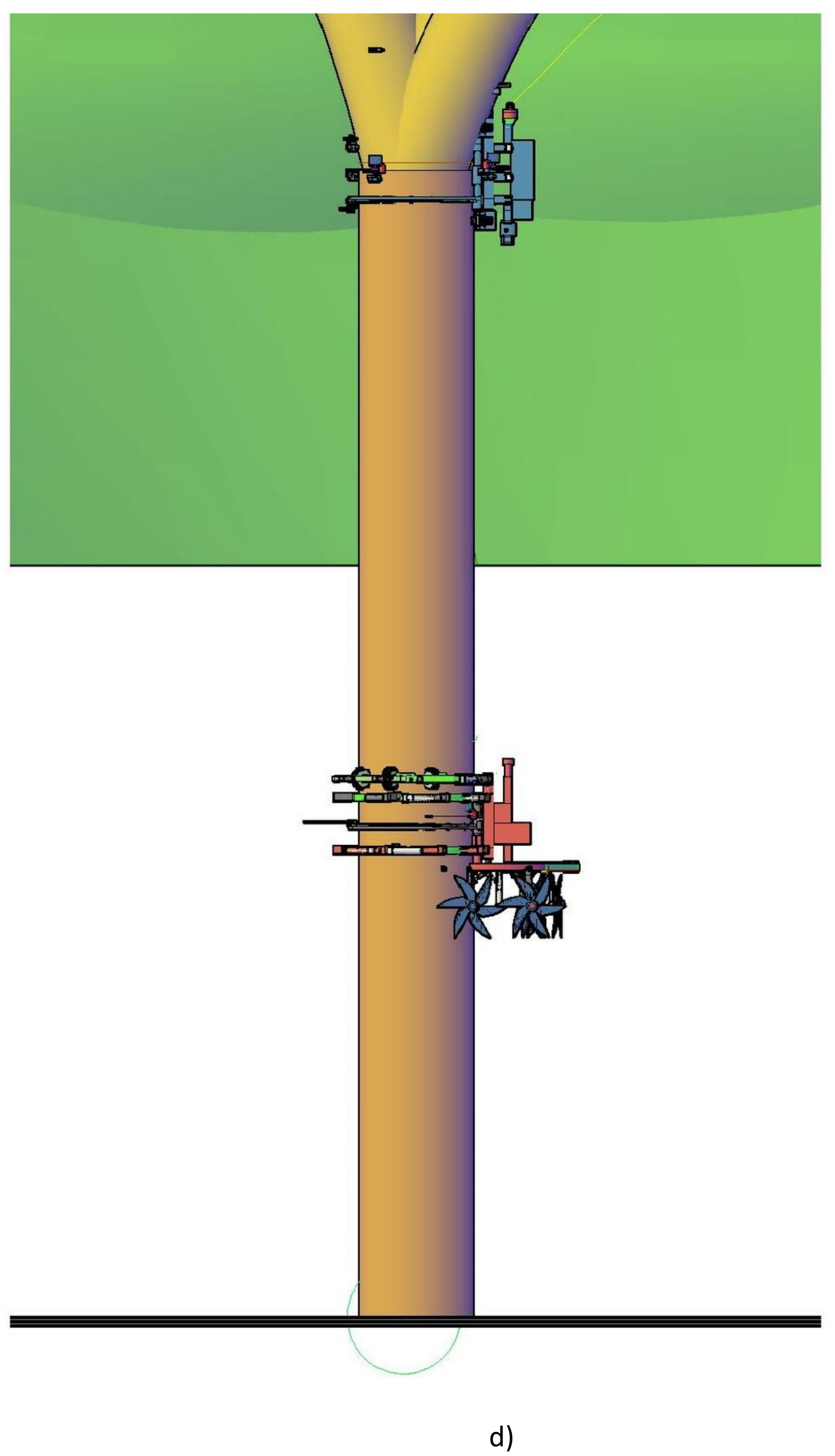

d)

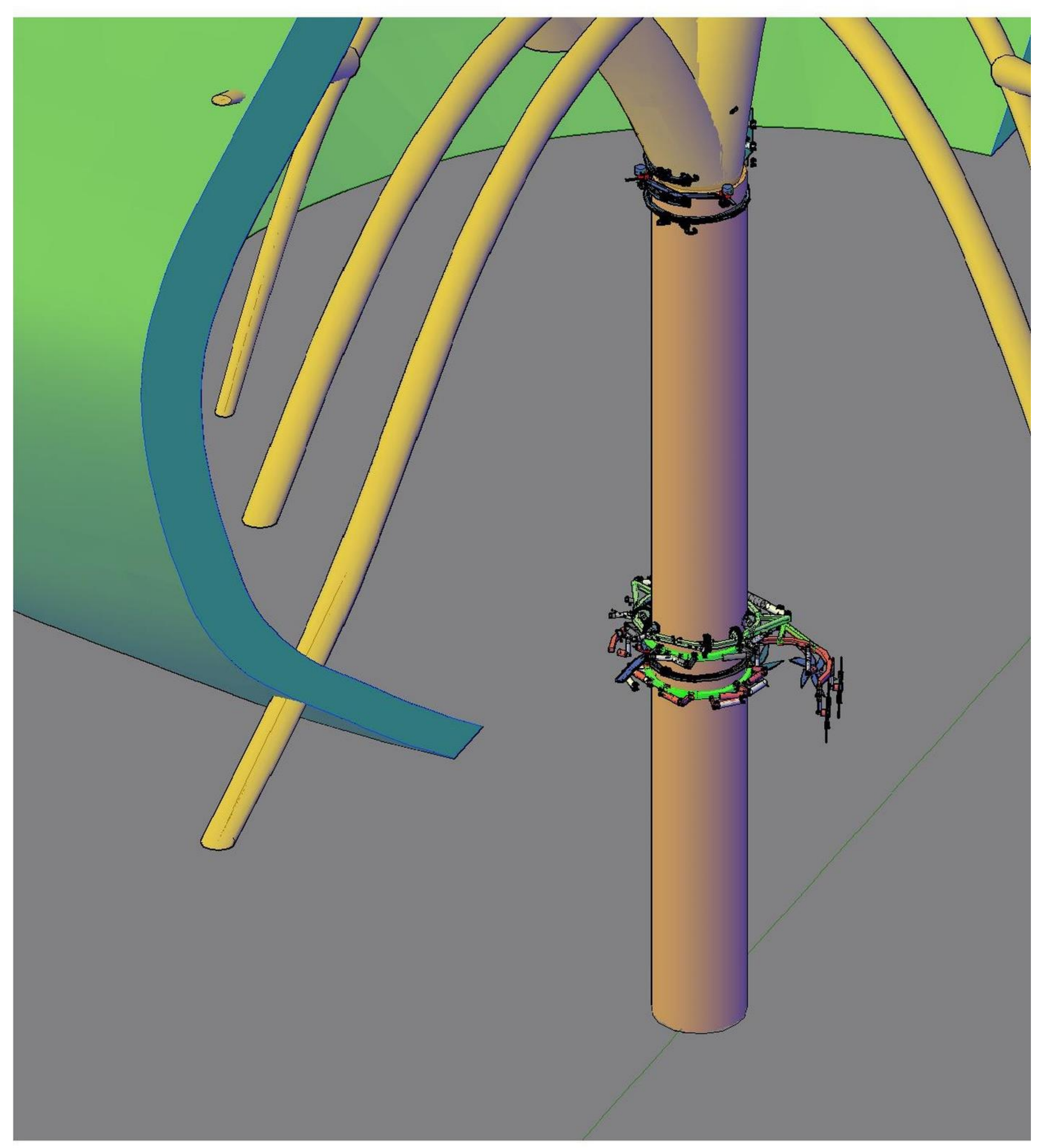

e)

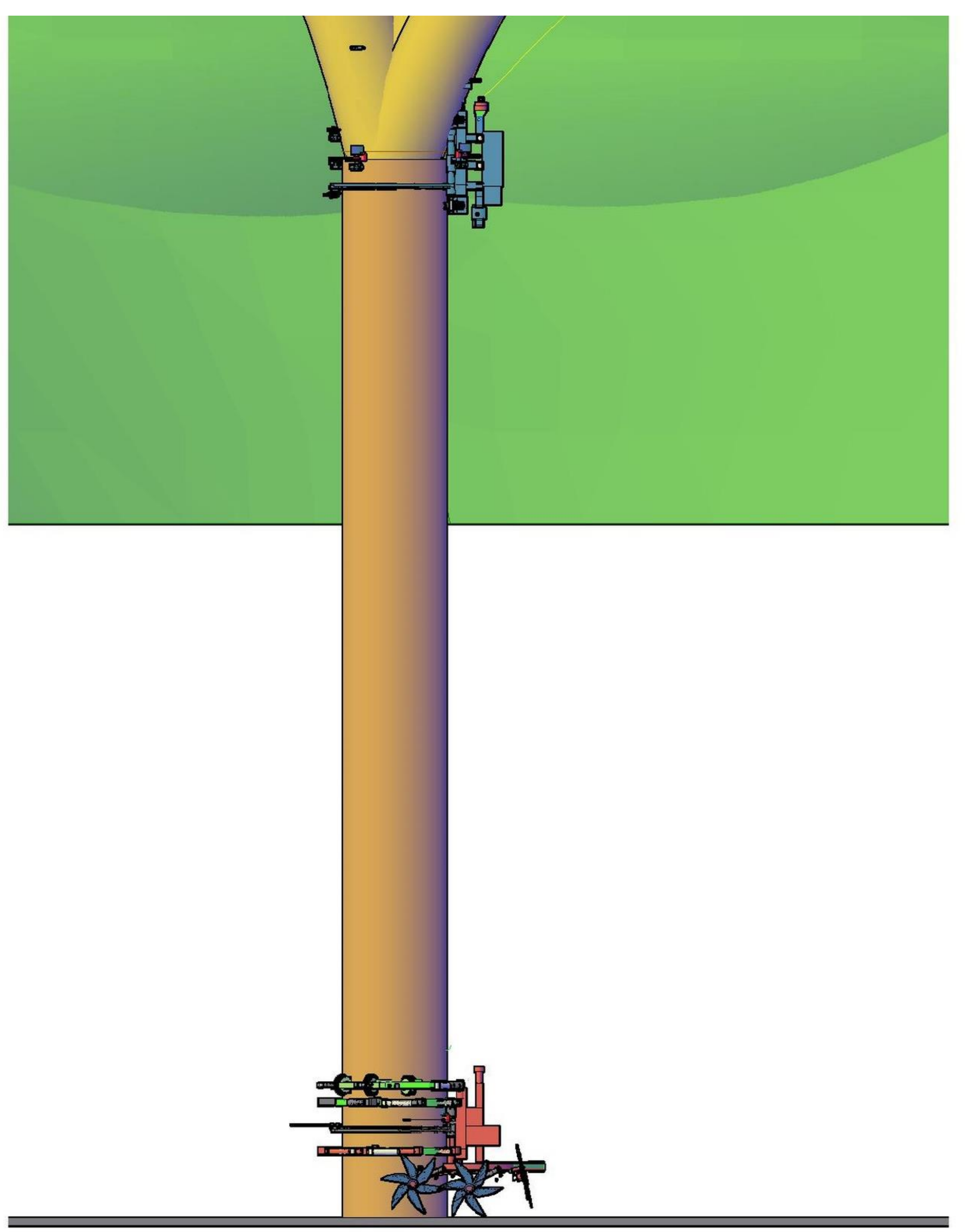

f)

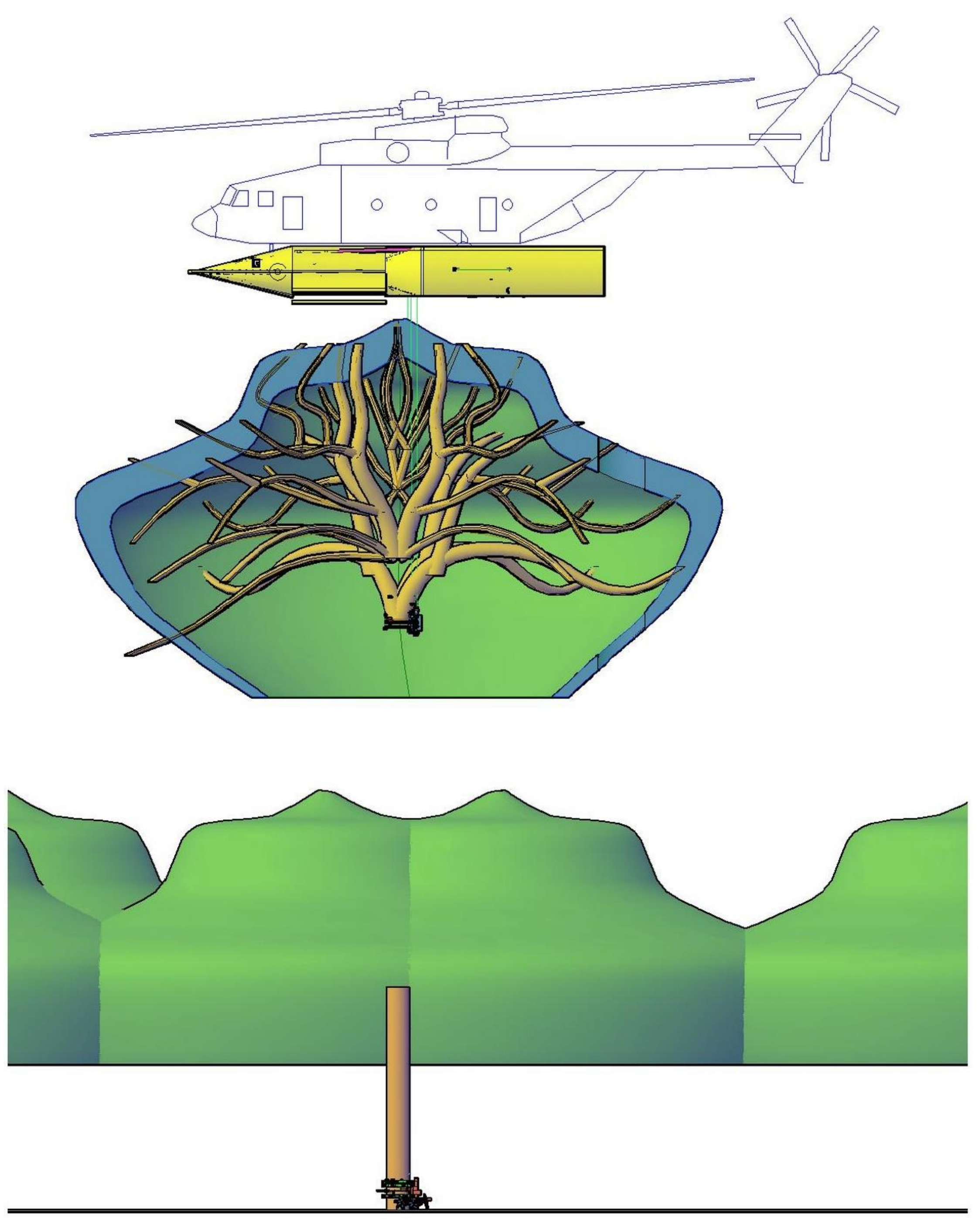

g)

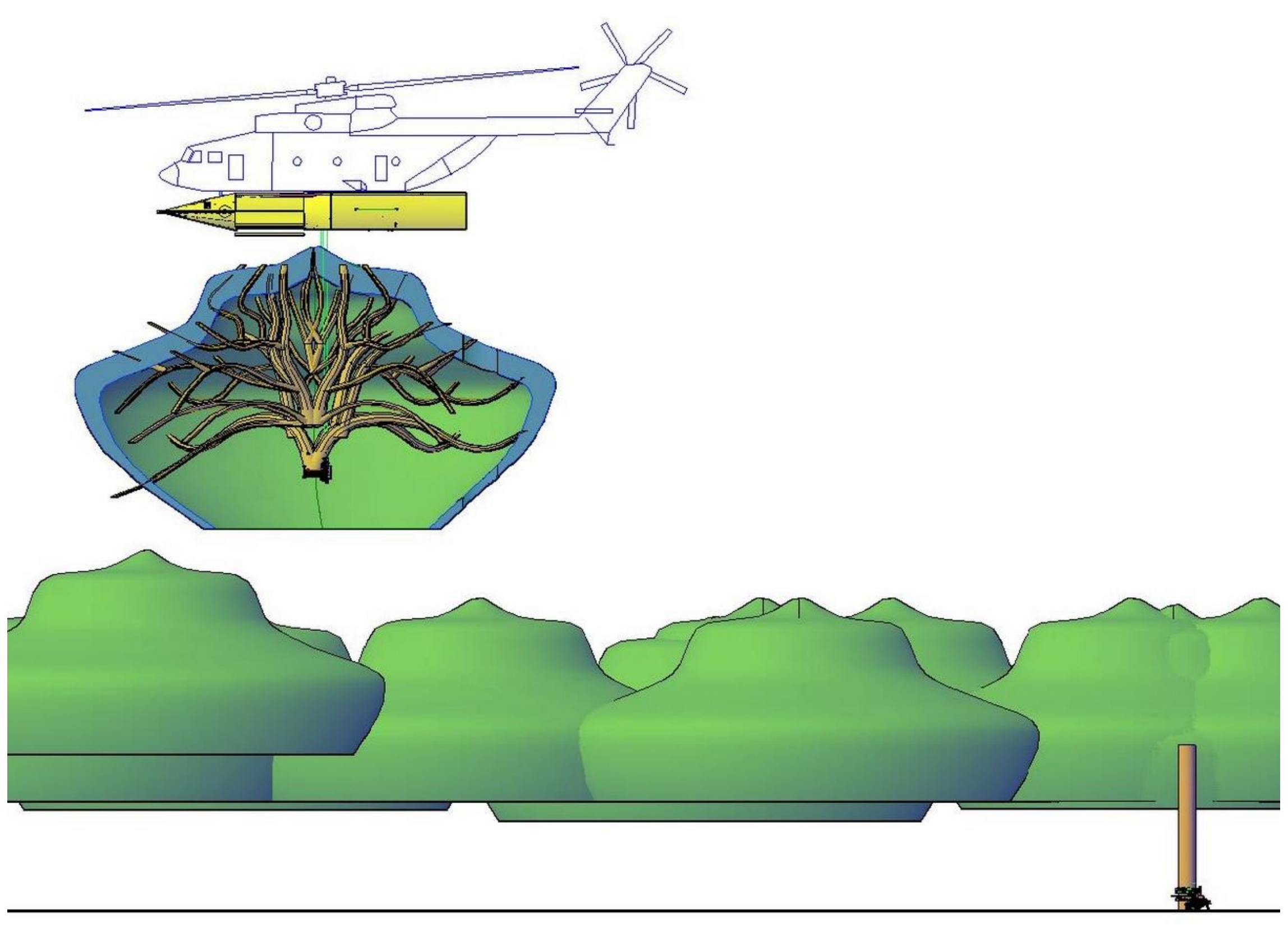

h)

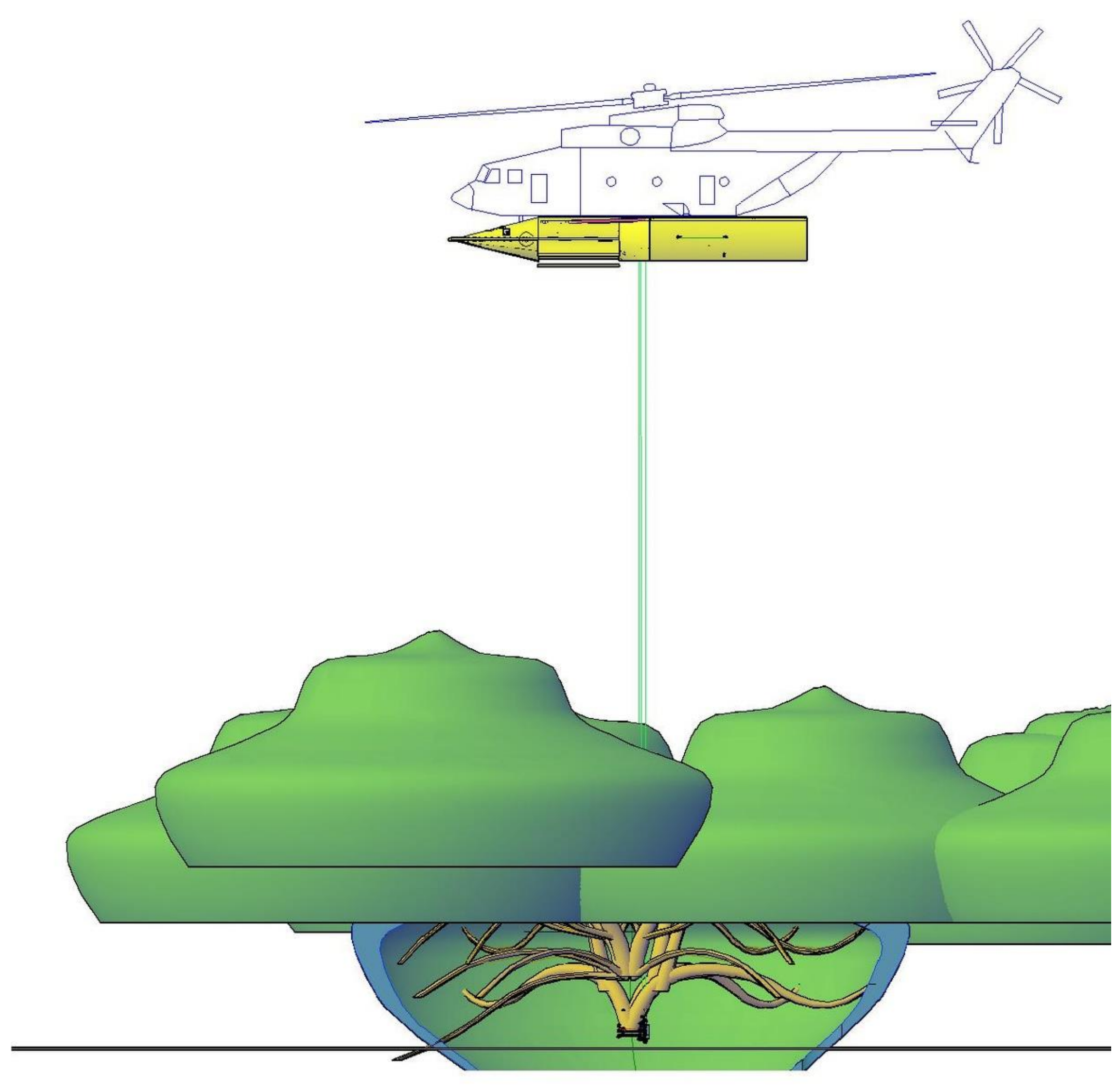

i)

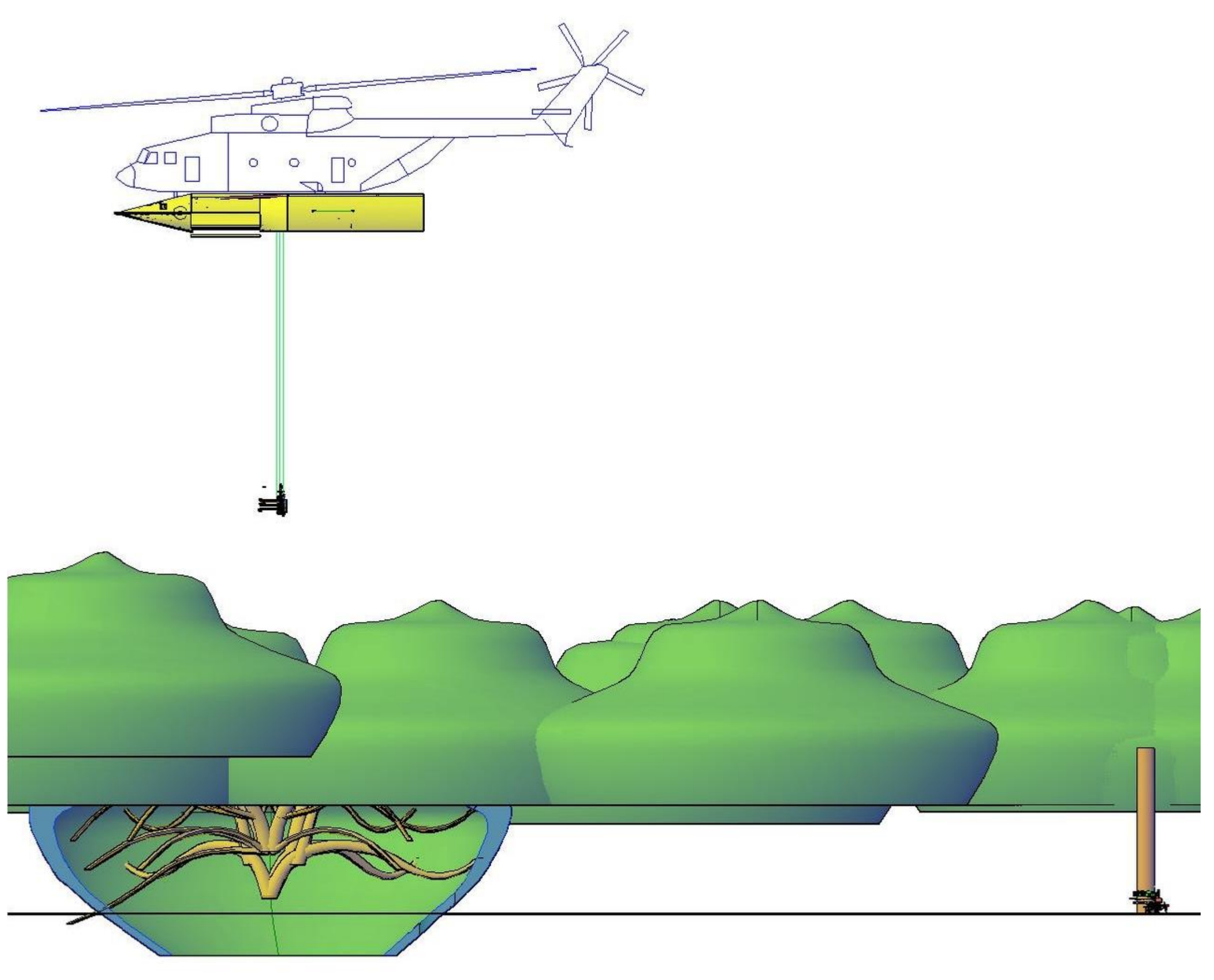

j)

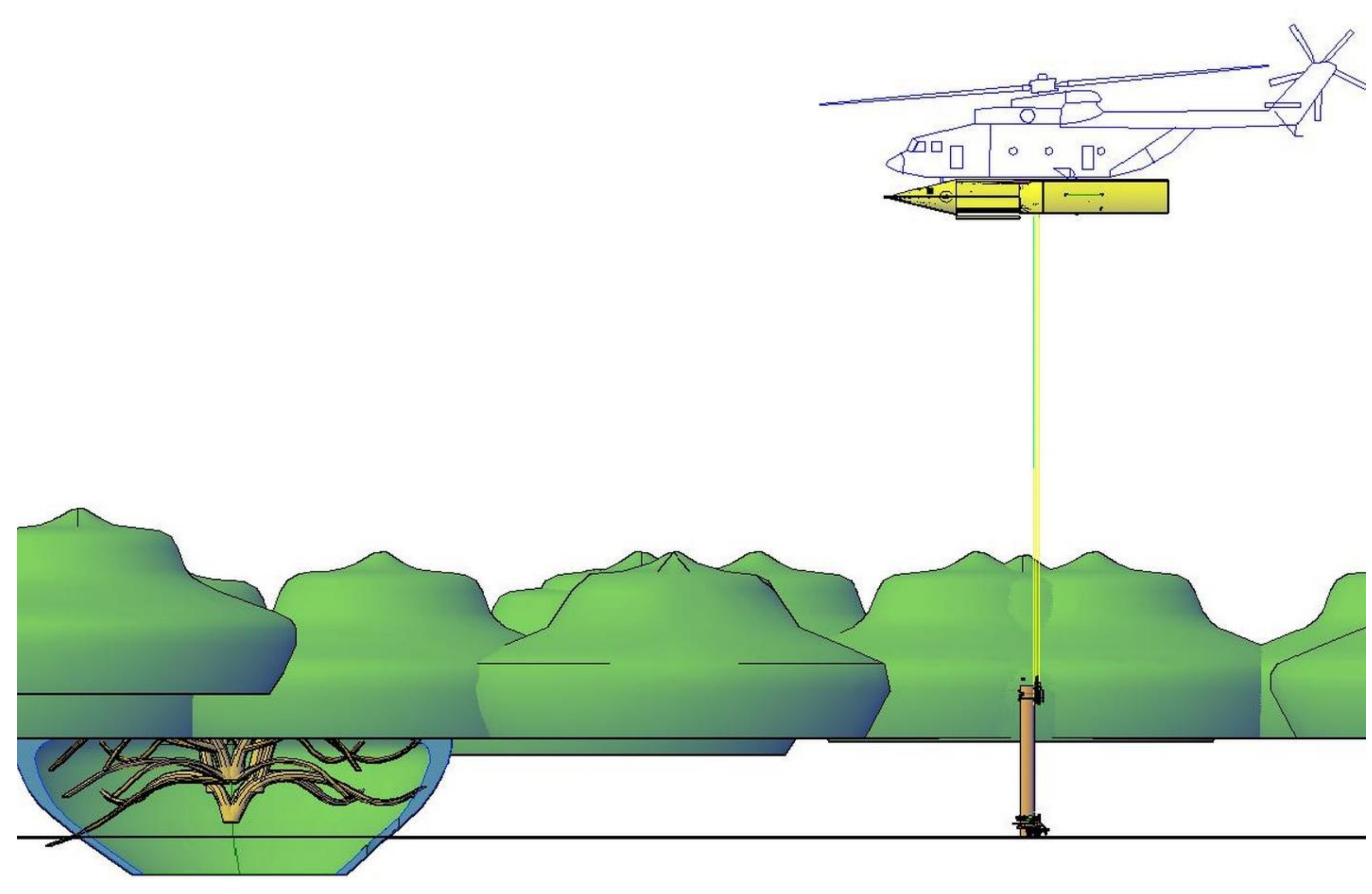

k)

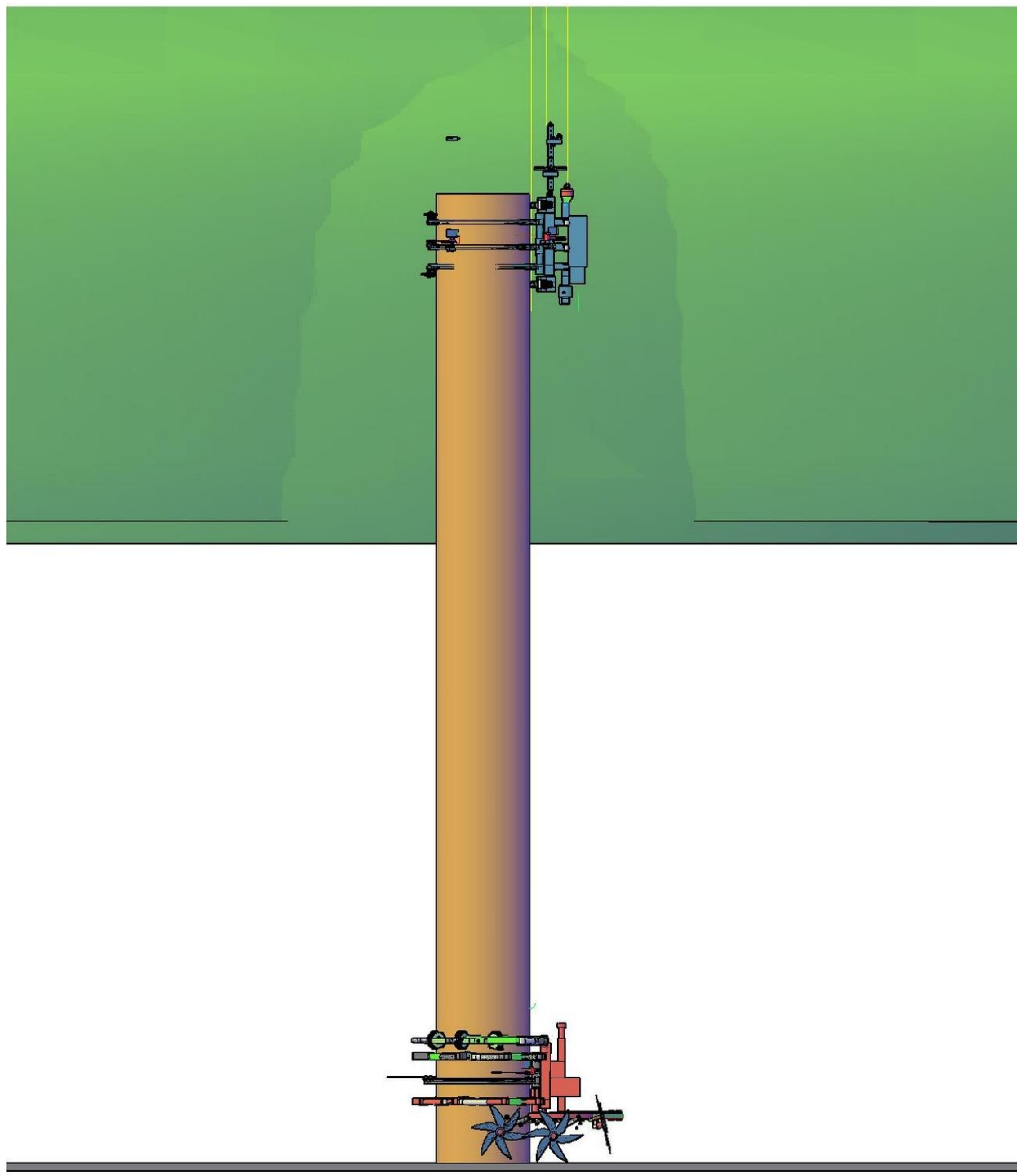

l)

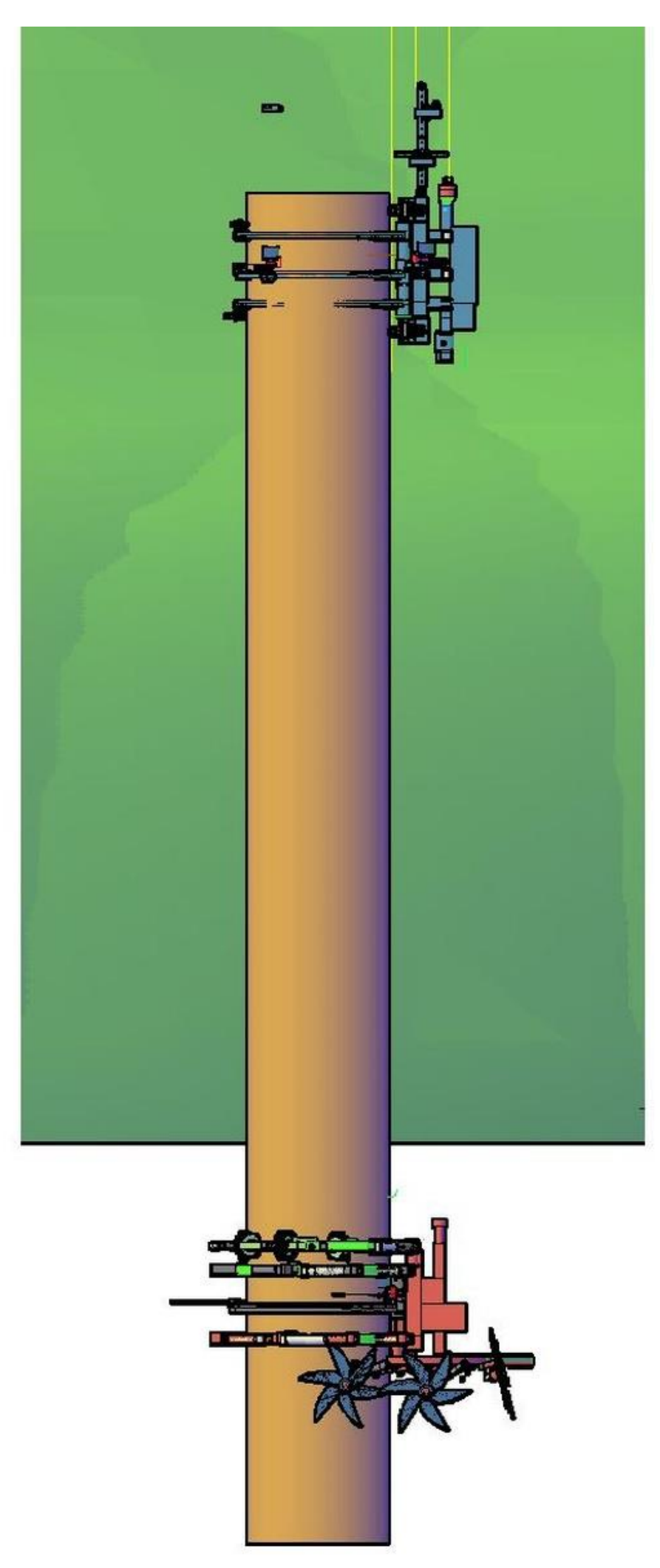

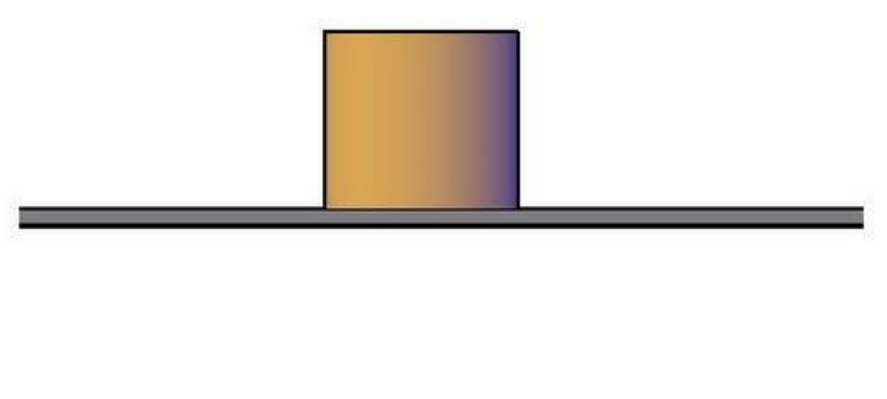

m)

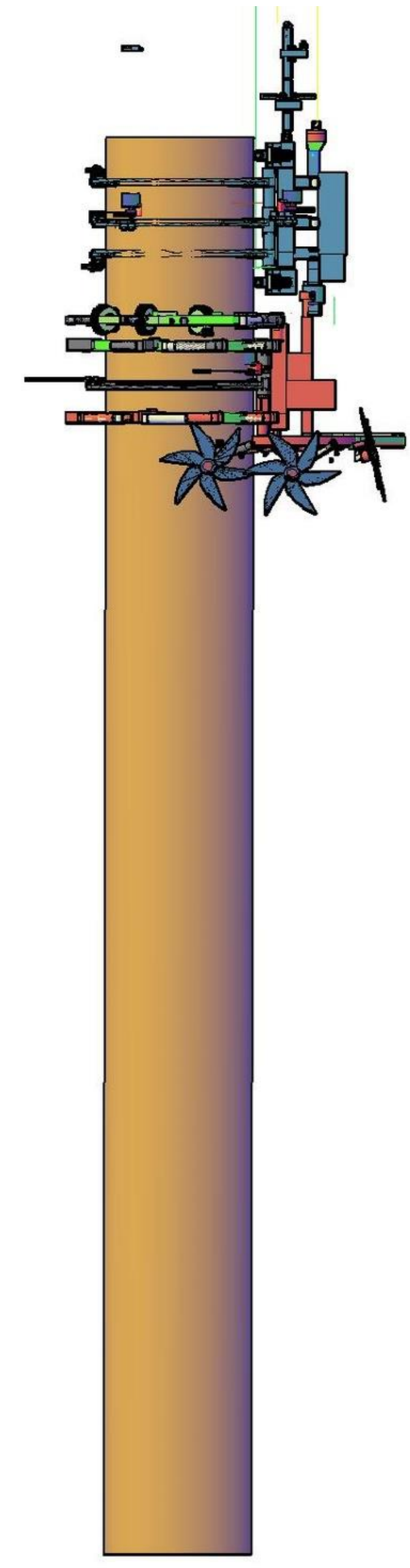

n)

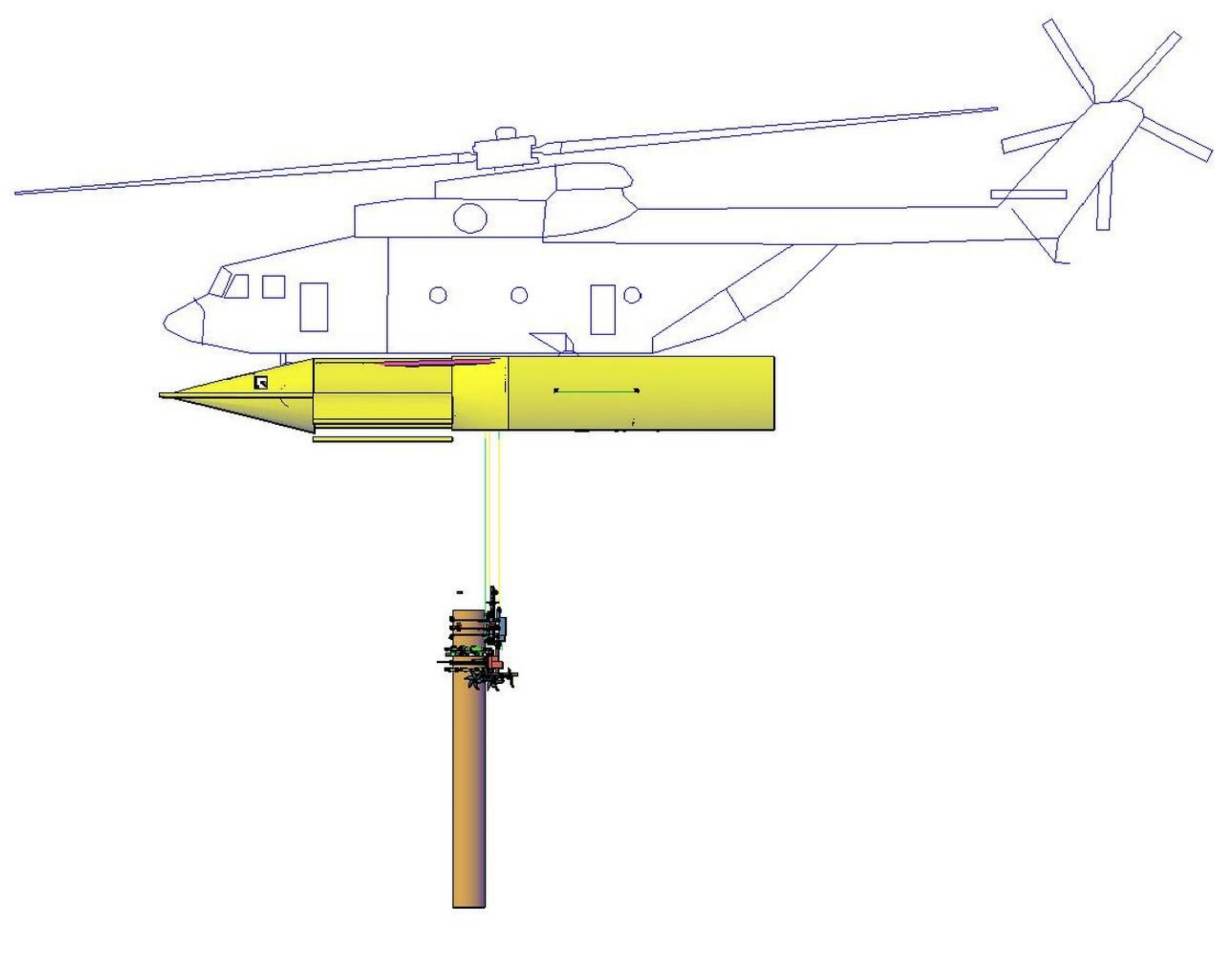

o)

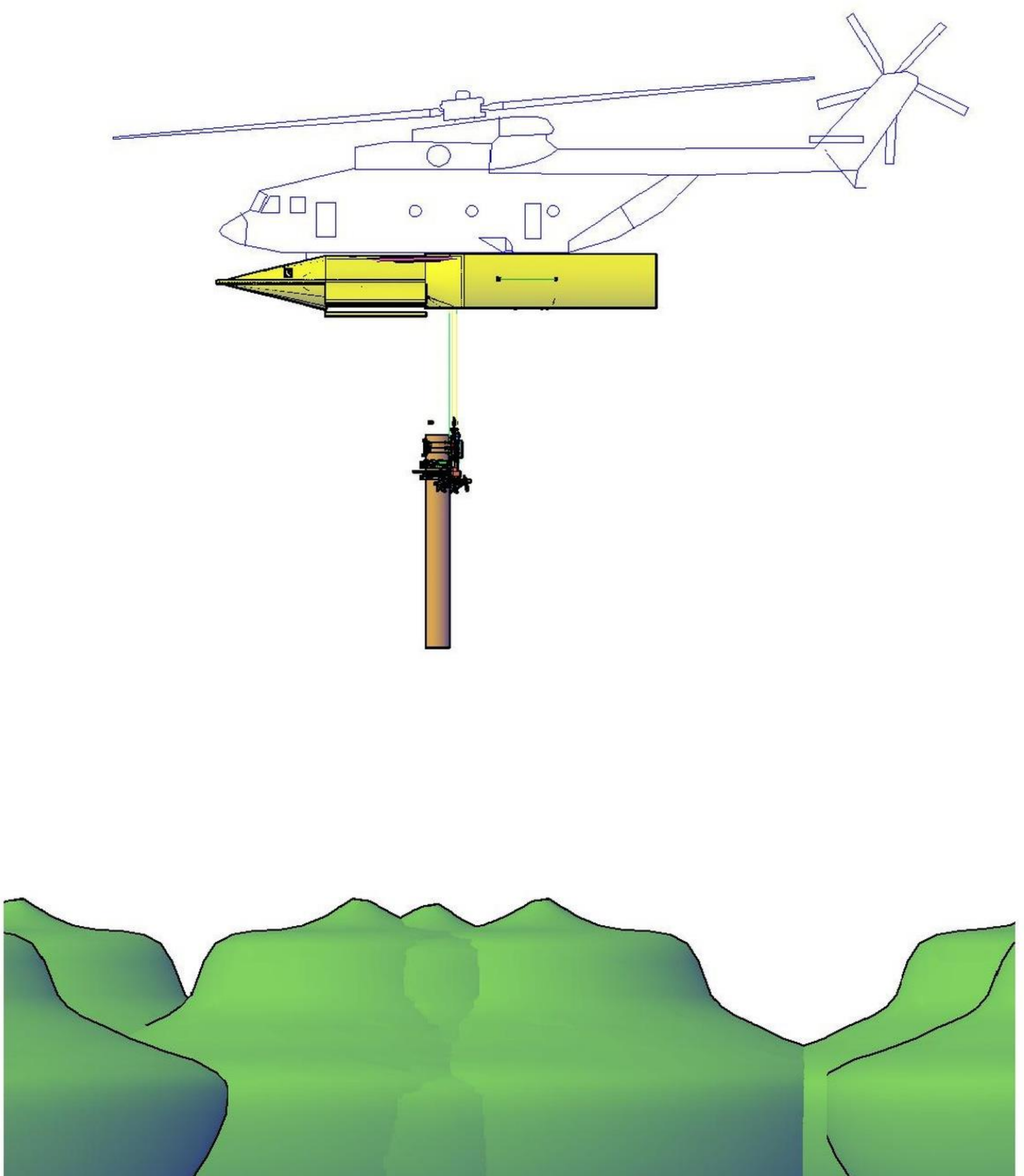

p)

Figure 38. MH Module on-the-go: a) MH Module about to close on the trunk; b) MH attached to the trunk; c) View of the MH Module attached to the trunk, note the loose steel cables stabilizing the Pod; d) Decoupling of the decoupling subsystem from the trunk cutting subsystem; e) Perspective view of the trunk cutting subsystem descending to the base of the tree; f) Decoupling system cutting the crown and securing the sections with the clamps; g) URIEL system maneuvering to raise the crown and stabilization subsystem performing compensatory force operations to stabilize the cut crown; h) URIEL system moving the crown to a site for encapsulated crown deposition on the forest floor; i) Encapsulation of the cut crown in a predefined region with forest space for encapsulation of the crown on the ground without damaging surrounding trees; j) Decoupling of the decoupling subsystem and elevation above the forest canopy; k) Return of the URIEL System to the target trunk region; l) Coupling of the decoupling subsystem with the upper part of the trunk; m) The trunk cutting subsystem performed the cut and the stabilization subsystem performs the elevation of the trunk with walking upwards from the trunk cutting subsystem; n) The trunk cutting subsystem couples to the decoupling subsystem; o) View of the URIEL System with the MH securing the elevated trunk to a safe position on the URIEL Pod; p) General view of the URIEL System with the trunk elevated above the forest canopy, note the base of the cut target tree.

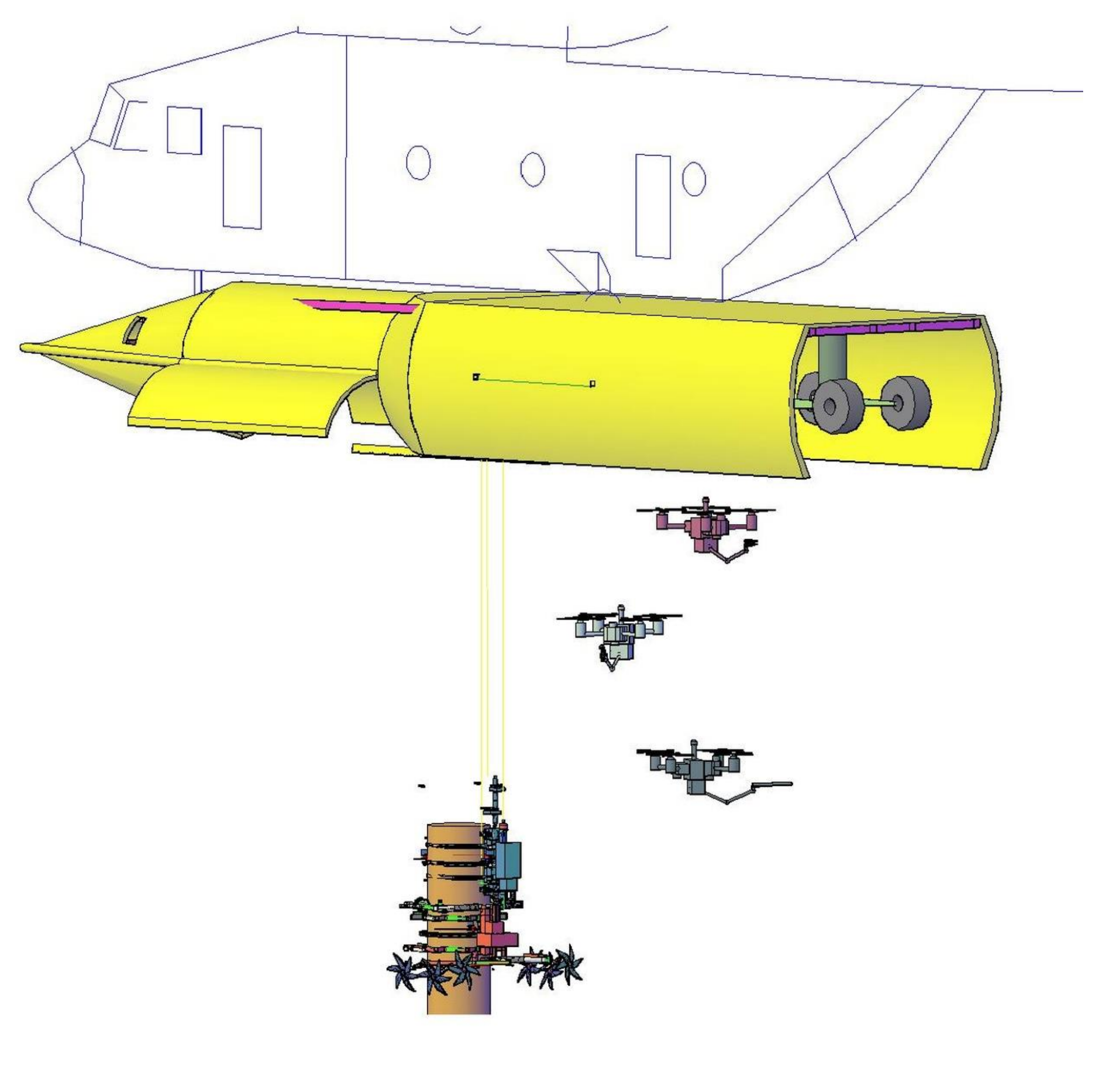

a)

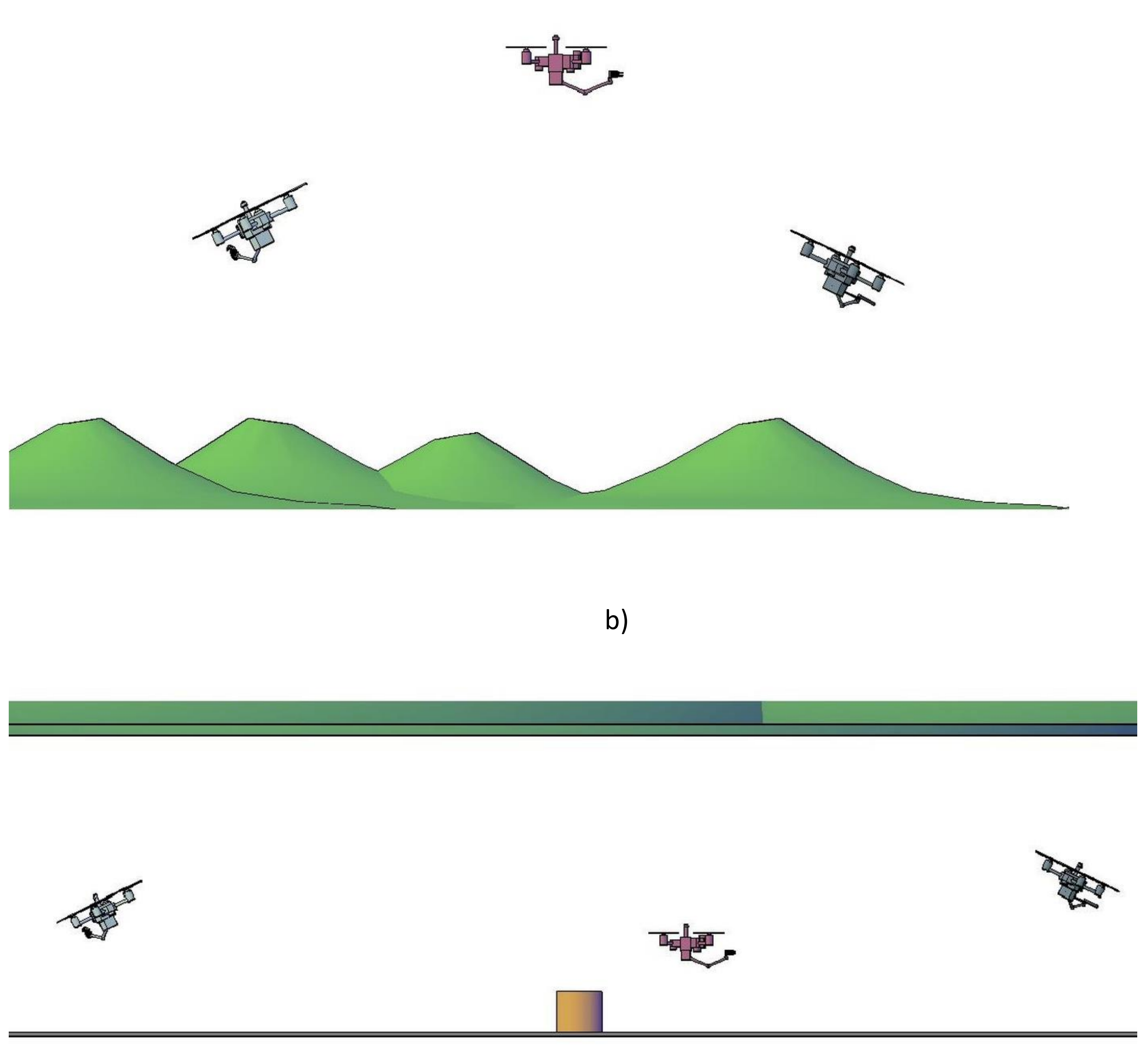

b)

c)

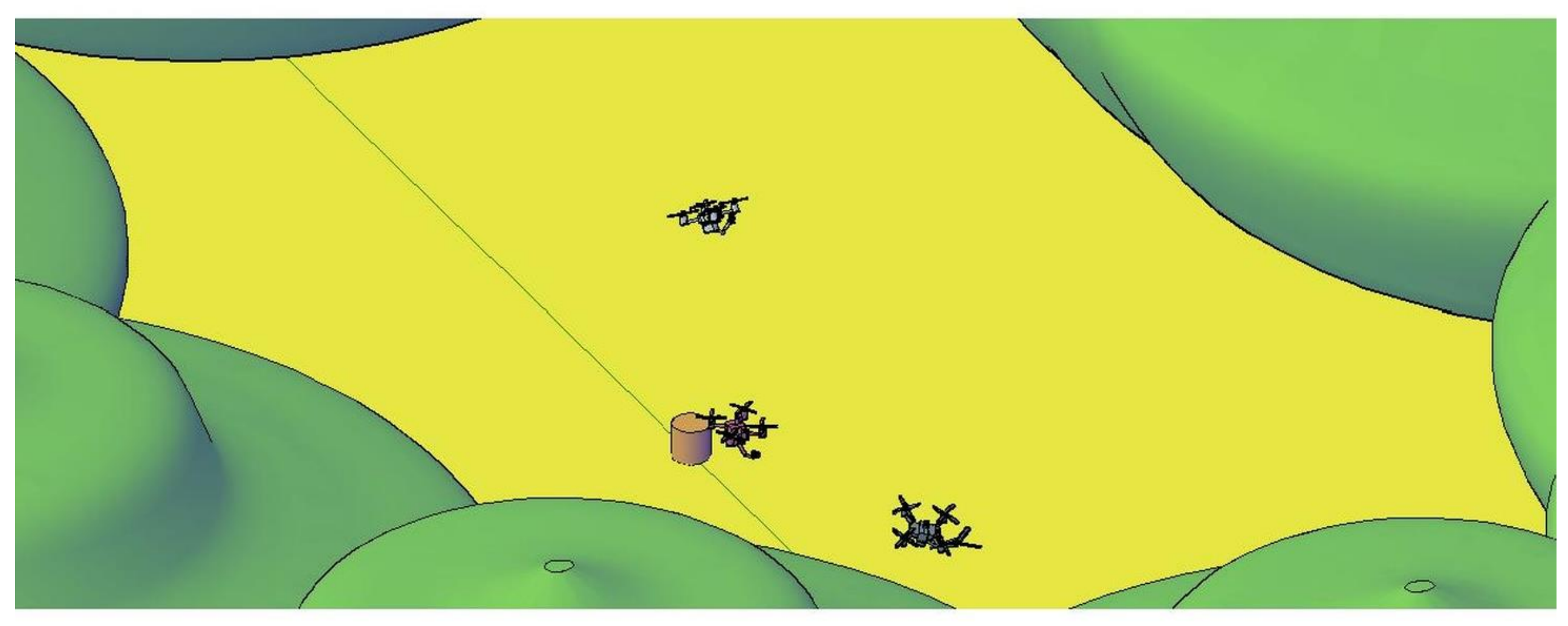

d)

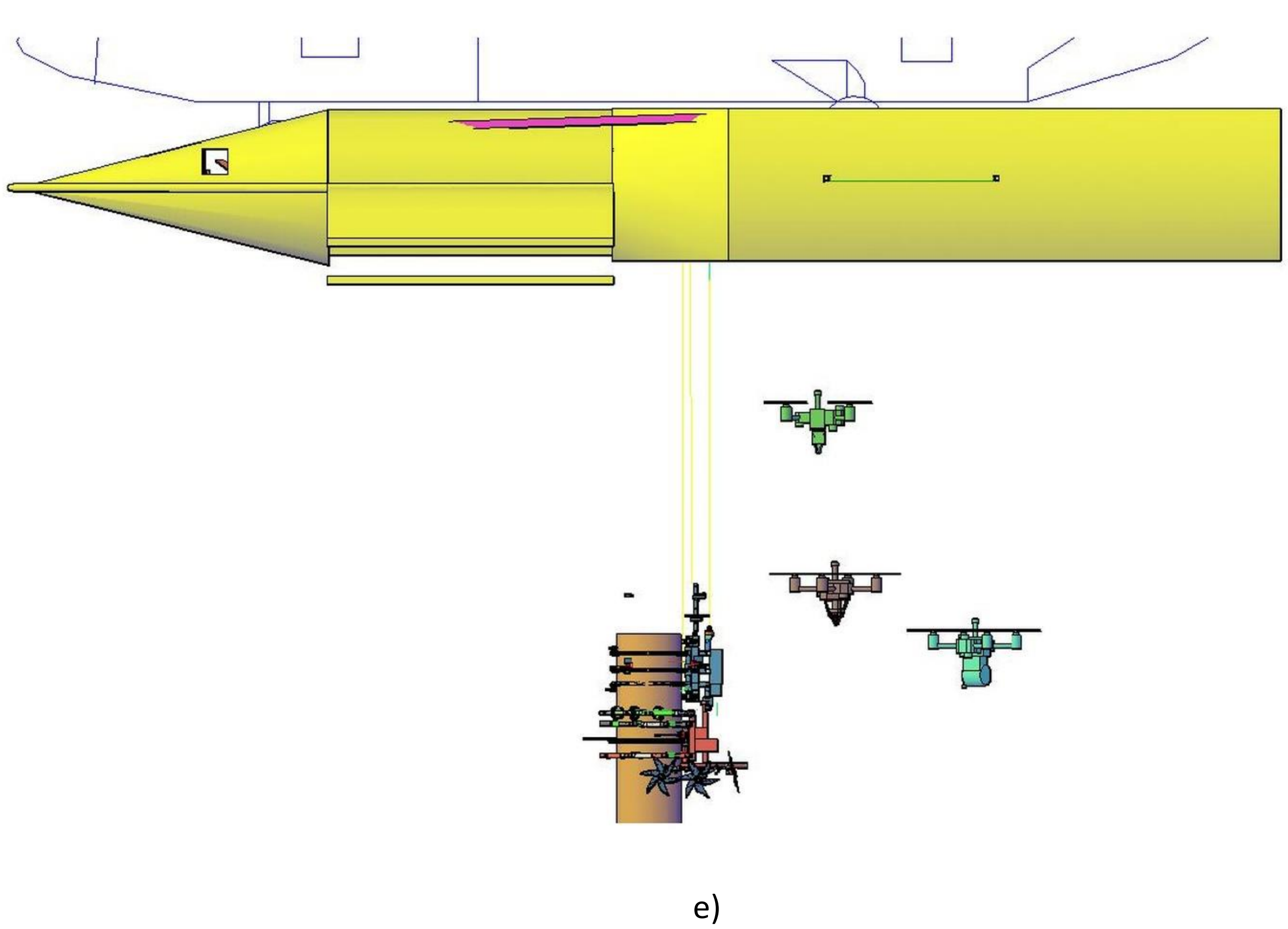

e)

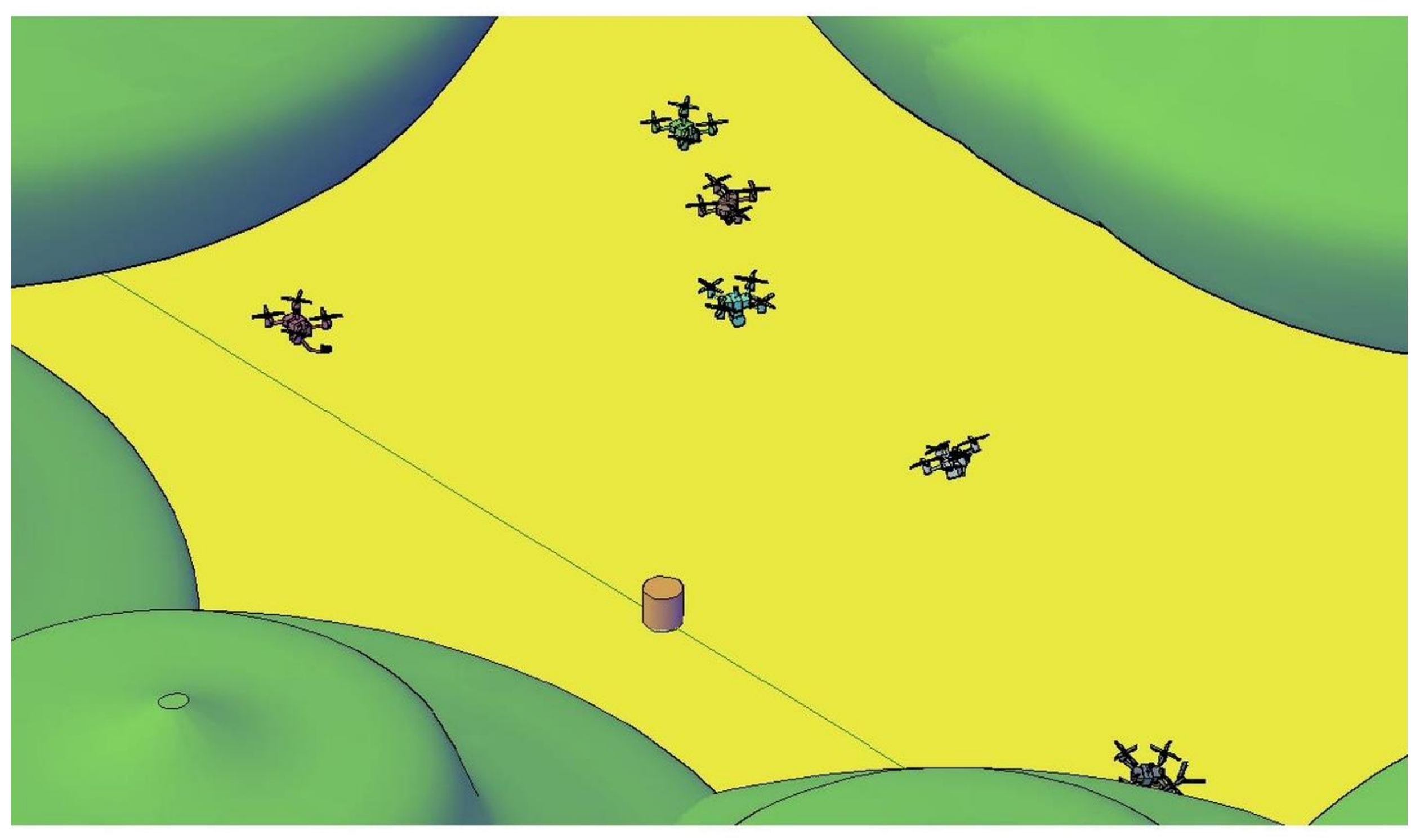

f)

g)

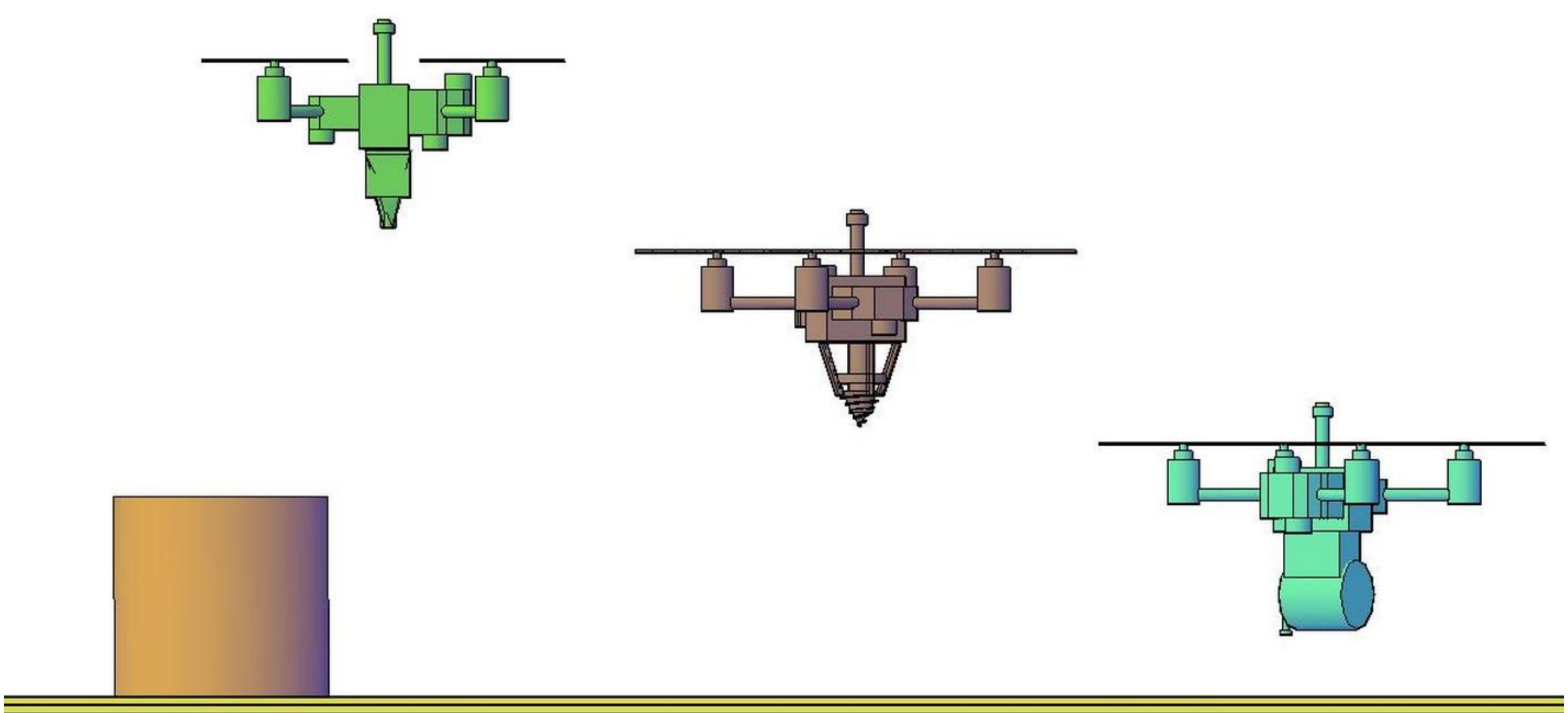

h)

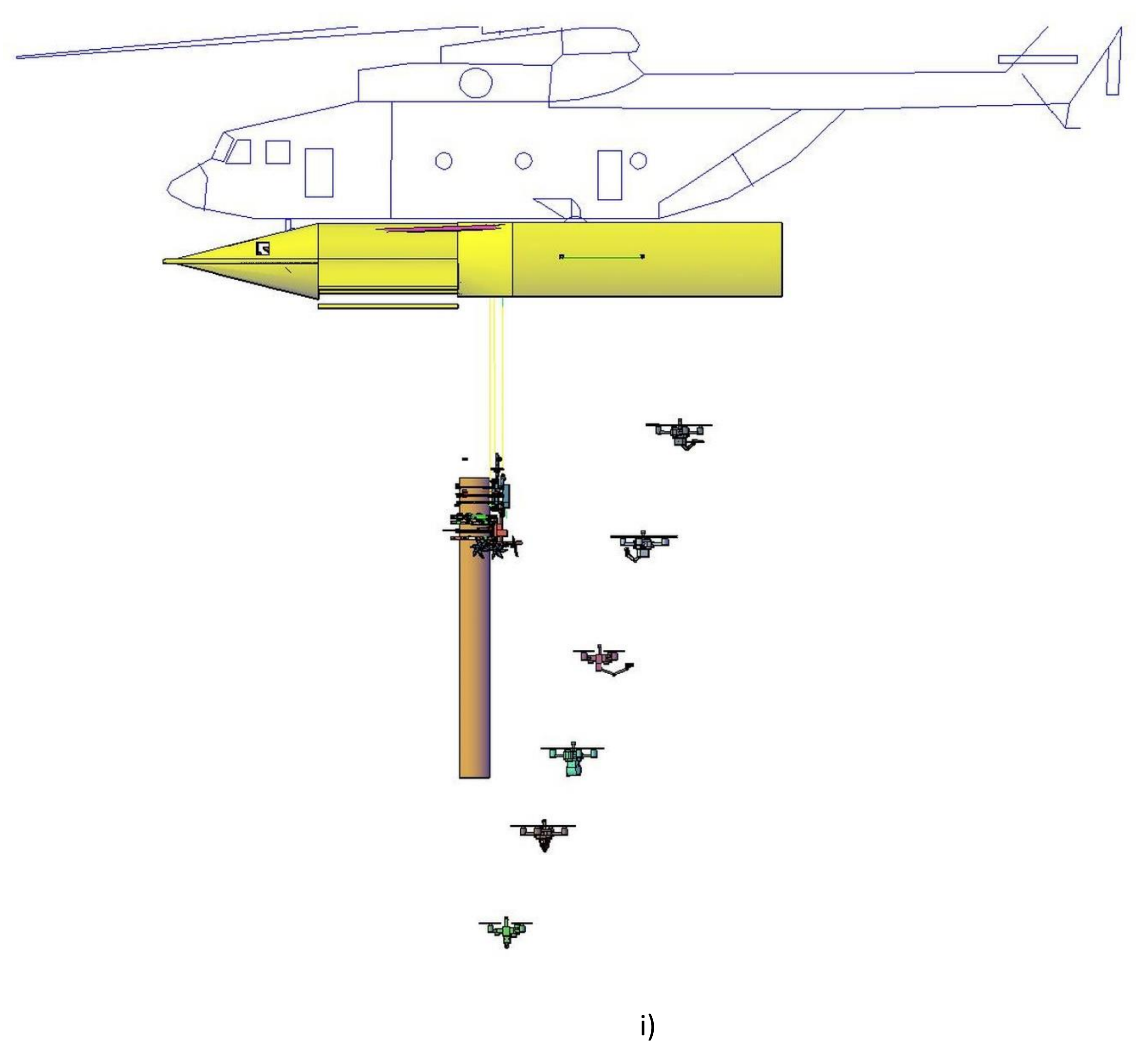

i)

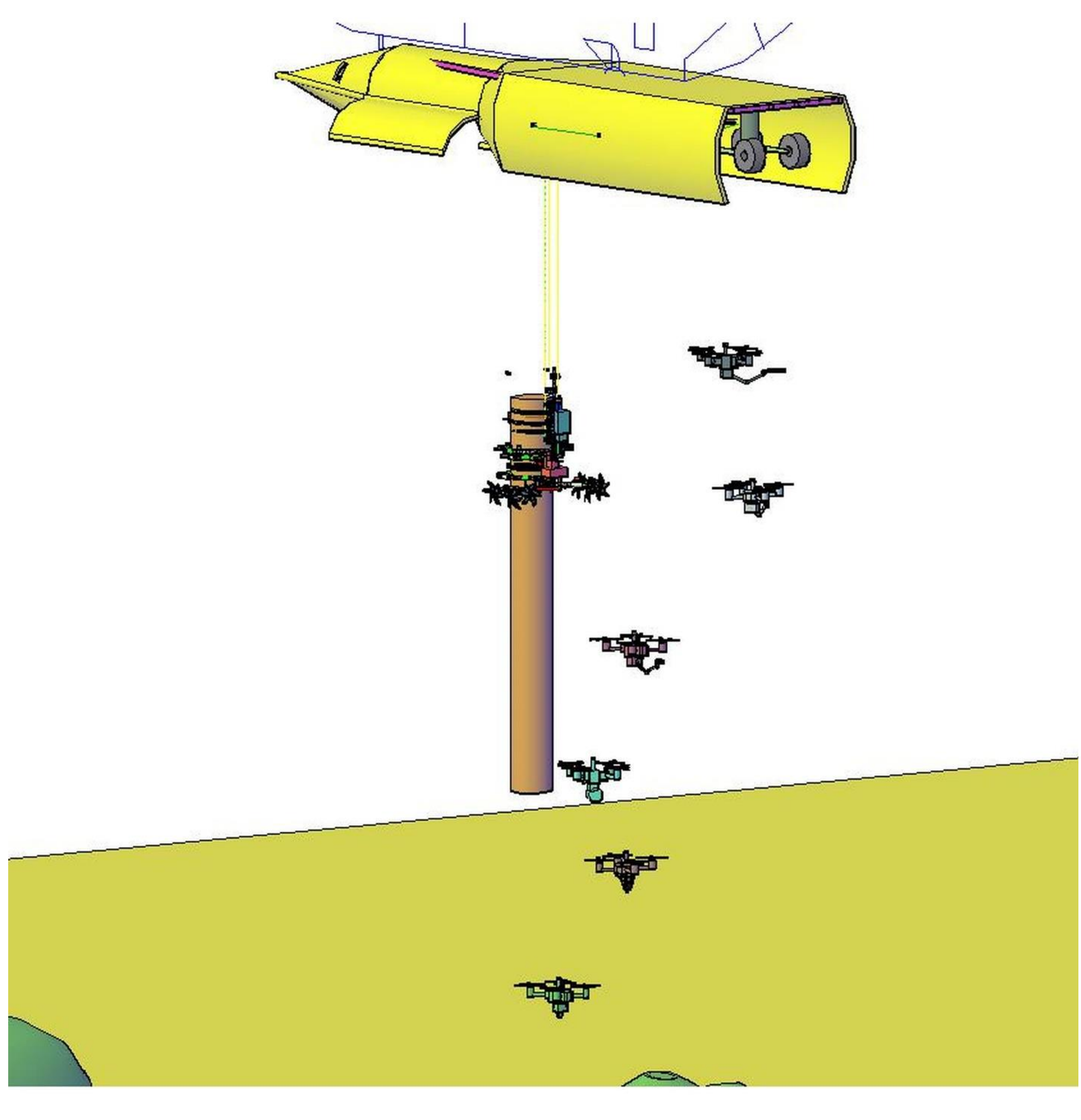

j)

Figure 39. MHST on-the-go module: a) Drone squadron launching for area clearing (vine cutting drone, secondary tree cutting drone, and tree girdling drone); b) Travel below the level of the URIEL Pod by the clearing squadron drones; c) Clearing squadron performing vine cutting, secondary tree cutting, and tree girdling actions; d) View from the URIEL Pod of the clearing squadron operating in the clearing formed by the removal of the target tree; e) Drone squadron launching for tree planting (irrigation drone, pitting drone, and planting drone); f) View from the URIEL Pod of the drone squadrons in action;

g) Planting squadron in action while the clearing squadron finishes its work; h) Planting squadron in operation: irrigation drone wets the soil at the planting position, then the planting drone digs a shallow or deep hole depending on the species, then the planting drone drops a seed or seedling into the hole, and finally the irrigation drone sprays water over the seedling or seed; i) Return of the squadrons to the MHST Module; j) Perspective view of the drones returning to the URIEL Pod.

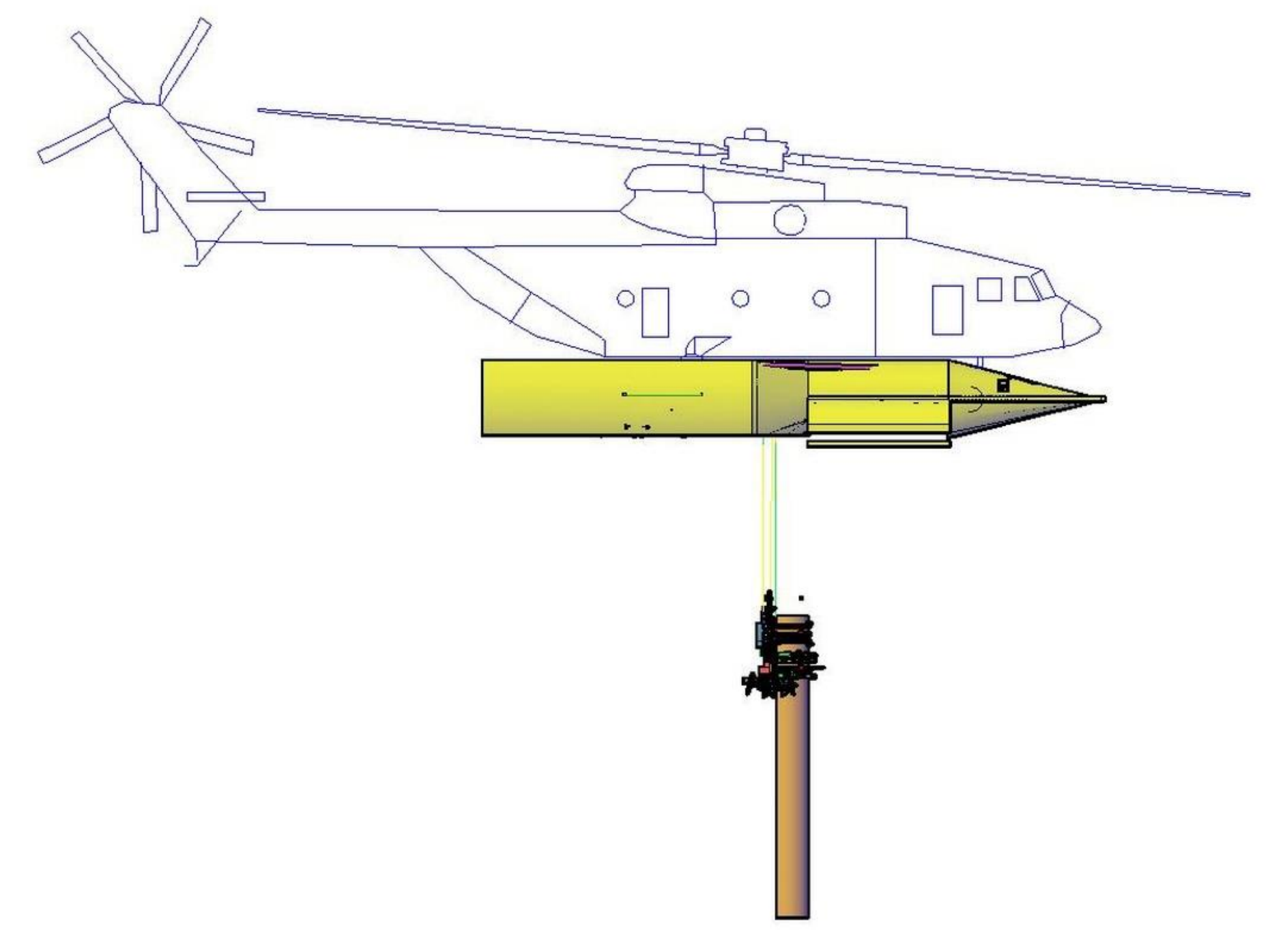

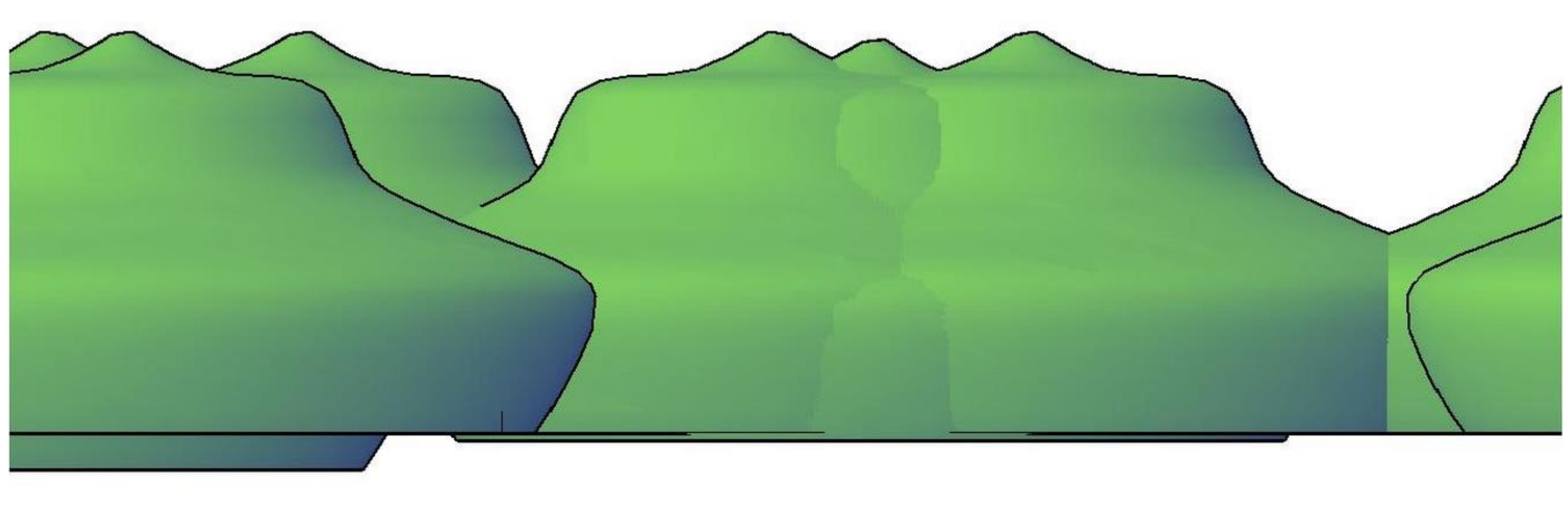

Figure 40. URIEL go home system with harvested stem.

“E pur si muove…”